\documentclass{article}
\usepackage{ijcai19}

\usepackage{times}
\usepackage{soul}
\usepackage{url}
\usepackage[hidelinks,draft]{hyperref}
\usepackage[utf8]{inputenc}
\usepackage[small]{caption}
\usepackage{graphicx}
\usepackage{amsmath}
\usepackage{multirow}
\usepackage{amssymb}
\usepackage{verbatim}
\usepackage{subcaption}
\usepackage{caption}
\usepackage{booktabs}
\usepackage{algorithm}
\usepackage{algorithmic}
\usepackage{xcolor}
\usepackage{natbib}
\usepackage{xspace}
\usepackage{cleveref}

\title{On the Effectiveness of Low Frequency Perturbations}
\usepackage{authblk}

\author{Yash Sharma}
\author{Gavin Weiguang Ding}
\author{Marcus A. Brubaker}
\affil{Borealis AI}

\begin{document}

\maketitle

\begin{abstract}
Carefully crafted, often imperceptible, adversarial perturbations have been shown to cause state-of-the-art models to yield extremely inaccurate outputs, rendering them unsuitable for safety-critical application domains. In addition, recent work has shown that constraining the attack space to a low frequency regime is particularly effective. Yet, it remains unclear whether this is due to generally constraining the attack search space or specifically removing high frequency components from consideration. By systematically controlling the frequency components of the perturbation, evaluating against the top-placing defense submissions in the NeurIPS 2017 competition, we empirically show that performance improvements in both the white-box and black-box transfer settings are yielded only when low frequency components are preserved. In fact, the defended models based on adversarial training are roughly as vulnerable to low frequency perturbations as undefended models, suggesting that the purported robustness of state-of-the-art ImageNet defenses is reliant upon adversarial perturbations being high frequency in nature. We do find that under $\ell_\infty$ $\epsilon=16/255$, the competition distortion bound, low frequency perturbations are indeed perceptible. This questions the use of the $\ell_\infty$-norm, in particular, as a distortion metric, and, in turn, suggests that explicitly considering the frequency space is promising for learning robust models which better align with human perception.
\end{abstract}

\section{Introduction}

Despite the impressive performance deep neural networks have shown, researchers have discovered that they are, in some sense, `brittle'; small carefully crafted `adversarial' perturbations to their inputs can result in wildly different outputs \citep{szegedy2013intriguing}. Even worse, these perturbations have been shown to \emph{transfer}: learned models can be successfully manipulated by adversarial perturbations generated by attacking distinct models. An attacker can discover a model's vulnerabilities even without access to it.

The goal of this paper is to investigate the relationship between a perturbation's frequency properties and its effectiveness, and is motivated by recent work showing the effectiveness of low frequency perturbations in particular. \citet{guo2018lowfreq} shows that constraining the perturbation to the low frequency subspace improves the query efficiency of the decision-based gradient-free boundary attack \citep{brendel2017boundary}. \citet{zhou2018transferable} achieves improved transferability by suppressing high frequency components of the perturbation. Similarly, \citet{caadcompetition} applied a 2D Gaussian filter on the gradient w.r.t. the input image during the iterative optimization process to win the CAAD 2018 competition\footnote{Competition on Adversarial Attacks and Defenses: \url{http://hof.geekpwn.org/caad/en/index.html}}.

However, two questions still remain unanswered:
\begin{enumerate}
\item is the effectiveness of low frequency perturbations simply due to the \emph{reduced search space} or specifically due to the use of \emph{low frequency components}? and
\item under what conditions are low frequency perturbations more effective than unconstrained perturbations?
\end{enumerate}

To answer these questions, we design systematic experiments to test the effectiveness of perturbations manipulating specified frequency components, utilizing the discrete cosine transform (DCT). Testing against state-of-the-art ImageNet~\citep{deng2009imagenet} defense methods, we show that, when perturbations are constrained to the low frequency subspace, they are 1) generated faster; and are 2) more transferable. These results mirror the performance obtained when applying spatial smoothing or downsampling-upsampling operations. However, if perturbations are constrained to other frequency subspaces, they perform worse in general. This confirms that the effectiveness of low frequency perturbations is due to the application of a low-pass filter in the frequency domain of the perturbation rather than a general reduction in the dimensionality of the search space.

On the other hand, we also notice that the improved effectiveness of low frequency perturbations is only significant for defended models, but not for clean models. In fact, the state-of-the-art ImageNet defenses in test are roughly as vulnerable to low frequency perturbations as undefended models, suggesting that their purported robustness is reliant upon the assumption that adversarial perturbations are high frequency in nature. As we show, this issue is not shared by the state-of-the-art on CIFAR-10~\citep{madry}, as the dataset is too low-dimensional for there to be a diverse frequency spectrum. Finally, based on the perceptual difference between the unconstrained and low frequency attacks, we discuss the problem of using the commonly used $\ell_\infty$ norm as a perceptual metric for quantifying robustness, illustrating the promise in utilizing frequency properties to learn robust models which better align with human perception.

\section{Background}
\label{sec:related_work}
Generating adversarial examples is an optimization problem, while generating transferable adversarial examples is a generalization problem. The optimization variable is the perturbation, and the objective is to fool the model, while constraining (or minimizing) the magnitude of the perturbation. $\ell_p$ norms are typically used to quantify the strength of the perturbation; though they are well known to be poor perceptual metrics~\citep{zhang2018perceptual}. Constraint magnitudes used in practice are assumed to be small enough such that the ball is a subset of the imperceptible region. 

Adversarial perturbations can be crafted in not only the \textit{white-box} setting~\citep{carlini2017towards,ead} but in limited access settings as well~\citep{chen2017zoo,alzantot2018genattack}, when solely query access is allowed. When even that is not possible, attacks operate in the \textit{black-box} setting, and must rely on transferability. Finally, adversarial perturbations are not a continuous phenomenon, recent work has shown applications in discrete settings (e.g. natural language)~\citep{alzantot2018nlp,lei2018nlp}.

Numerous approaches have been proposed as defenses, to limited success. Many have been found to be easily circumvented~\citep{carlini2017bypass,ead_feature,athalye2018obfuscated}, while others have been unable to scale to high-dimensional complex datasets, e.g. ImageNet~\citep{smith2018uncertainty,papernot2018deepk,li2018generative,schott2018generative}. Adversarial training, training the model with adversarial examples~\citep{goodfellow2014explaining,tramer2017ensemble,madry,ding2018margin}, has demonstrated improvement, but is limited to the properties of the perturbations used, e.g. training exclusively on $\ell_\infty$ does not provide robustness to perturbations generated under other distortion metrics~\citep{ead_madry,schott2018generative}. In the NeurIPS 2017 ImageNet competition, winning defenses built upon these trained models to reduce their vulnerabilities~\citep{nipscompetition,xie2018denoising}.

\section{Methods}
\label{sec:methods}
\subsection{Attacks}

We consider $\ell_\infty$-norm constrained perturbations, where the perturbation $\delta$ satisfies $\|\delta\|_\infty \leq \epsilon$ with $\epsilon$ being the maximum perturbation magnitude, as the NeurIPS 2017 competition bounded $\delta$ with $\epsilon=16$. The Fast Gradient Sign Method (FGSM) \citep{goodfellow2014explaining} provides a simple, one-step gradient-based perturbation of $\ell_\infty$ $\epsilon$ size as follows:  
\begin{align}
\delta_\text{FGSM} = s \cdot \epsilon\cdot\text{sign}(\nabla_{x}J(x,y))
\end{align}
where $x$ is the input image, $J$ is the classification loss function, $
\text{sign}(\cdot)$ is the element-wise sign function\footnote{$\text{sign} = 1$ if $x>0$, $\text{sign} = -1$ if $x<0$, $\text{sign} = 0$, if $x=0$.}. When $y$ is the true label of $x$ and $s=+1$, $\delta$ is the \emph{non-targeted} attack for misclassification; when $y$ is a \emph{target} label other than the true label of $x$ and $s=-1$, $\delta$ is the \emph{targeted} attack for manipulating the network to wrongly predict $y$.

FGSM suffers from an ``underfitting'' problem when applied to non-linear loss function, as its formulation is dependent on a linearization of $J$ about $x$. The Basic Iterative Method (BIM)~\citep{kurakin2016adversarial,madry}, otherwise known as PGD (without random starts), runs FGSM for multiple iterations to rectify this problem. The top-placing attack in the previously mentioned NeurIPS 2017 competition, the Momentum Iterative Method (MIM)~\citep{dong2017momentum}, replaces the gradient $\nabla_{x}J(x,y)$ with a ``momentum'' term to prevent the ``overfitting'' problem, caused by poor local optima, in order to improve transferability. Thus, we use this method for our NeurIPS 2017 defense evaluation.

\subsection{Frequency Constraints}

\begin{figure}[t]
    \centering
    \includegraphics[width=\linewidth]{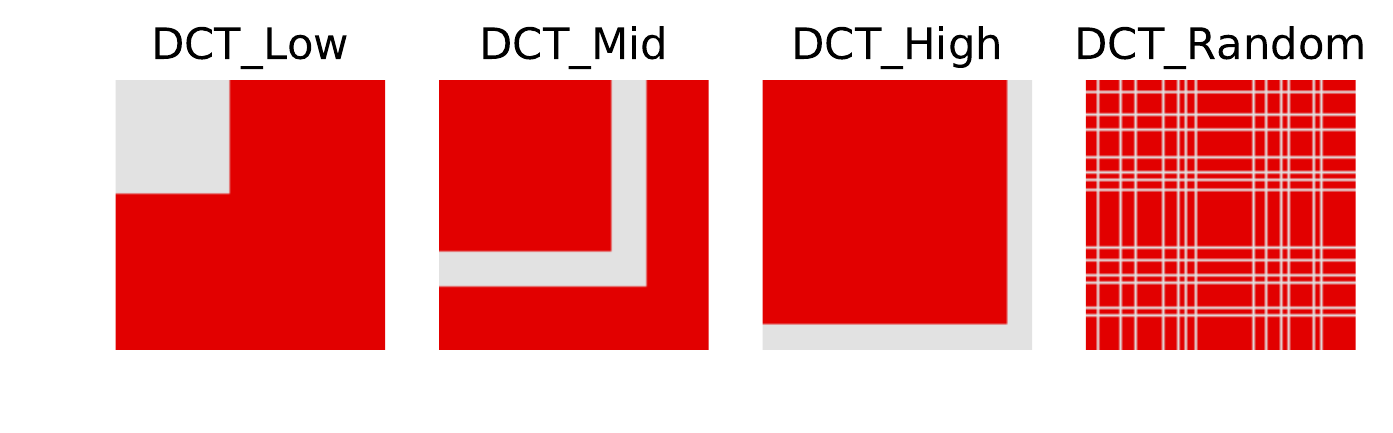}
    \caption{Masks used to constrain the frequency space where $n=128$ and $d=299$ (ImageNet). Red denotes frequency components of the perturbation which will be masked when generating the adversarial example, both during and after the optimization process.}
    \label{dct}
\end{figure}
\begin{table}[!t]
    \centering
    \resizebox{\linewidth}{!}{\begin{tabular}{c|c}
    \toprule
    Cln\_1 & [\texttt{InceptionV3}] \\
    \multirow{2}{*}{Cln\_3} & \multirow{2}{*}{\shortstack{[\texttt{InceptionV3, InceptionV4,} \\\texttt{ResNetV2\_101}]}} \\
    & \\
    Adv\_1 & [\texttt{AdvInceptionV3}] \\
    \multirow{2}{*}{Adv\_3} & \multirow{2}{*}{\shortstack{[\texttt{AdvInceptionV3, Ens3AdvInceptionV3,} \\\texttt{Ens4AdvInceptionV4}]}} \\
    & \\
    \bottomrule
    \end{tabular}}
    \caption{Models used for generating black-box transfer attacks.}
    \label{blackbox_config}
\end{table}

Our goal is to examine whether the effectiveness of low frequency perturbations is due to a reduced search space in general or due to the specific use of a low-pass filter in the frequency domain of the perturbation. To achieve this, we use the \emph{discrete cosine transform} (DCT) \citep{rao2014discrete} to constrain the perturbation to only modify certain frequency components of the input. 

The DCT decomposes a signal into cosine wave components with different frequencies and amplitudes. Given a 2D image (or perturbation) $x\in\mathbb{R}^{d\times d}$, the DCT Transform of $x$ is $v=\text{DCT}(x)$,
where the entry $v_{i,j}$ is the magnitude of its corresponding basis functions.

The numerical values of $i$ and $j$ represent the frequencies, i.e. smaller values represent lower frequencies and vice versa. The DCT is invertible, with an inverse transform $x=\text{IDCT}(v)$\footnote{DCT / IDCT is applied to each color channel independently.}.

We remove certain frequency components of the perturbation $\delta$ by applying a mask to its DCT transform $\text{DCT}(\delta)$. We then reconstruct the perturbation by applying IDCT on the masked DCT transform. Specifically, the mask, $m \in \{0, 1\}^{d \times d}$, is a 2D matrix image whose pixel values are 0's and 1's, and the ``masking'' is done by element-wise product.

We can then reconstruct the ``transformed'' perturbation by applying the IDCT to the masked $\text{DCT}(\delta)$. The entire transformation can then be represented as:
\begin{align}
\text{FreqMask}(\delta)=\text{IDCT}(\text{Mask}(\text{DCT}(\delta)))~.
\end{align}

Accordingly in our attack, we use the following gradient
$$
\nabla_\delta J(x+\text{FreqMask}(\delta),y)~.
$$

\newcommand{\dcthigh}{\texttt{DCT\_High}\xspace}
\newcommand{\dctlow}{\texttt{DCT\_Low}\xspace}
\newcommand{\dctmid}{\texttt{DCT\_Mid}\xspace}
\newcommand{\dctrand}{\texttt{DCT\_Rand}\xspace}

We use 4 different types of FreqMask to constrain the perturbations, as shown in \Cref{dct}. \dcthigh only preserves high frequency components; \dctlow only preserves low frequency components; \dctmid only preserves mid frequency components; and \dctrand preserves randomly sampled components. For reduced dimensionality $n$, we preserve $n \times n$ components. Recall that $v=\text{DCT}(x)$, \dctlow preserves components $v_{i,j}$ if $1\leq i,j\leq n$; \dcthigh masks components if $1\leq i,j\leq \sqrt{d^2 - n^2}$; \dctmid and \dctrand also preserve $n \times n$ components, the detailed generation processes can be found in the appendix. \Cref{dct} visualizes the masks when $d=299$ (e.g. ImageNet) and $n=128$. Note that when $n=128$, we only preserve $128^2 / 299^2 \approx 18.3\%$ of the frequency components, a small fraction of the original unconstrained perturbation.

\section{Results and Analyses}
\label{sec:results}

To evaluate the effectiveness of perturbations under different frequency constraints, we test against models and defenses from the NeurIPS 2017 Adversarial Attacks and Defences Competition \citep{nipscompetition}.

\begin{figure*}[!h]
    \centering
    \includegraphics[width=\textwidth]{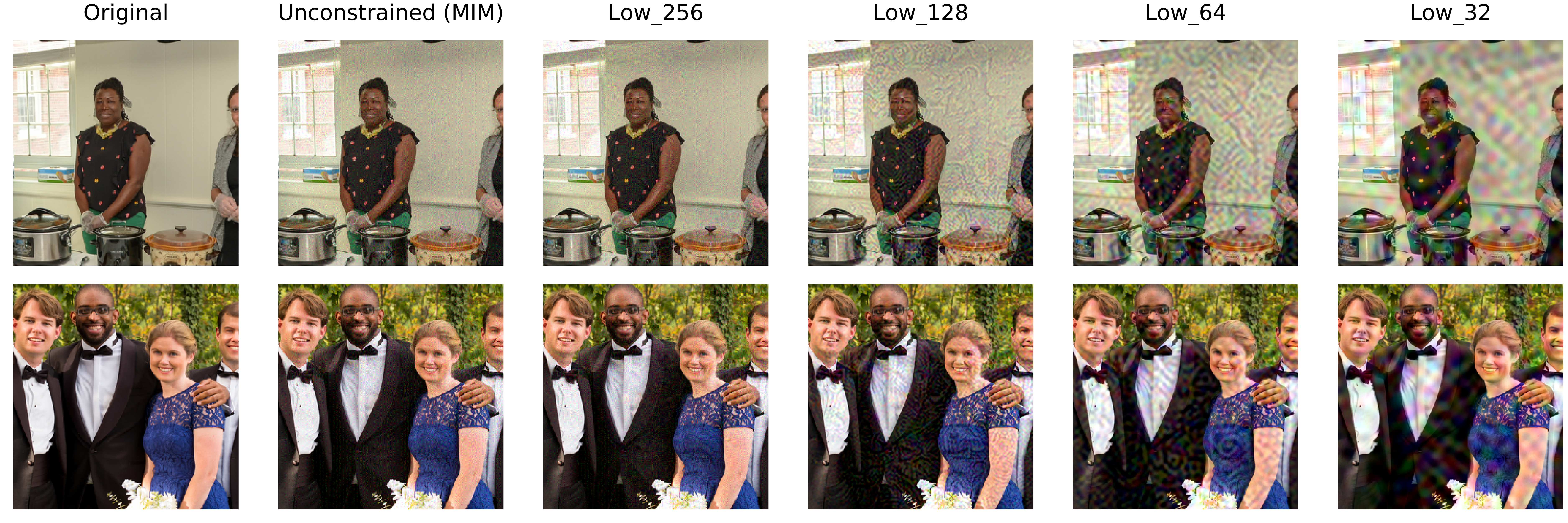}
    \caption{Adversarial examples generated with $\ell_\infty$ $\epsilon=32/255$ distortion}
    \label{32_perception}
\end{figure*}

\paragraph{Threat Models:} We evaluate attacks in both the non-targeted and targeted case, and measure the attack success rate (ASR) on 1000 test examples from the NeurIPS 2017 development toolkit\footnote{\url{https://www.kaggle.com/c/6864/download/dev_toolkit.zip}}. We test on $\epsilon=16/255$ (competition distortion bound) and $\text{iterations}=[1,10]$ for the non-targeted case; $\epsilon=32/255$ and $\text{iterations}=10$ for the targeted case. The magnitude for the targeted case is larger since targeted attacks, particularly on ImageNet (1000 classes), are significantly harder. As can be seen in \Cref{perception} and \ref{32_perception}, unconstrained adversarial perturbations generated under these distortion bounds are still imperceptible.

\paragraph{Attacks:} As described in \Cref{sec:methods}, we experiment with the original unconstrained MIM and frequency constrained MIM with masks shown in Figure~\ref{dct}.
For each mask type, we test $n=[256,128,64,32]$ with $d = 299$. For \dctrand, we average results over $3$ random seeds.

To describe the attack settings, we specify model placeholders $A$ and $B$. We call an attack \textit{white-box}, when we attack model $A$ with the perturbation generated from $A$ itself. We call an attack \textit{grey-box}, when the perturbation is generated from $A$, but used to attack a ``defended'' $A$, where a defense module is prepended to $A$. We call an attack \textit{black-box} (transfer), when the perturbation generated from $A$ is used to attack distinct $B$, where $B$ can be defended or not. Note that this is distinct from the black-box setting discussed in \citep{guo2018lowfreq}, in which query access is allowed.

\paragraph{Target Models and Defenses for Evaluation:}

We evaluate each of the attack settings against the top defense solutions in the NeurIPS 2017 competition \citep{nipscompetition}. Each of the top-4 NeurIPS 2017 defenses prepend a tuned (or trained) preprocessor to an ensemble of classifiers, which for all of them included the strongest available adversarially trained model: \texttt{EnsAdvInceptionResNetV2}\footnote{\url{https://github.com/tensorflow/models/tree/master/research/adv_imagenet_models}} \citep{tramer2017ensemble}. Thus, we use \texttt{EnsAdvInceptionResNetV2} to benchmark the robustness\footnote{\texttt{EnsAdvInceptionResNetV2} is to be attacked.} of adversarially trained models. 

We then prepend the preprocessors from the top-4 NeurIPS 2017 defenses to \texttt{EnsAdvInceptionResNetV2}, and denote the defended models as D1, D2, D3, and D4, respectively. 
Regarding the preprocessors, 
D1 uses a trained denoiser where the loss function is defined as the difference between the target model’s outputs activated by the clean image and denoised image~\citep{liao2017guided}; D2 uses random resizing and random padding~\citep{xie2017random}; D3 uses a number of image transformations: shear, shift, zoom, and rotation~\citep{nips3}; and D4 simply uses median smoothing~\citep{nipscompetition}.

For our representative cleanly trained model, we evaluate against the state-of-the-art \texttt{NasNetLarge\_331}\footnote{https://github.com/tensorflow/models/tree/master/research/slim} \citep{zoph2017nas}. We denote \texttt{EnsAdvInceptionResNetV2} as EnvAdv and \texttt{NasNetLarge\_331} as NasNet for brevity.

\paragraph{Source Models for Perturbation Generation:}

For white-box attacks, we evaluate perturbations generated from NasNet and EnsAdv to attack themselves respectively. For grey-box attacks, we use perturbations generated from EnsAdv to attack D1, D2, D3, and D4 respectively. For black-box attacks, since the models for generating the perturbations need to be distinct from the ones being attacked, we use 4 different sources (ensembles) which vary in ensemble size and whether the models are adversarially trained or cleanly trained, as shown in \Cref{blackbox_config}. In summary, for black-box attacks, perturbations generated from Adv\_1, Adv\_3, Cln\_1, and Cln\_3 are used to attack NasNet, EnsAdv, D1, D2, D3, and D4.
\begin{figure*}[!h]
\begin{subfigure}{0.49\linewidth}
    \centering
    \includegraphics[width=\linewidth]{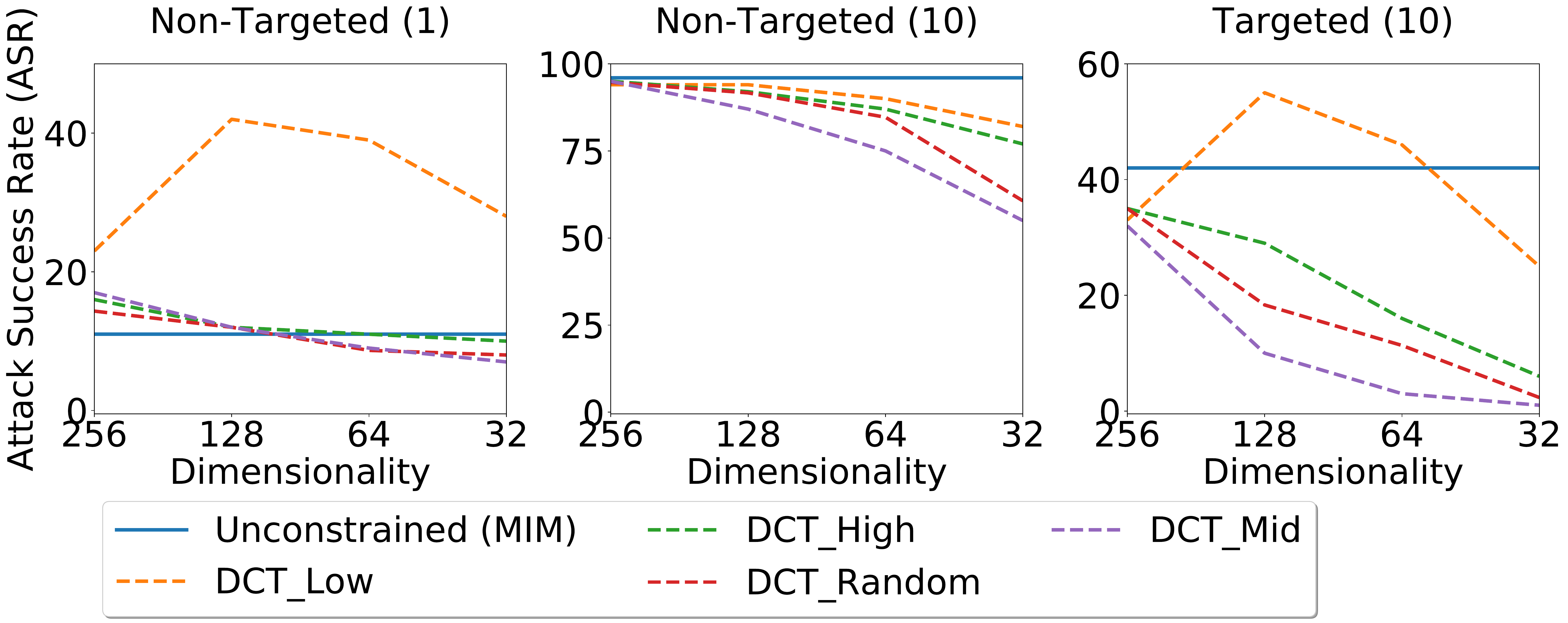}
    \caption{\textbf{White-box} attack on adversarially trained model, EnsAdv.}
    \label{whitebox}
\end{subfigure}
~~~~
\begin{subfigure}{0.49\linewidth}
    \centering
    \includegraphics[width=\linewidth]{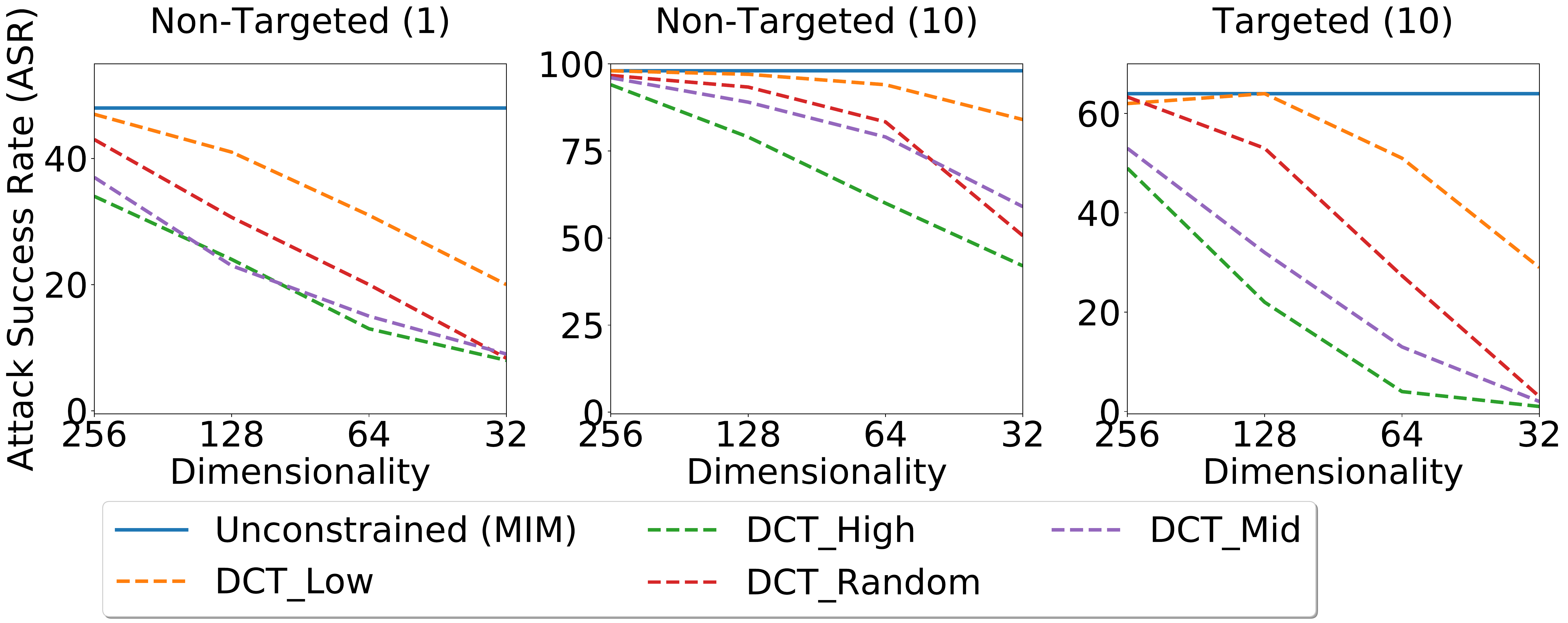}
    \caption{\textbf{White-box} attack on standard cleanly trained model, NasNet.}
    \label{clnwhitebox}
\end{subfigure}
\hfill
\begin{subfigure}{0.49\linewidth}
    \centering
    \includegraphics[width=\linewidth]{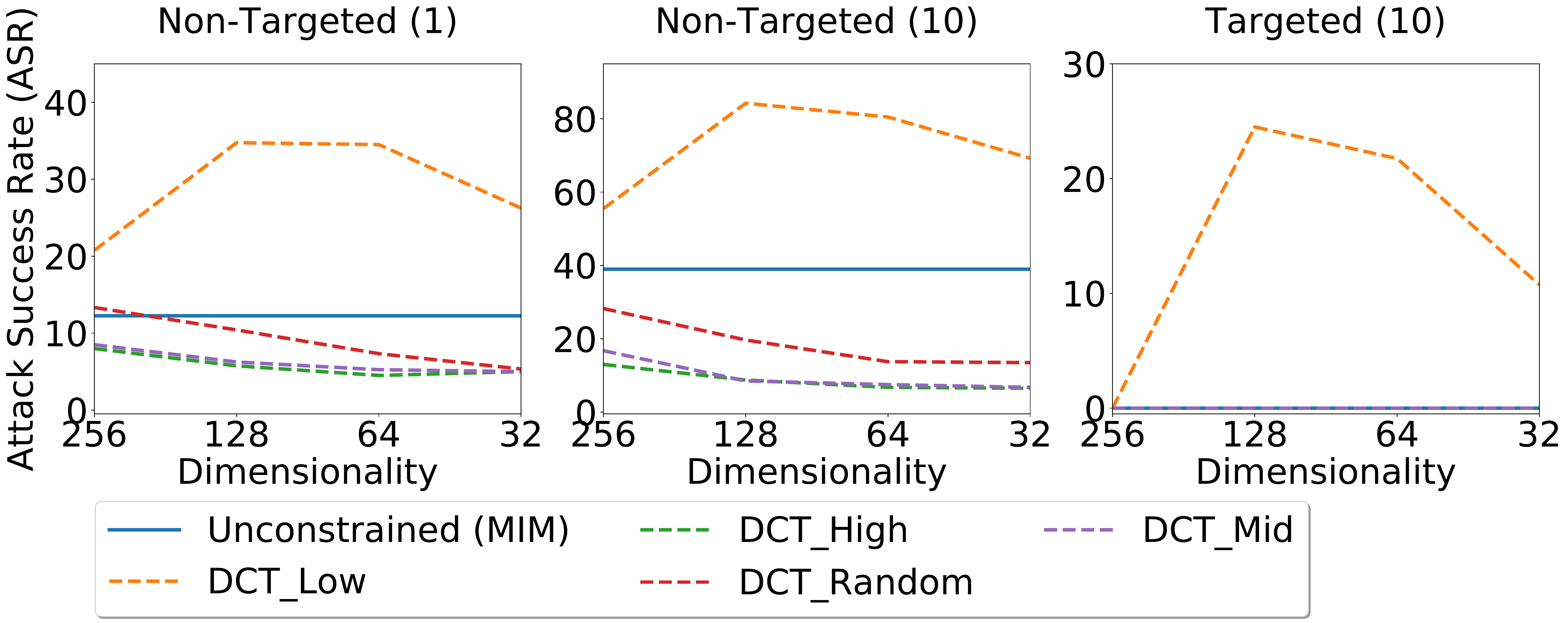}
    \caption{\textbf{Grey-box} attack on top-4 NeurIPS 2017 defenses prepended to adversarially trained model.}%\\~\\~}
    \label{greybox}
\end{subfigure}
~~~~
\begin{subfigure}{0.49\linewidth}
    \centering
    \includegraphics[width=\linewidth]{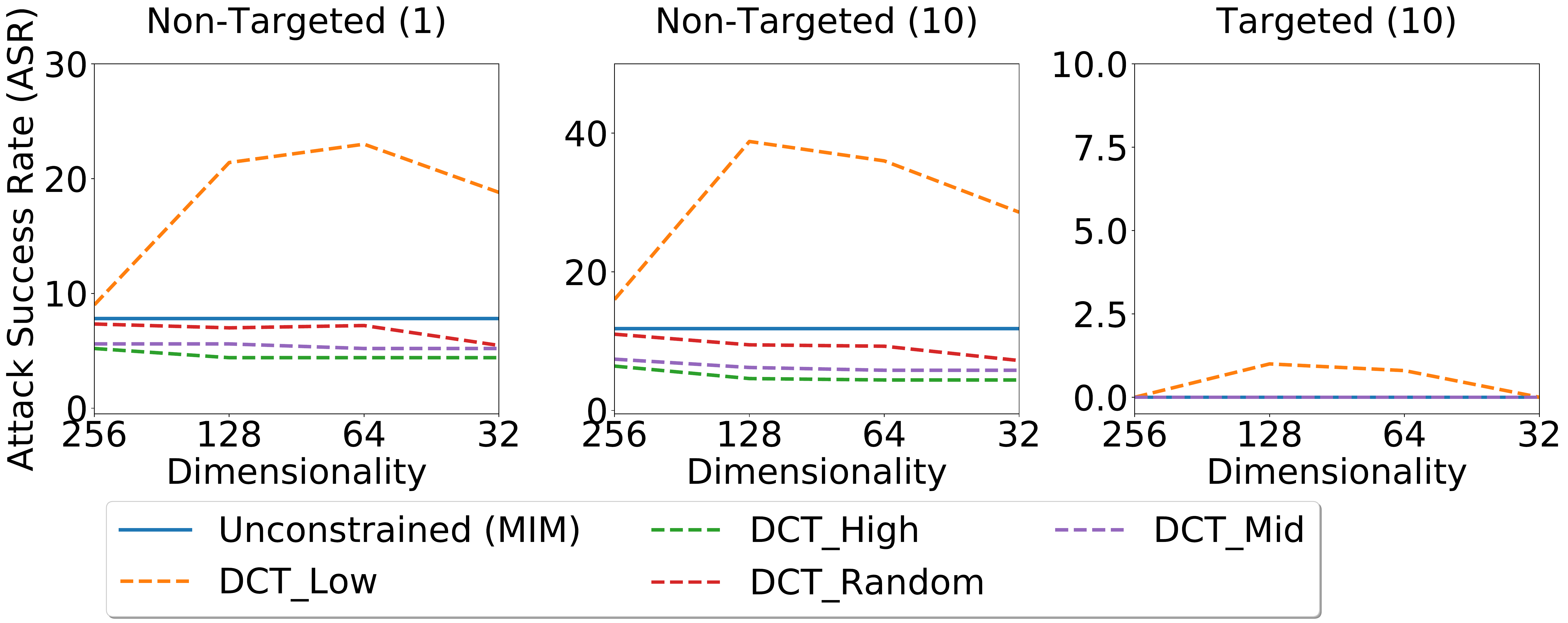}
    \caption{\textbf{Black-box} attack on sources (Table~\ref{blackbox_config}) transferred to defenses (EnsAdv + D1$\sim$4)}
    \label{blackbox}
\end{subfigure}
\caption{Number of iterations in parentheses. Non-targeted with $\epsilon=16/255$, targeted with $\epsilon=32/255$.}
\end{figure*}

\subsection{Overview of the Results} 

As described, we test the unconstrained and constrained perturbations in the white-box, grey-box, and black-box scenarios. Representative results are shown in \Cref{whitebox}, \ref{clnwhitebox}, \ref{greybox}, and \ref{blackbox}. In each of these plots, the vertical axis is attack success rate (ASR), while the horizontal indicates the number of frequency components kept (Dimensionality). Unconstrained MIM is shown as a horizontal line across the dimensionality axis for ease of comparison. In each figure, the plots are, from left to right, non-targeted attack with $\text{iterations}=1$, non-targeted with $\text{iterations}=10$, and targeted with $\text{iterations}=10$. From these figures, we can see that \dctlow always outperforms the other frequency constraints, including \dcthigh, \dctmid and \dctrand. 

In the appendix, we show results where the perturbation is constrained using a spatial smoothing filter or a downsampling-upsampling filter (perturbation resized with bilinear interpolation). The performance mirrors that of \Cref{whitebox}, \ref{clnwhitebox}, \ref{greybox}, and \ref{blackbox}, further confirming that the effectiveness of low frequency perturbations is not due to a general restriction of search space, but due to the low frequency regime itself. Thus, in our remaining experiments, we focus on low frequency constrained perturbations induced with \dctlow.

\begin{figure*}[!h]
\begin{minipage}[c]{0.49\linewidth}
\begin{subfigure}{0.49\linewidth}
\includegraphics[width=\linewidth]{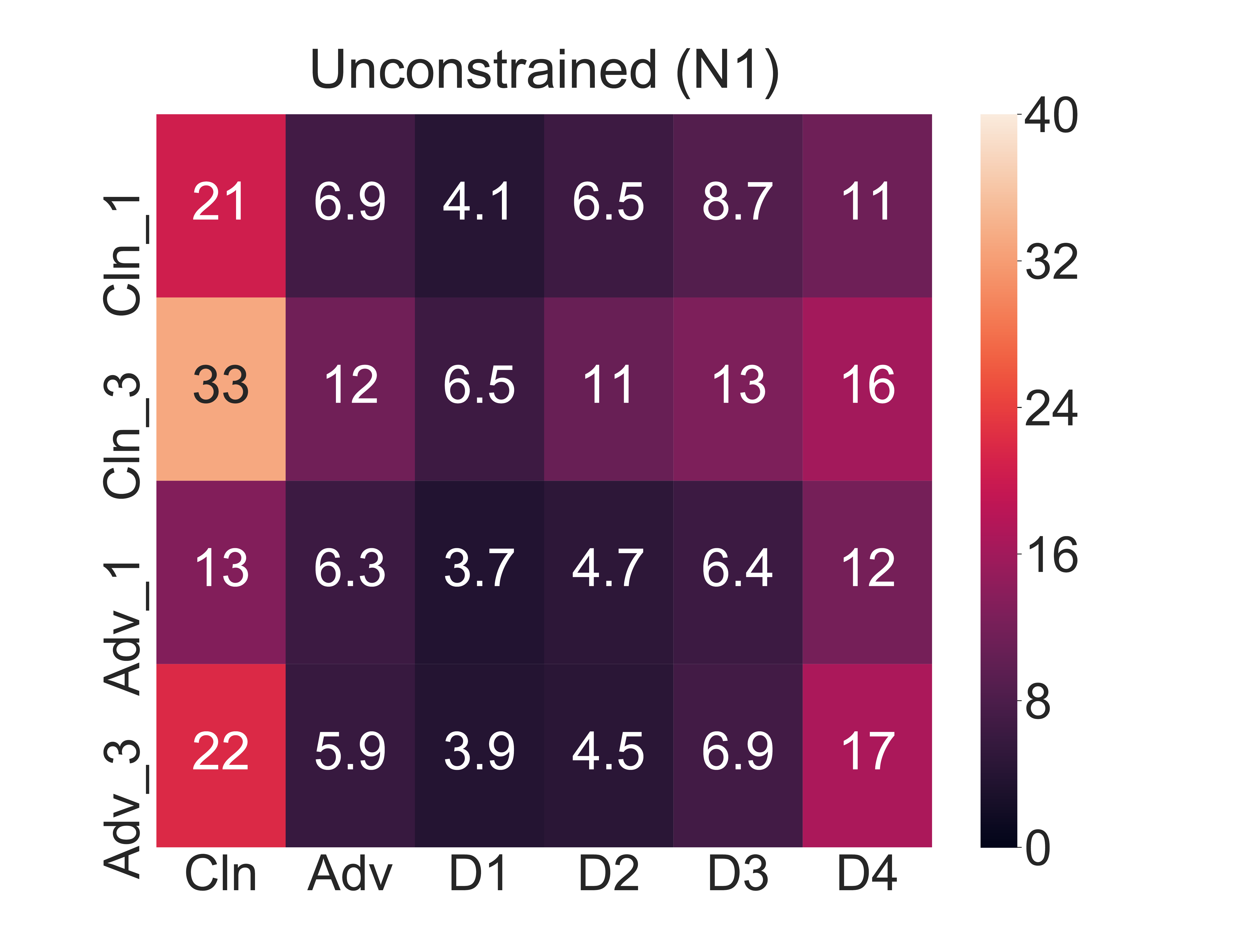}
\caption{}
\label{m11}
\end{subfigure}
\begin{subfigure}{0.49\linewidth}
\includegraphics[width=\linewidth]{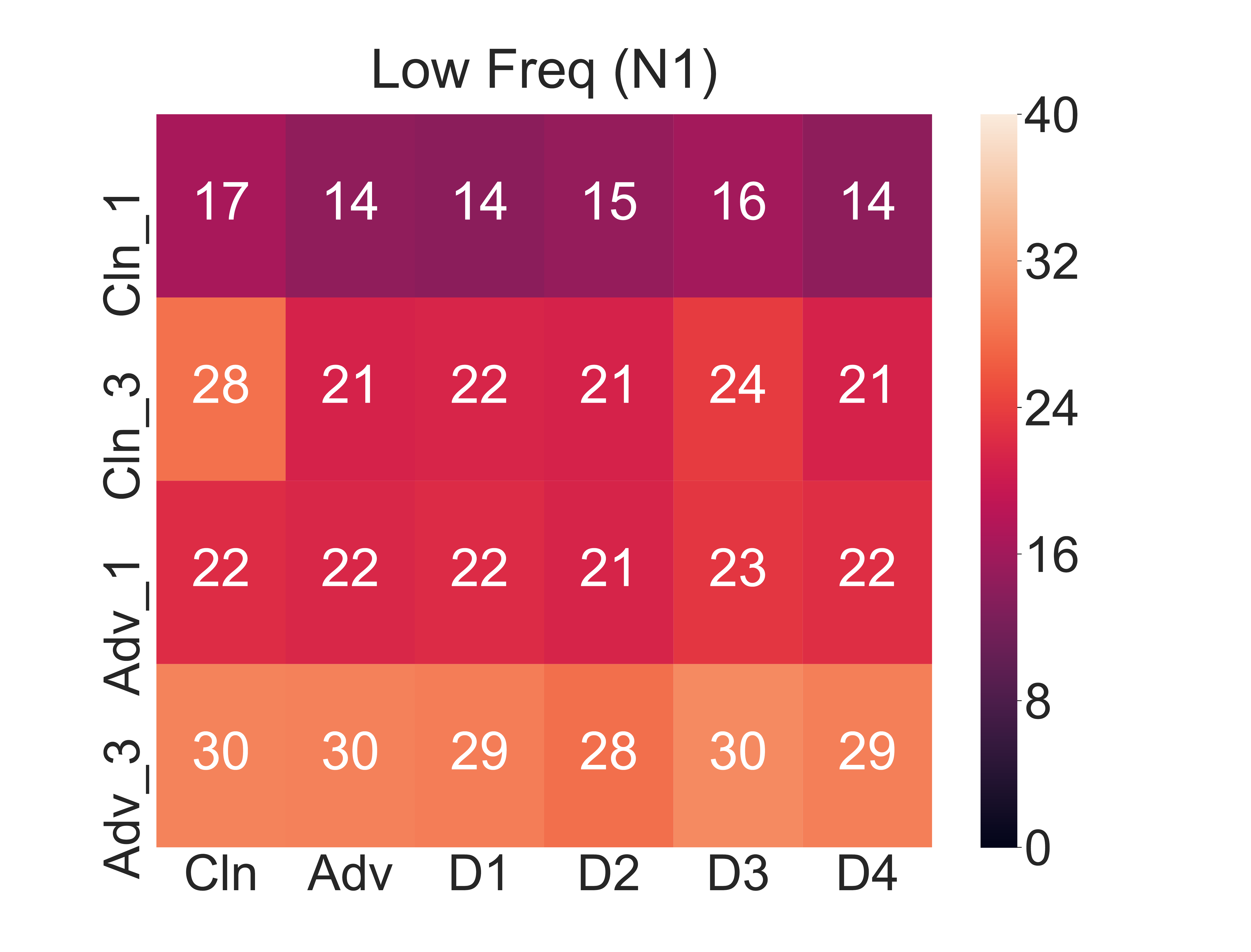}
\caption{}
\label{m12}
\end{subfigure}
\begin{subfigure}{0.49\linewidth}
\includegraphics[width=\linewidth]{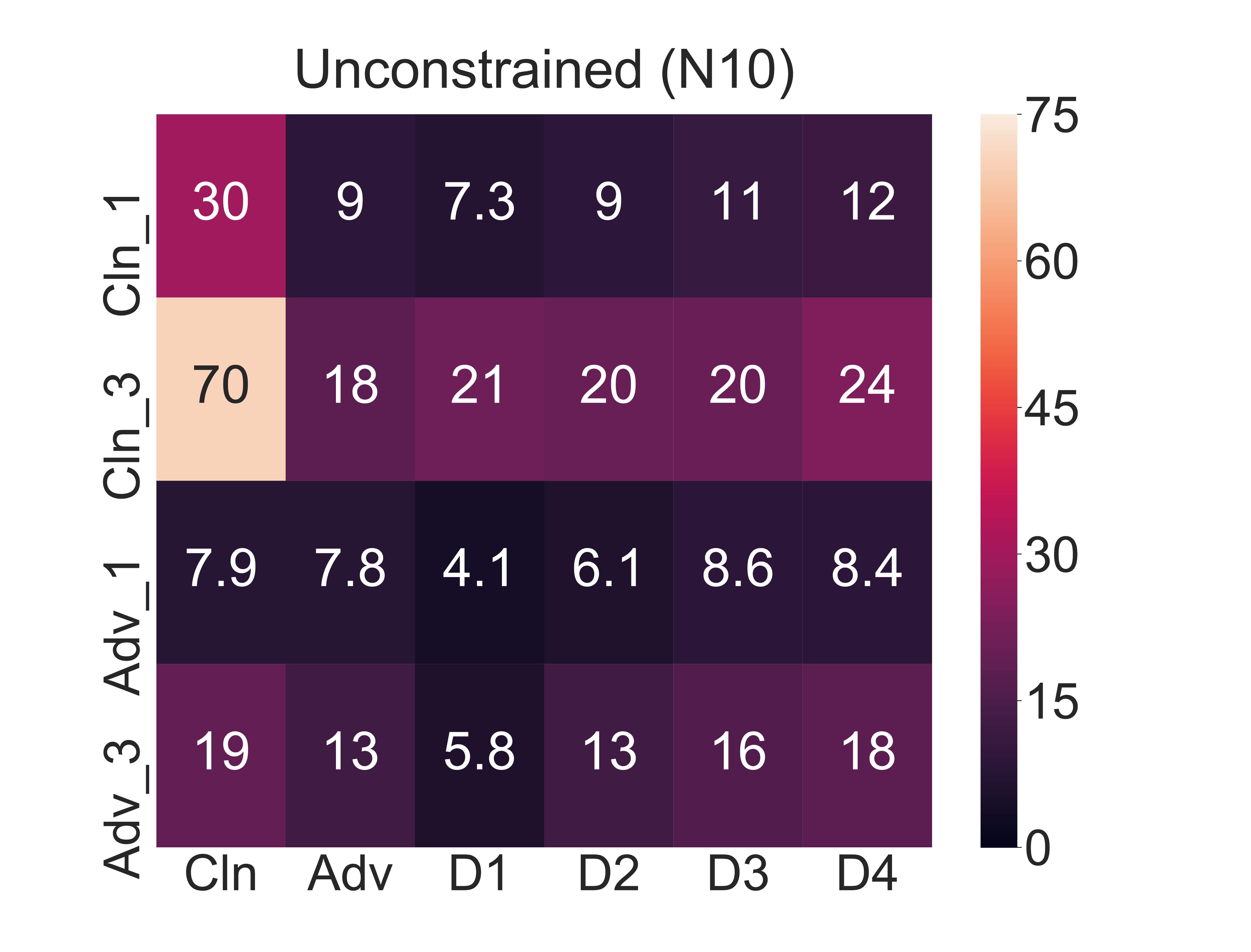}
\caption{}
\label{m13}
\end{subfigure}
\begin{subfigure}{0.49\linewidth}
\includegraphics[width=\linewidth]{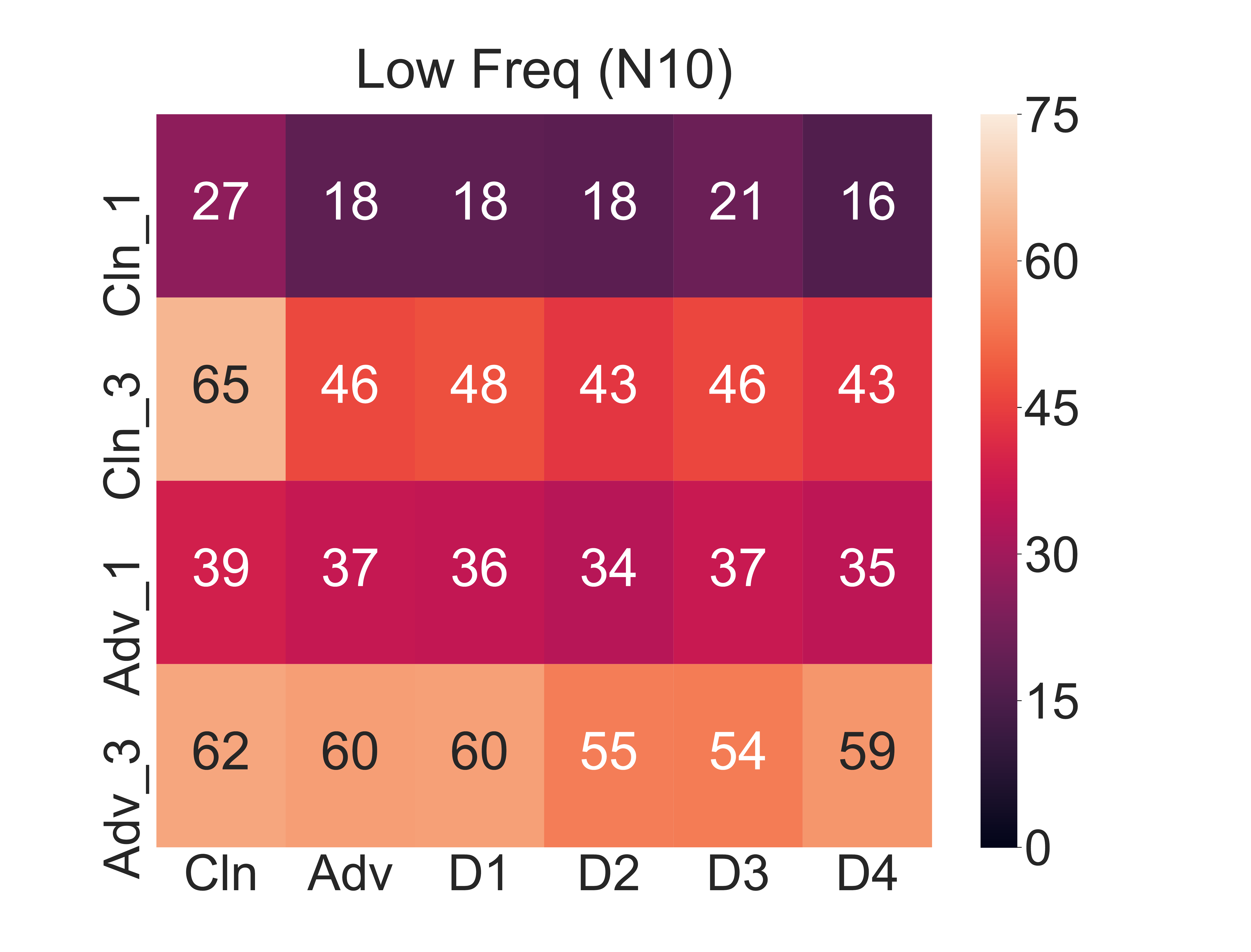}
\caption{}
\label{m14}
\end{subfigure}
\caption{
Transferability matrices with attack success rates (ASRs), comparing unconstrained MIM with low frequency constrained \texttt{DCT\_Low} ($n=128$) in the non-targeted case. First row is with $\text{iterations}=1$, second is with $\text{iterations}=10$. The column Cln is NasNet, Adv is EnsAdv.
\\~}
\label{fig:matrix_asr}% Overall figure caption
\end{minipage}
\hfill
\begin{minipage}[c]{0.49\linewidth}
\begin{subfigure}{0.49\linewidth}
\includegraphics[width=\linewidth]{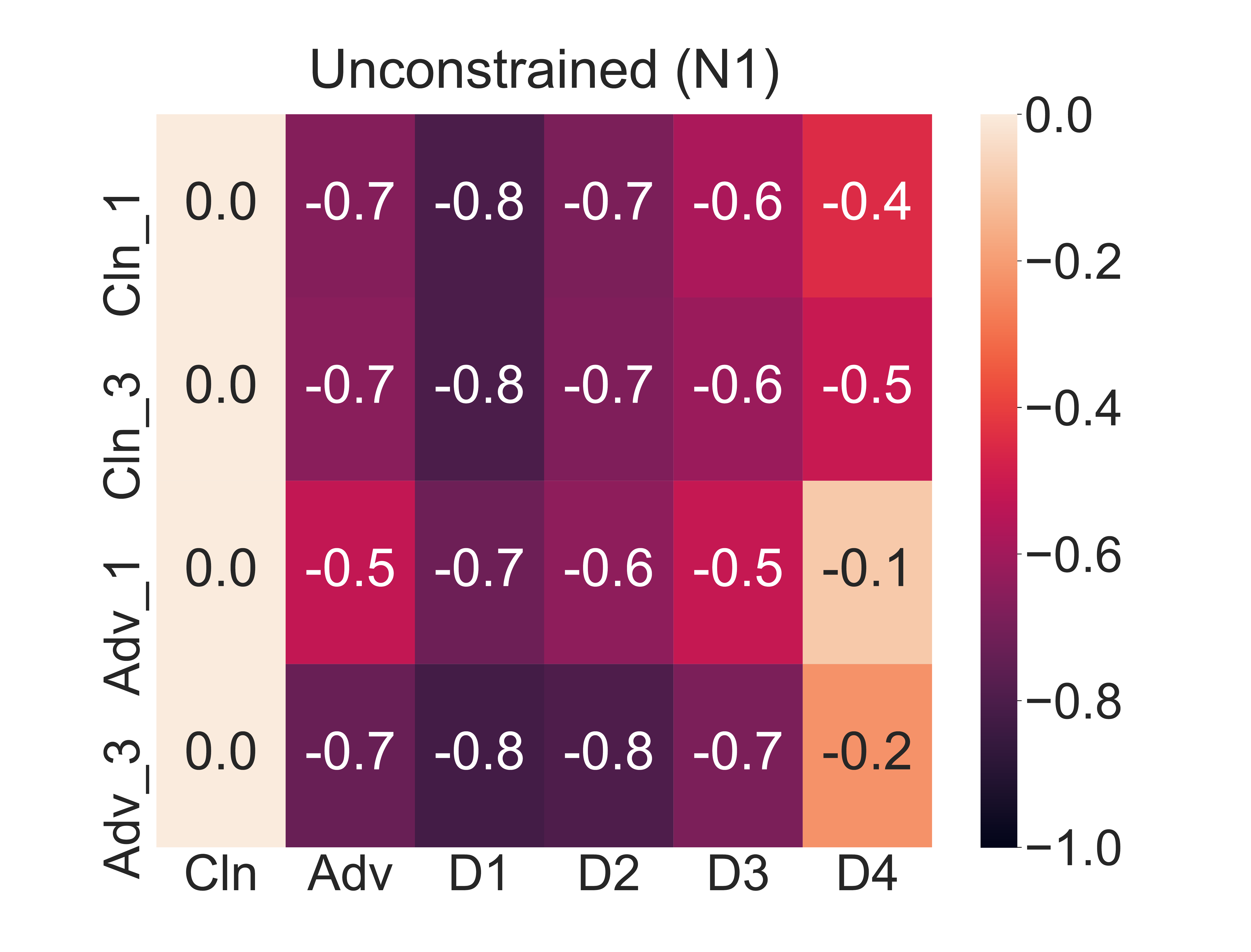}
\caption{}
\label{m21}
\end{subfigure}
\begin{subfigure}{0.49\linewidth}
\includegraphics[width=\linewidth]{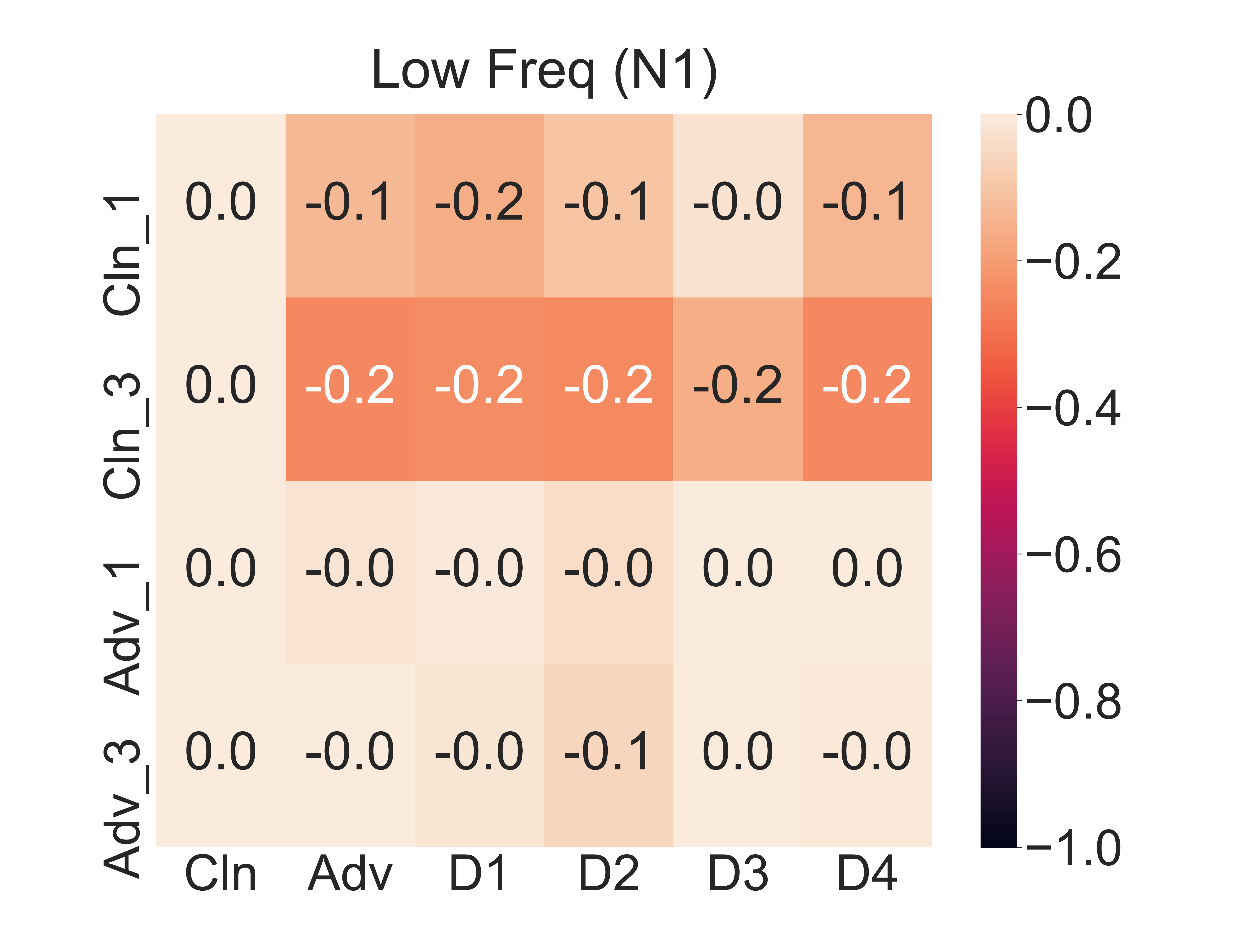}
\caption{}
\label{m22}
\end{subfigure}
\begin{subfigure}{0.49\linewidth}
\includegraphics[width=\linewidth]{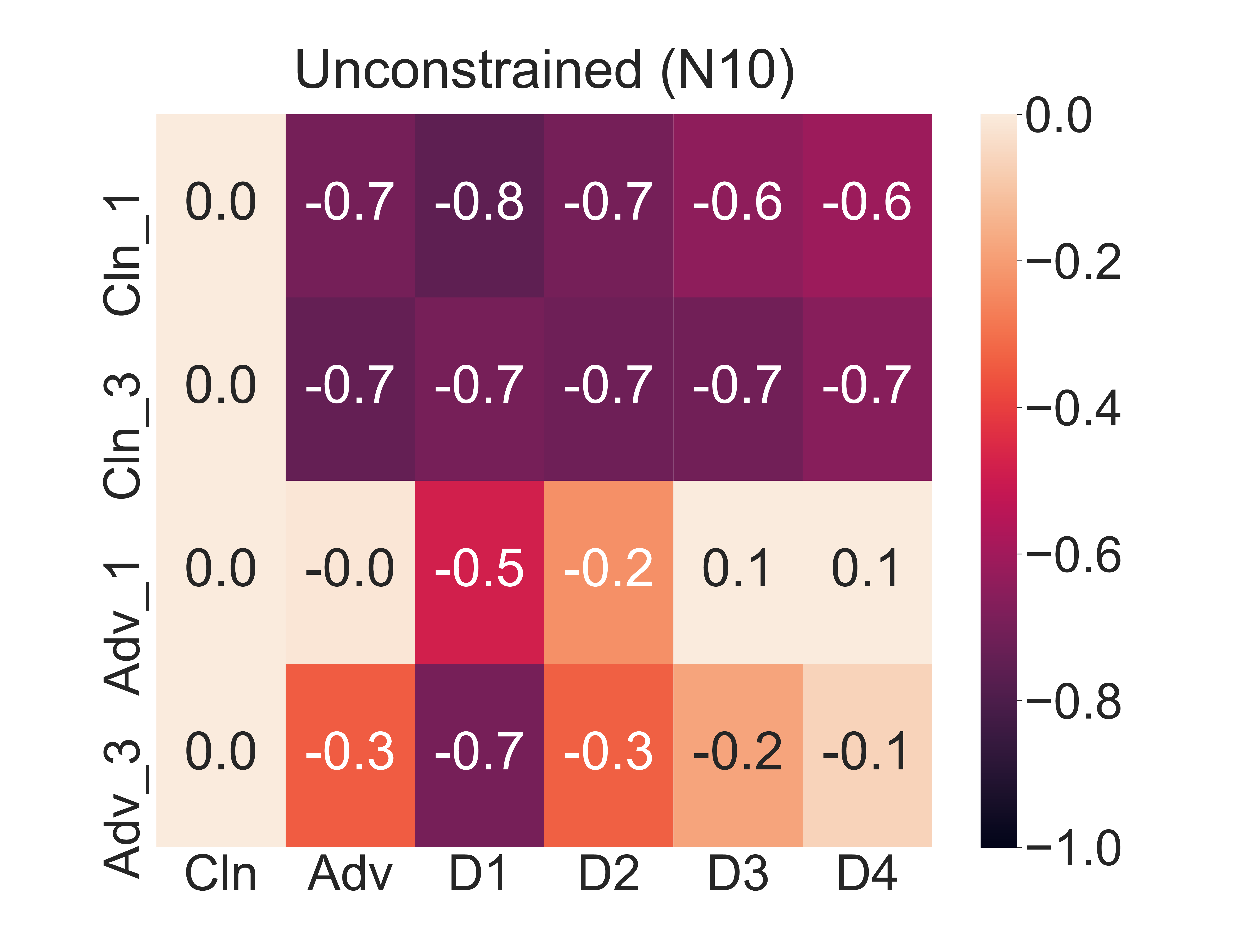}
\caption{}
\label{m23}
\end{subfigure}
\begin{subfigure}{0.49\linewidth}
\includegraphics[width=\linewidth]{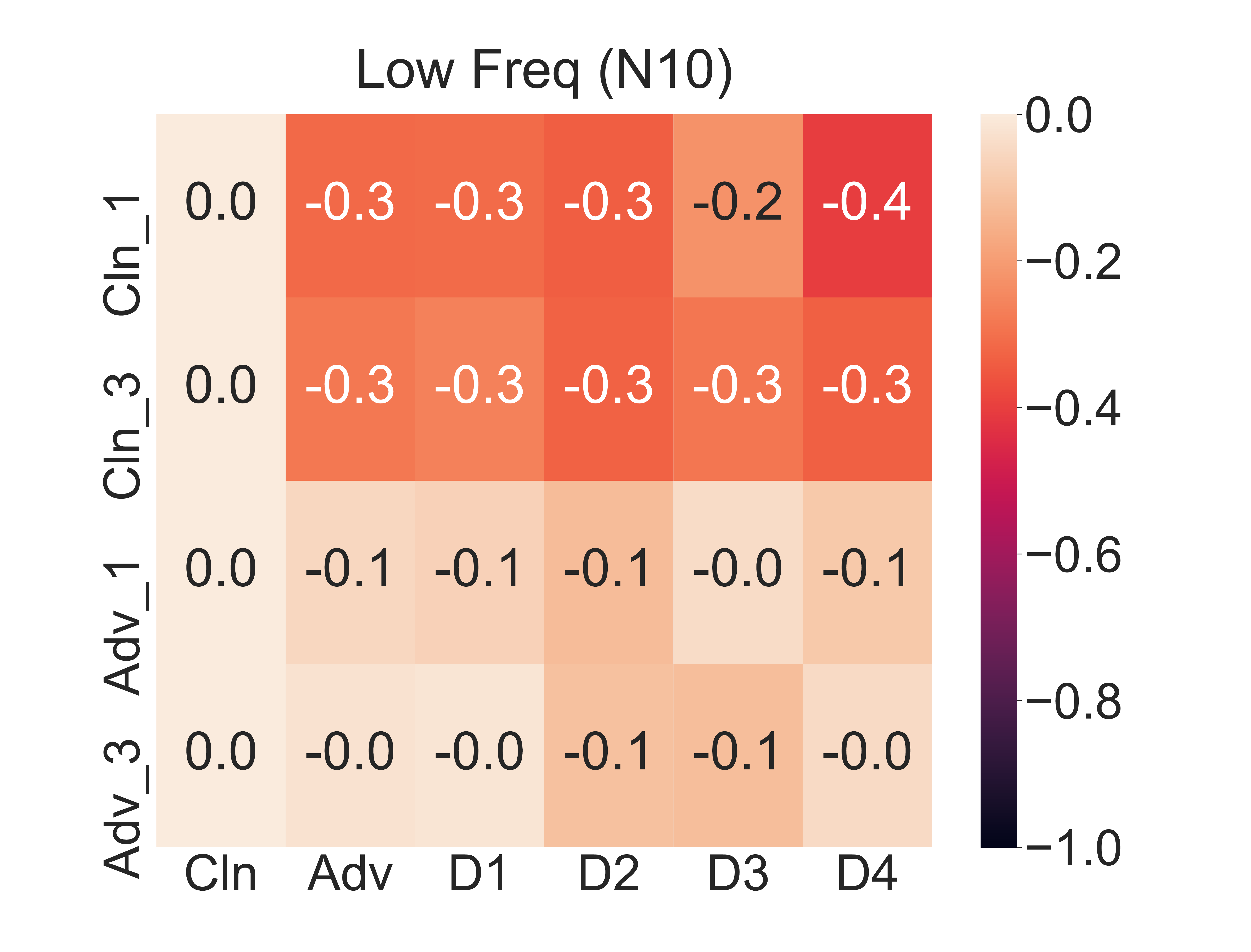}
\caption{}
\label{m24}
\end{subfigure}
\caption{Transferability matrices with attack relative difference in ASR with the Cln model (first column). Rows and columns in each subfigure is indexed in the same way as \Cref{fig:matrix_asr}.\\~\\~\\~}
\label{fig:matrix_percent}% Overall figure caption
\end{minipage}
\end{figure*}

We compare ASR and relative changes across all black-box transfer pairs between standard unconstrained MIM and MIM constrained with \dctlow $n=128$, on non-targeted attacks with both $\text{iterations}=1$ and $\text{iterations}=10$. This comparison is visualized in \Cref{fig:matrix_asr} and \ref{fig:matrix_percent}. We also show that these results do not transfer to the significantly lower-dimensional CIFAR-10 dataset ($d=32$, minimum $n$ used in ImageNet experiments), as the rich frequency spectrum of natural images is no longer present.
\subsection{Observations and Analyses}

\paragraph{\dctlow generates effective perturbations faster on adversarially trained models, but not on cleanly trained models.}
\Cref{whitebox} and \ref{clnwhitebox} show the white-box ASRs on EnsAdv and NasNet respectively. For EnsAdv, we can see that \dctlow improves ASR in the non-targeted case with $\text{iterations}=1$ and in the targeted case with $\text{iterations}=10$, but not in the non-targeted case with $\text{iterations}=10$. However, in this case, \dctlow still outperforms other frequency constraints and does not significantly deviate from unconstrained MIM's performance. When the number of iterations is large enough that unconstrained MIM can succeed consistently, constraining the space only limits the attack, but otherwise, the low frequency prior is effective. Therefore, low frequency perturbations are more ``iteration efficient'', as they can be found more easily with a less exhaustive search, which is practically helpful computationally.

However, for white-box attacks on NasNet in \Cref{clnwhitebox}, we see that although \dctlow still outperforms the other frequency constraints, it does perform worse than unconstrained MIM. Comparing \Cref{whitebox} and \ref{clnwhitebox}, it is clear that \dctlow performs similarly against the adversarially trained model as with the cleanly trained model, the difference here is due to unconstrained MIM performing significantly better against the cleanly trained model than against the adversarially trained model. This implies that the low frequency prior is useful against defended models, in particular, since it exploits the space where adversarial training, which is necessarily imperfect, fails to reduce vulnerabilities.

\paragraph{\dctlow bypasses defenses prepended to the adversarially trained model.} As previously mentioned, in the grey-box case, we generate the perturbations from the undefended EnsAdv and use them to attack D1, D2, D3 and D4 (which include preprocessors prepended to EnsAdv). \Cref{greybox} shows the ASR results averaged over D1$\sim$4. \dctlow outperforms unconstrained MIM by large margins in all cases. Comparing \Cref{whitebox} with \Cref{greybox}, the larger difference between unconstrained MIM and \dctlow in the grey-box case reflects the fact that the top NeurIPS 2017 defenses are not nearly as effective against low frequency perturbations as they are against standard unconstrained attacks. In fact, \dctlow yields the same ASR on D1, the winning defense submission in the NeurIPS 2017 competition, as on the adversarially trained model without the preprocessor prepended; the preprocessors are not effective (at all) at protecting the model from low frequency perturbations, even in the targeted case, where success is only yielded if the model is fooled to predict, out of all 1000 class labels, the specified target label. Results against the individual defenses are presented in the appendix.

\paragraph{\dctlow helps black-box transfer to defended models.}
For assessing black-box transferability, we use Cln\_1, Cln\_3, Adv\_1, Adv\_3 in \Cref{blackbox_config} as the source models for generating perturbations, and attack EnsAdv and D1$\sim$4, resulting in 20 source-target pairs in total. The average ASR results over these pairs are reported in \Cref{blackbox}. In the non-targeted case, we again see that \dctlow significantly outperforms unconstrained MIM. However, in the targeted case, constraining to the low frequency subspace does not enable MIM to succeed in transferring to distinct black-box defended models due to the difficult nature of targeted transfer.

Next, we look at individual source-target pairs. For each pair, we compare \dctlow ($n=128$) with unconstrained MIM in the non-targeted case with $\text{iterations}=1$ and $\text{iterations}=10$. Results for all frequency configurations with varied dimensionality are provided in the appendix. \Cref{fig:matrix_asr} shows the transferability matrices for all source-target pairs, where for each subplot, the row indices denote source models, and the column indices denote target models. The value (and associated color) in each gridcell represent the ASR for the specified source-target pair. For \Cref{fig:matrix_percent}, the gridcell values represent the relative difference in ASR between the target model and the cleanly trained model (Cln)\footnote{The relative difference for the target model = (ASR on the target model - ASR on Cln) / ASR on Cln.}, using the source model of the corresponding row. 

Comparing (a) to (b) and (c) to (d) in \Cref{fig:matrix_asr}, it is clear that low frequency perturbations are much more effective than unconstrained MIM against defended models. Specifically, we can see that \dctlow is significantly more effective than unconstrained MIM against EnsAdv, and D1$\sim$4 provide almost no additional robustness to EnsAdv when low frequency perturbations are applied.

\paragraph{\dctlow is not effective when transferring between undefended cleanly trained models.}
However, we do observe that \dctlow does not improve black-box transfer between undefended cleanly trained models, which can be seen by comparing indices (Cln\_1,Cln) and (Cln\_3,Cln) between \Cref{fig:matrix_asr} (a) and (b), as well as (c) and (d). As discussed when comparing white-box performance against cleanly trained and adversarially trained models, low frequency constraints are not generally more effective, but instead exploit the vulnerabilities in currently proposed defenses.
\begin{figure*}[!h]
    \centering
    \includegraphics[width=\textwidth]{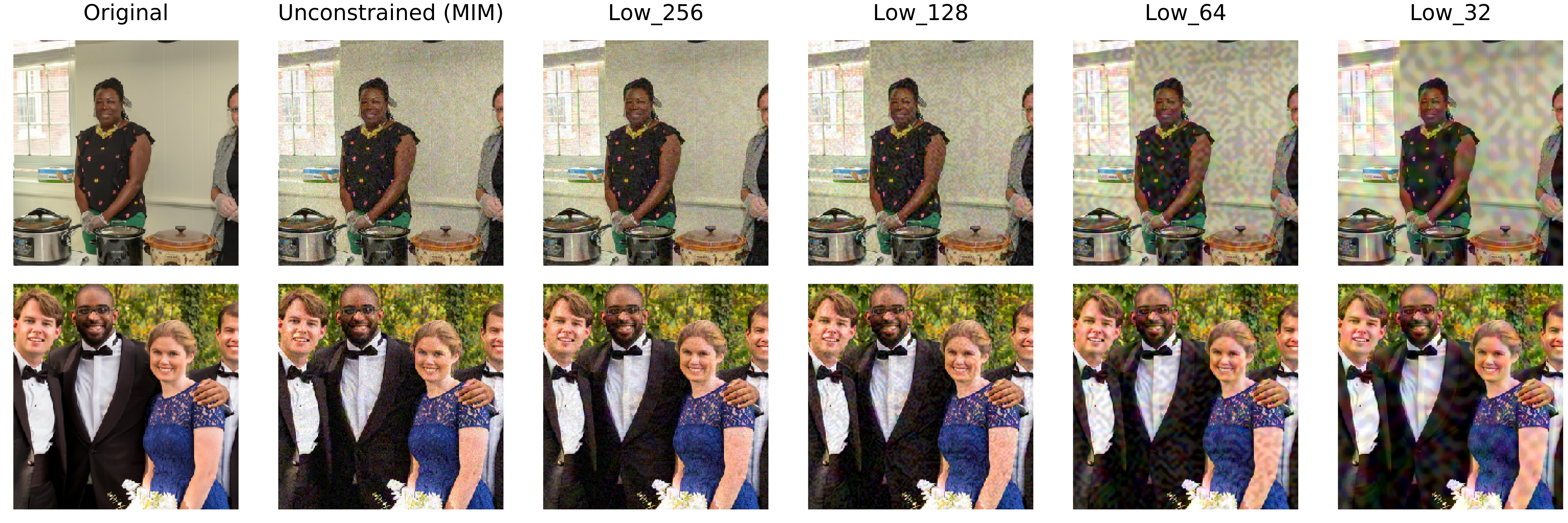}
    \caption{Adversarial examples generated with $\ell_\infty$ $\epsilon=16/255$ distortion}
    \label{perception}
\end{figure*}
\subsection{Effectiveness of Low Frequency on Undefended Models v.s. Defended Models}

In the last section, we showed that \dctlow is highly effective against adversarially trained models and top-performing preprocessor-based defenses, in the white-box, grey-box and black-box cases. However, low frequency does not help when only cleanly trained models are involved, i.e. white-box on clean models and black-box transfer between clean models. To explain this phenomenon, we hypothesize that the state-of-the-art ImageNet defenses considered here do not reduce vulnerabilities within the low frequency subspace, and thus \dctlow is roughly as effective against defended models as against clean models, a property not seen when evaluating with standard, unconstrained attacks.

This can be most clearly seen in \Cref{fig:matrix_percent}, which presents the normalized difference between ASR on each of the target models with ASR on the cleanly trained model. The difference is consistently smaller for \dctlow than for unconstrained MIM, and nearly nonexistent when the perturbations were generated against adversarially trained (defended) models (Adv\_1,Adv\_3). Thus, as discussed, defended models are roughly as vulnerable as undefended models when encountered by low frequency perturbations.
\begin{table}[!h]
\begin{tabular}{l|lll}
\midrule
Dim & White (Adv) & Black (Adv) & Black (Cln) \\ \midrule
32             & 54.6      & 38.1            & 14.4            \\
24             & 48.1      & 33.1            & 14.4            \\
16             & 46.4      & 28.8            & 14.4            \\
8              & 37.0      & 25.4            & 14.4            \\
4              & 26.5      & 20.0            & 14.0 \\
\end{tabular}
\caption{Non-targeted attack success rate (ASR) with $\text{iterations}=10$ and $\epsilon=8/255$ of \texttt{DCT\_Low} in the white-box and black-box settings (transfer from distinct adversarially trained and cleanly trained models of the same architecture) against adversarially trained model with 12.9\% test error~\citep{madry}.}
\label{madry}
\end{table}
\subsection{Effectiveness of Low Frequency on CIFAR-10}
We test the effectiveness of low frequency perturbations on the much lower-dimensional than ImageNet, CIFAR-10 dataset ($d=299$ to $d=32$), attacking the state-of-the-art adversarially trained model~\citep{madry}. Experiment results with 1000 test examples can be seen in \Cref{madry}. Constraining the adversary used for training (non-targeted PGD~\citep{kurakin2016adversarial,madry}; $\text{iterations}=10$ and $\epsilon=8/255$) with \texttt{DCT\_Low}, and evaluating both in the white-box and black-box settings (transfer from distinct adversarially trained and cleanly trained models of the same architecture), we observe that dimensionality reduction only hurts performance. This suggests that the notion of low frequency perturbations is not only constrained to the computer vision domain, but also only induces problems for robustness in the realm of high-dimensional natural images. 
\section{Discussion}
Our experiments show that the results seen in recent work on the effectiveness of constraining the attack space to low frequency components \citep{guo2018lowfreq,zhou2018transferable,caadcompetition} are not due to generally reducing the size of the attack search space. When evaluating against state-of-the-art adversarially trained models and winning defense submissions in the NeurIPS 2017 competition in the white-box, grey-box, and black-box settings, significant improvements are only yielded when low frequency components of the perturbation are preserved. Low frequency perturbations are so effective that they render state-of-the-art ImageNet defenses to be roughly as vulnerable as undefended, cleanly trained models under attack.

However, we also noticed that low frequency perturbations do not improve performance when defended models are not involved, seen in evaluating white-box performance against and black-box transfer between cleanly trained models. Low frequency perturbations do not yield faster white-box attacks on clean models, nor do they provide more effective transfer between clean models.

Our results suggest that the state-of-the-art ImageNet defenses, based on necessarily imperfect adversarial training, only significantly reduce vulnerability outside of the low frequency subspace, but not so much within.
Against defenses, low frequency perturbations are more effective than unconstrained ones since they exploit the vulnerabilities which purportedly robust models share. Against undefended models, constraining to a subspace of significantly reduced dimensionality is unhelpful, since undefended models share vulnerabilities beyond the low frequency subspace. Understanding whether this observed vulnerability in defenses is caused by an intrinsic difficulty to being robust in the low frequency subspace, or simply due to the (adversarial) training procedure rarely sampling from the low frequency region is an interesting direction for further work. 

\paragraph{Low frequency perturbations are perceptible (under $\ell_\infty$-norm  $\epsilon=16/255$).} Our results show that the robustness of currently proposed ImageNet defenses relies on the assumption that adversarial perturbations are high frequency in nature. Though the adversarial defense problem is not constrained to achieving robustness to imperceptible perturbations, this is a reasonable first step. Thus, in Figure~\ref{perception}, we visualize low frequency constrained adversarial examples under the competition $\ell_\infty$-norm constraint $\epsilon=16/255$. Though the perturbations do not significantly change human perceptual judgement, e.g. the top example still appears to be a standing woman, the perturbations with $n\leq 128$ are indeed perceptible.
% , although non of these perturbations actually changes humans' perception of ``a standing woman''.

Although it is well-known that $\ell_p$-norms (in input space) are far from metrics aligned with human perception, exemplified by their widespread use, it is still assumed that with a small enough bound (e.g. $\ell_\infty$  $\epsilon=16/255$), the resulting ball will constitute a subset of the imperceptible region. The fact that low frequency perturbations are fairly visible challenges this common belief. In addition, if the goal is robustness to imperceptible perturbations, our study suggests this might be achieved, without adversarial training, by relying on low frequency components, yielding a much more computationally practical training procedure. In all, we hope our study encourages researchers to not only consider the frequency space, but perceptual priors in general, when bounding perturbations and proposing tractable, reliable defenses.
\bibliographystyle{named}
\bibliography{ijcai19}
\newpage
\appendix
\section{Construction Process}
For a specified reduced dimensionality $n$, and original dimensionality $d$, we consider the frequency subspace $\mathbb{R}^{n\times n}$. For the low frequency domain, \texttt{DCT\_Low}, we preserve components $v_{i,j}$ if $1\leq i,j\leq n$.

For the high frequency domain, \texttt{DCT\_High}, we do the opposite, masking the lowest frequency components such that $n^2$ components are preserved: $1\leq i,j\leq d(1-r_{h})$. Thus $r_{h}d$ bands (rows/columns in $V$) are preserved. To ensure the number of preserved components is equal between the differently constructed masks, we specify $r_{l}=\frac{n}{d}$, and solve the following equation for $r_{h}$:
\begin{align}
\begin{split}
d^{2}-(1-r_{h})^{2}d^{2}&=r_{l}^{2}d^{2}\\
1-(1-r_{h})^2&=r_{l}^{2}\\
2r_{h}-r_{h}^2&=r_{l}^{2}\\
r_{h}^2-2r_{h}+r_{l}^{2}&=0
\end{split}
\label{lowtohigh}
\end{align}
Solving the quadratic equation for $0\leq r_{h}\leq1$, $r_{h}=1-\frac{\sqrt{4(1-r_{l}^{2})}}{2}$.

For the middle frequency band, \texttt{DCT\_Mid}, we would like to ensure we mask an equal number of low and high frequency components. We thus solve the following equation for $r_{ml}$:
\begin{align}
\begin{split}
r_{ml}^{2}d^{2}&=\frac{d^{2}-r_{l}^{2}d^{2}}{2}\\
r_{ml}&=\sqrt{\frac{1-r_{l}^{2}}{2}}
\end{split}
\end{align}
Thus, we mask components $v_{i,j}$ if $1\leq i,j\leq r_{ml}d$, and if $d(1-r_{mh})\leq i,j\leq d$. Compute $r_{mh}$ from $r_{ml}$ with equation~\ref{lowtohigh}. 

For our representative random frequency mask, \texttt{DCT\_Random}, much like \texttt{DCT\_High}, rows/columns are chosen, except in this case randomly rather than the highest frequency bands. To ensure that $n^2$ components are preserved, $r_{h}d$ rows/columns are chosen, which are then preserved in both the $x$ and $y$ directions.
\section{Spatial Smoothing \& Downsampling-Upsampling Filters}
\begin{figure*}[!h]
\begin{subfigure}{0.49\linewidth}
    \centering
    \includegraphics[width=\linewidth]{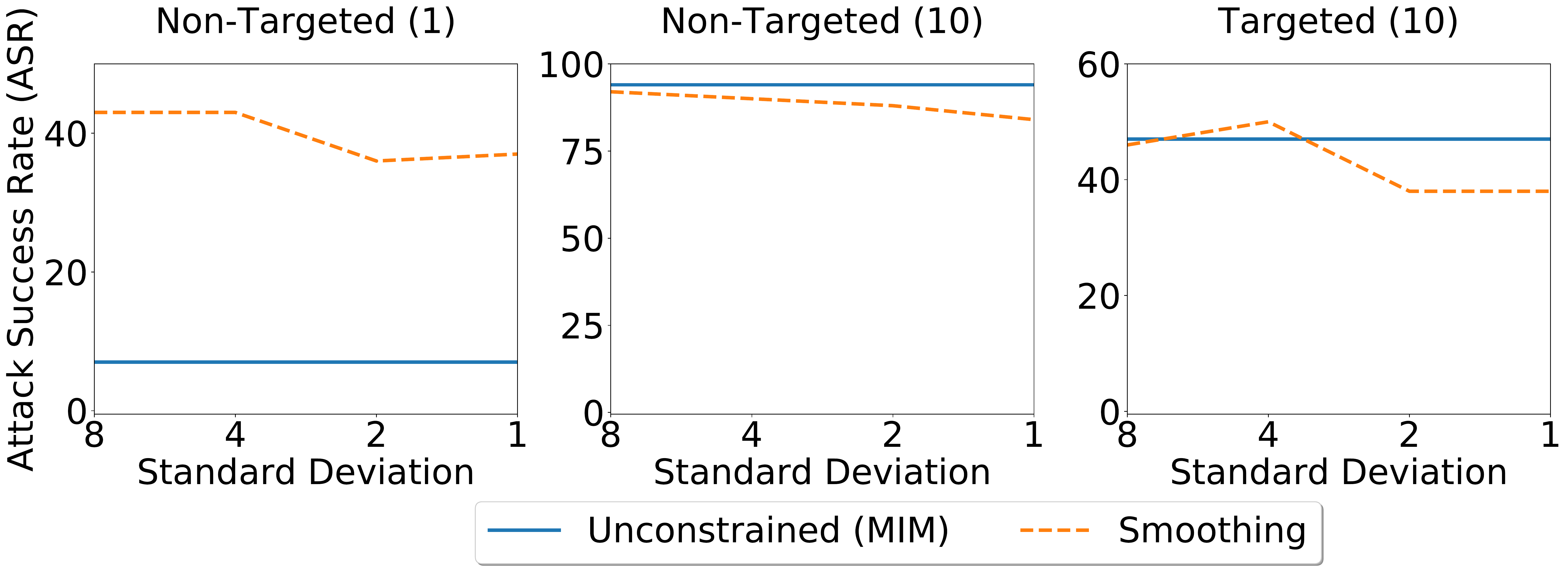}
    \caption{\textbf{White-box} attack on adversarially trained model, EnsAdv.}
    \label{sswhitebox}
\end{subfigure}
~~~~
\begin{subfigure}{0.49\linewidth}
    \centering
    \includegraphics[width=\linewidth]{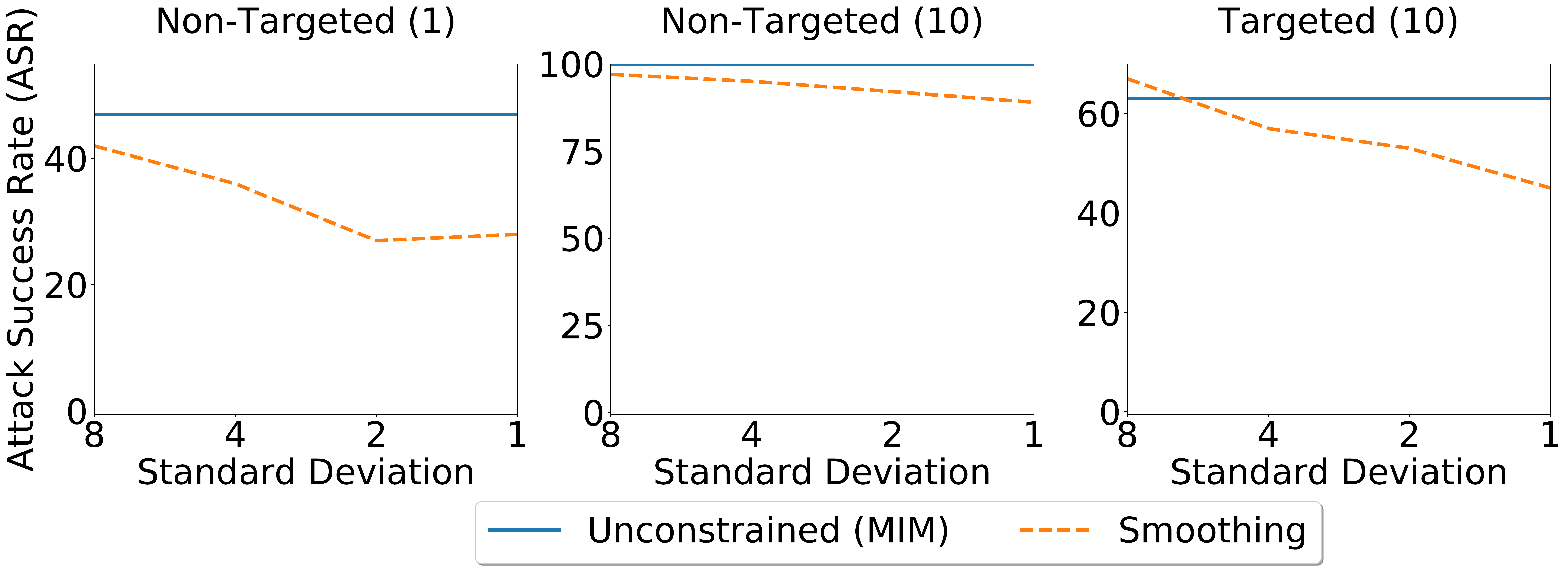}
    \caption{\textbf{White-box} attack on standard cleanly trained model, NasNet.}
    \label{ssclnwhitebox}
\end{subfigure}
\hfill
\begin{subfigure}{0.49\linewidth}
    \centering
    \includegraphics[width=\linewidth]{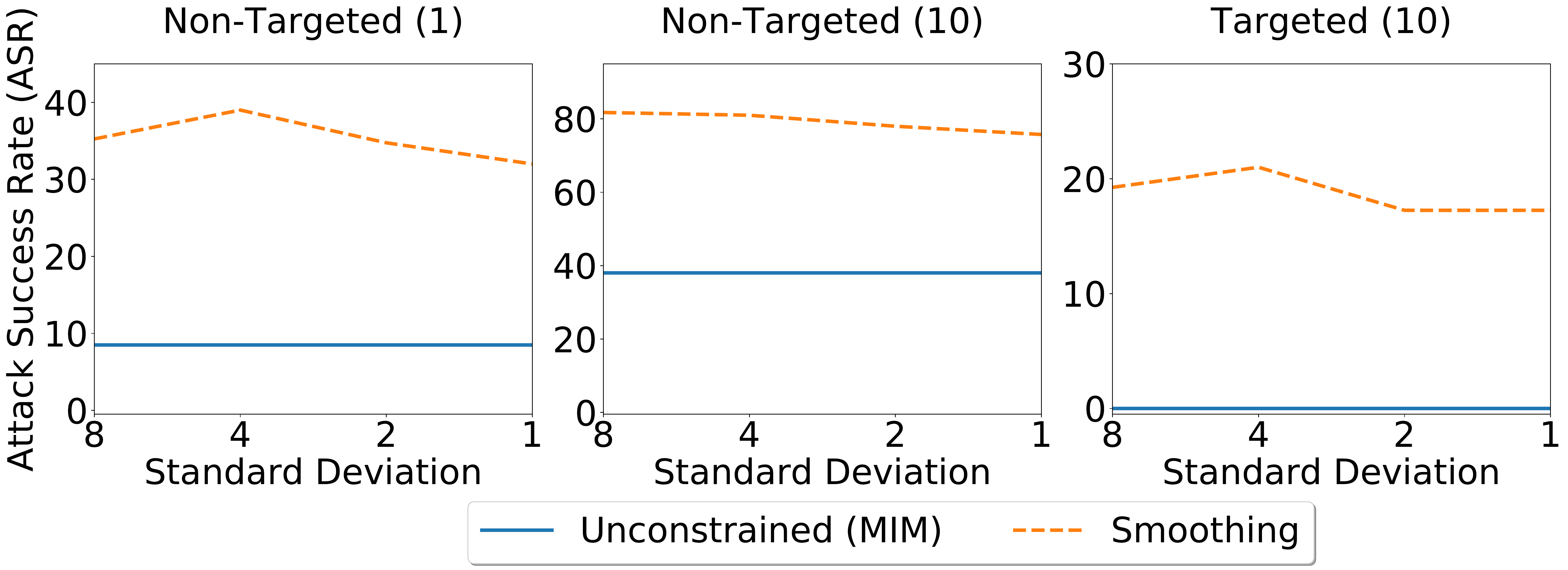}
    \caption{\textbf{Grey-box} attack on top-4 NeurIPS 2017 defenses prepended to adversarially trained model.}%\\~\\~}
    \label{ssgreybox}
\end{subfigure}
~~~~
\begin{subfigure}{0.49\linewidth}
    \centering
    \includegraphics[width=\linewidth]{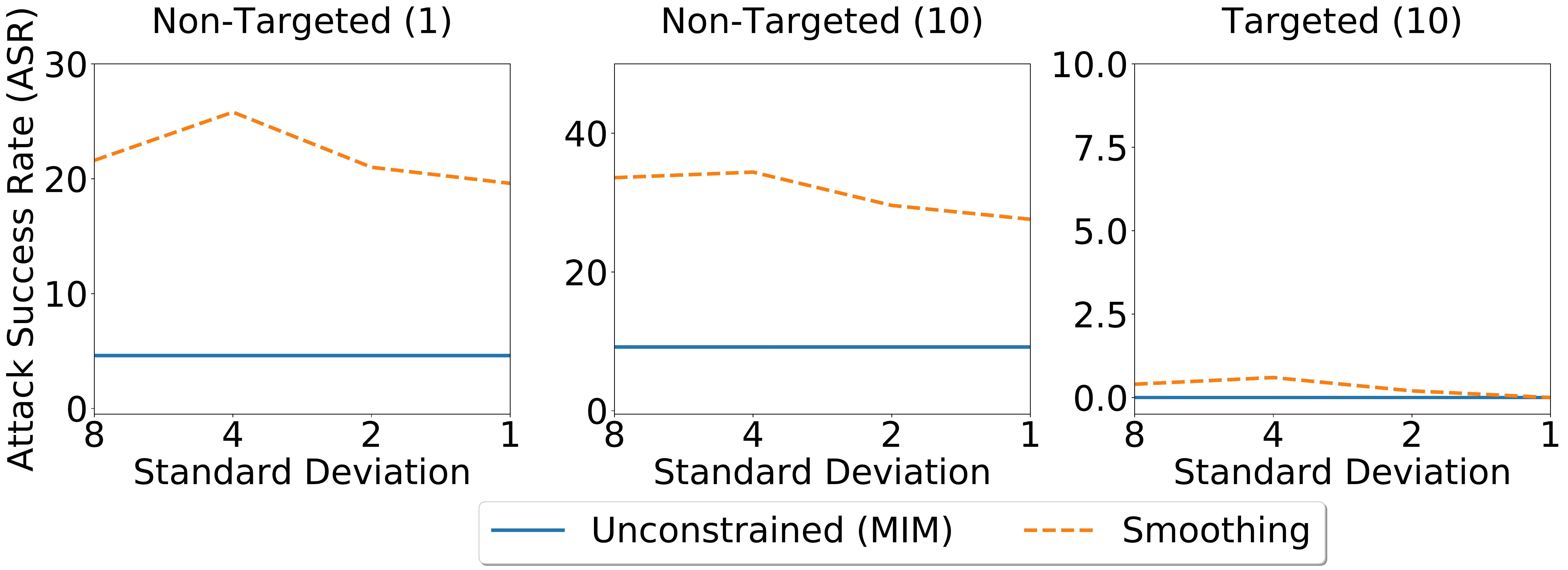}
    \caption{\textbf{Black-box} attack on sources transferred to defenses (EnsAdv + D1$\sim$4)}
    \label{ssblackbox}
\end{subfigure}
\caption{\textbf{Spatial Smoothing} filter; number of iterations in parentheses, non-targeted with $\epsilon=16/255$, targeted with $\epsilon=32/255$.}
\label{ssplot}
\end{figure*}

\begin{figure*}[!h]
\begin{subfigure}{0.49\linewidth}
    \centering
    \includegraphics[width=\linewidth]{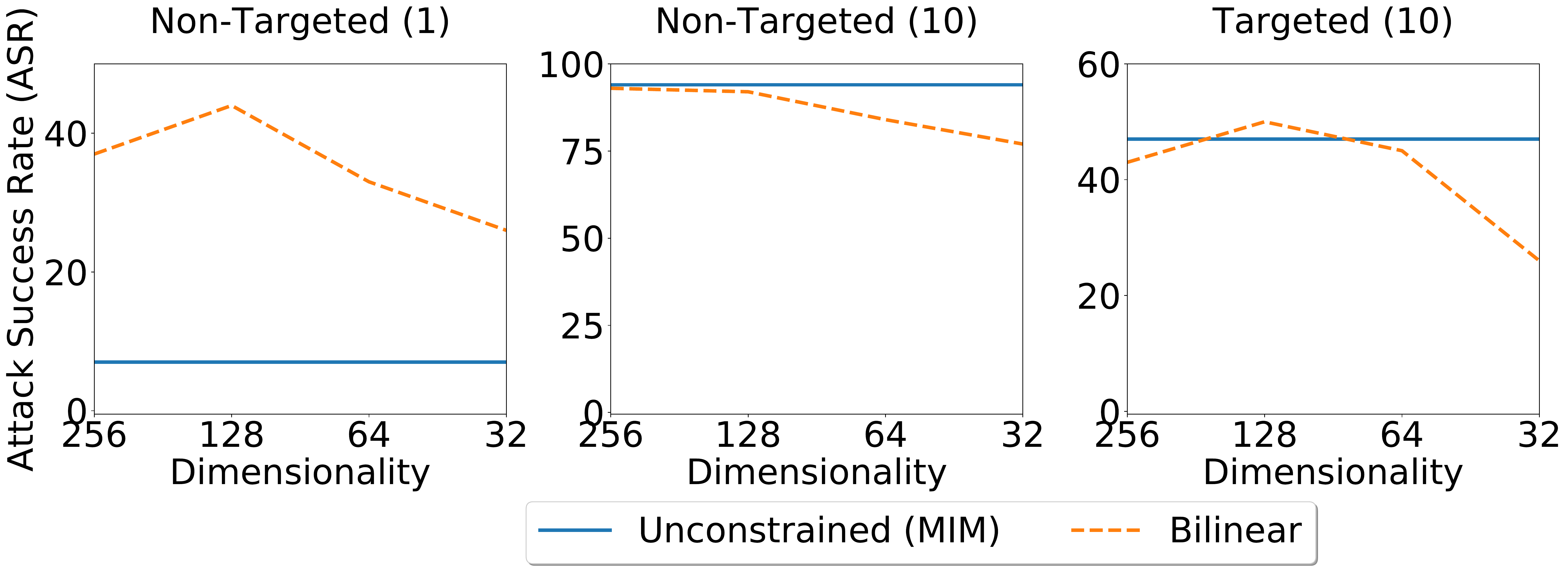}
    \caption{\textbf{White-box} attack on adversarially trained model, EnsAdv.}
    \label{duwhitebox}
\end{subfigure}
~~~~
\begin{subfigure}{0.49\linewidth}
    \centering
    \includegraphics[width=\linewidth]{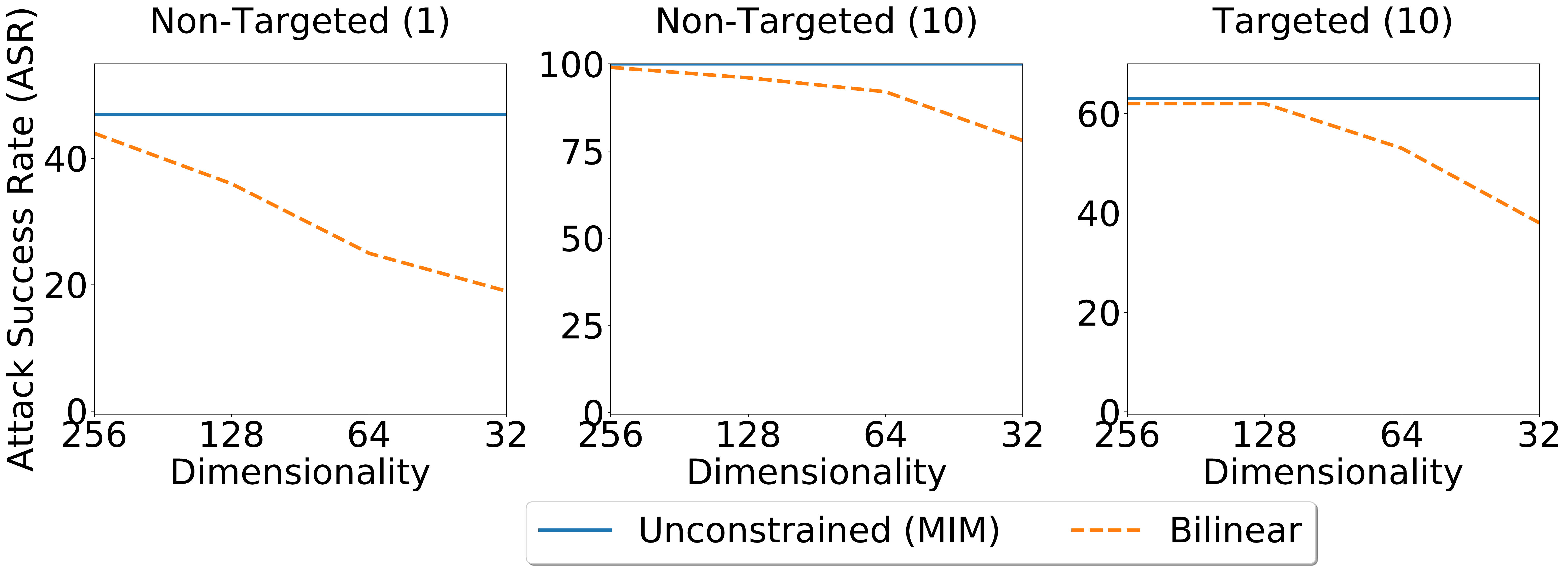}
    \caption{\textbf{White-box} attack on standard cleanly trained model, NasNet.}
    \label{duclnwhitebox}
\end{subfigure}
\hfill
\begin{subfigure}{0.49\linewidth}
    \centering
    \includegraphics[width=\linewidth]{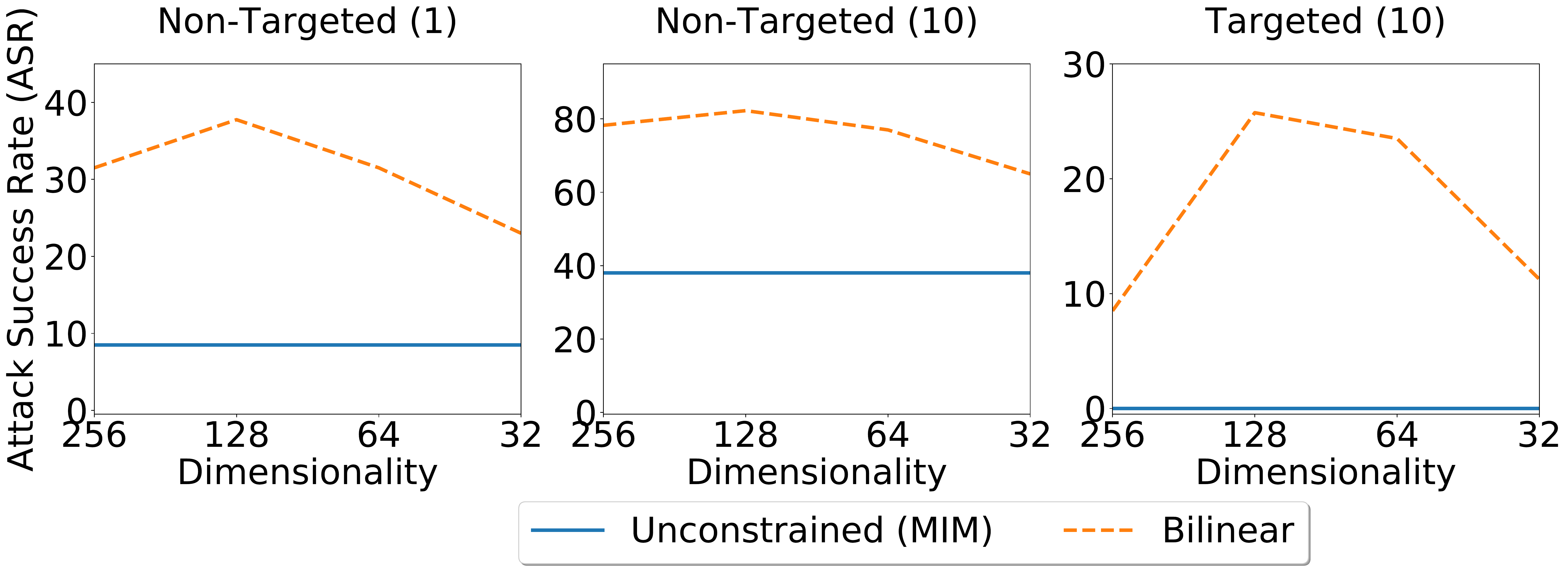}
    \caption{\textbf{Grey-box} attack on top-4 NeurIPS 2017 defenses prepended to adversarially trained model.}%\\~\\~}
    \label{dugreybox}
\end{subfigure}
~~~~
\begin{subfigure}{0.49\linewidth}
    \centering
    \includegraphics[width=\linewidth]{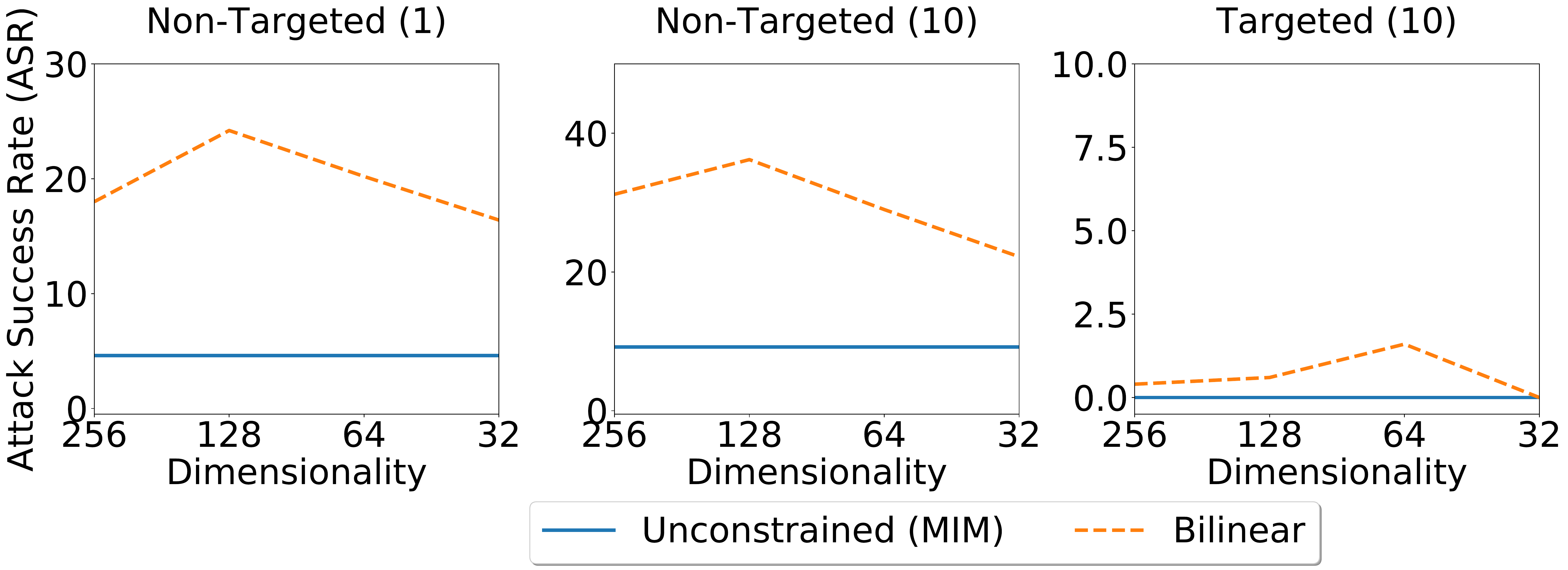}
    \caption{\textbf{Black-box} attack on sources transferred to defenses (EnsAdv + D1$\sim$4)}
    \label{dublackbox}
\end{subfigure}
\caption{\textbf{Downsampling-Upsampling} filter; number of iterations in parentheses, non-targeted with $\epsilon=16/255$, targeted with $\epsilon=32/255$.}
\label{duplot}
\end{figure*}

In the main paper, we show that \texttt{DCT\_Low} significantly outperforms all other frequency configurations (\texttt{DCT\_High}, \texttt{DCT\_Mid}, \texttt{DCT\_Random}) in the white-box, grey-box, and black-box settings. Specifically, we observed that \texttt{DCT\_Low} generates effective perturbations faster than without constraint on adversarially trained models (but not so on clean models), bypasses defenses prepended to the adversarially trained model, helps black-box transfer to defended models, but is not effective when transferring between undefended cleanly trained models. We observe mirrored results when constraining the perturbation with both spatial smoothing and downsampling-upsampling filters; shown in Figure~\ref{ssplot} and \ref{duplot}.

For the \textit{downsampling-upsampling} filter, we resize the perturbation with bilinear interpolation, and decrease the dimensionality from $299$ to $32$, as with DCT\_Low. For the \textit{spatial smoothing} filter, we smooth the perturbation with a gaussian filter of fixed kernel size ($7x7$), decreasing the standard deviation to strengthen the constraint. As can be seen, despite the differing parameters, the trends of each of the low frequency perturbation methods are the same.
\section{Complete Heatmap}
\begin{figure*}[!t]
\includegraphics[width=0.48\textwidth]{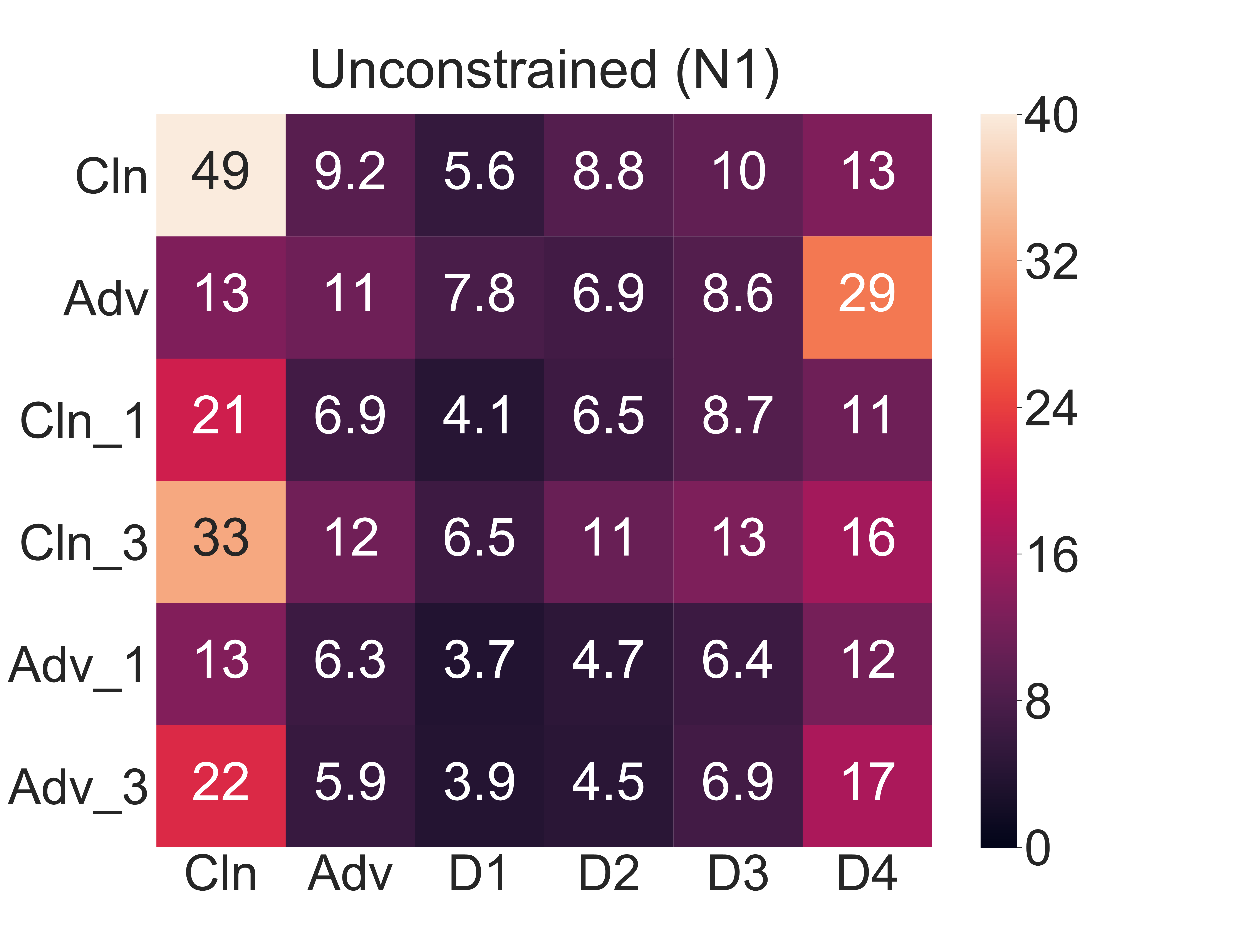}
\includegraphics[width=0.48\textwidth]{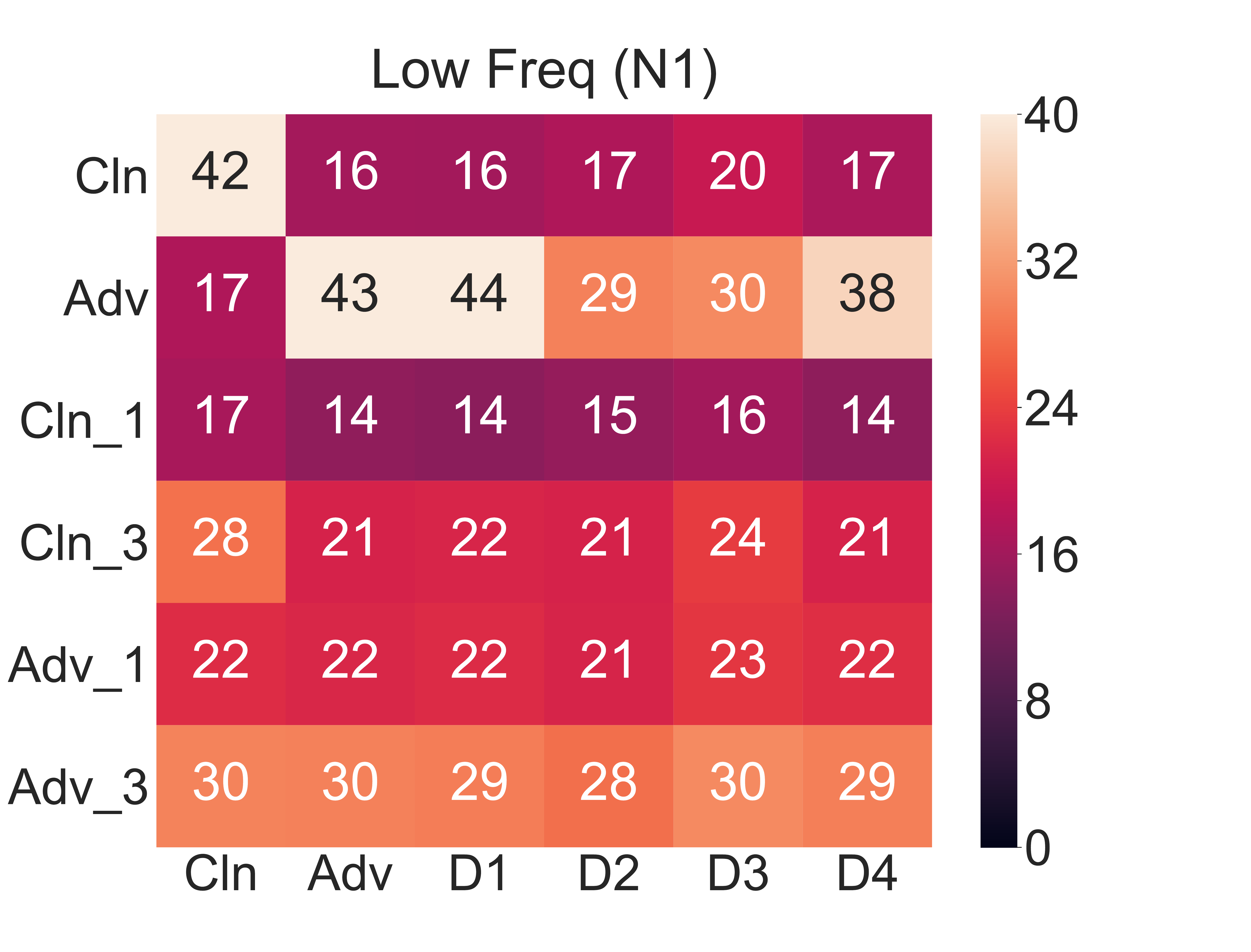}
\includegraphics[width=0.48\textwidth]{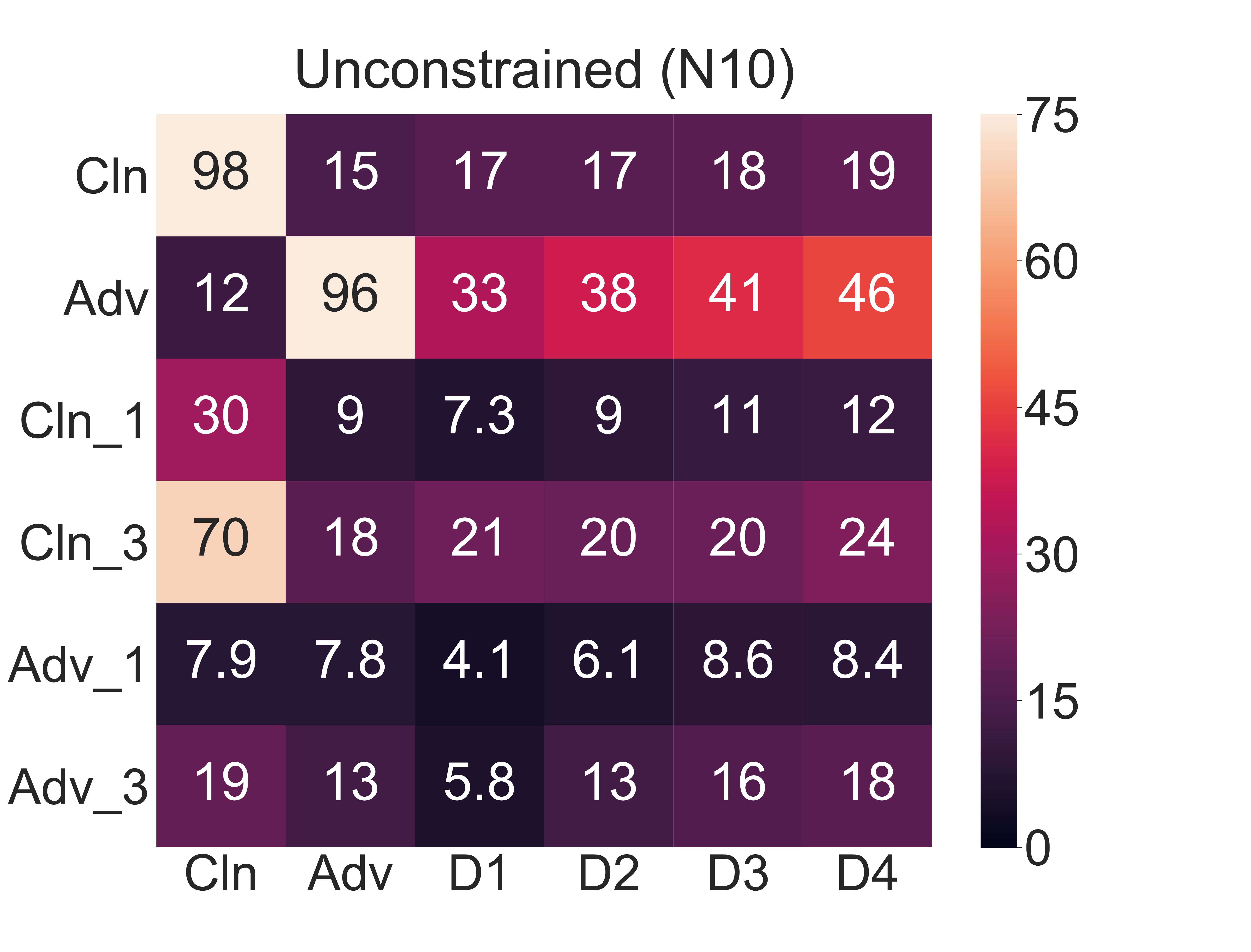}
\includegraphics[width=0.48\textwidth]{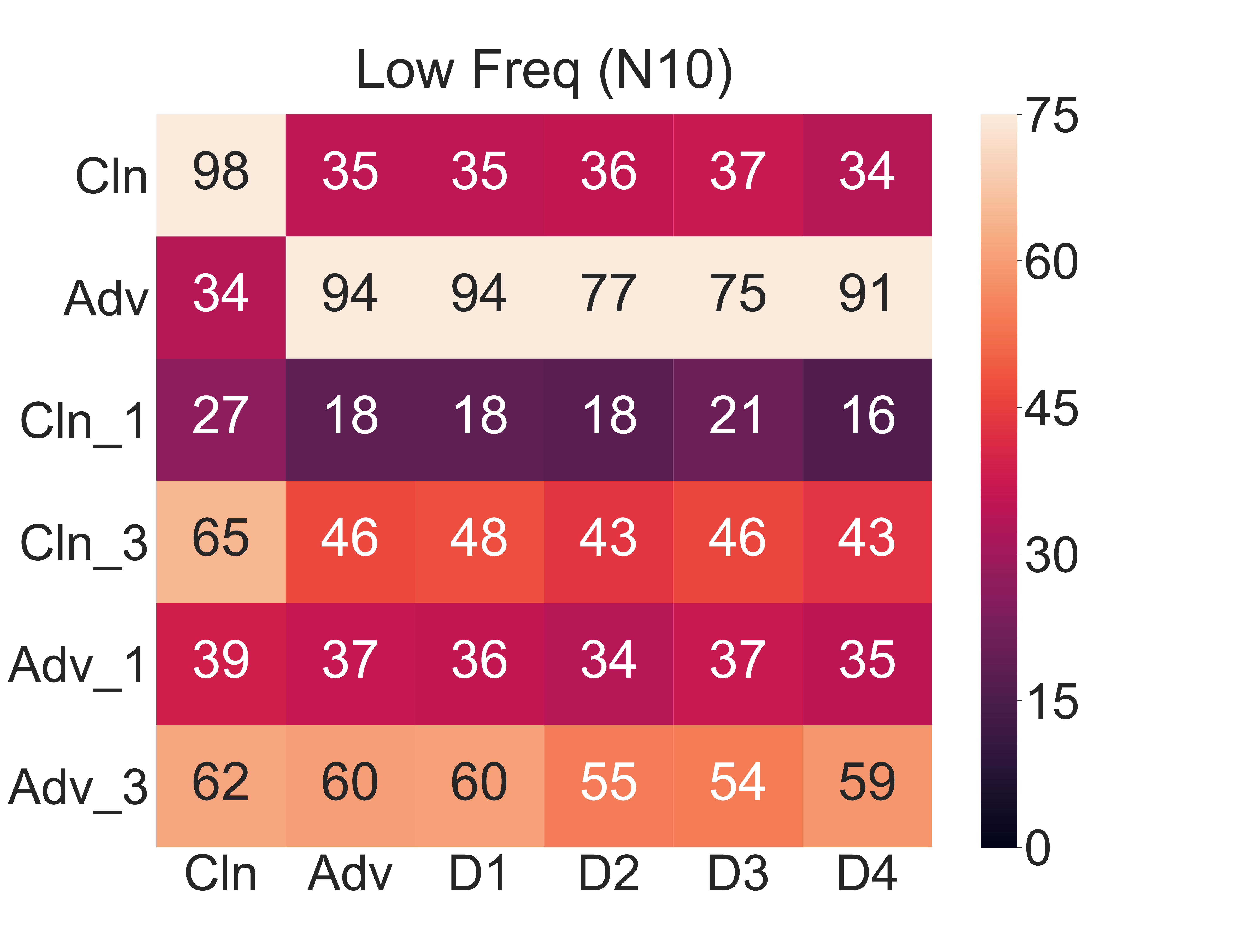}
\includegraphics[width=0.48\textwidth]{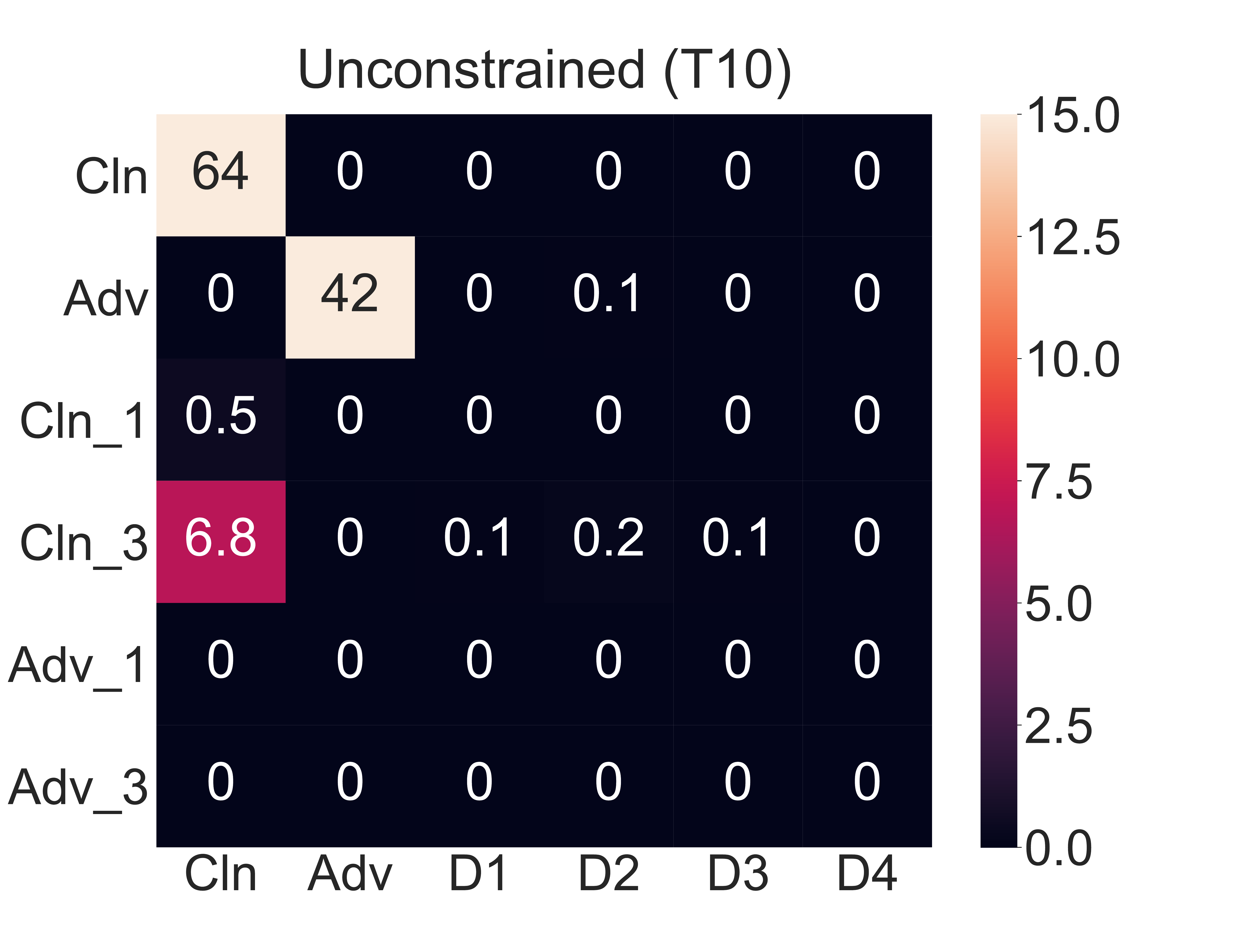}
\includegraphics[width=0.48\textwidth]{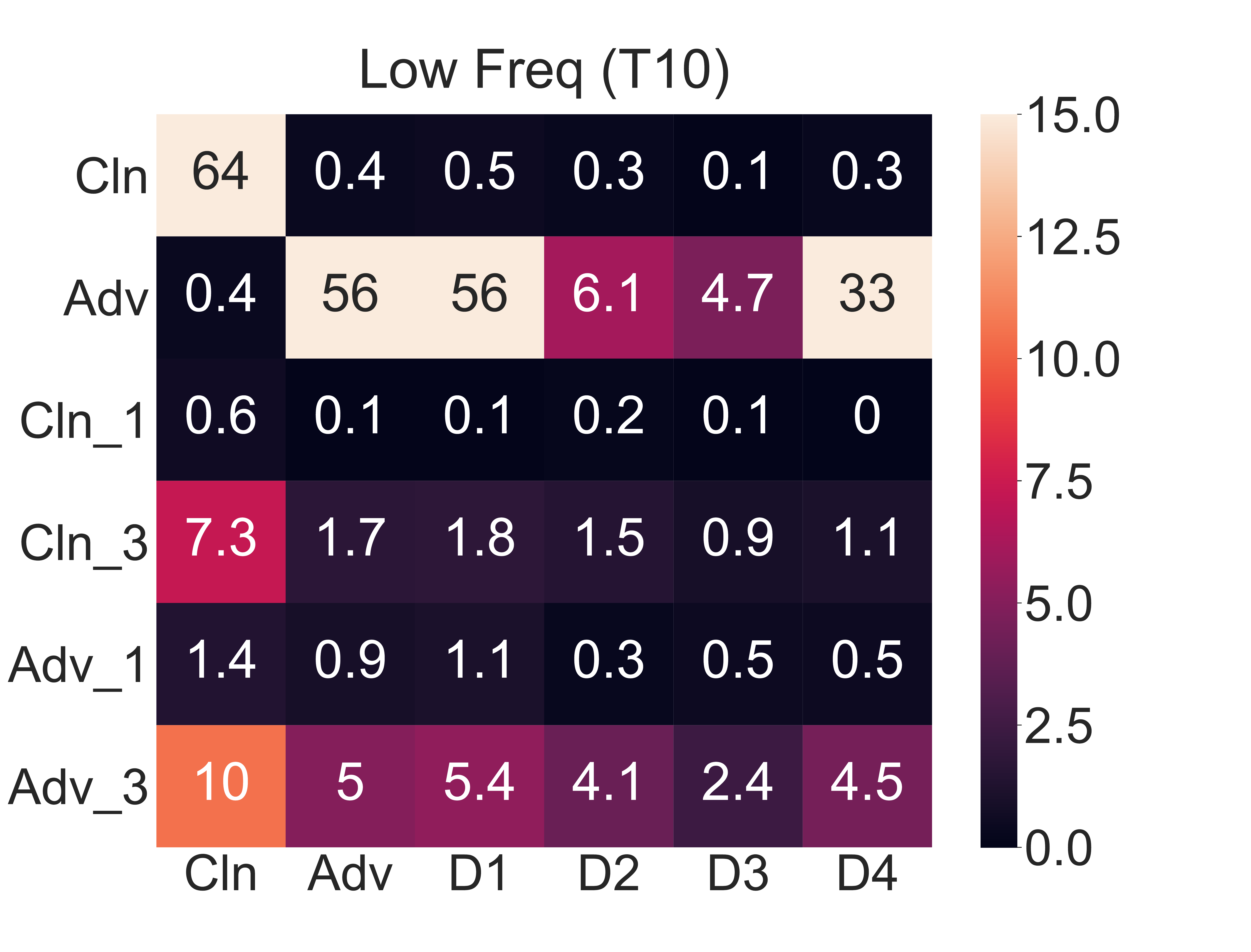}
\caption{Transferability matrix comparing standard unconstrained MIM with low frequency constrained \texttt{DCT\_Low} ($n=128$). First, second, and third rows are non-targeted with $\text{iterations}=1$, non-targeted with $\text{iterations}=10$, and targeted with $\text{iterations}=10$}
\label{matrix1}% Overall figure caption
\end{figure*}

We summarize our attack success rate results with \texttt{DCT\_Low} in Figure~\ref{matrix1}. The rows correspond to sources, and columns corresponds to targets. The sources include [Cln, Adv, Cln\_1, Adv\_1, Cln\_3, Adv\_3], where Cln is \texttt{NasNetLarge\_331}, Adv is \texttt{EnsAdvInceptionResNetV2}; Cln\_1, Adv\_1, Cln\_3, Adv\_3 are summarized in the main text. The targets include [Cln, Adv, D1, D2, D3, D4], where D1$\sim$4 are defenses summarized in the main text. Thus (Cln,Cln) and (Adv,Adv) summarize white-box results, (Adv,D1$\sim$4) summarizes grey-box results, and the rest of the entries summarize black-box results. Note that the low frequency configuration is \texttt{DCT\_Low} with $n=128$.
\section{All Plots}
\begin{figure*}
\begin{subfigure}{0.3\linewidth}
    \centering
    \includegraphics[width=\textwidth]{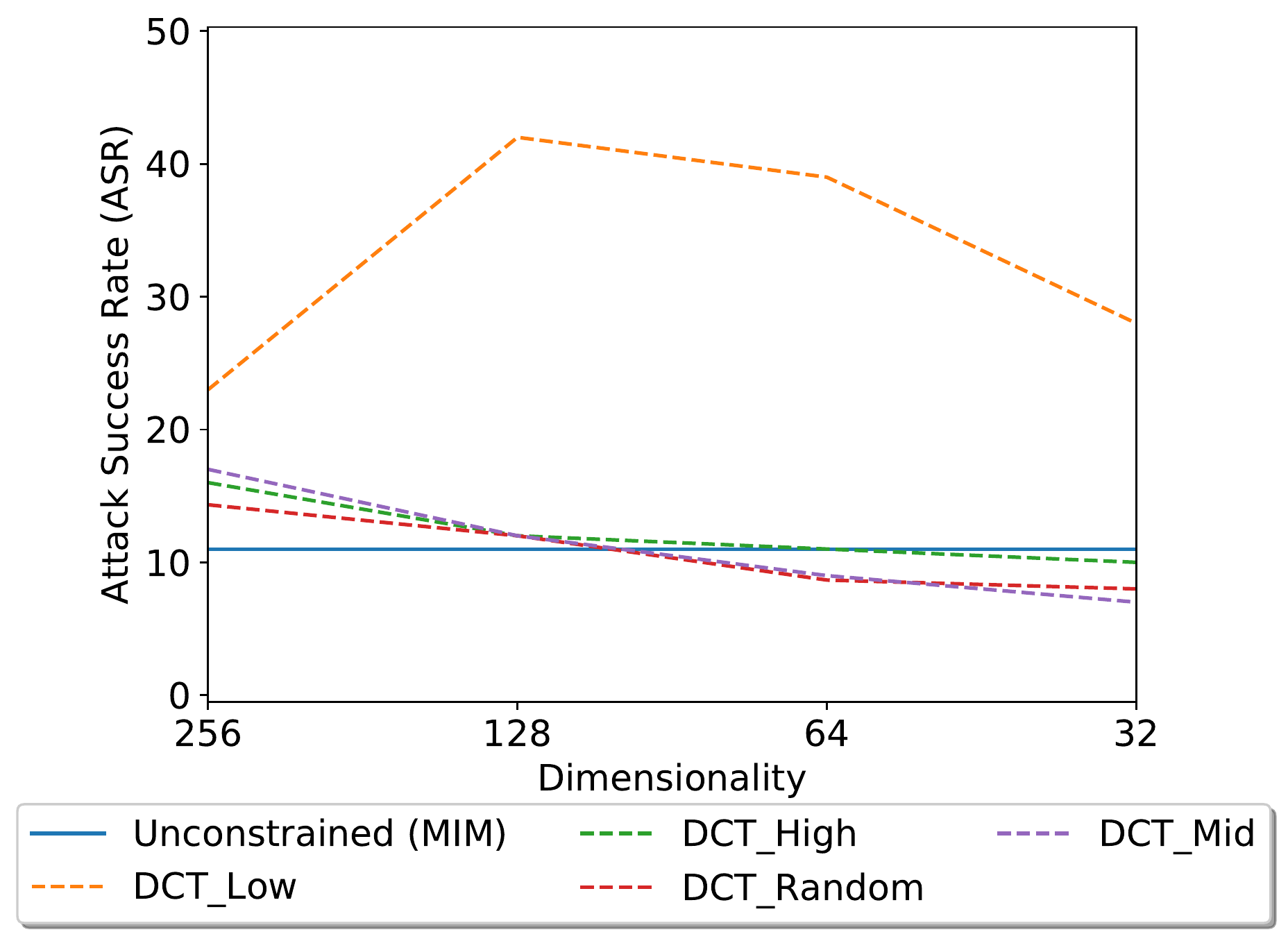}
    \caption{Non-targeted with $\epsilon=16$ and $\text{iterations}=1$.}
    \label{whitebox1}
\end{subfigure}
~~~
\begin{subfigure}{0.3\linewidth}
    \centering
    \includegraphics[width=\textwidth]{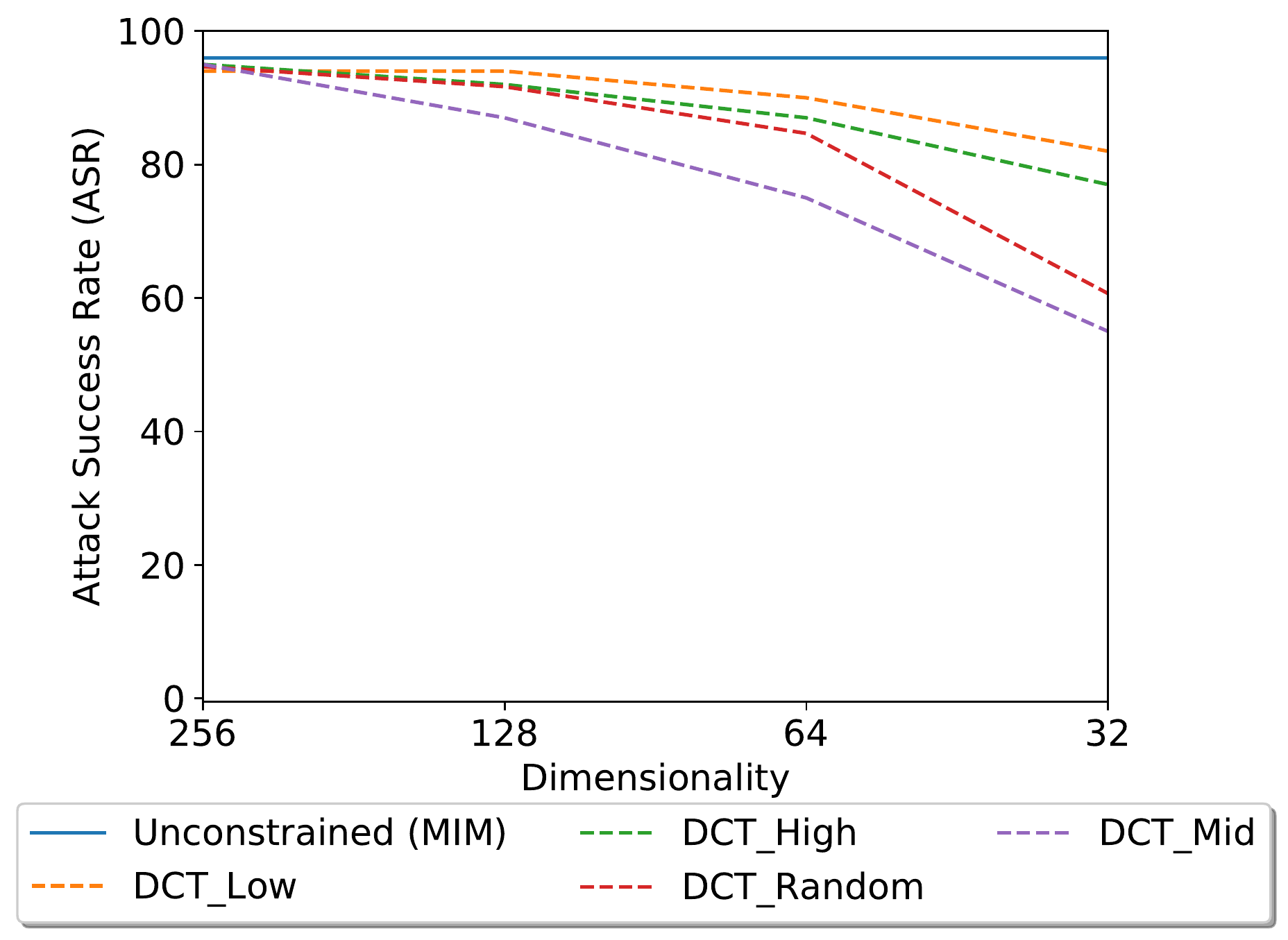}
    \caption{Non-targeted with $\epsilon=16$ and $\text{iterations}=10$.}
    \label{whitebox2}
\end{subfigure}
~~~
\begin{subfigure}{0.3\linewidth}
    \centering
    \includegraphics[width=\textwidth]{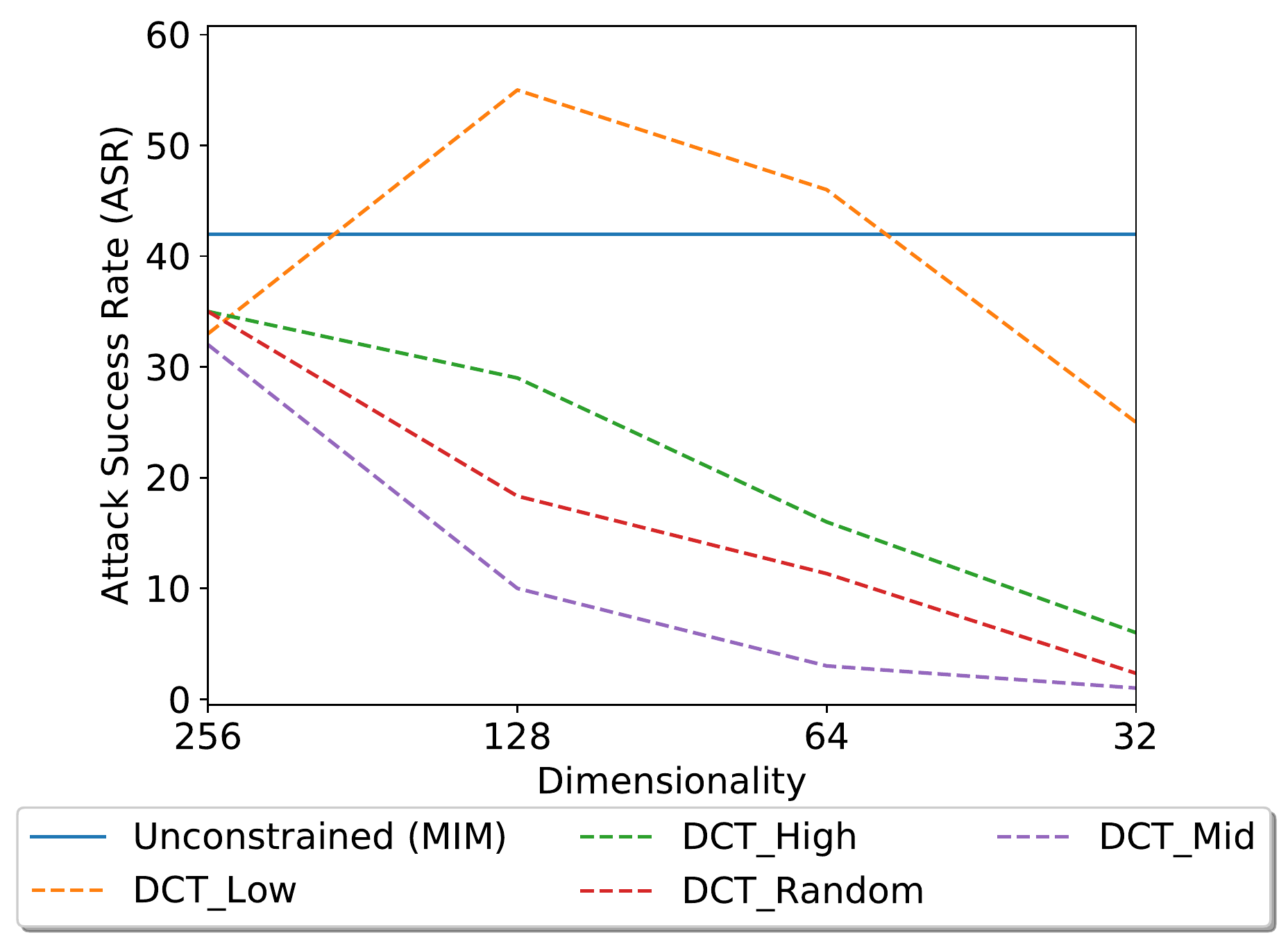}
    \caption{Targeted with $\epsilon=32$ and $\text{iterations}=10$.}
    \label{whitebox3}
\end{subfigure}
\caption{\textbf{White-box} attack on adversarially trained model.}
\label{whiteboxs}
\end{figure*}

\begin{figure*}
\begin{subfigure}{0.3\linewidth}
    \centering
    \includegraphics[width=\textwidth]{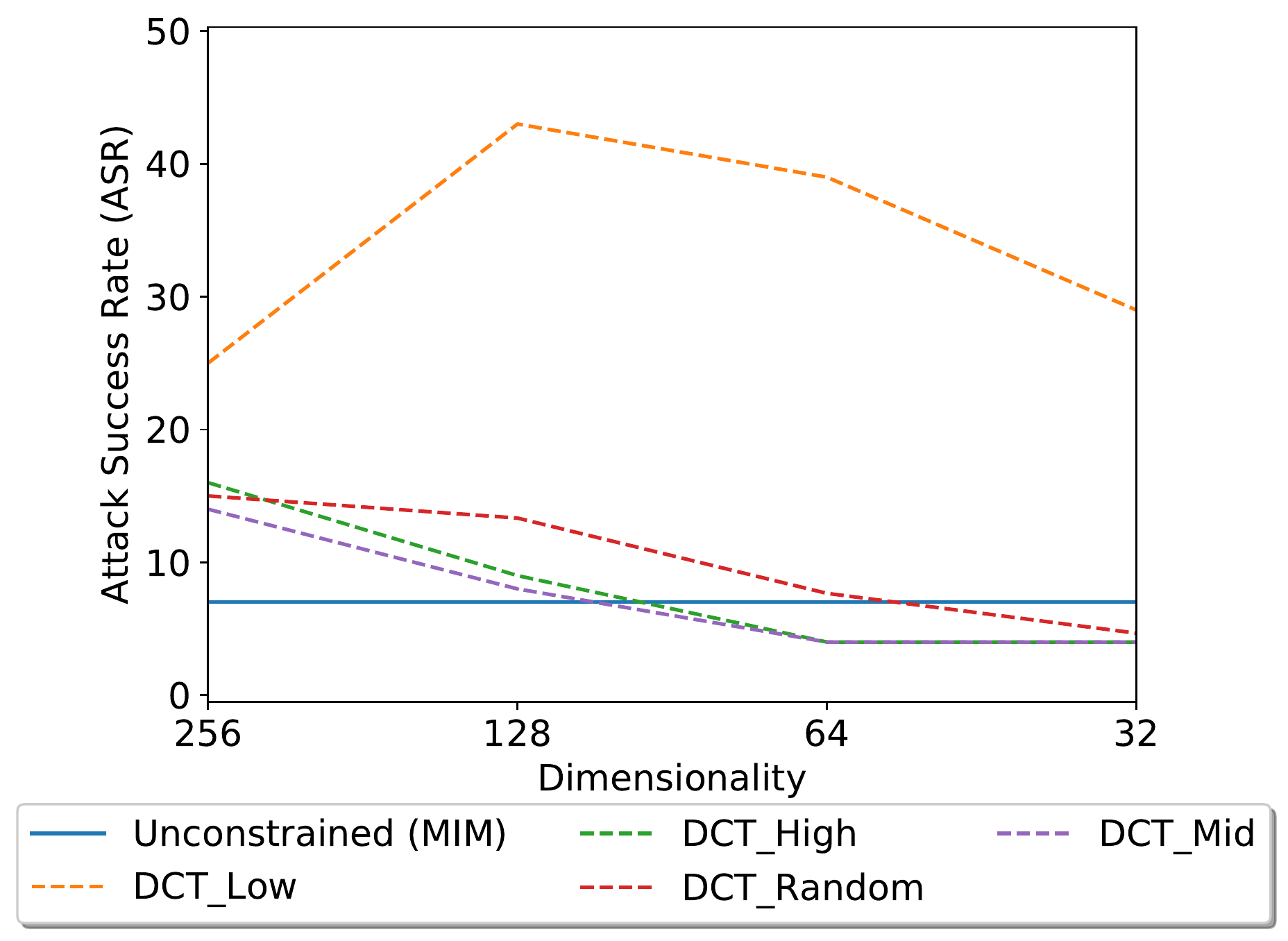}
    \caption{Non-targeted with $\epsilon=16$ and $\text{iterations}=1$.}
    \label{greybox11}
\end{subfigure}
~~~
\begin{subfigure}{0.3\linewidth}
    \centering
    \includegraphics[width=\textwidth]{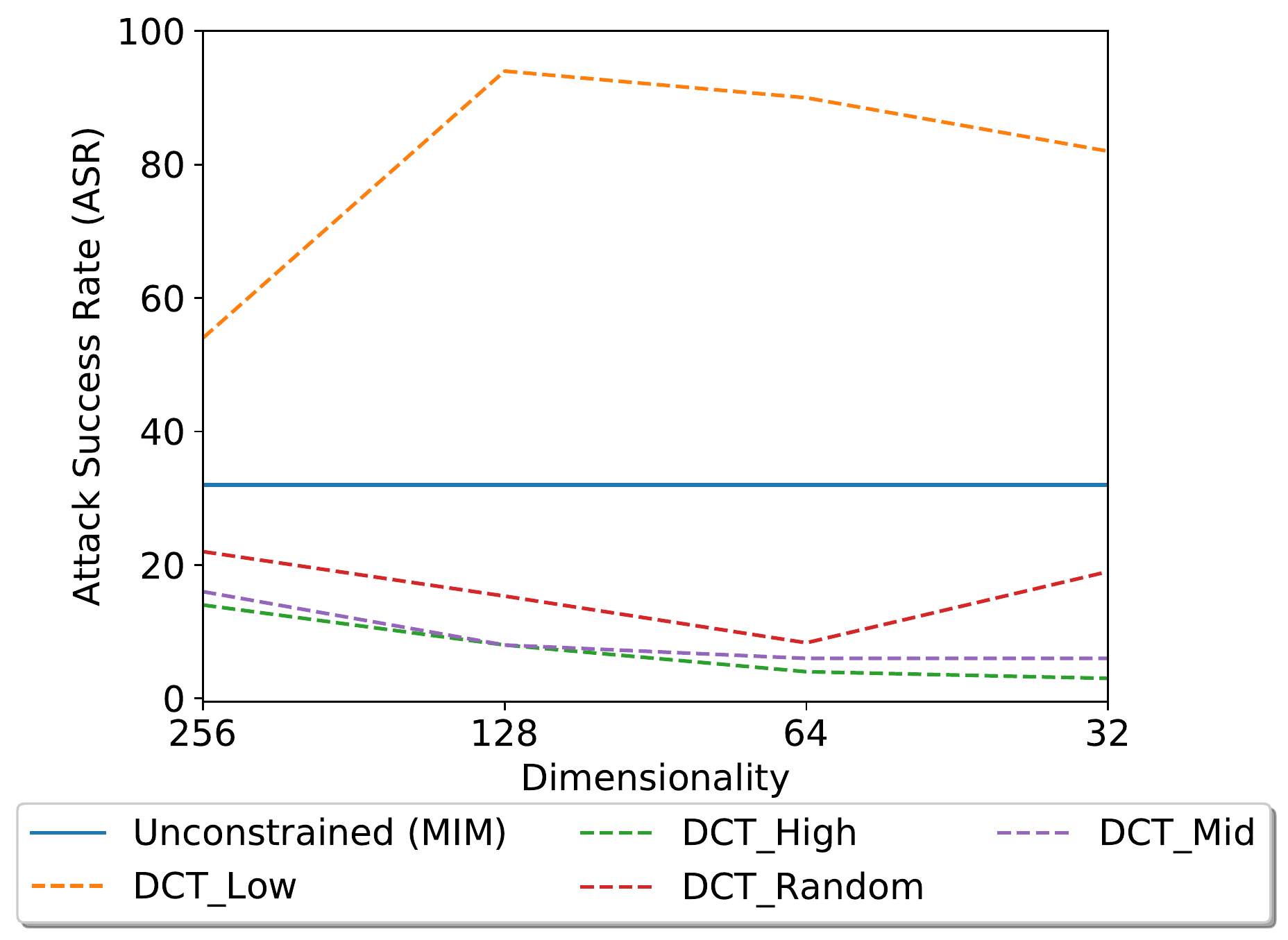}
    \caption{Non-targeted with $\epsilon=16$ and $\text{iterations}=10$.}
    \label{greybox21}
\end{subfigure}
~~~
\begin{subfigure}{0.3\linewidth}
    \centering
    \includegraphics[width=\textwidth]{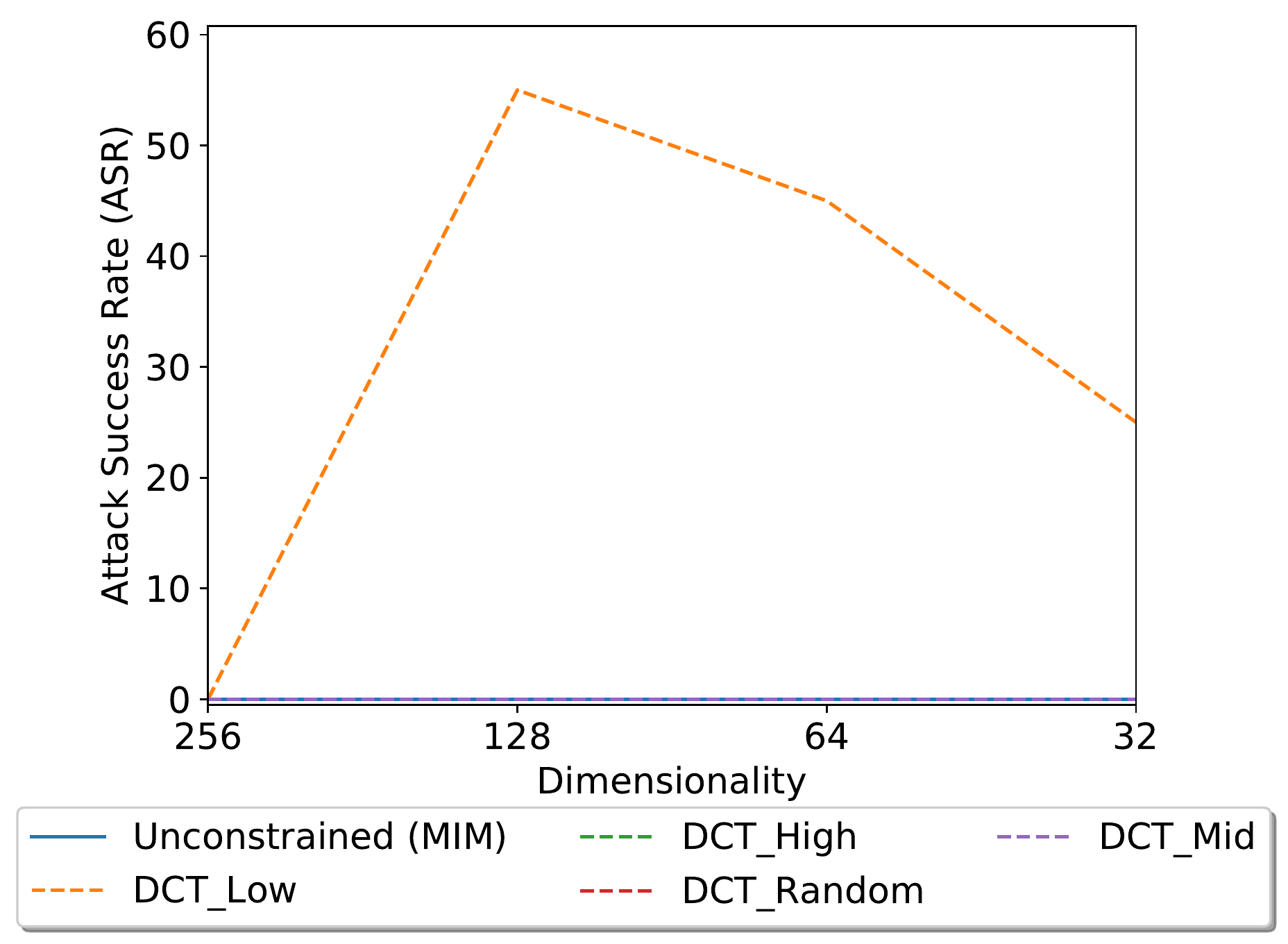}
    \caption{Targeted with $\epsilon=32$ and $\text{iterations}=10$.}
    \label{greybox31}
\end{subfigure}
\caption{\textbf{Grey-box} attack on D1.}
\label{greybox1}
\end{figure*}

\begin{figure*}
\begin{subfigure}{0.3\linewidth}
    \centering
    \includegraphics[width=\textwidth]{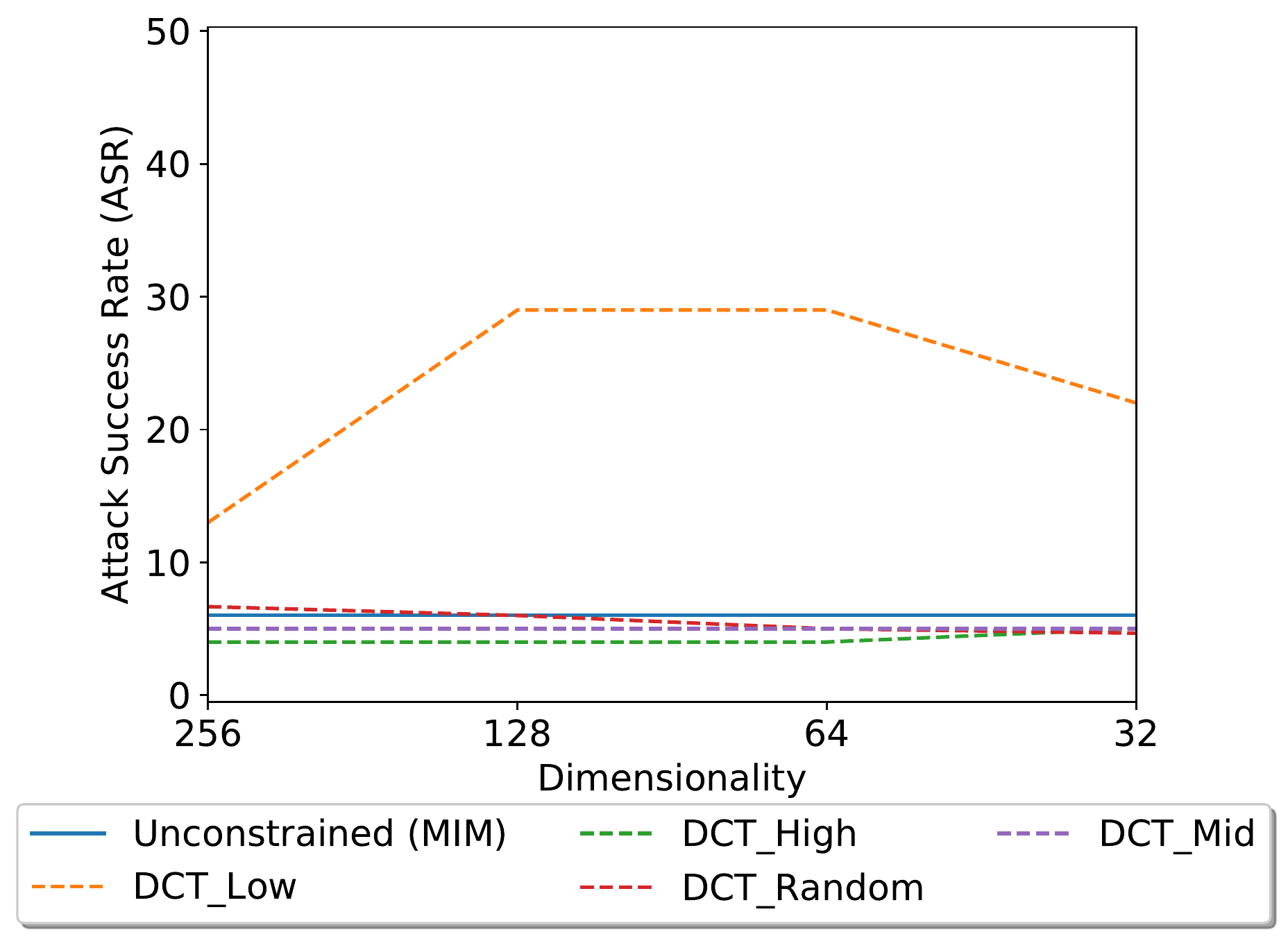}
    \caption{Non-targeted with $\epsilon=16$ and $\text{iterations}=1$.}
    \label{greybox12}
\end{subfigure}
~~~
\begin{subfigure}{0.3\linewidth}
    \centering
    \includegraphics[width=\textwidth]{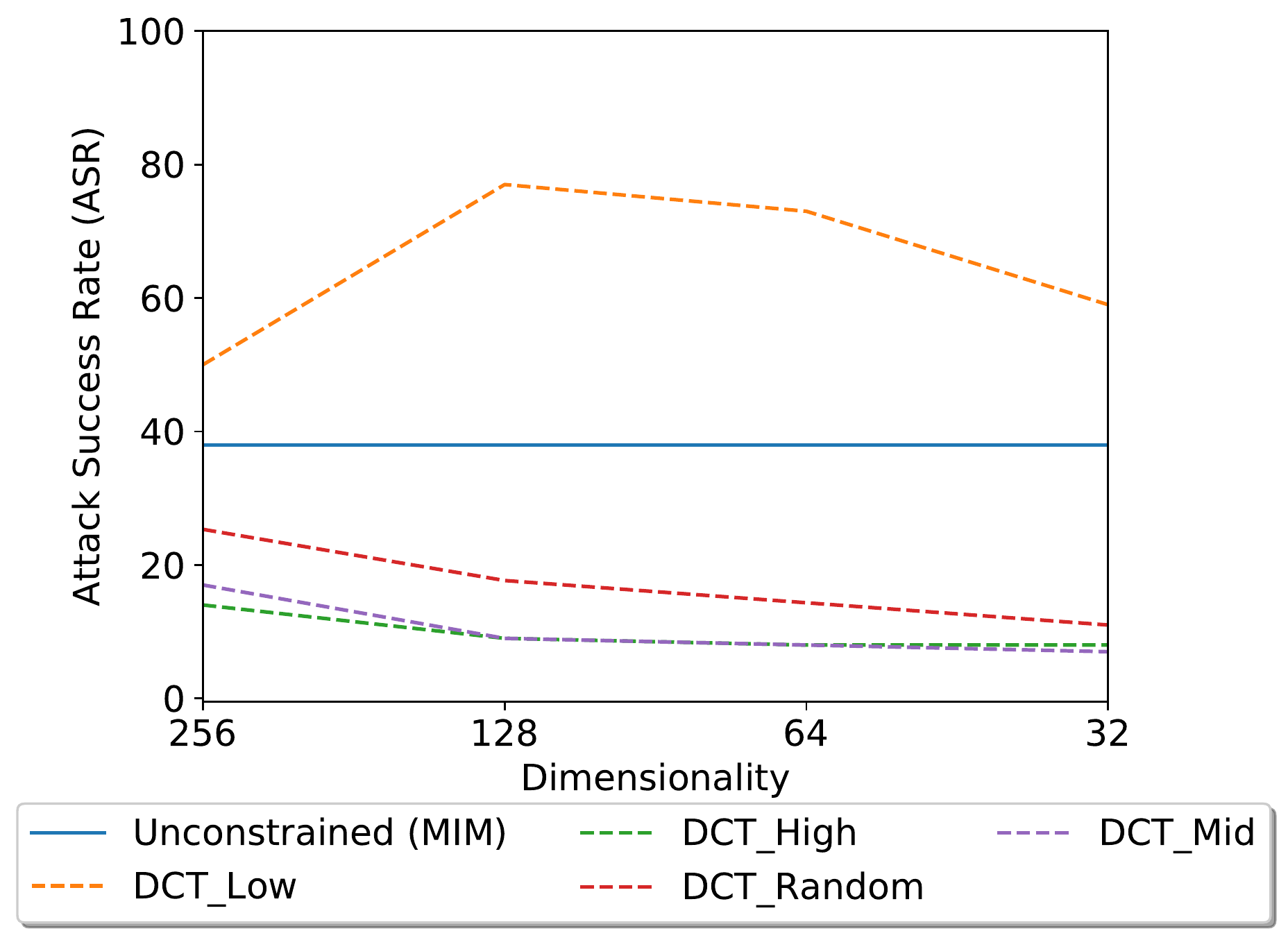}
    \caption{Non-targeted with $\epsilon=16$ and $\text{iterations}=10$.}
    \label{greybox22}
\end{subfigure}
~~~
\begin{subfigure}{0.3\linewidth}
    \centering
    \includegraphics[width=\textwidth]{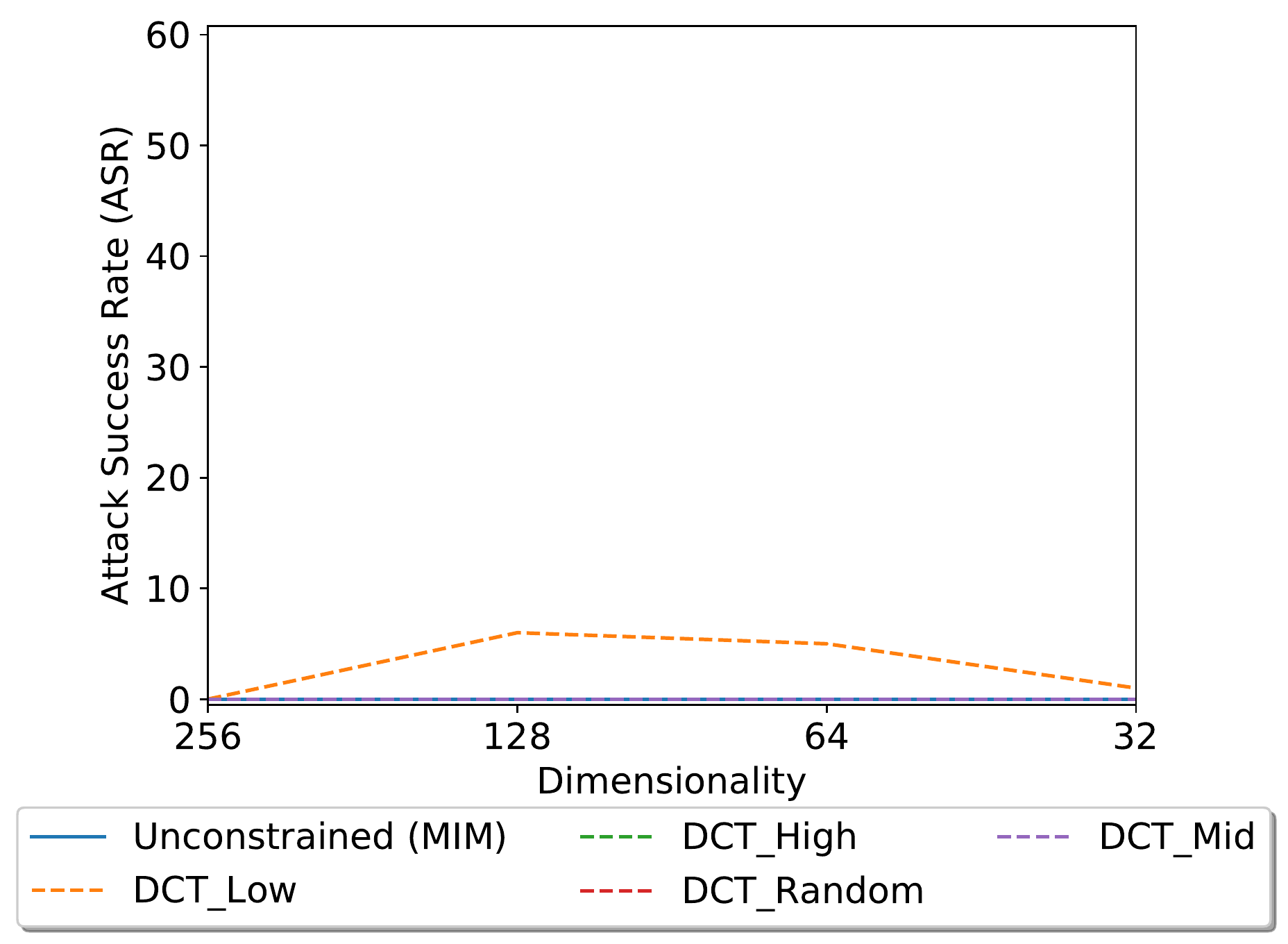}
    \caption{Targeted with $\epsilon=32$ and $\text{iterations}=10$.}
    \label{greybox32}
\end{subfigure}
\caption{\textbf{Grey-box} attack on D2.}
\label{greybox2}
\end{figure*}

\begin{figure*}
\begin{subfigure}{0.3\linewidth}
    \centering
    \includegraphics[width=\textwidth]{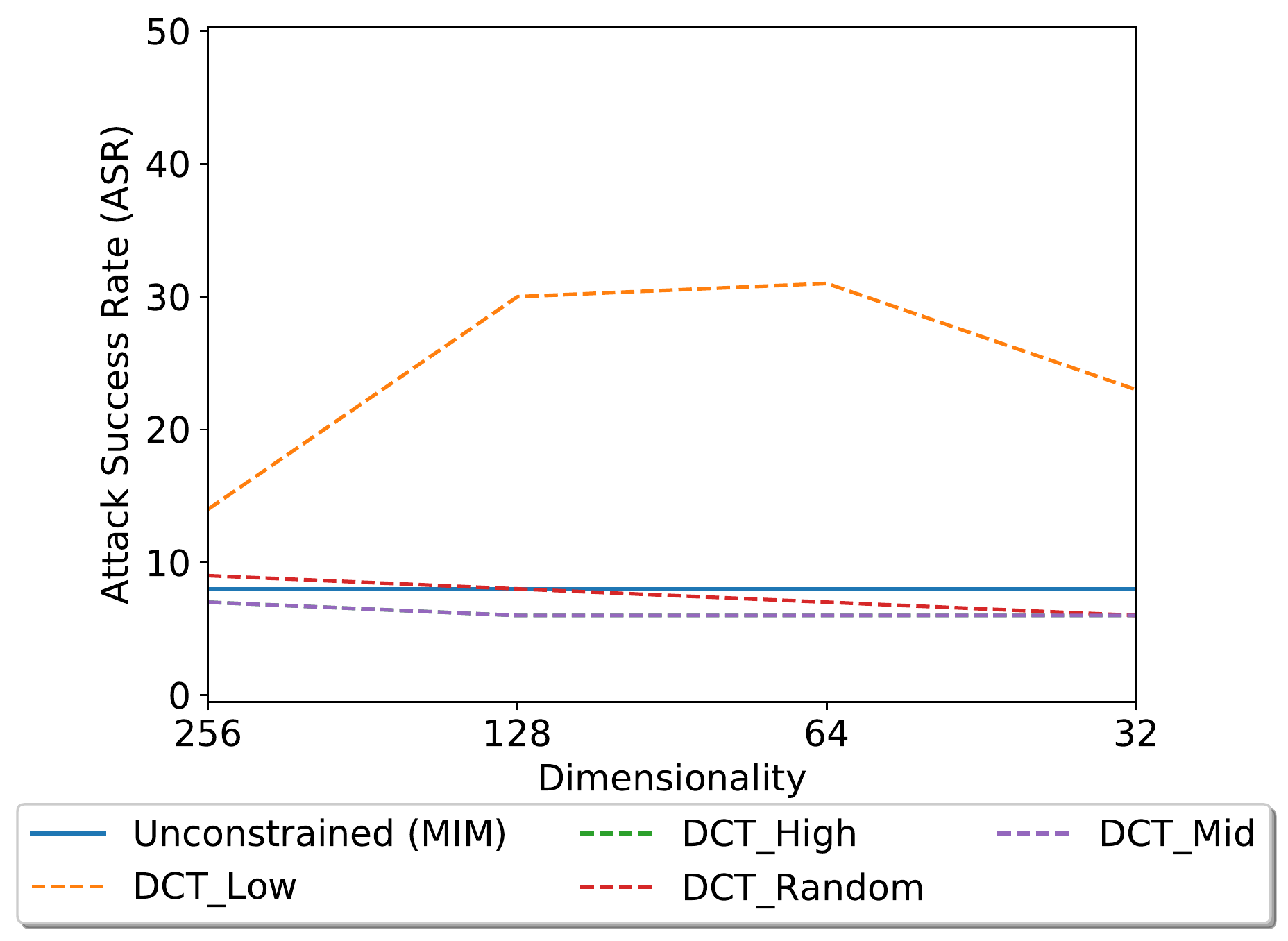}
    \caption{Non-targeted with $\epsilon=16$ and $\text{iterations}=1$.}
    \label{greybox13}
\end{subfigure}
~~~
\begin{subfigure}{0.3\linewidth}
    \centering
    \includegraphics[width=\textwidth]{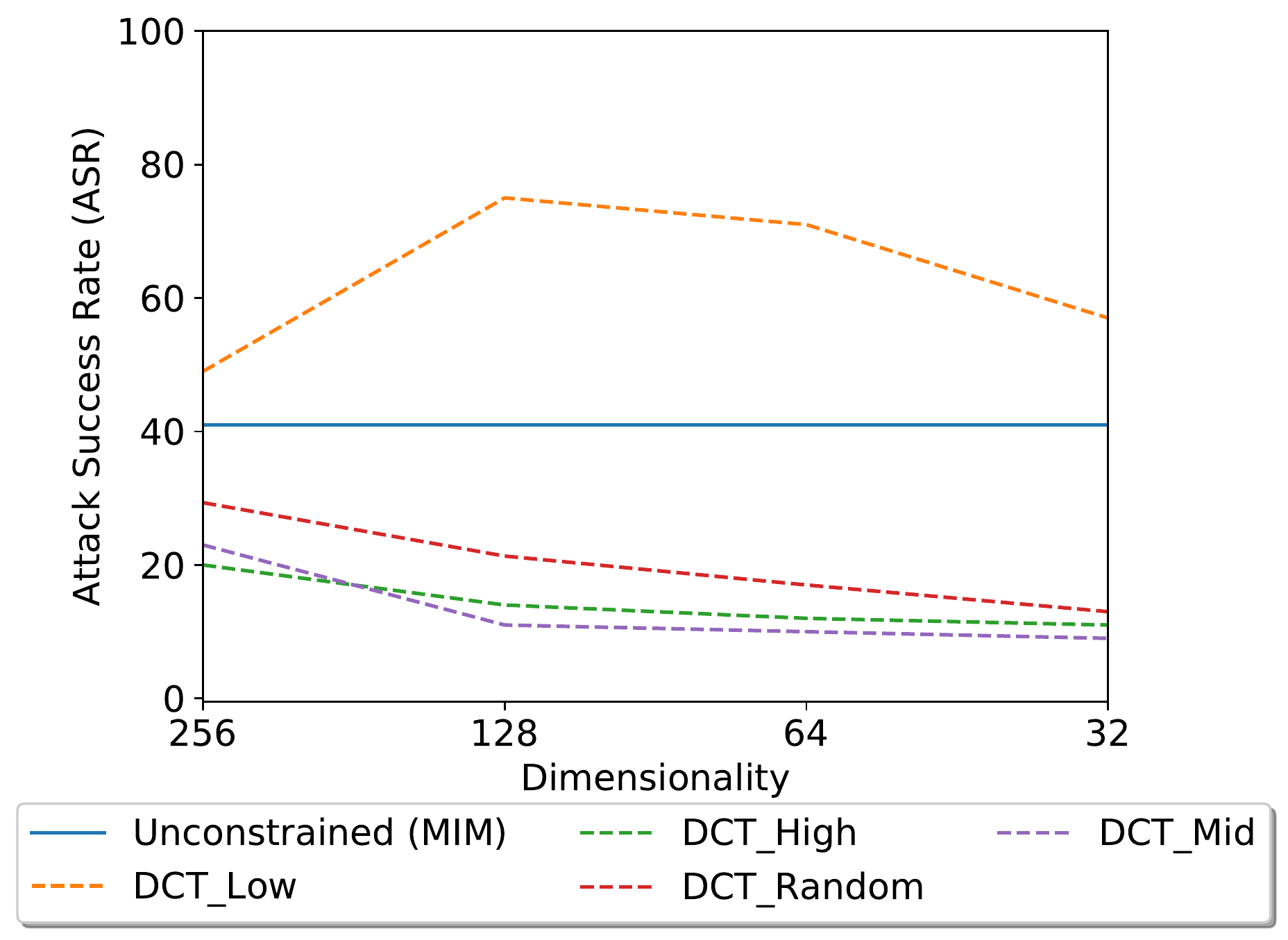}
    \caption{Non-targeted with $\epsilon=16$ and $\text{iterations}=10$.}
    \label{greybox23}
\end{subfigure}
~~~
\begin{subfigure}{0.3\linewidth}
    \centering
    \includegraphics[width=\textwidth]{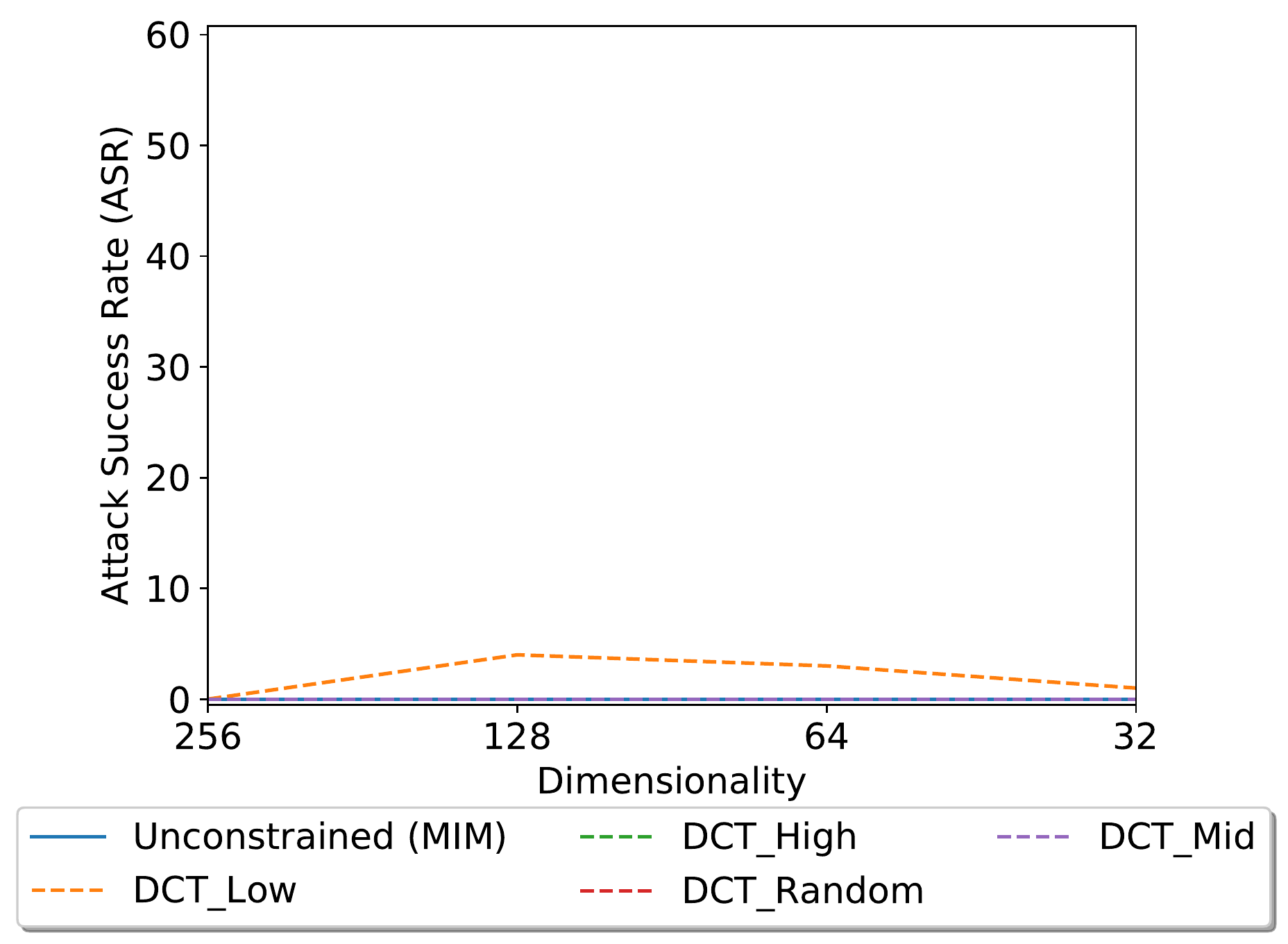}
    \caption{Targeted with $\epsilon=32$ and $\text{iterations}=10$.}
    \label{greybox33}
\end{subfigure}
\caption{\textbf{Grey-box} attack on D3.}
\label{greybox3}
\end{figure*}

\begin{figure*}
\begin{subfigure}{0.3\linewidth}
    \centering
    \includegraphics[width=\textwidth]{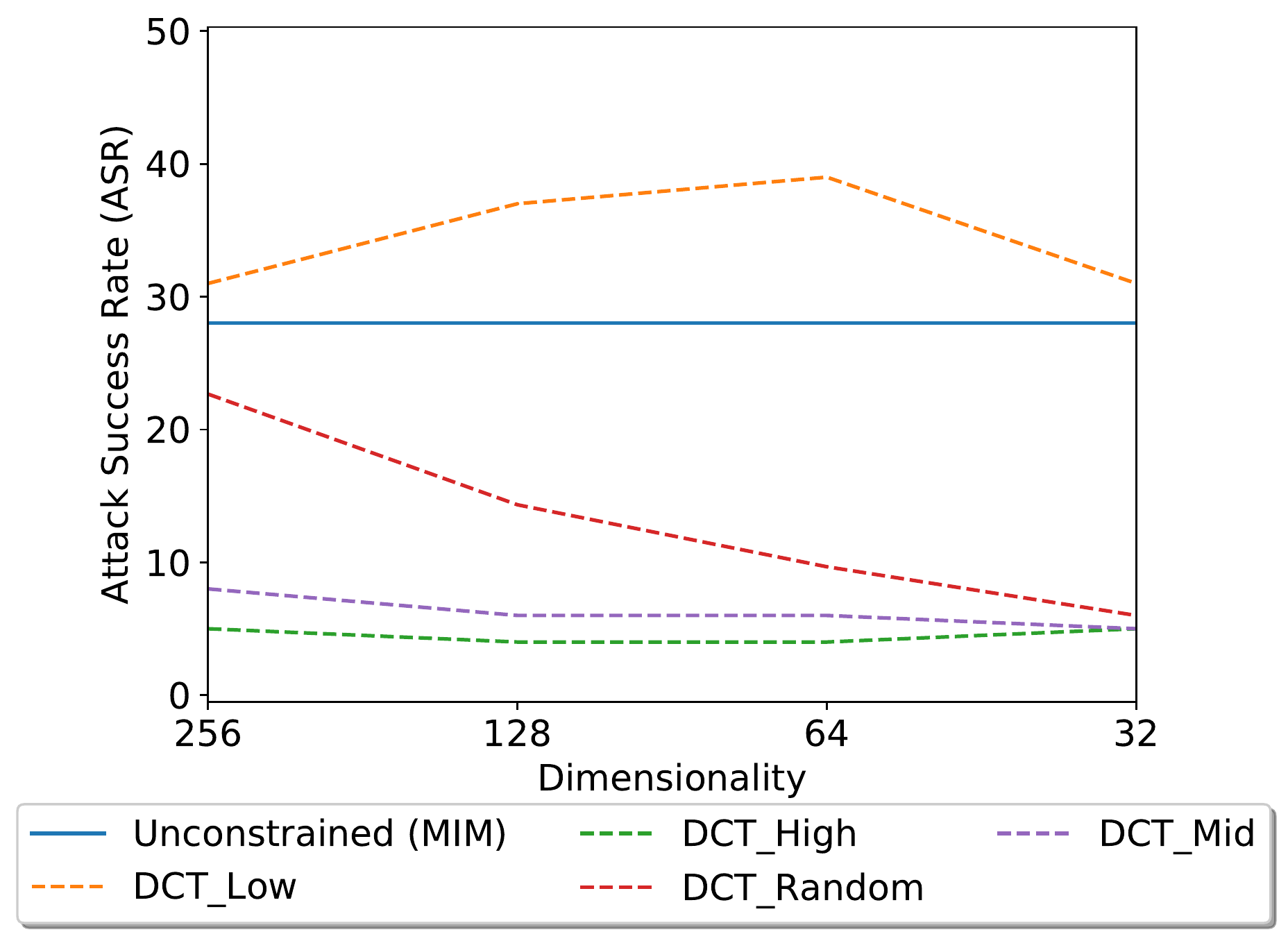}
    \caption{Non-targeted with $\epsilon=16$ and $\text{iterations}=1$.}
    \label{greybox14}
\end{subfigure}
~~~
\begin{subfigure}{0.3\linewidth}
    \centering
    \includegraphics[width=\textwidth]{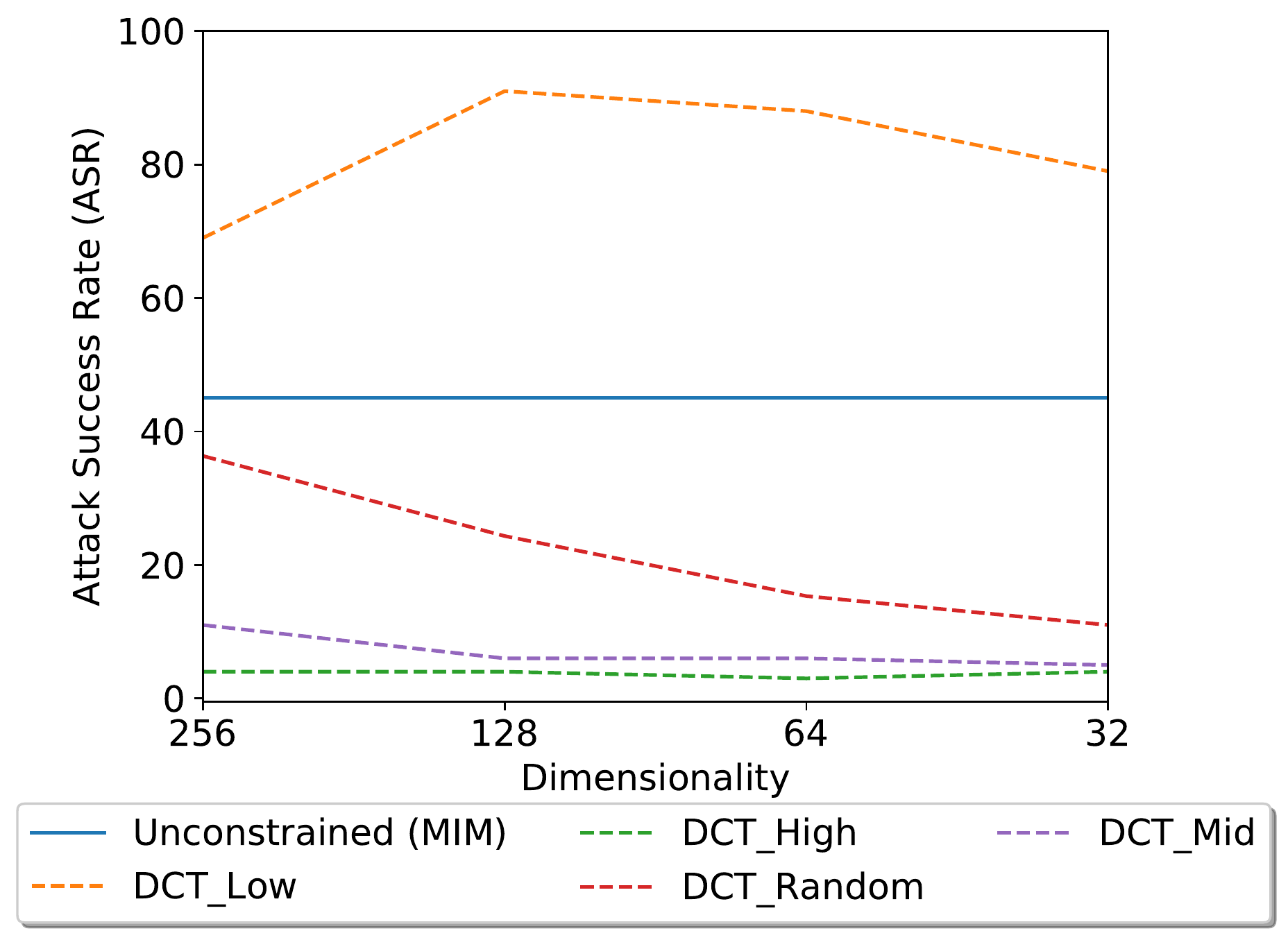}
    \caption{Non-targeted with $\epsilon=16$ and $\text{iterations}=10$.}
    \label{greybox24}
\end{subfigure}
~~~
\begin{subfigure}{0.3\linewidth}
    \centering
    \includegraphics[width=\textwidth]{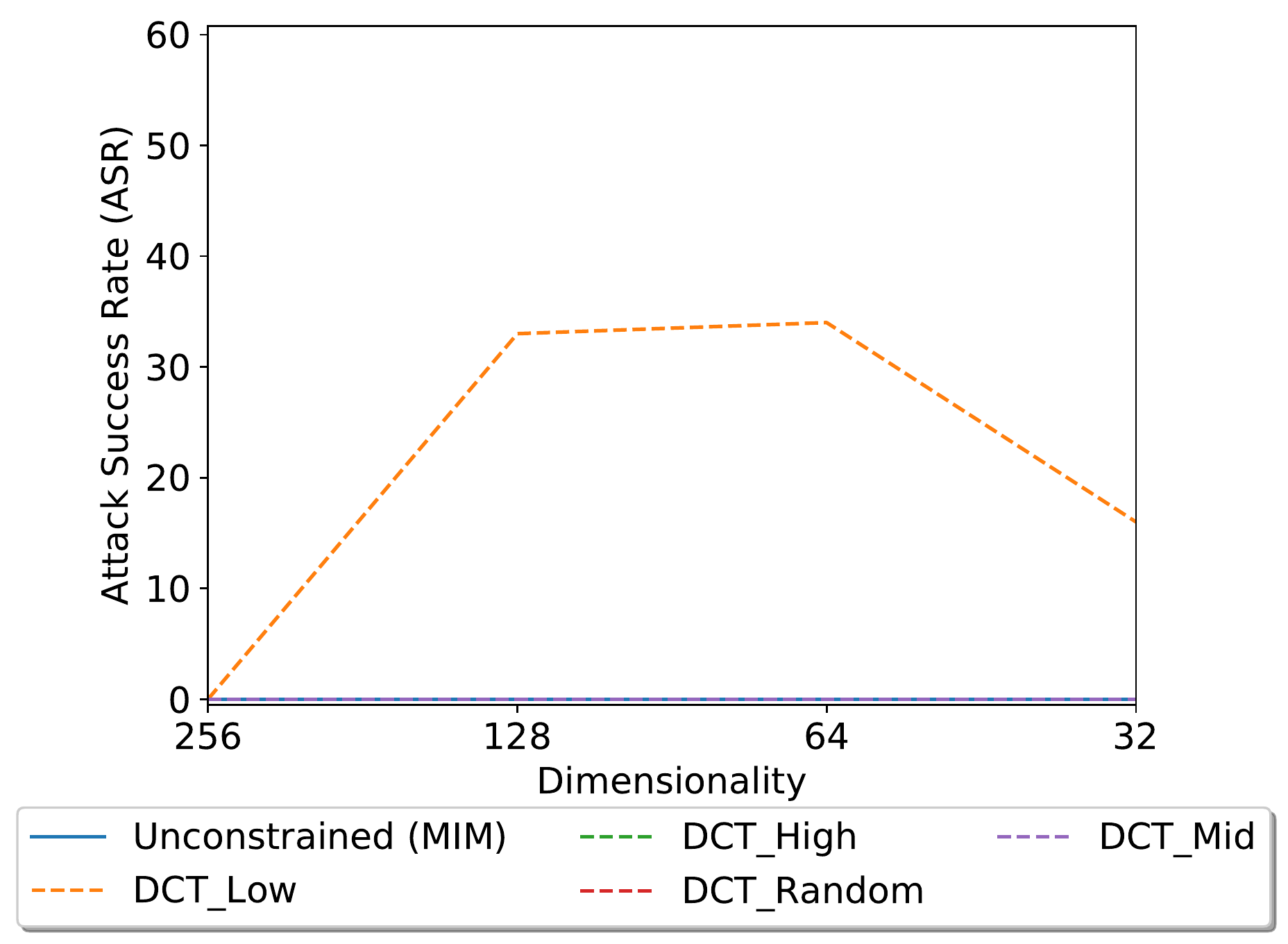}
    \caption{Targeted with $\epsilon=32$ and $\text{iterations}=10$.}
    \label{greybox34}
\end{subfigure}
\caption{\textbf{Grey-box} attack on D4.}
\label{greybox4}
\end{figure*}

\begin{figure*}
\begin{subfigure}{0.3\linewidth}
    \centering
    \includegraphics[width=\textwidth]{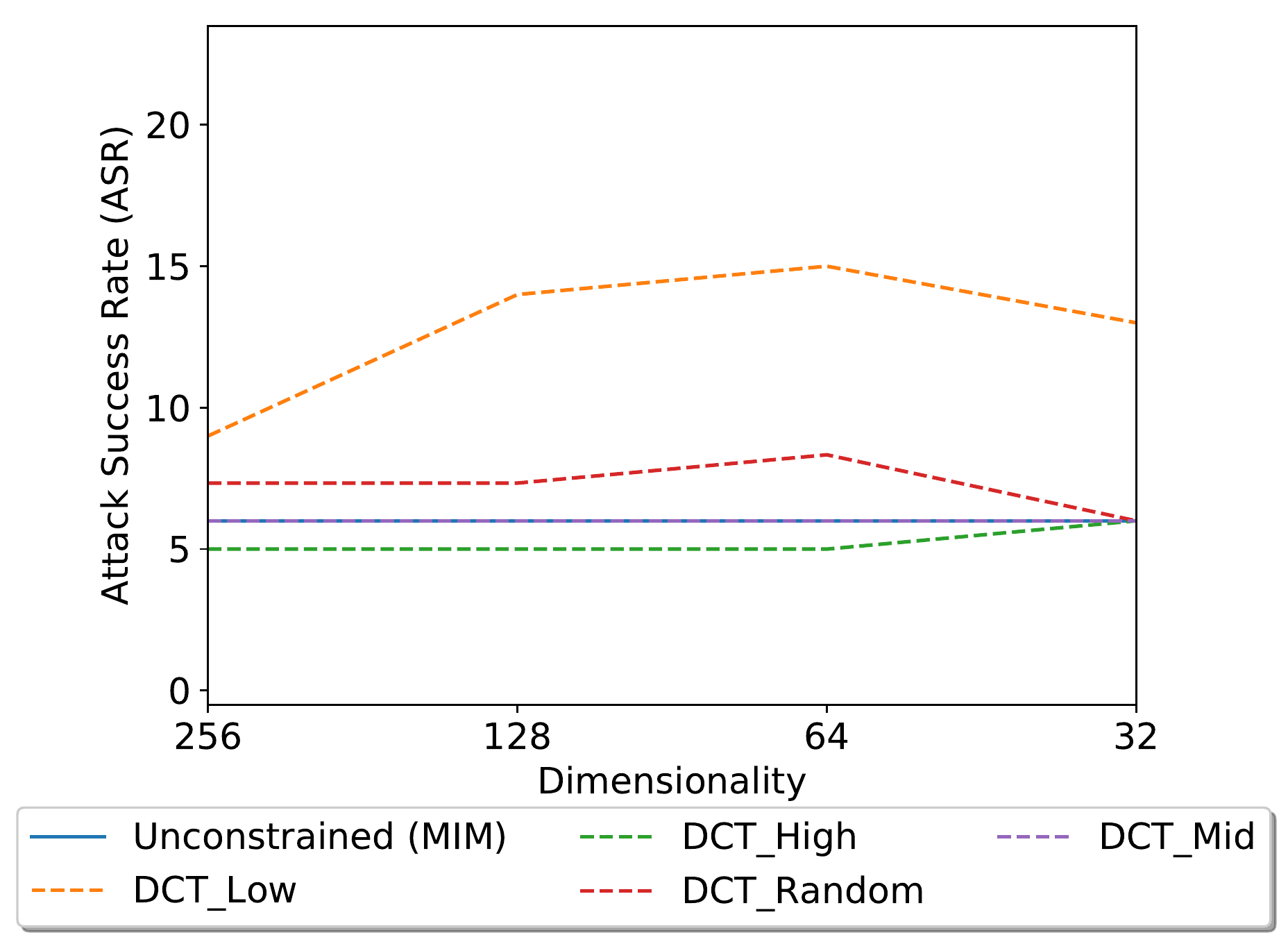}
    \caption{Non-targeted with $\epsilon=16$ and $\text{iterations}=1$.}
    \label{greybox14}
\end{subfigure}
~~~
\begin{subfigure}{0.3\linewidth}
    \centering
    \includegraphics[width=\textwidth]{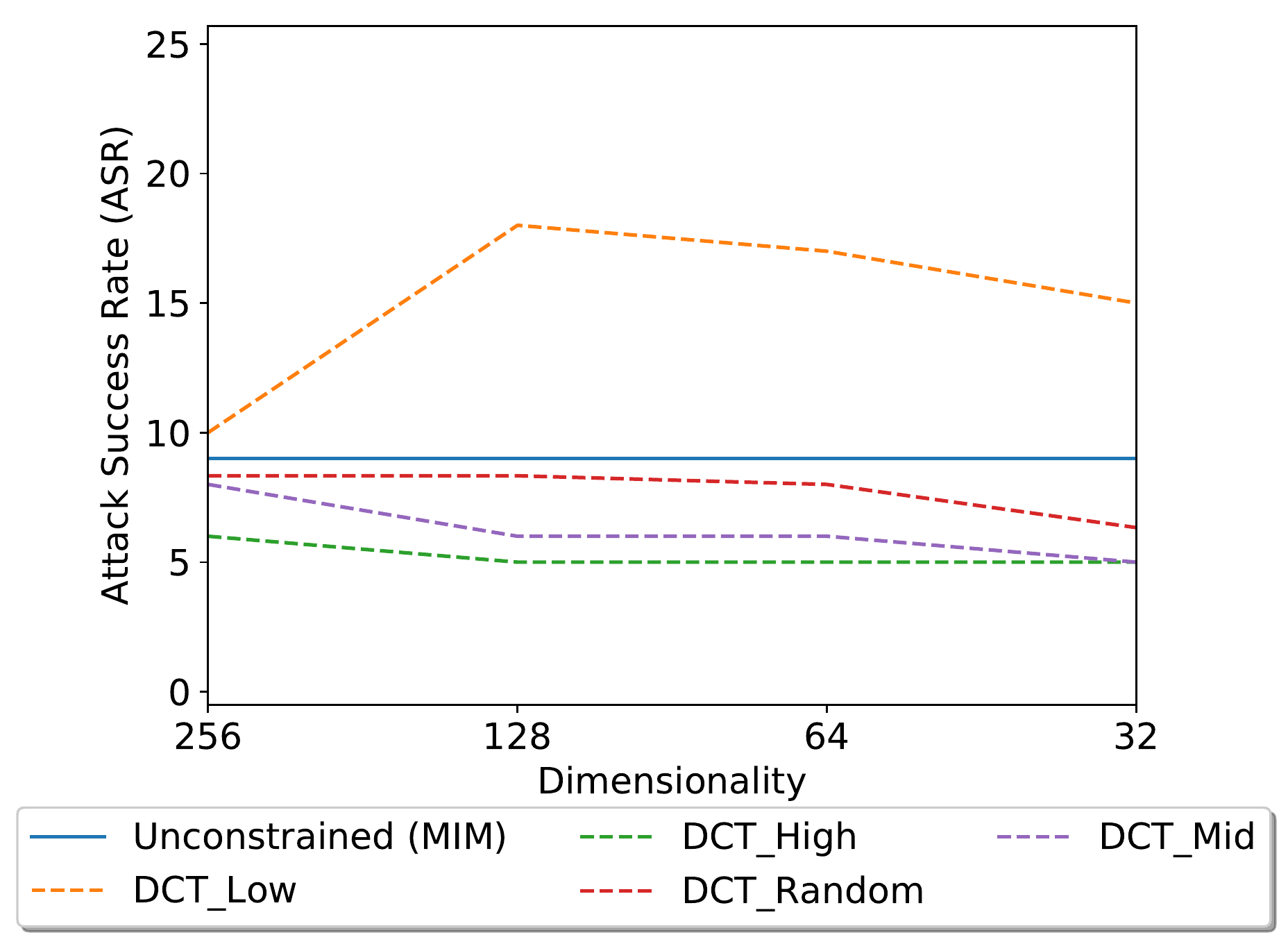}
    \caption{Non-targeted with $\epsilon=16$ and $\text{iterations}=10$.}
    \label{greybox24}
\end{subfigure}
~~~
\begin{subfigure}{0.3\linewidth}
    \centering
    \includegraphics[width=\textwidth]{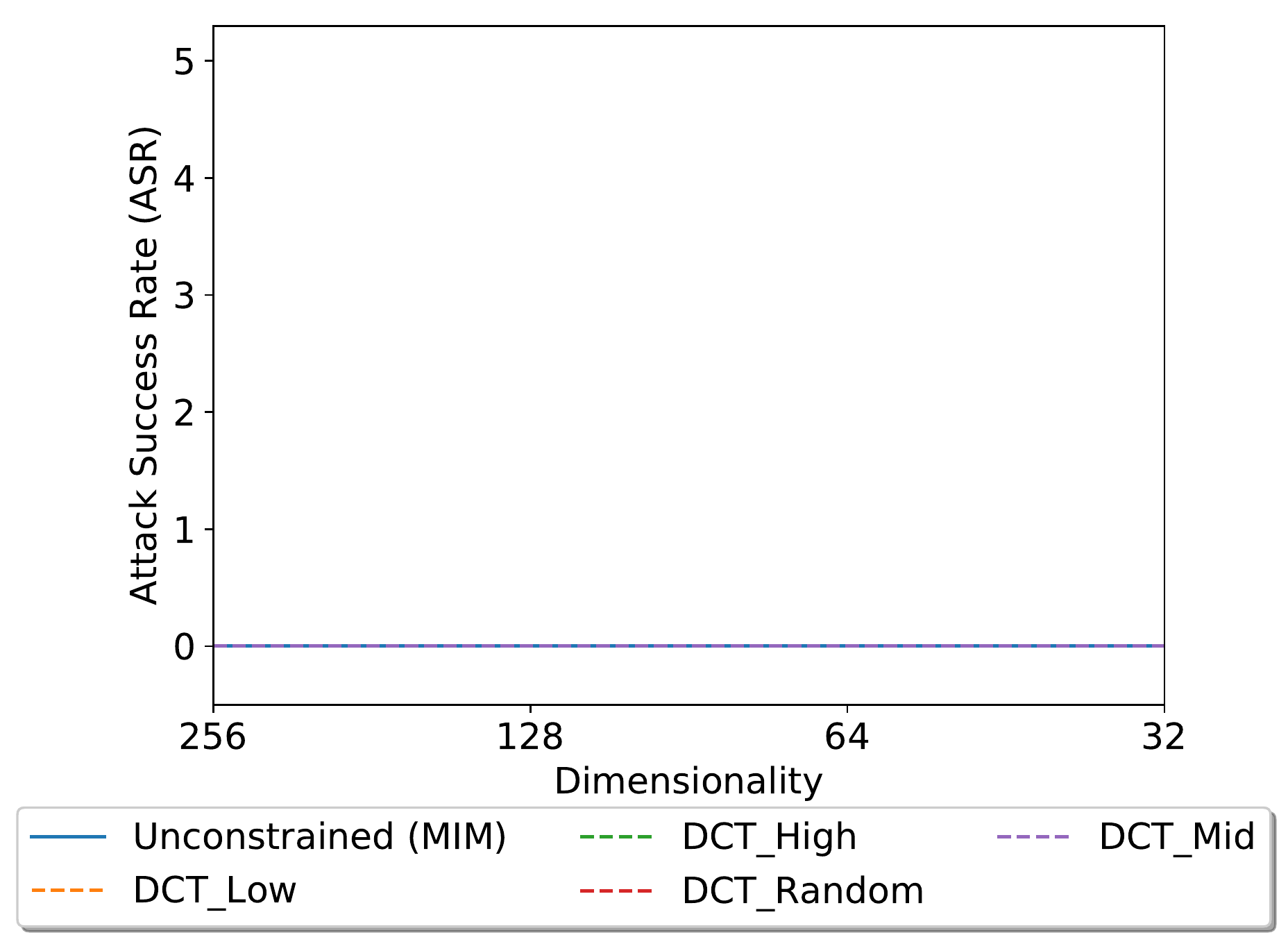}
    \caption{Targeted with $\epsilon=32$ and $\text{iterations}=10$.}
    \label{greybox34}
\end{subfigure}
\caption{\textbf{Black-box} attack from Cln\_1 to EnsAdv.}
\label{blackbox11}
\end{figure*}

\begin{figure*}
\begin{subfigure}{0.3\linewidth}
    \centering
    \includegraphics[width=\textwidth]{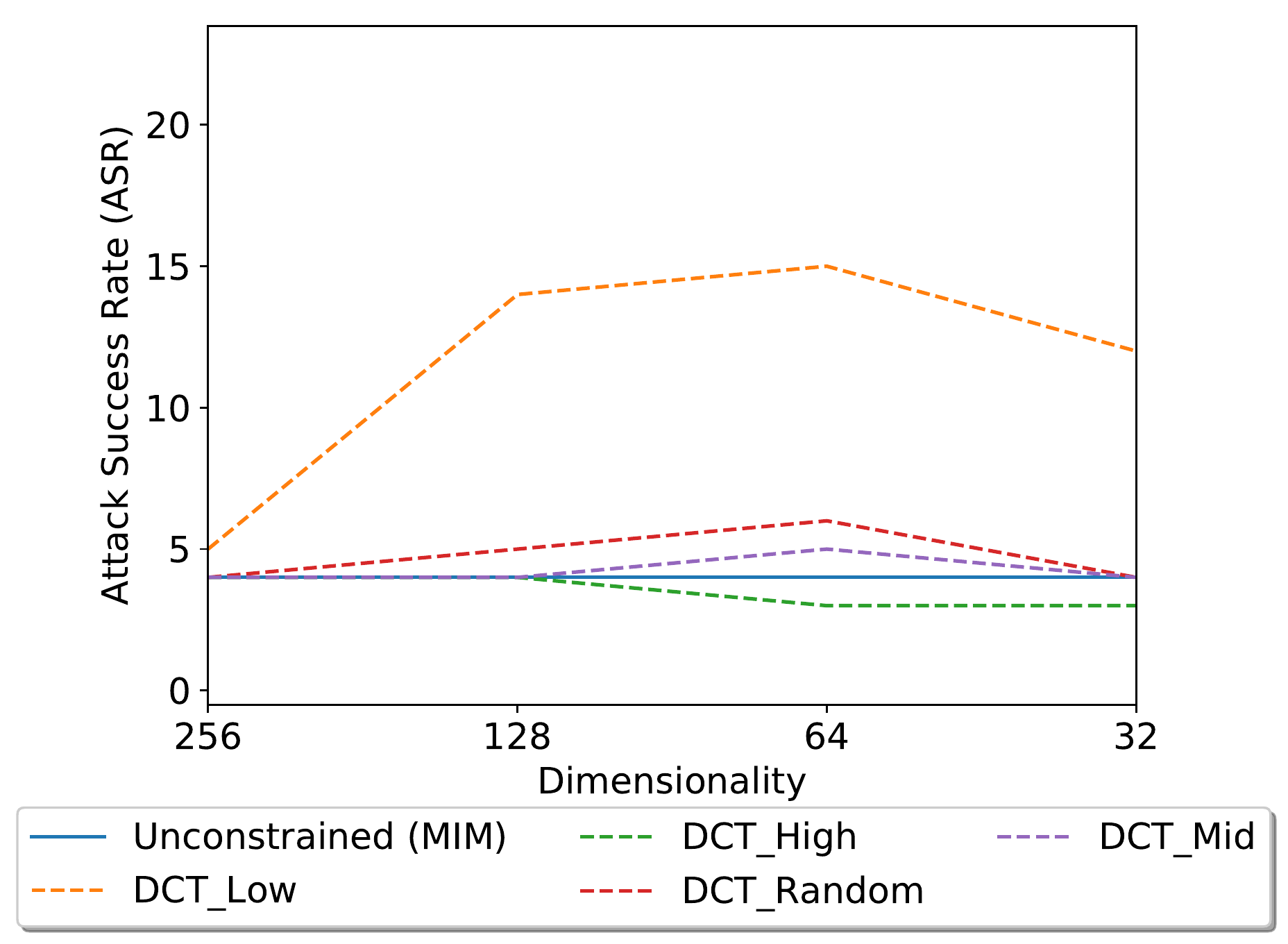}
    \caption{Non-targeted with $\epsilon=16$ and $\text{iterations}=1$.}
    \label{greybox14}
\end{subfigure}
~~~
\begin{subfigure}{0.3\linewidth}
    \centering
    \includegraphics[width=\textwidth]{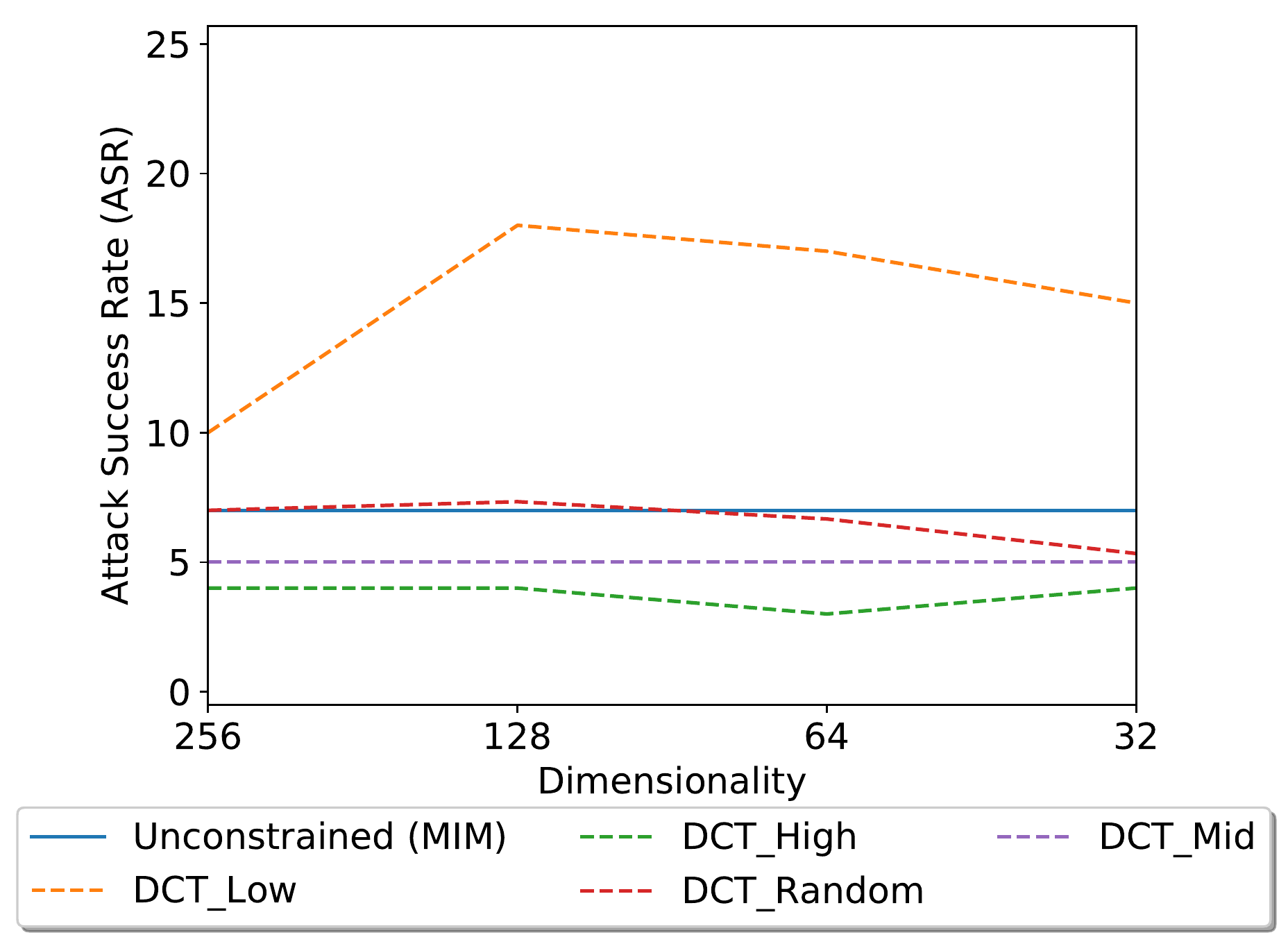}
    \caption{Non-targeted with $\epsilon=16$ and $\text{iterations}=10$.}
    \label{greybox24}
\end{subfigure}
~~~
\begin{subfigure}{0.3\linewidth}
    \centering
    \includegraphics[width=\textwidth]{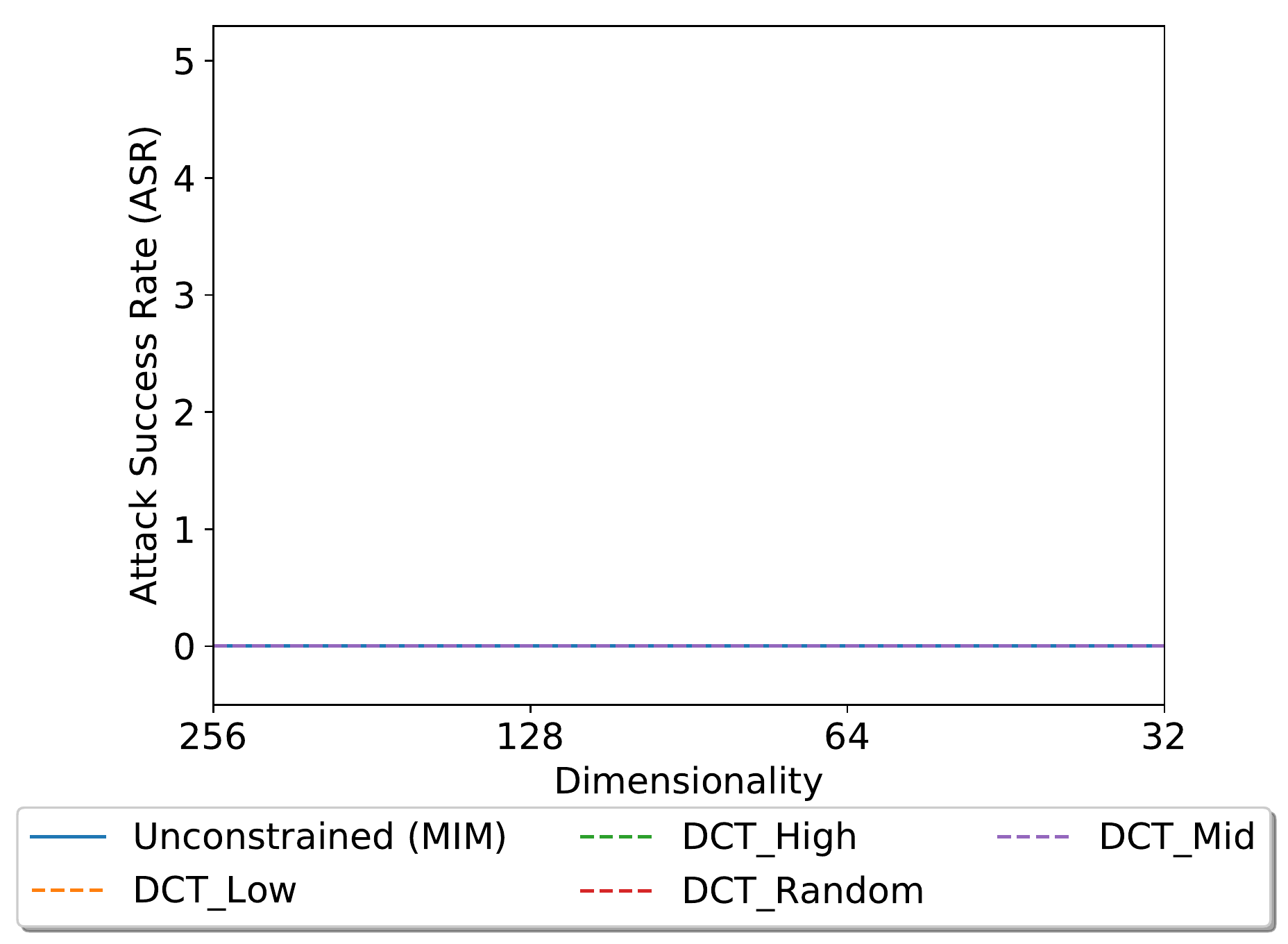}
    \caption{Targeted with $\epsilon=32$ and $\text{iterations}=10$.}
    \label{greybox34}
\end{subfigure}
\caption{\textbf{Black-box} attack from Cln\_1 to D1.}
\label{blackbox12}
\end{figure*}

\begin{figure*}
\begin{subfigure}{0.3\linewidth}
    \centering
    \includegraphics[width=\textwidth]{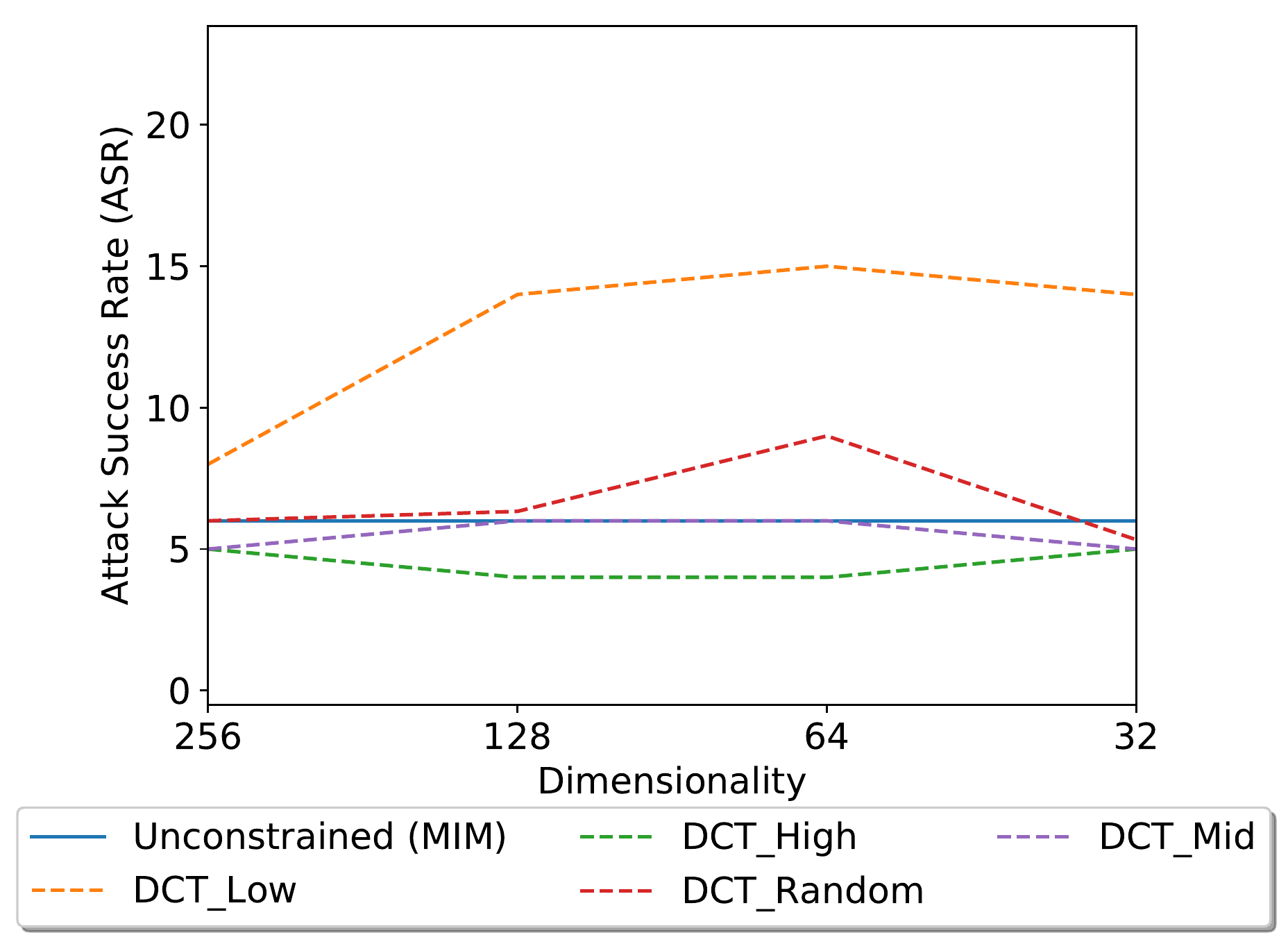}
    \caption{Non-targeted with $\epsilon=16$ and $\text{iterations}=1$.}
    \label{greybox14}
\end{subfigure}
~~~
\begin{subfigure}{0.3\linewidth}
    \centering
    \includegraphics[width=\textwidth]{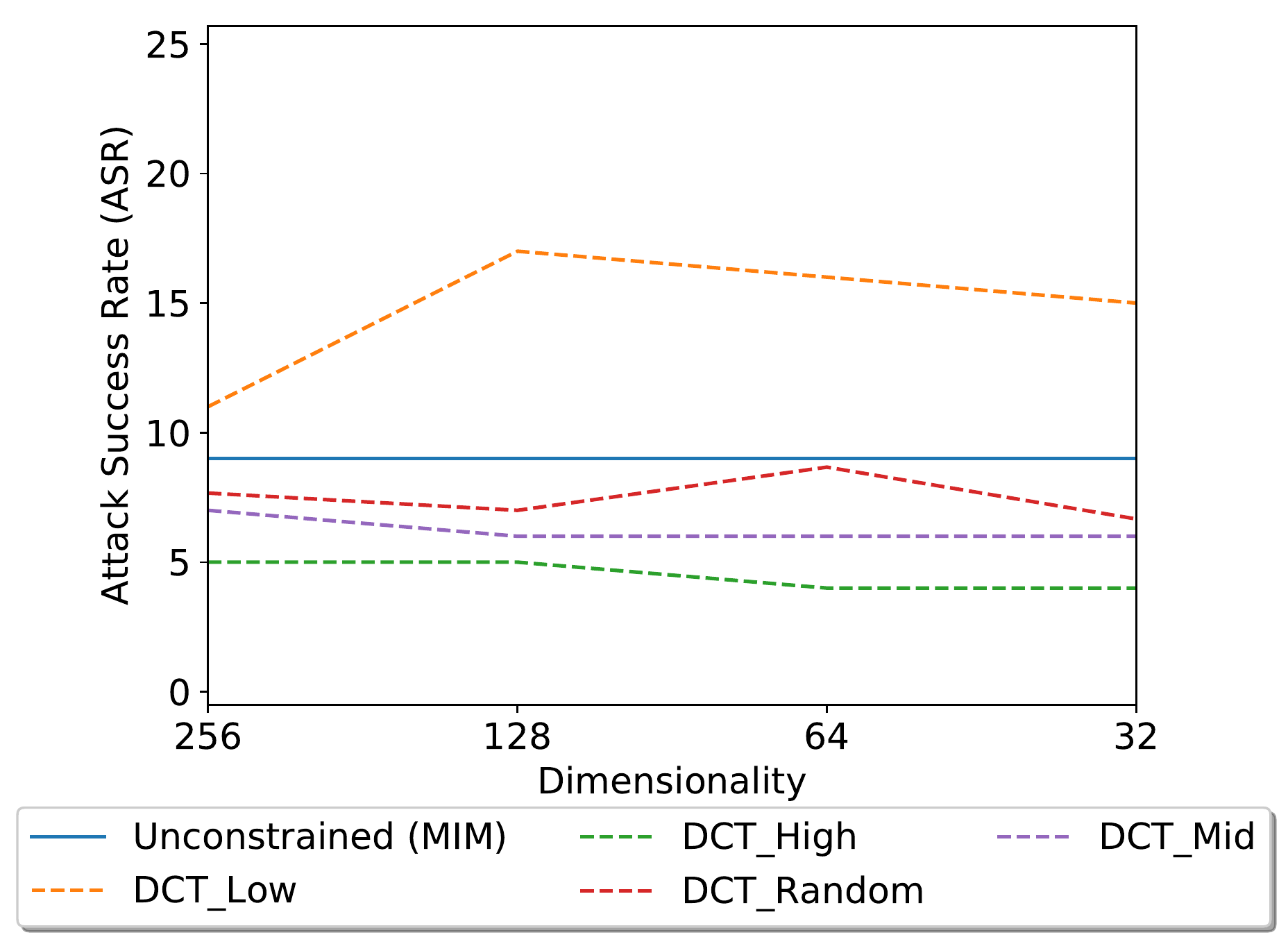}
    \caption{Non-targeted with $\epsilon=16$ and $\text{iterations}=10$.}
    \label{greybox24}
\end{subfigure}
~~~
\begin{subfigure}{0.3\linewidth}
    \centering
    \includegraphics[width=\textwidth]{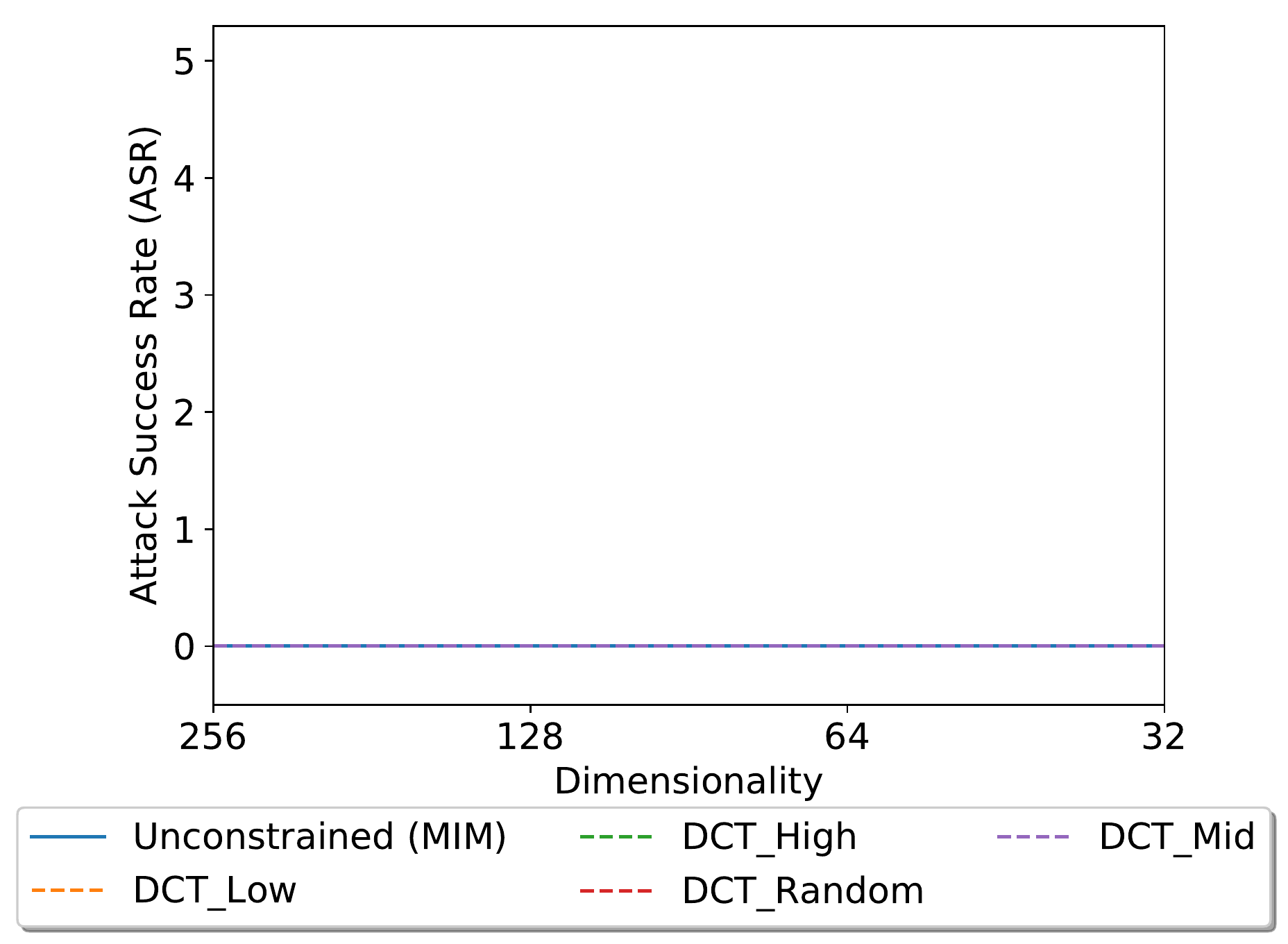}
    \caption{Targeted with $\epsilon=32$ and $\text{iterations}=10$.}
    \label{greybox34}
\end{subfigure}
\caption{\textbf{Black-box} attack from Cln\_1 to D2.}
\label{blackbox13}
\end{figure*}

\begin{figure*}
\begin{subfigure}{0.3\linewidth}
    \centering
    \includegraphics[width=\textwidth]{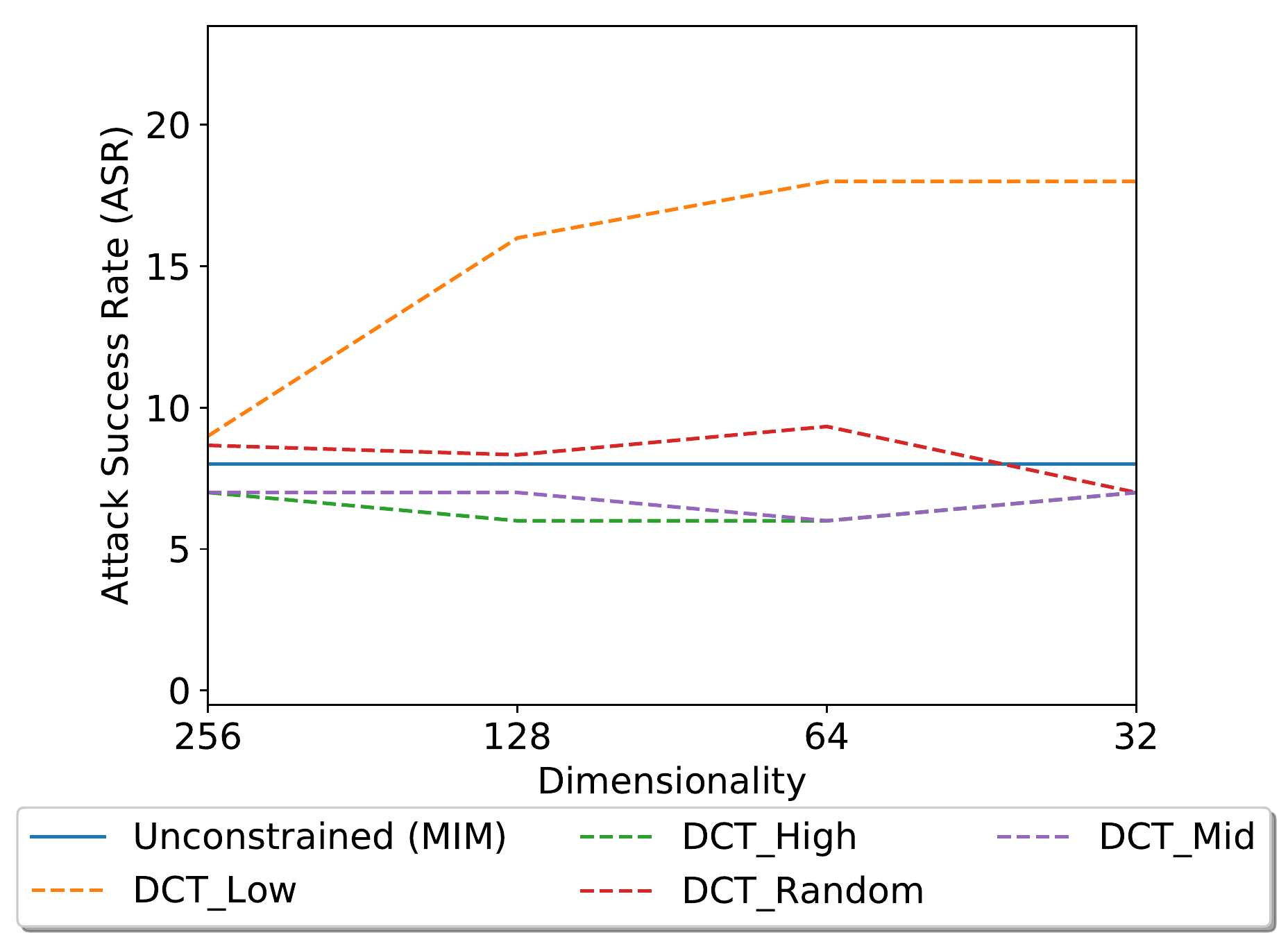}
    \caption{Non-targeted with $\epsilon=16$ and $\text{iterations}=1$.}
    \label{greybox14}
\end{subfigure}
~~~
\begin{subfigure}{0.3\linewidth}
    \centering
    \includegraphics[width=\textwidth]{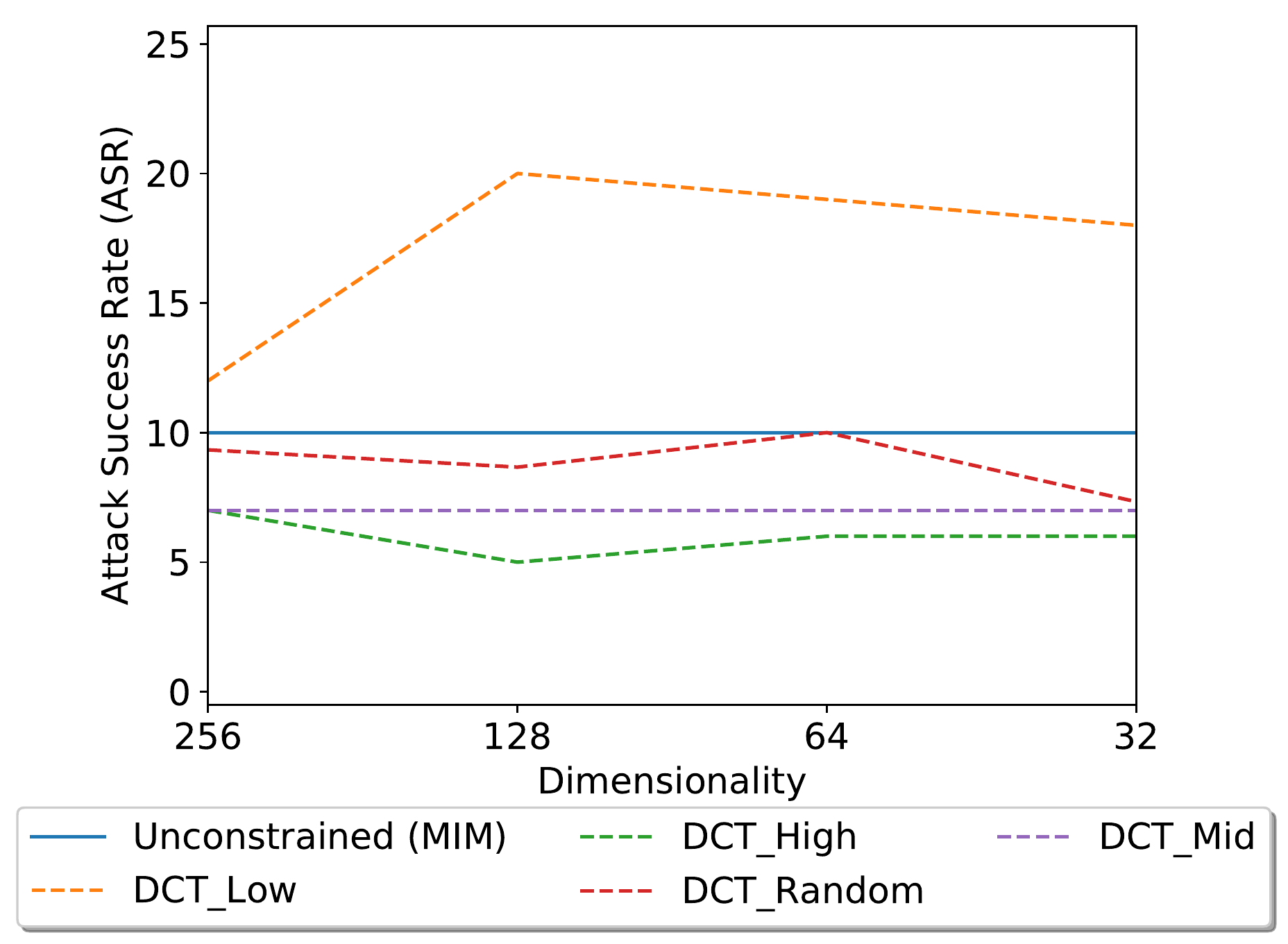}
    \caption{Non-targeted with $\epsilon=16$ and $\text{iterations}=10$.}
    \label{greybox24}
\end{subfigure}
~~~
\begin{subfigure}{0.3\linewidth}
    \centering
    \includegraphics[width=\textwidth]{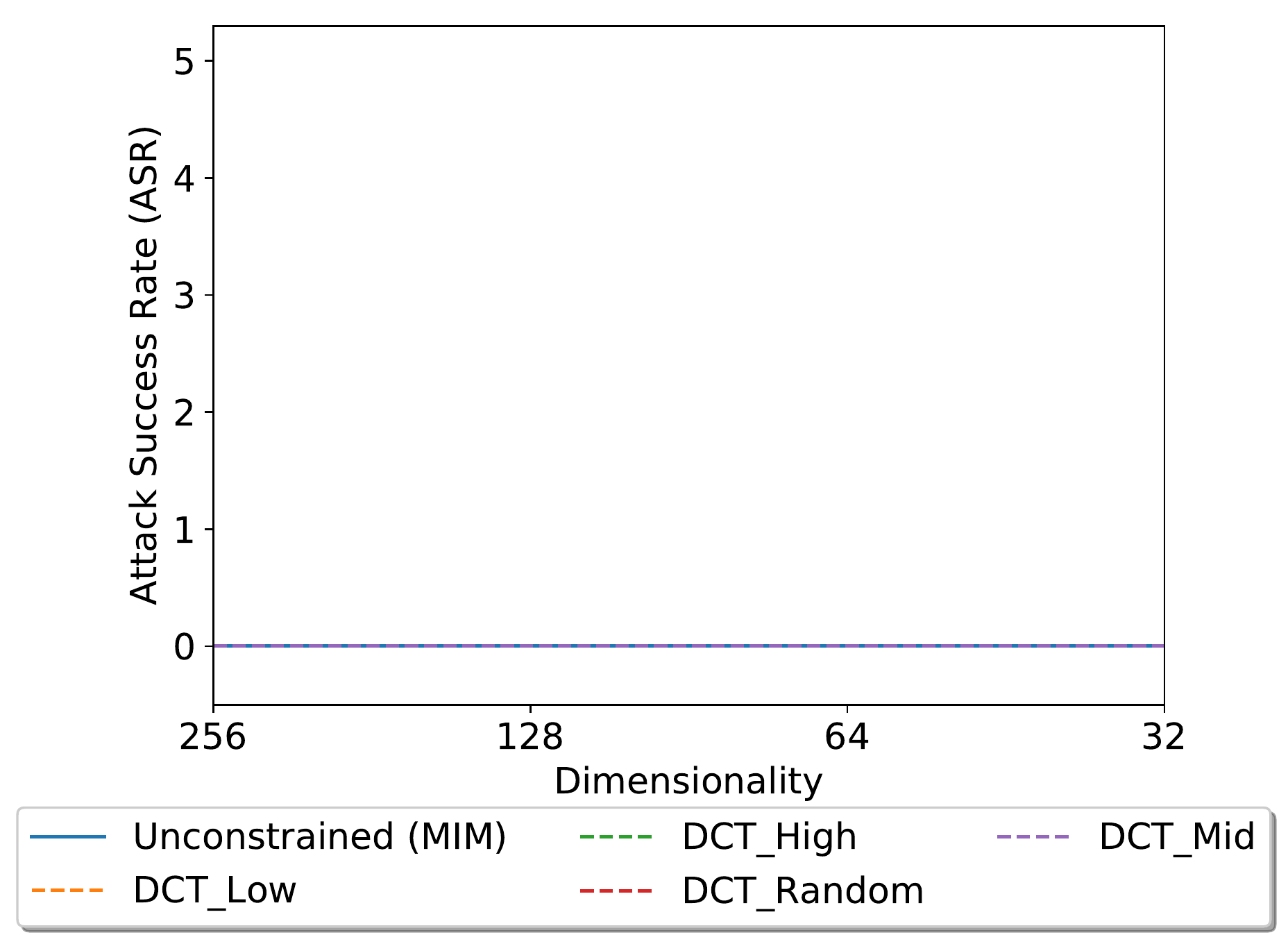}
    \caption{Targeted with $\epsilon=32$ and $\text{iterations}=10$.}
    \label{greybox34}
\end{subfigure}
\caption{\textbf{Black-box} attack from Cln\_1 to D3.}
\label{blackbox14}
\end{figure*}

\begin{figure*}
\begin{subfigure}{0.3\linewidth}
    \centering
    \includegraphics[width=\textwidth]{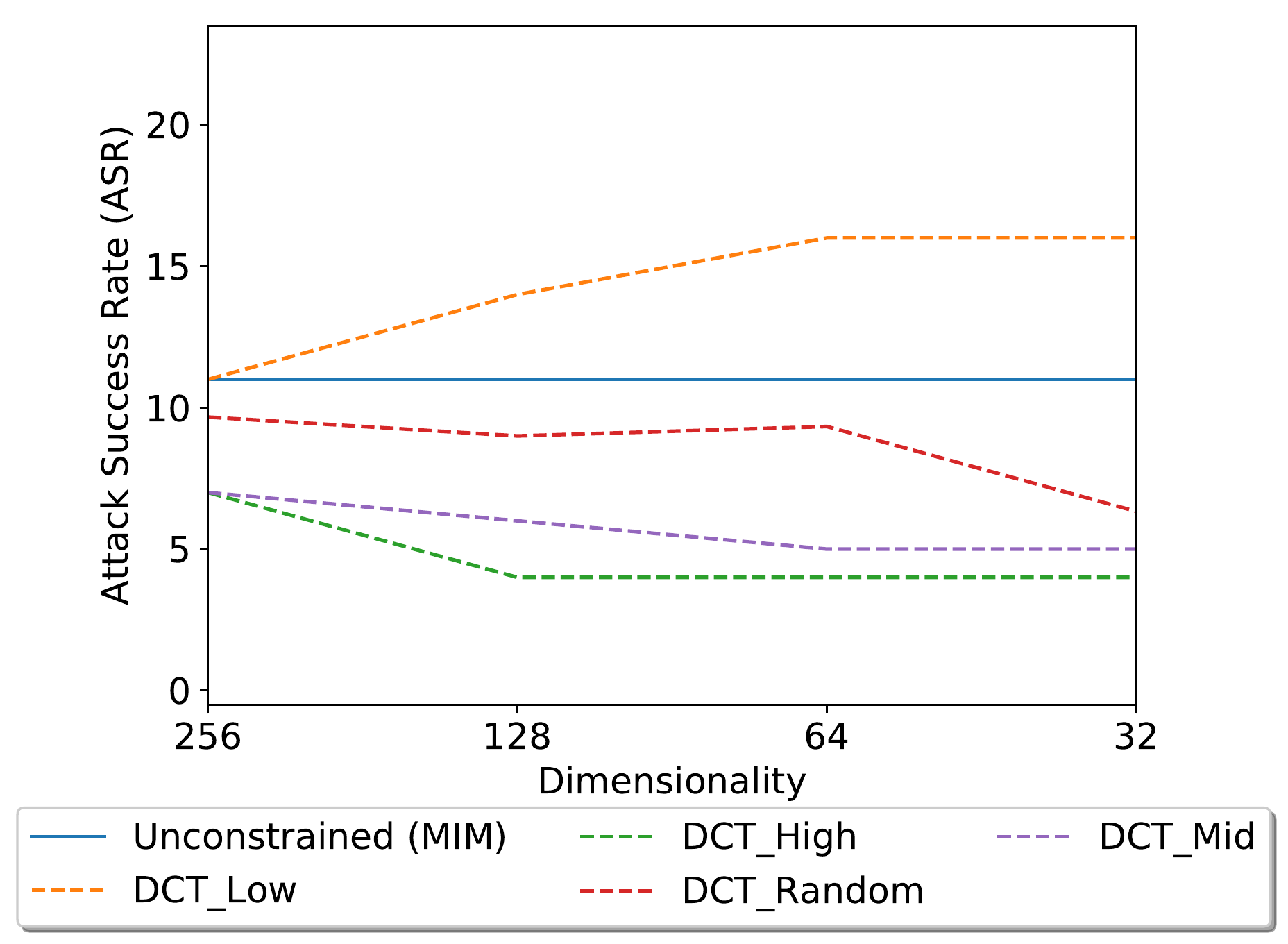}
    \caption{Non-targeted with $\epsilon=16$ and $\text{iterations}=1$.}
    \label{greybox14}
\end{subfigure}
~~~
\begin{subfigure}{0.3\linewidth}
    \centering
    \includegraphics[width=\textwidth]{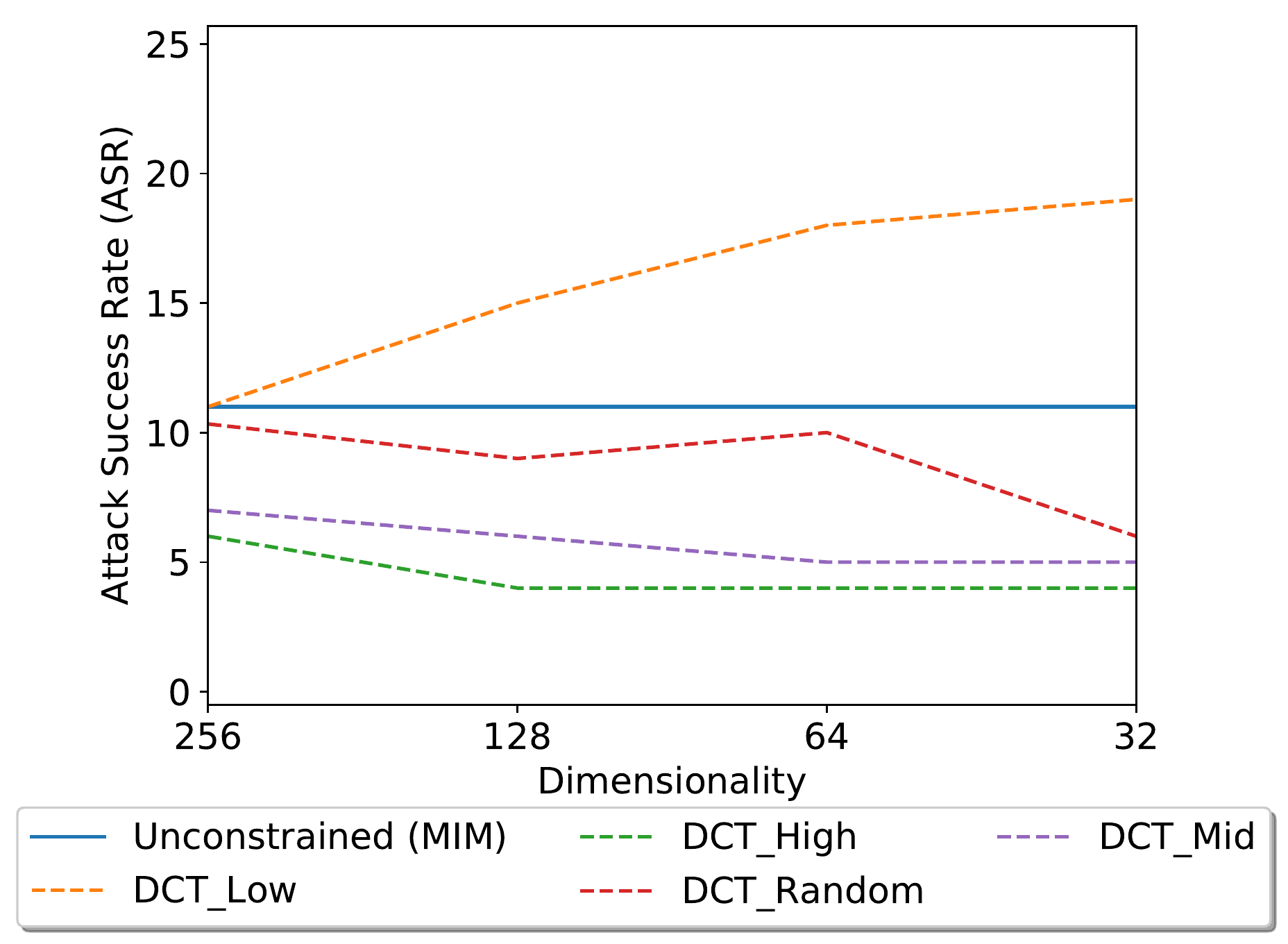}
    \caption{Non-targeted with $\epsilon=16$ and $\text{iterations}=10$.}
    \label{greybox24}
\end{subfigure}
~~~
\begin{subfigure}{0.3\linewidth}
    \centering
    \includegraphics[width=\textwidth]{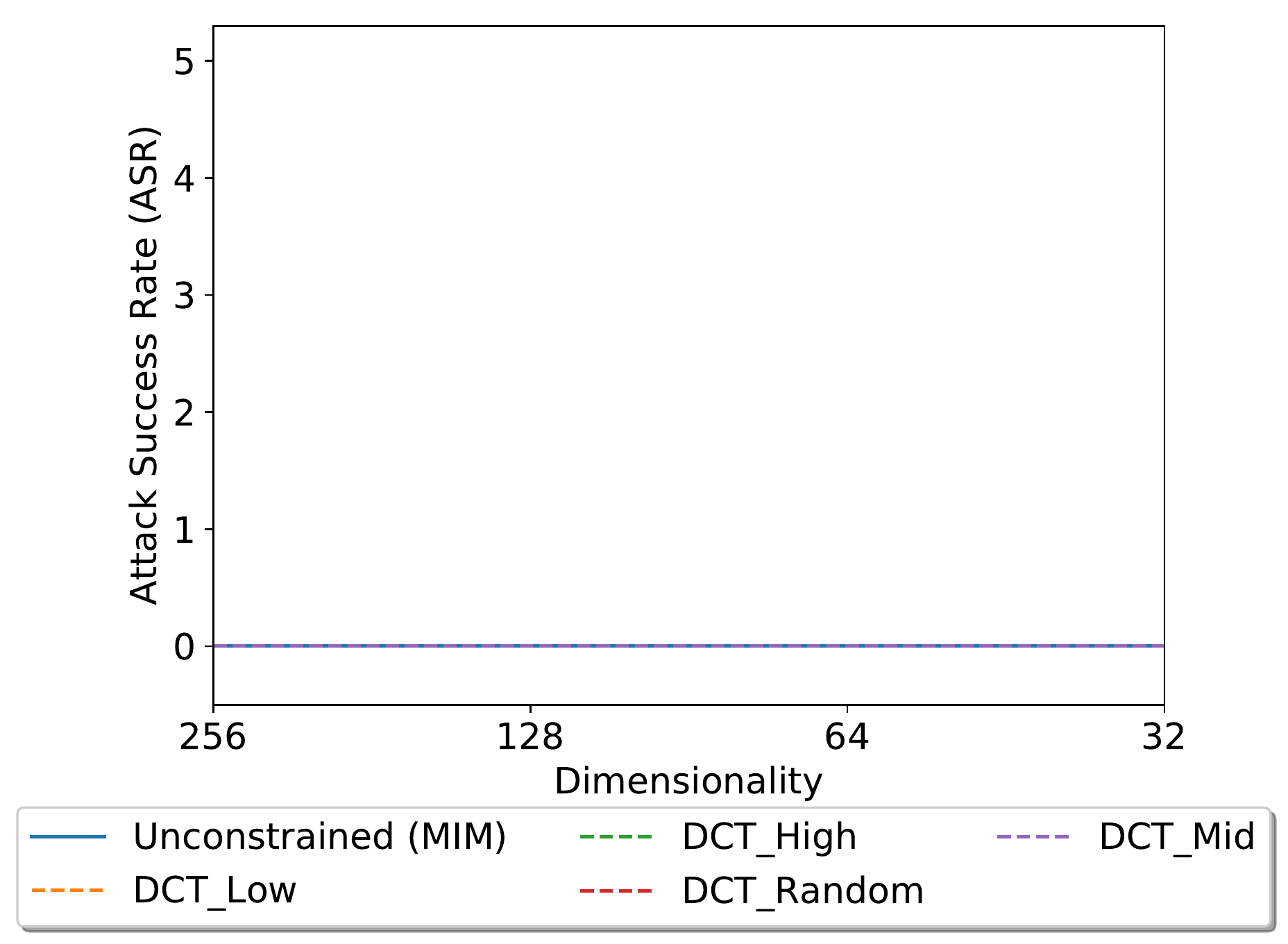}
    \caption{Targeted with $\epsilon=32$ and $\text{iterations}=10$.}
    \label{greybox34}
\end{subfigure}
\caption{\textbf{Black-box} attack from Cln\_1 to D4.}
\label{blackbox15}
\end{figure*}

\begin{figure*}
\begin{subfigure}{0.3\linewidth}
    \centering
    \includegraphics[width=\textwidth]{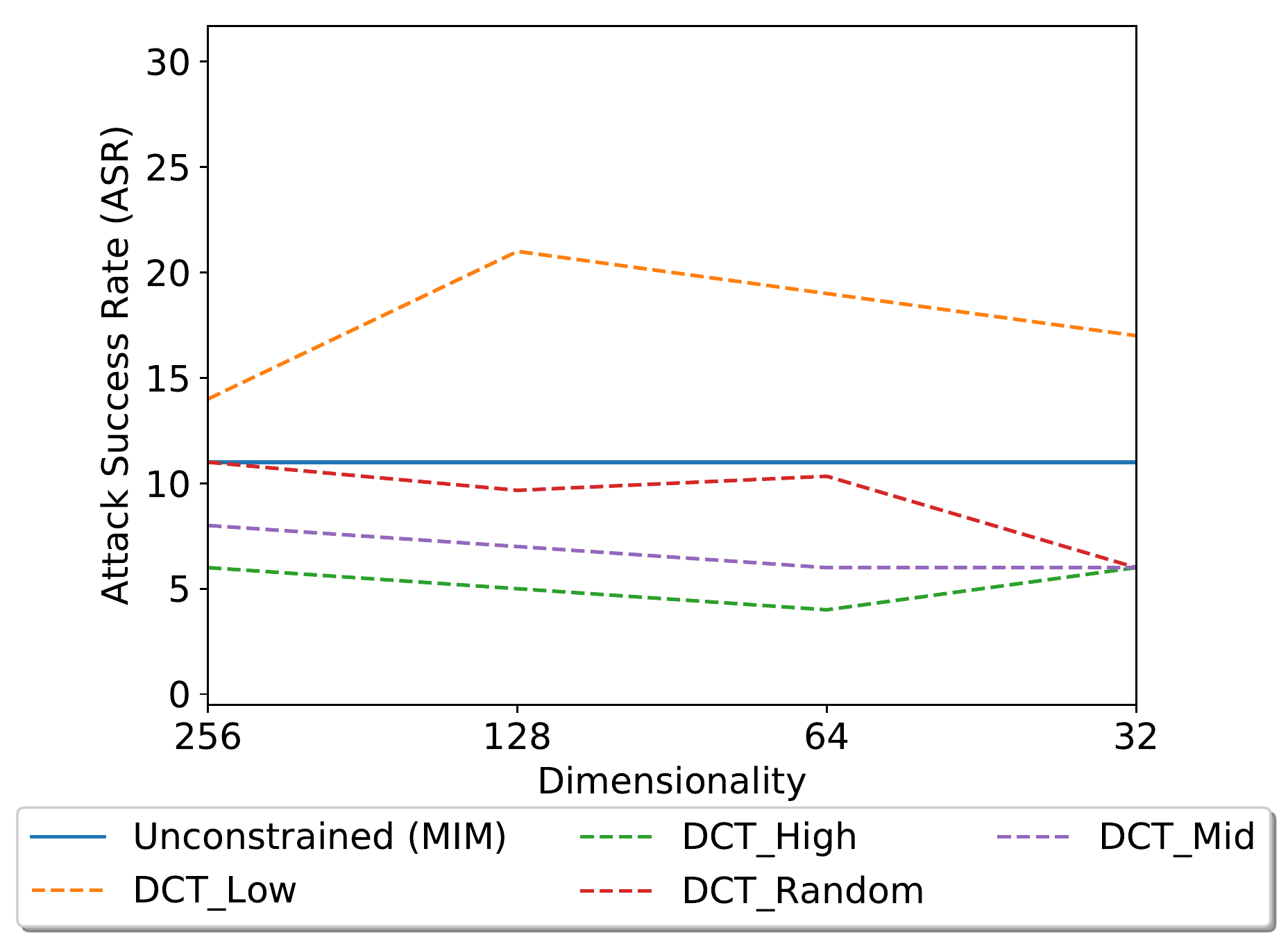}
    \caption{Non-targeted with $\epsilon=16$ and $\text{iterations}=1$.}
    \label{greybox14}
\end{subfigure}
~~~
\begin{subfigure}{0.3\linewidth}
    \centering
    \includegraphics[width=\textwidth]{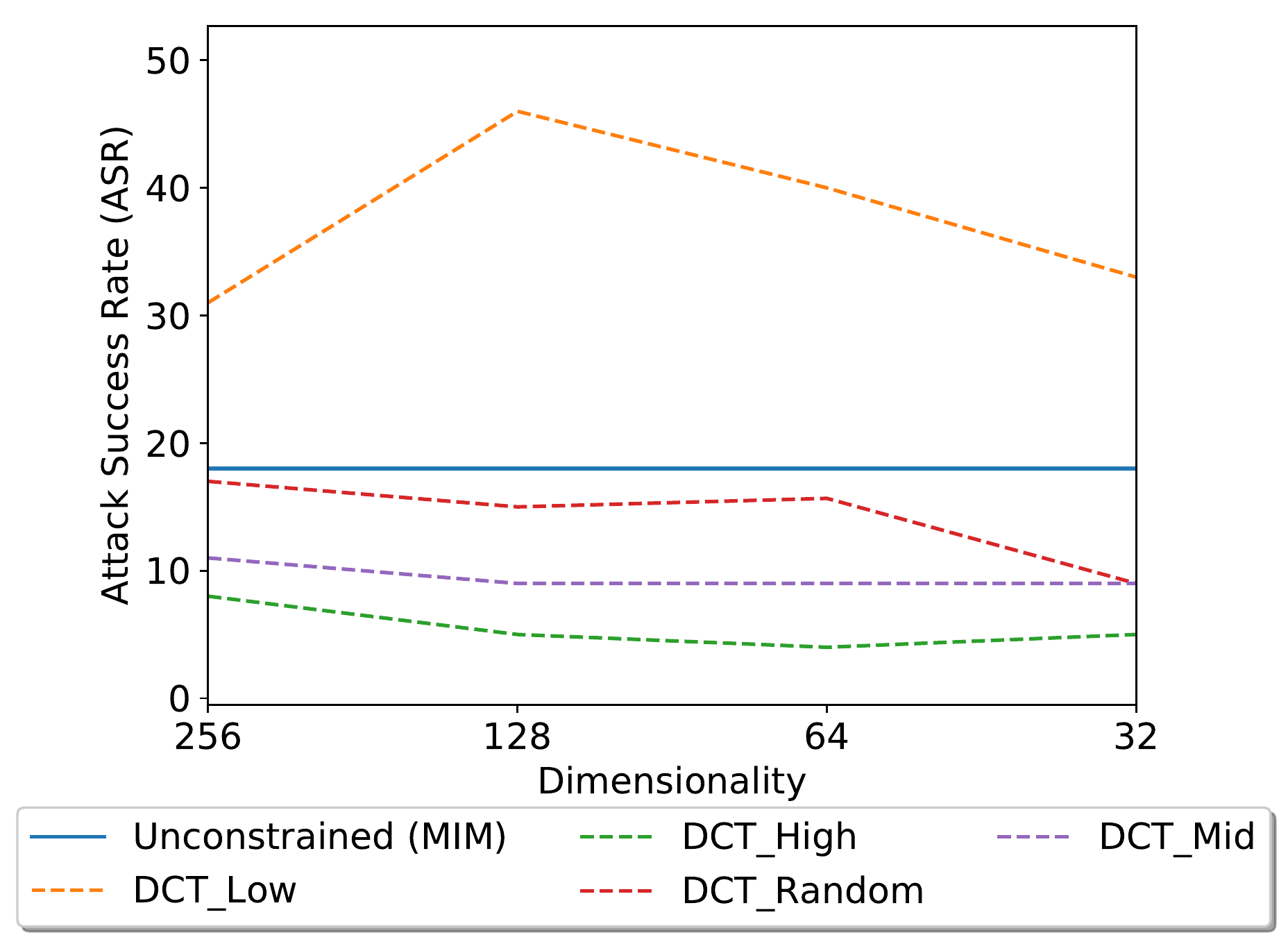}
    \caption{Non-targeted with $\epsilon=16$ and $\text{iterations}=10$.}
    \label{greybox24}
\end{subfigure}
~~~
\begin{subfigure}{0.3\linewidth}
    \centering
    \includegraphics[width=\textwidth]{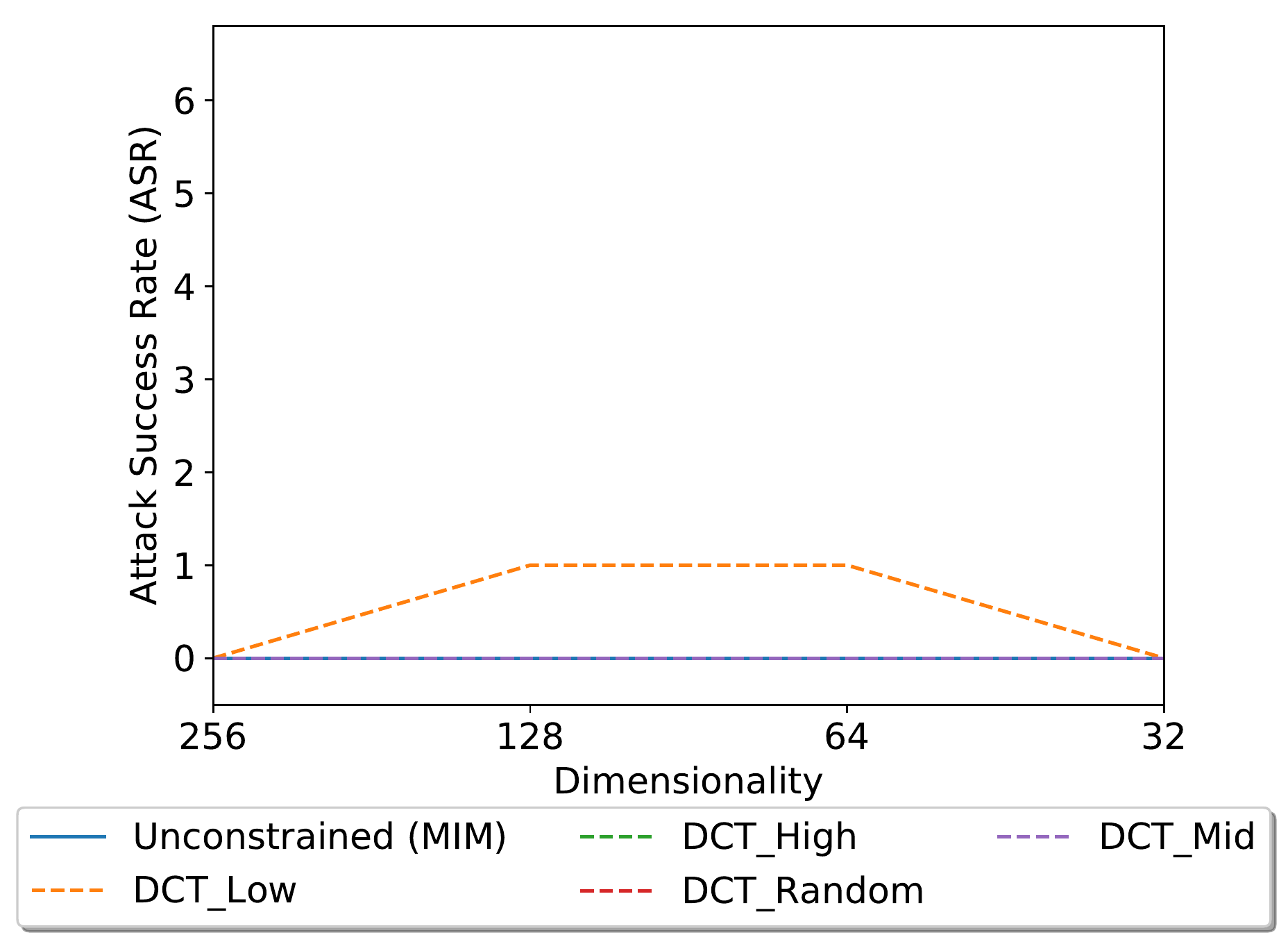}
    \caption{Targeted with $\epsilon=32$ and $\text{iterations}=10$.}
    \label{greybox34}
\end{subfigure}
\caption{\textbf{Black-box} attack from Cln\_3 to EnsAdv.}
\label{blackbox21}
\end{figure*}

\begin{figure*}
\begin{subfigure}{0.3\linewidth}
    \centering
    \includegraphics[width=\textwidth]{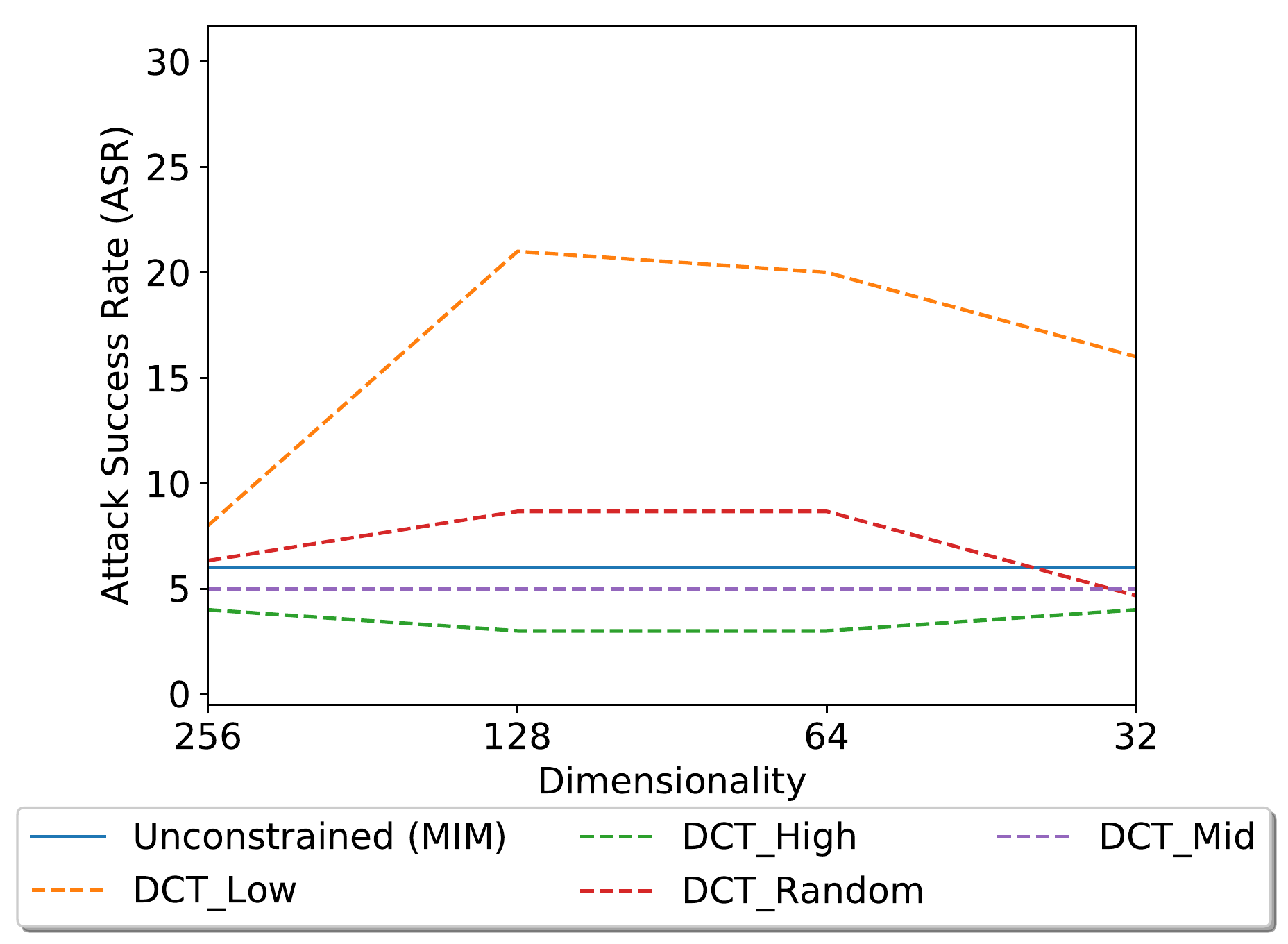}
    \caption{Non-targeted with $\epsilon=16$ and $\text{iterations}=1$.}
    \label{greybox14}
\end{subfigure}
~~~
\begin{subfigure}{0.3\linewidth}
    \centering
    \includegraphics[width=\textwidth]{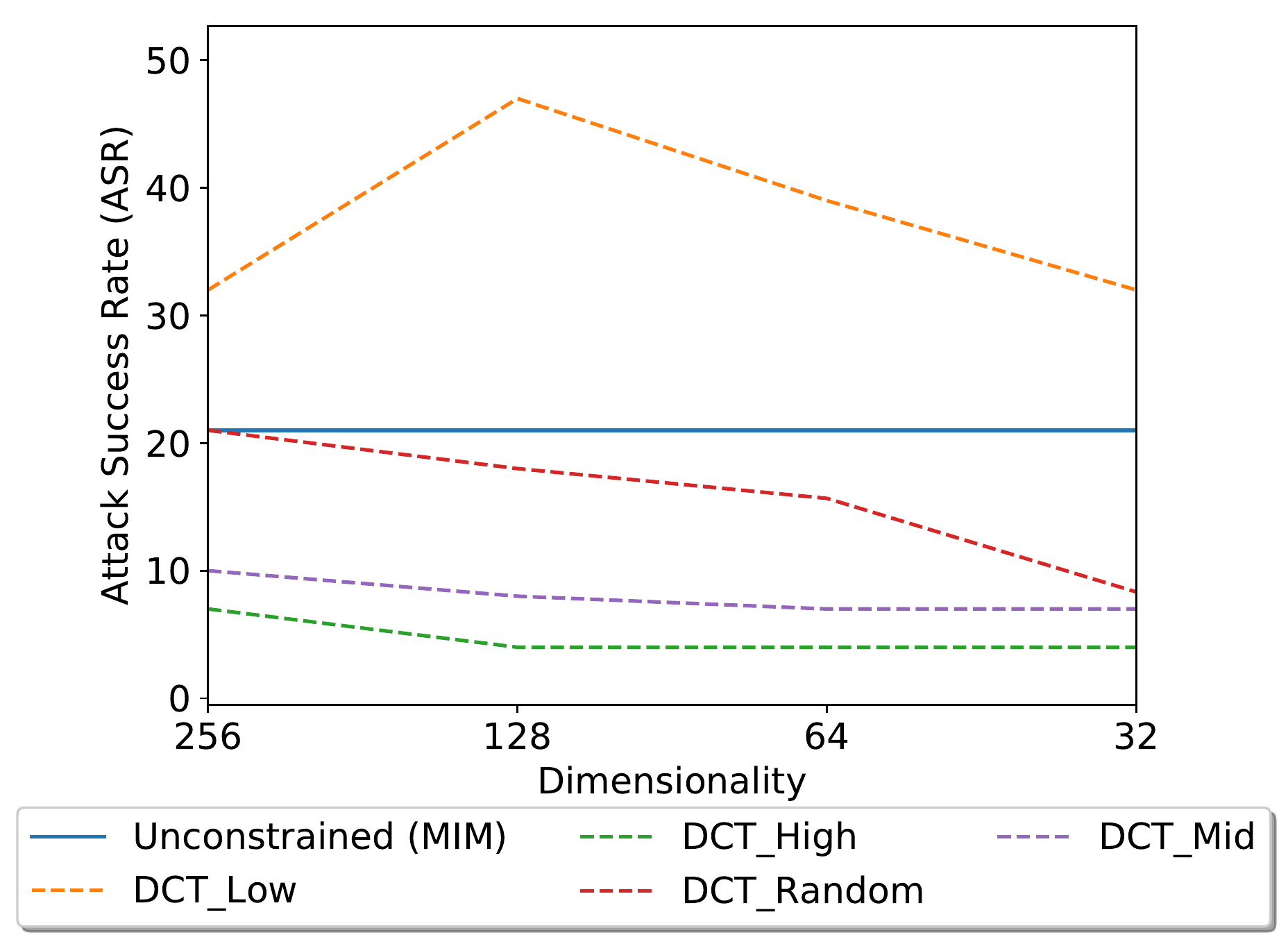}
    \caption{Non-targeted with $\epsilon=16$ and $\text{iterations}=10$.}
    \label{greybox24}
\end{subfigure}
~~~
\begin{subfigure}{0.3\linewidth}
    \centering
    \includegraphics[width=\textwidth]{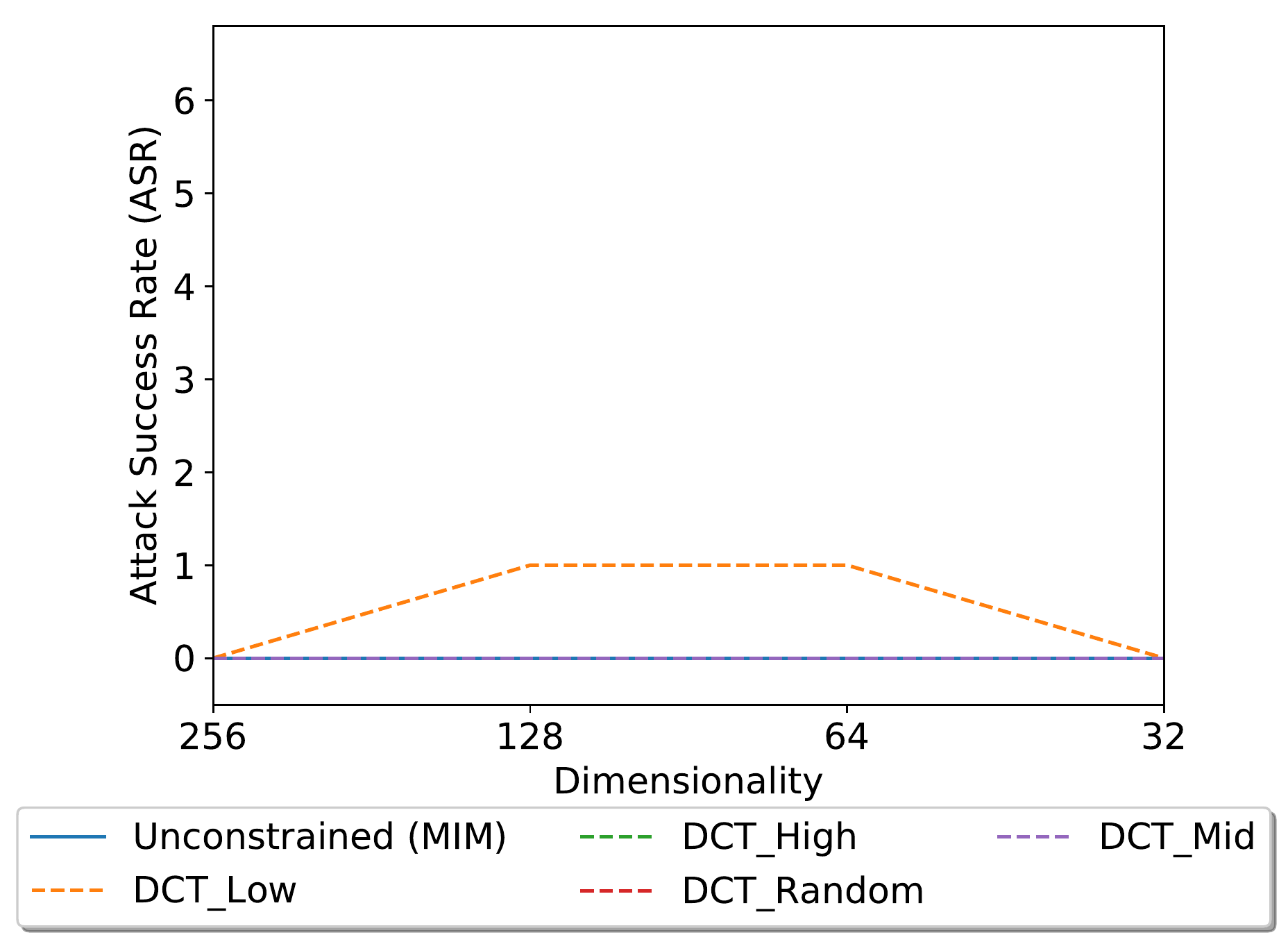}
    \caption{Targeted with $\epsilon=32$ and $\text{iterations}=10$.}
    \label{greybox34}
\end{subfigure}
\caption{\textbf{Black-box} attack from Cln\_3 to D1.}
\label{blackbox22}
\end{figure*}

\begin{figure*}
\begin{subfigure}{0.3\linewidth}
    \centering
    \includegraphics[width=\textwidth]{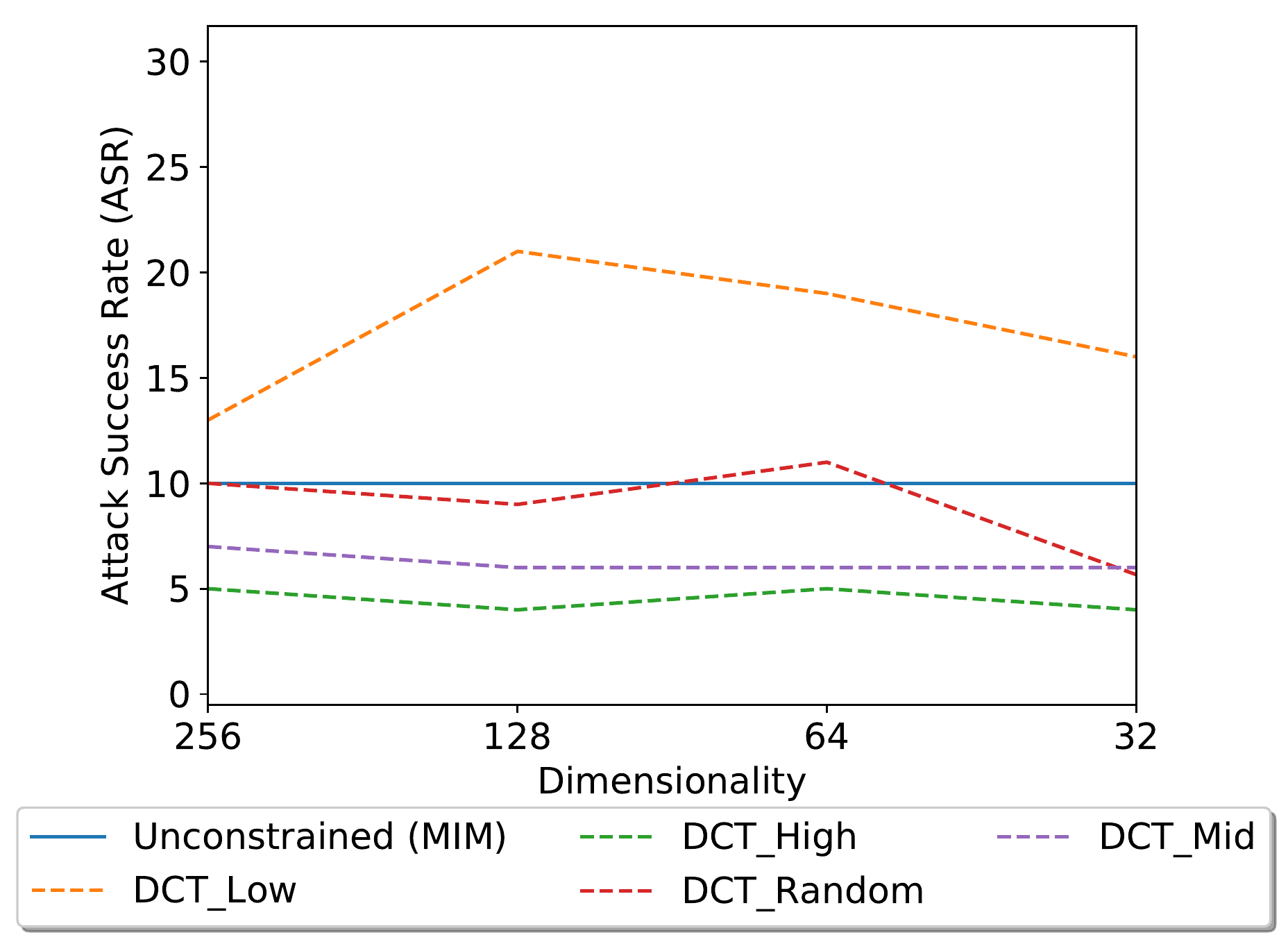}
    \caption{Non-targeted with $\epsilon=16$ and $\text{iterations}=1$.}
    \label{greybox14}
\end{subfigure}
~~~
\begin{subfigure}{0.3\linewidth}
    \centering
    \includegraphics[width=\textwidth]{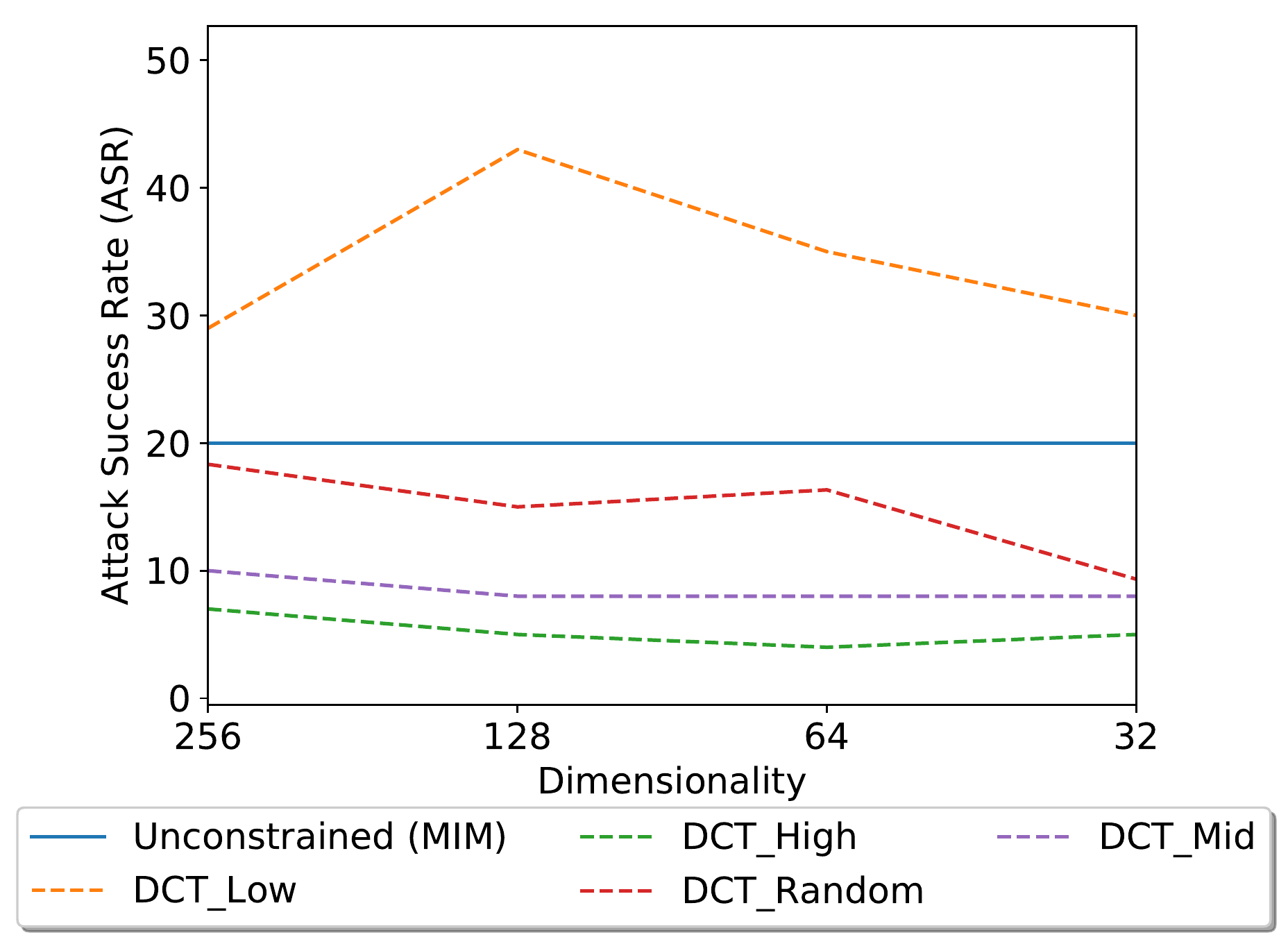}
    \caption{Non-targeted with $\epsilon=16$ and $\text{iterations}=10$.}
    \label{greybox24}
\end{subfigure}
~~~
\begin{subfigure}{0.3\linewidth}
    \centering
    \includegraphics[width=\textwidth]{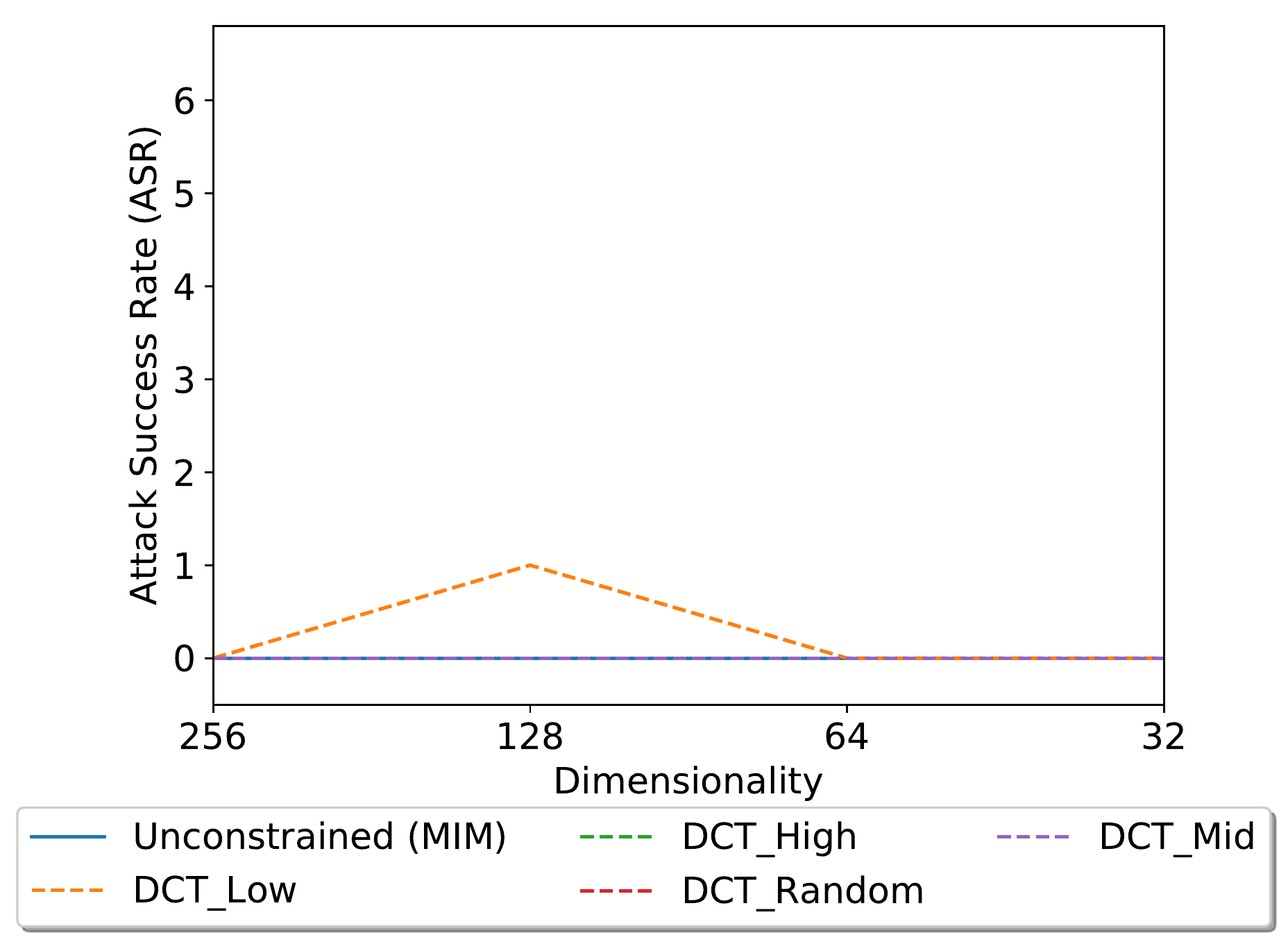}
    \caption{Targeted with $\epsilon=32$ and $\text{iterations}=10$.}
    \label{greybox34}
\end{subfigure}
\caption{\textbf{Black-box} attack from Cln\_3 to D2.}
\label{blackbox23}
\end{figure*}

\begin{figure*}
\begin{subfigure}{0.3\linewidth}
    \centering
    \includegraphics[width=\textwidth]{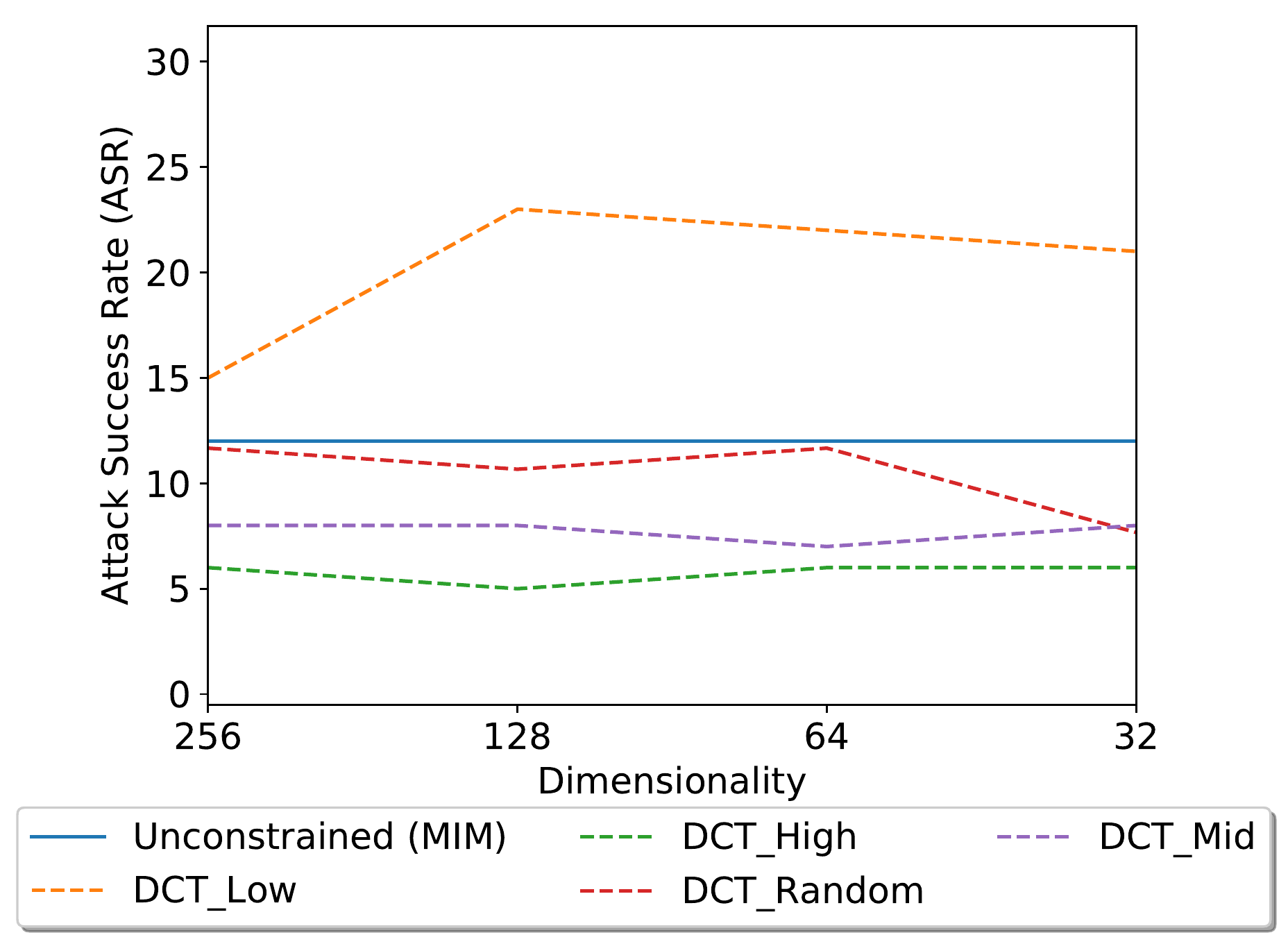}
    \caption{Non-targeted with $\epsilon=16$ and $\text{iterations}=1$.}
    \label{greybox14}
\end{subfigure}
~~~
\begin{subfigure}{0.3\linewidth}
    \centering
    \includegraphics[width=\textwidth]{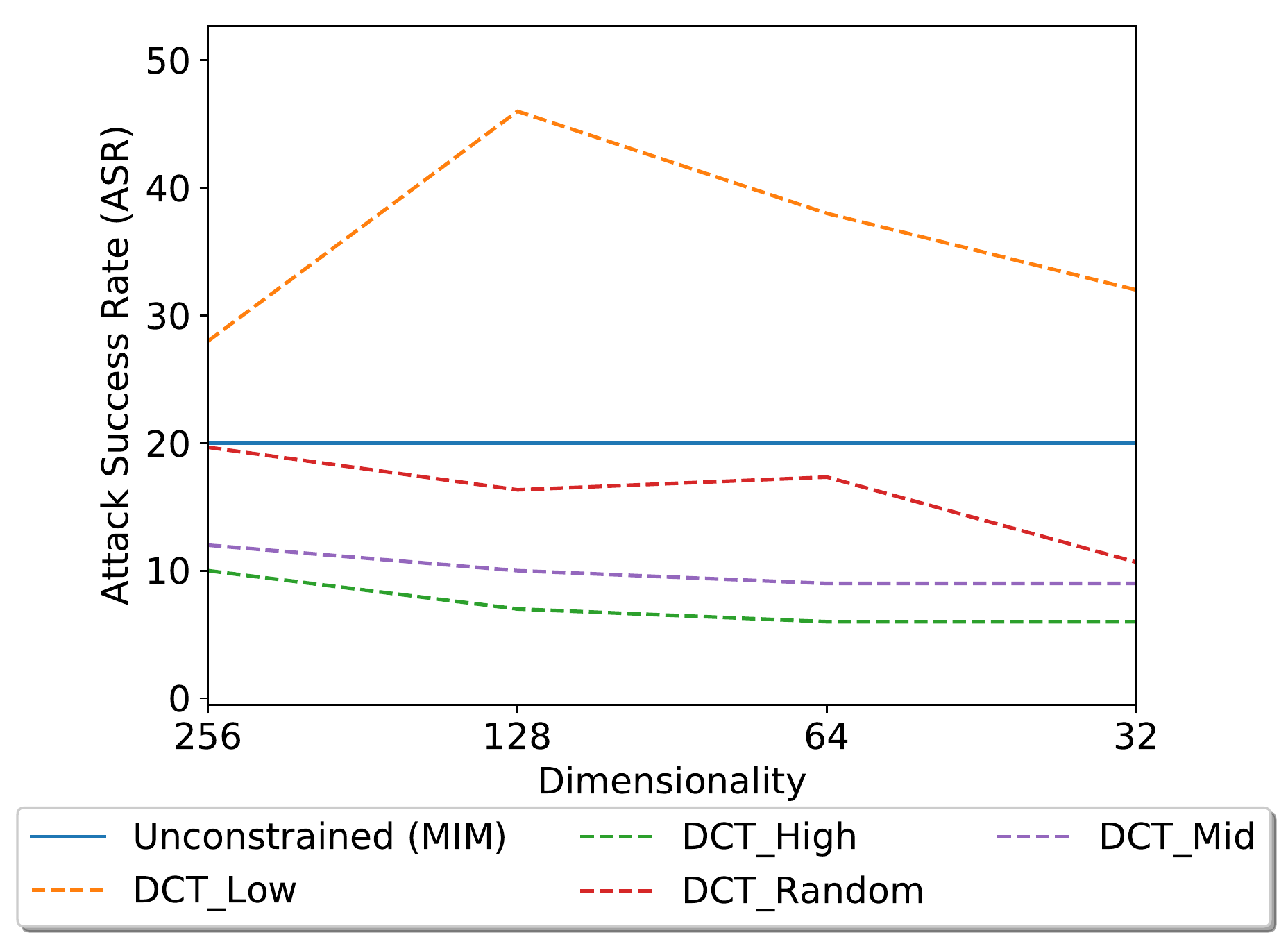}
    \caption{Non-targeted with $\epsilon=16$ and $\text{iterations}=10$.}
    \label{greybox24}
\end{subfigure}
~~~
\begin{subfigure}{0.3\linewidth}
    \centering
    \includegraphics[width=\textwidth]{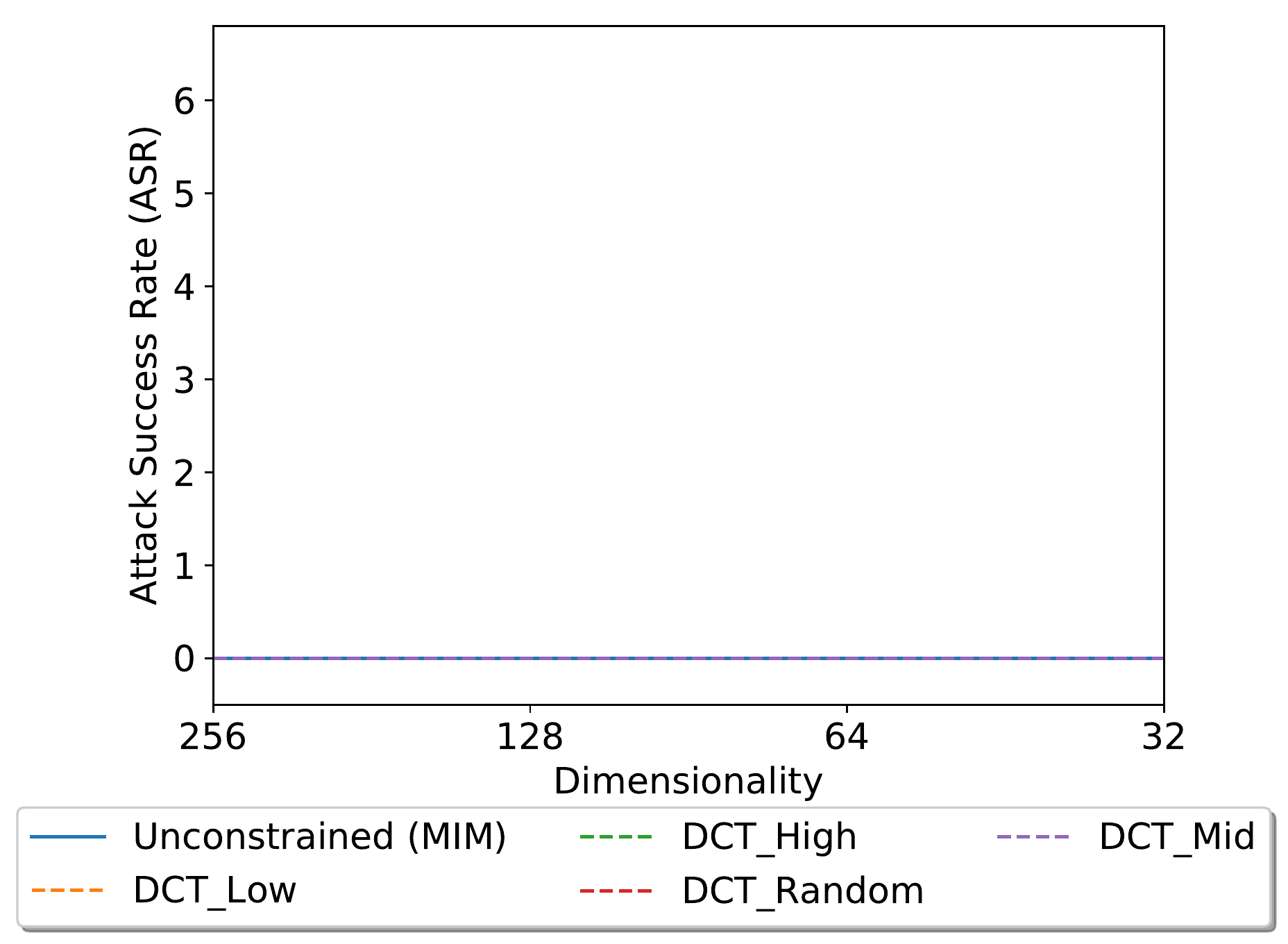}
    \caption{Targeted with $\epsilon=32$ and $\text{iterations}=10$.}
    \label{greybox34}
\end{subfigure}
\caption{\textbf{Black-box} attack from Cln\_3 to D3.}
\label{blackbox24}
\end{figure*}

\begin{figure*}
\begin{subfigure}{0.3\linewidth}
    \centering
    \includegraphics[width=\textwidth]{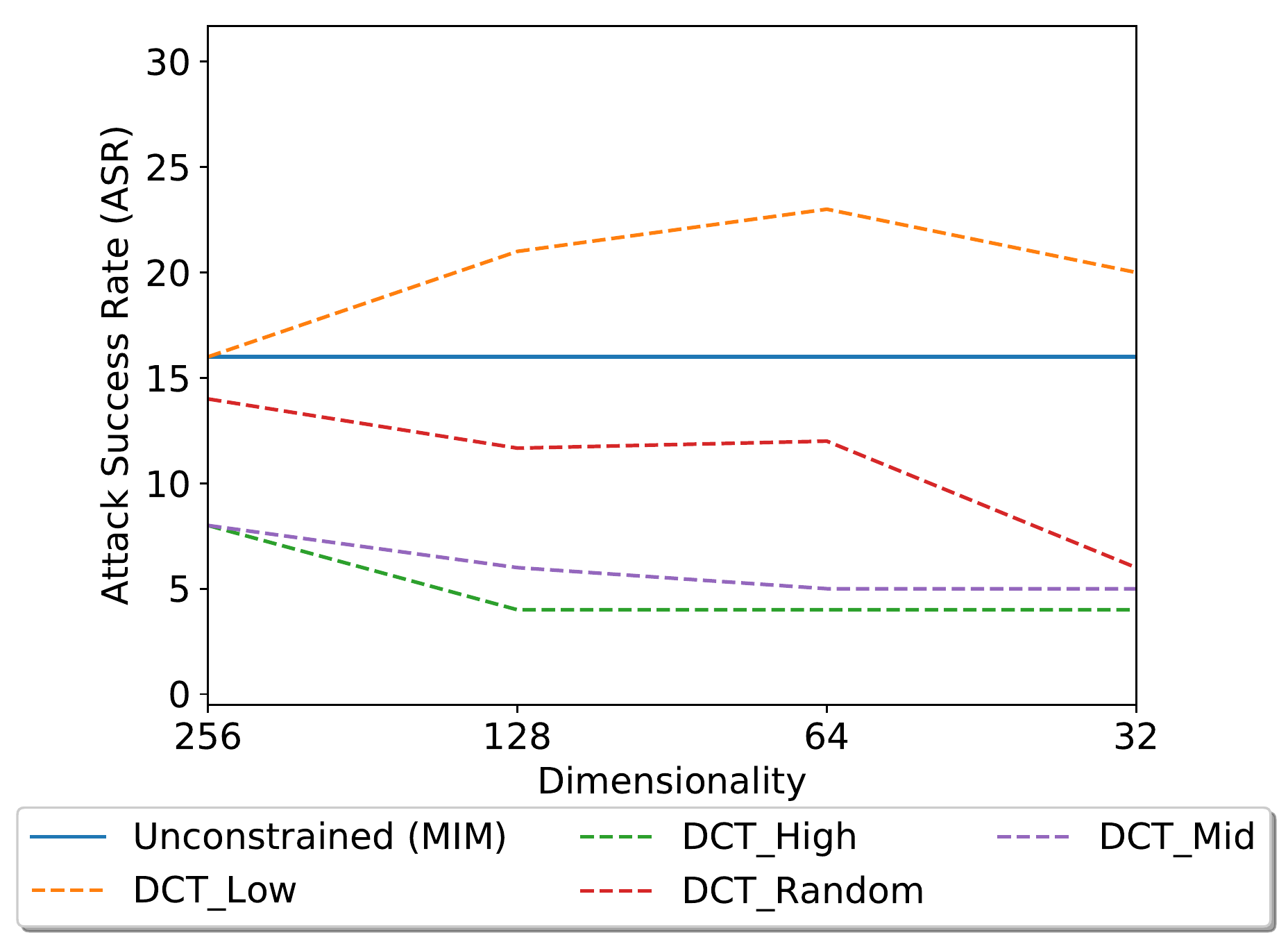}
    \caption{Non-targeted with $\epsilon=16$ and $\text{iterations}=1$.}
    \label{greybox14}
\end{subfigure}
~~~
\begin{subfigure}{0.3\linewidth}
    \centering
    \includegraphics[width=\textwidth]{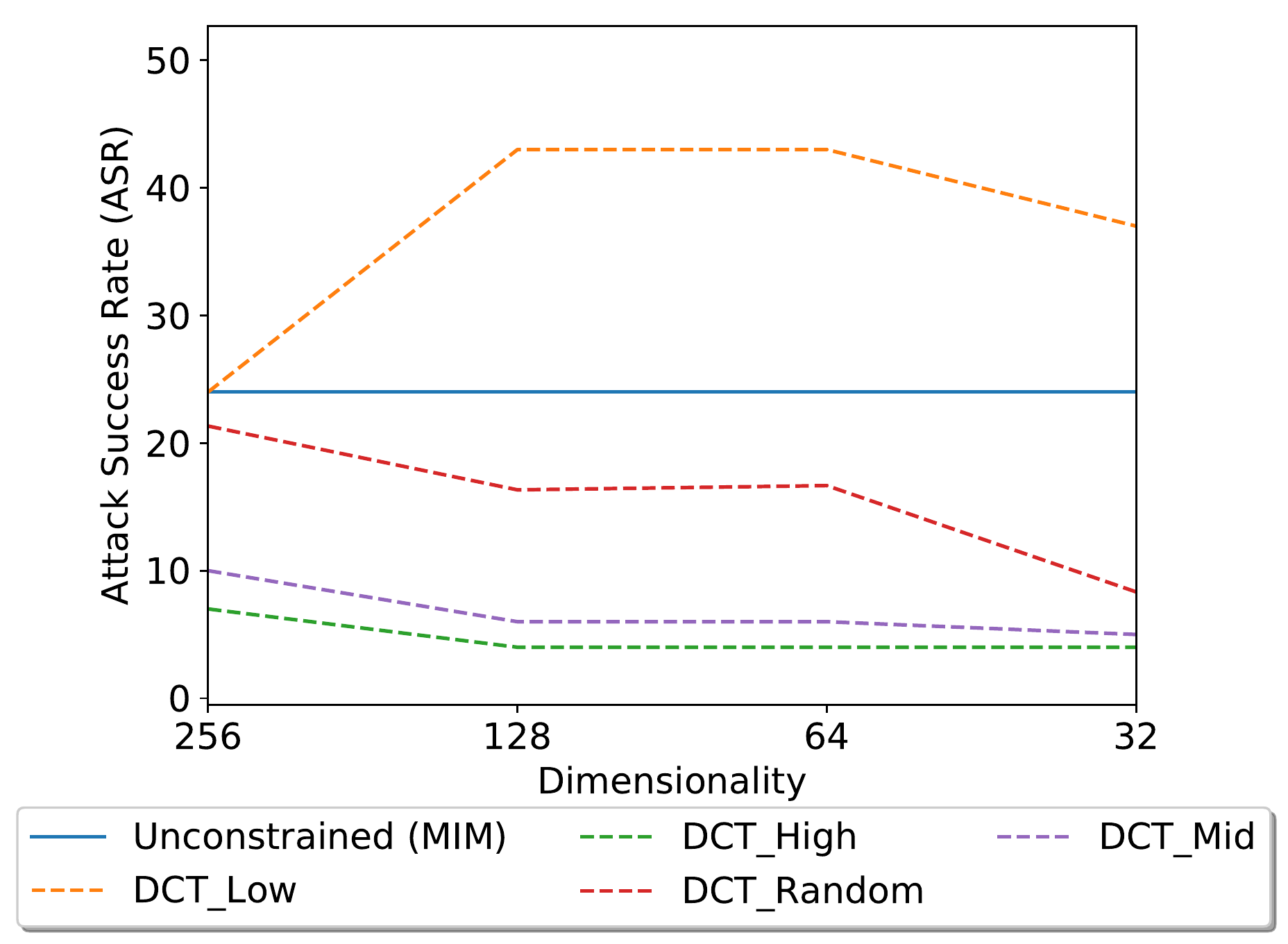}
    \caption{Non-targeted with $\epsilon=16$ and $\text{iterations}=10$.}
    \label{greybox24}
\end{subfigure}
~~~
\begin{subfigure}{0.3\linewidth}
    \centering
    \includegraphics[width=\textwidth]{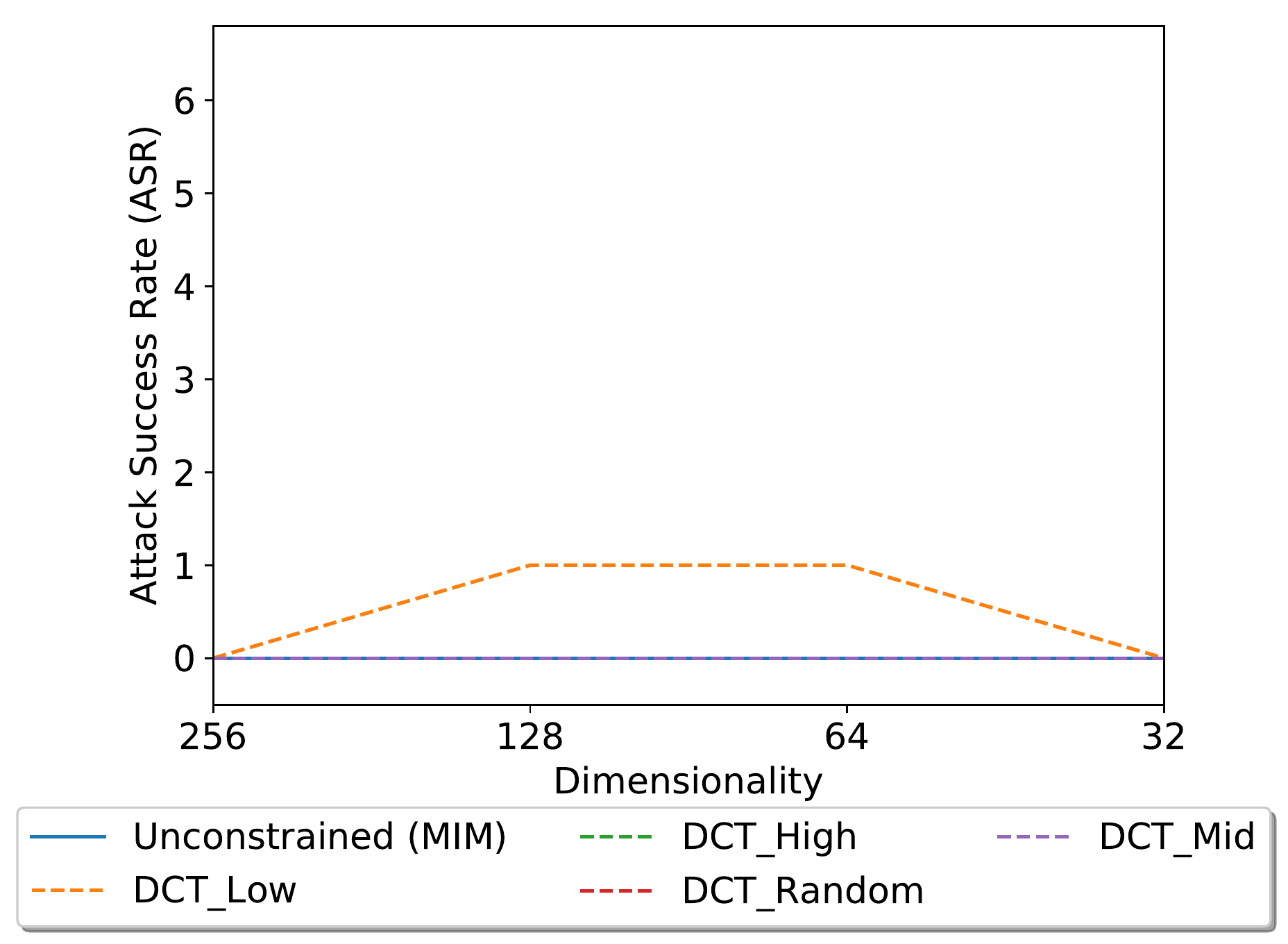}
    \caption{Targeted with $\epsilon=32$ and $\text{iterations}=10$.}
    \label{greybox34}
\end{subfigure}
\caption{\textbf{Black-box} attack from Cln\_3 to D4.}
\label{blackbox25}
\end{figure*}

\begin{figure*}
\begin{subfigure}{0.3\linewidth}
    \centering
    \includegraphics[width=\textwidth]{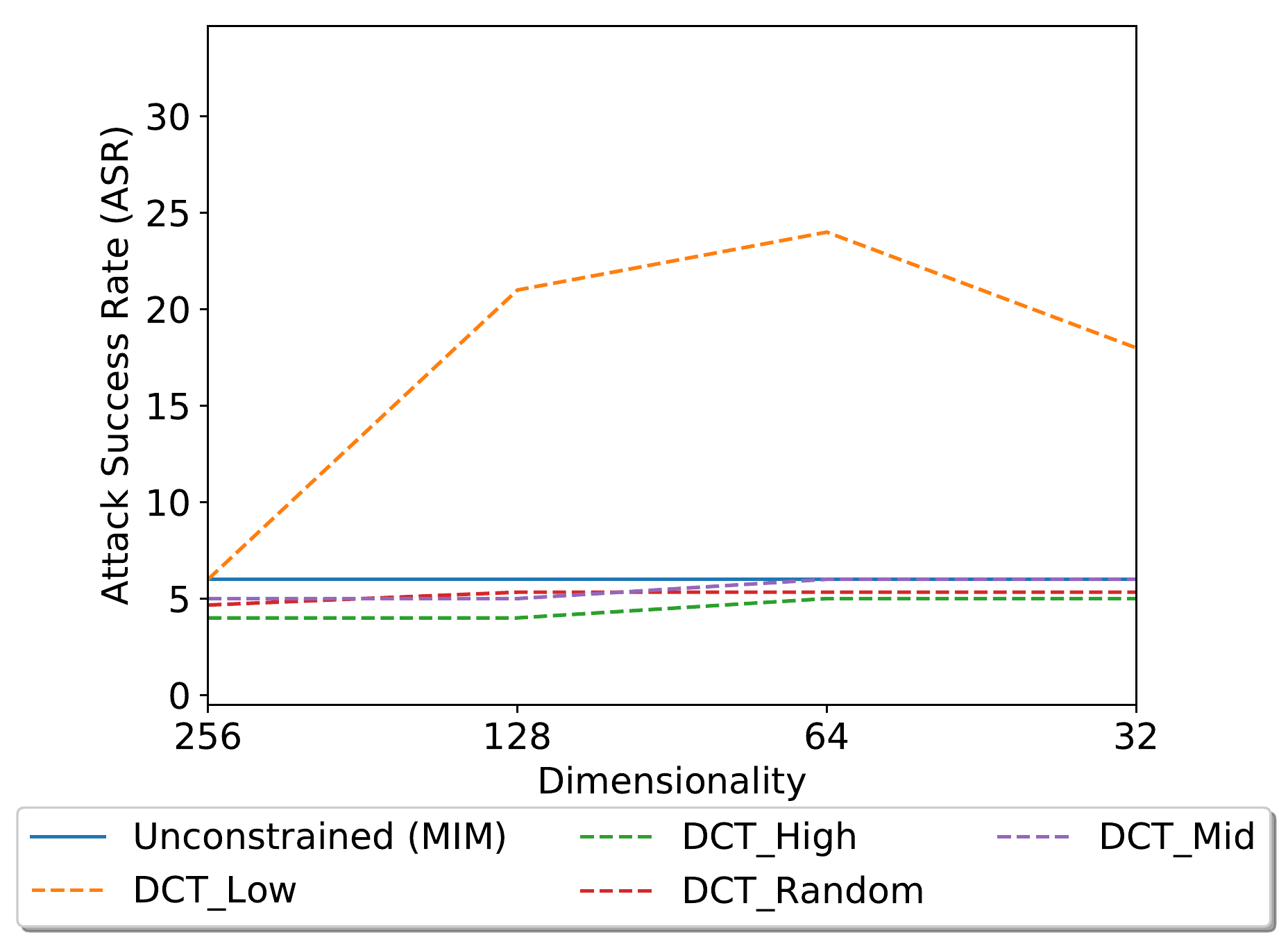}
    \caption{Non-targeted with $\epsilon=16$ and $\text{iterations}=1$.}
    \label{greybox14}
\end{subfigure}
~~~
\begin{subfigure}{0.3\linewidth}
    \centering
    \includegraphics[width=\textwidth]{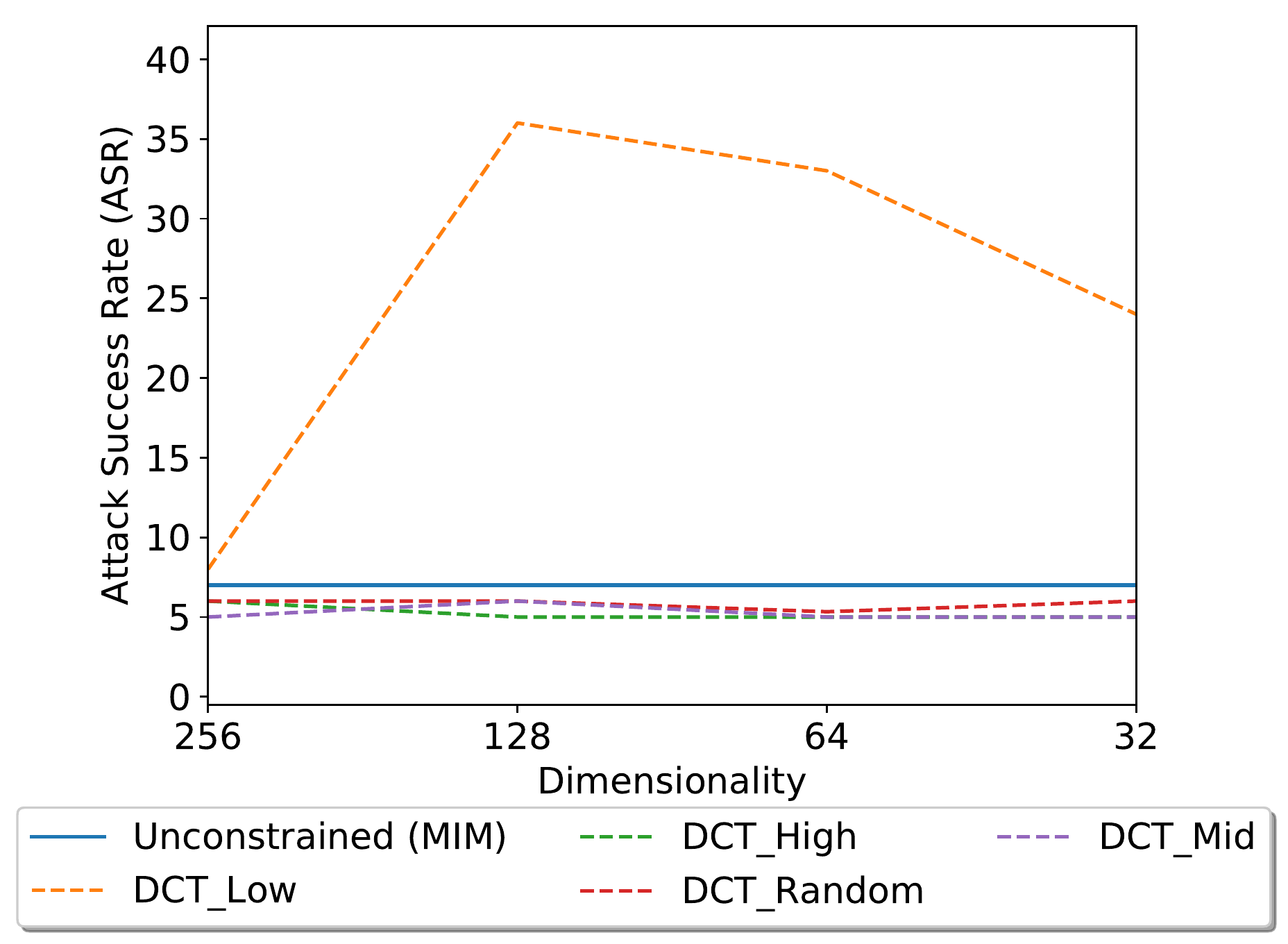}
    \caption{Non-targeted with $\epsilon=16$ and $\text{iterations}=10$.}
    \label{greybox24}
\end{subfigure}
~~~
\begin{subfigure}{0.3\linewidth}
    \centering
    \includegraphics[width=\textwidth]{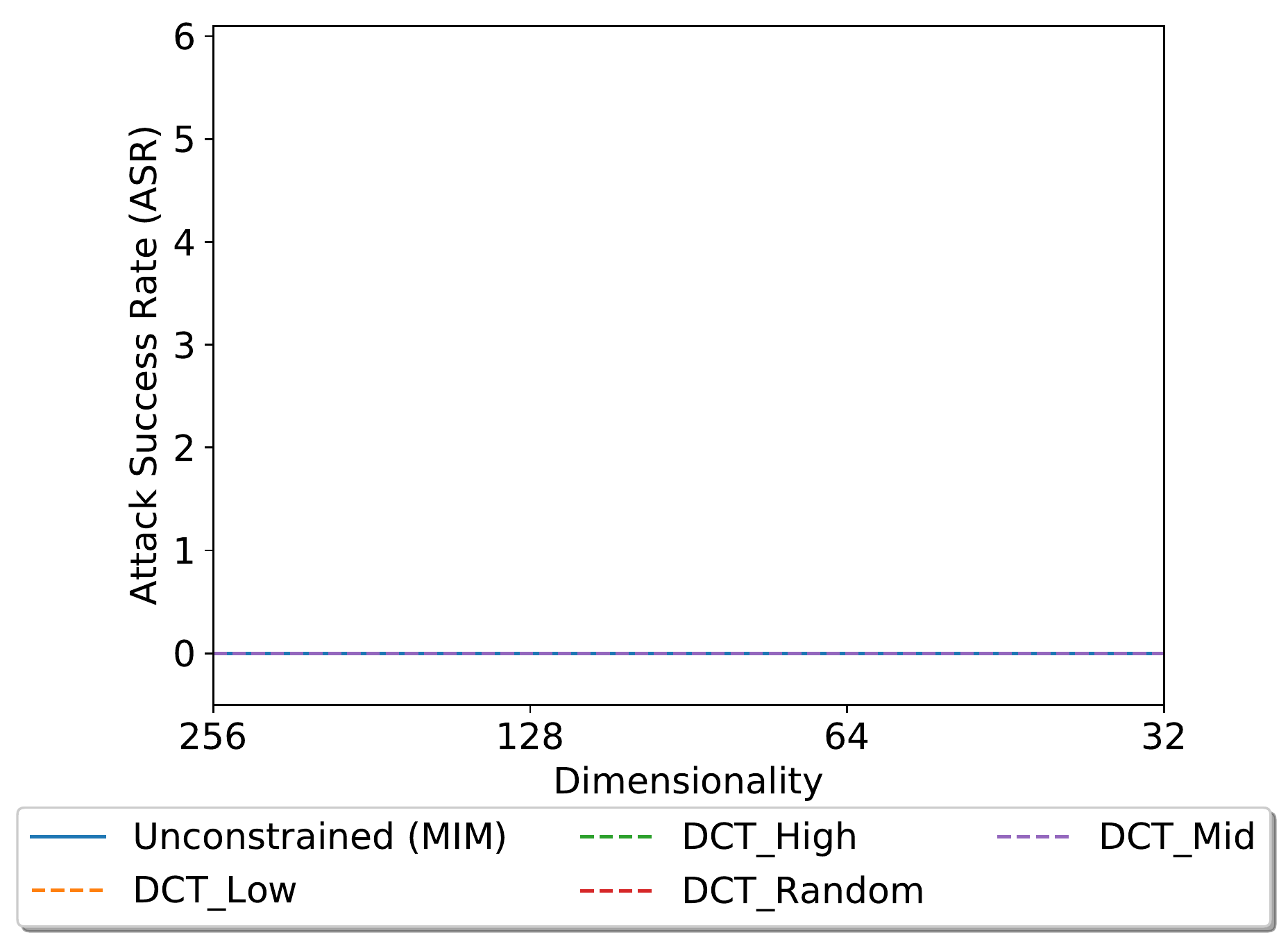}
    \caption{Targeted with $\epsilon=32$ and $\text{iterations}=10$.}
    \label{greybox34}
\end{subfigure}
\caption{\textbf{Black-box} attack from Adv\_1 to EnsAdv.}
\label{blackbox31}
\end{figure*}

\begin{figure*}
\begin{subfigure}{0.3\linewidth}
    \centering
    \includegraphics[width=\textwidth]{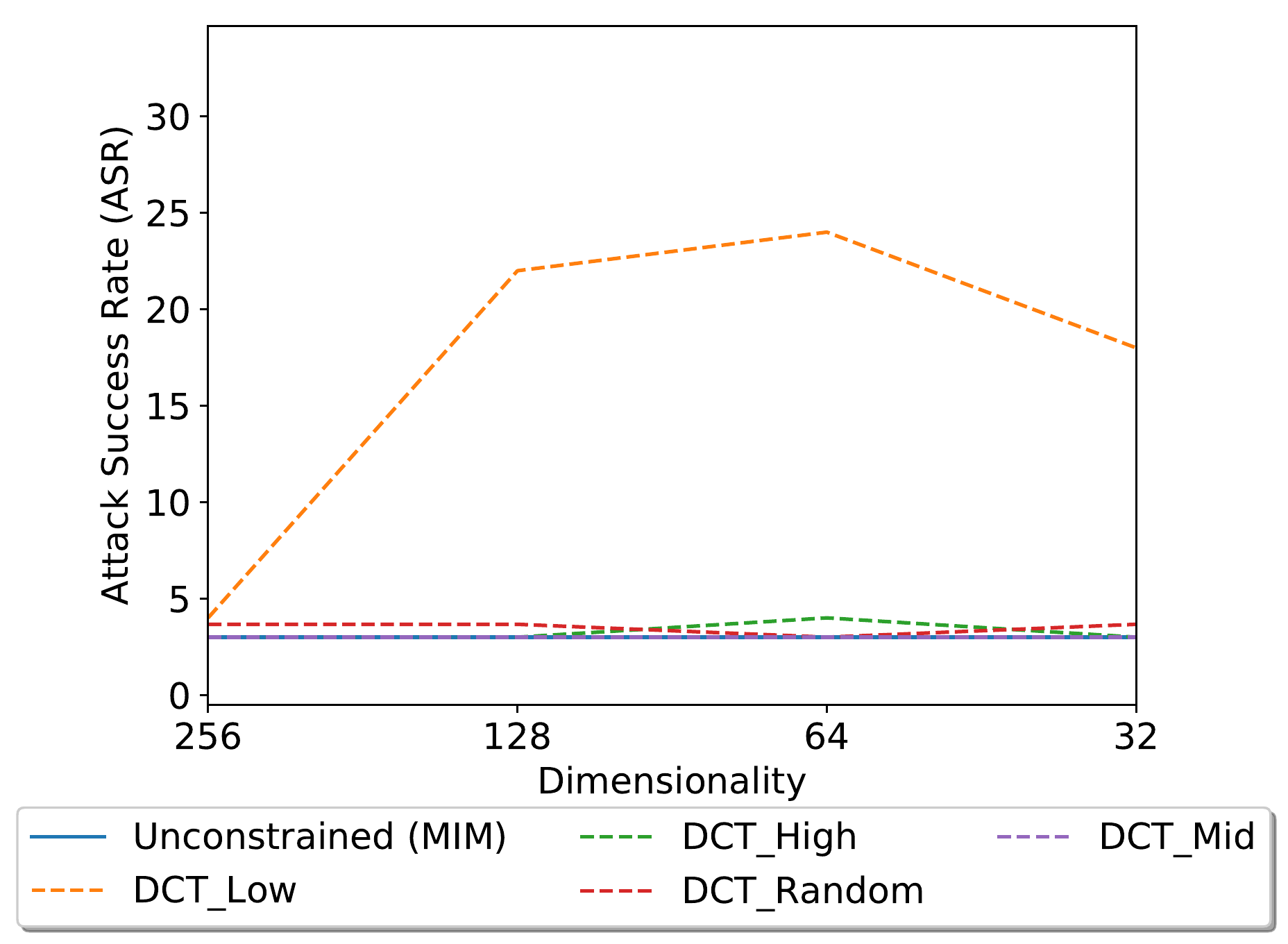}
    \caption{Non-targeted with $\epsilon=16$ and $\text{iterations}=1$.}
    \label{greybox14}
\end{subfigure}
~~~
\begin{subfigure}{0.3\linewidth}
    \centering
    \includegraphics[width=\textwidth]{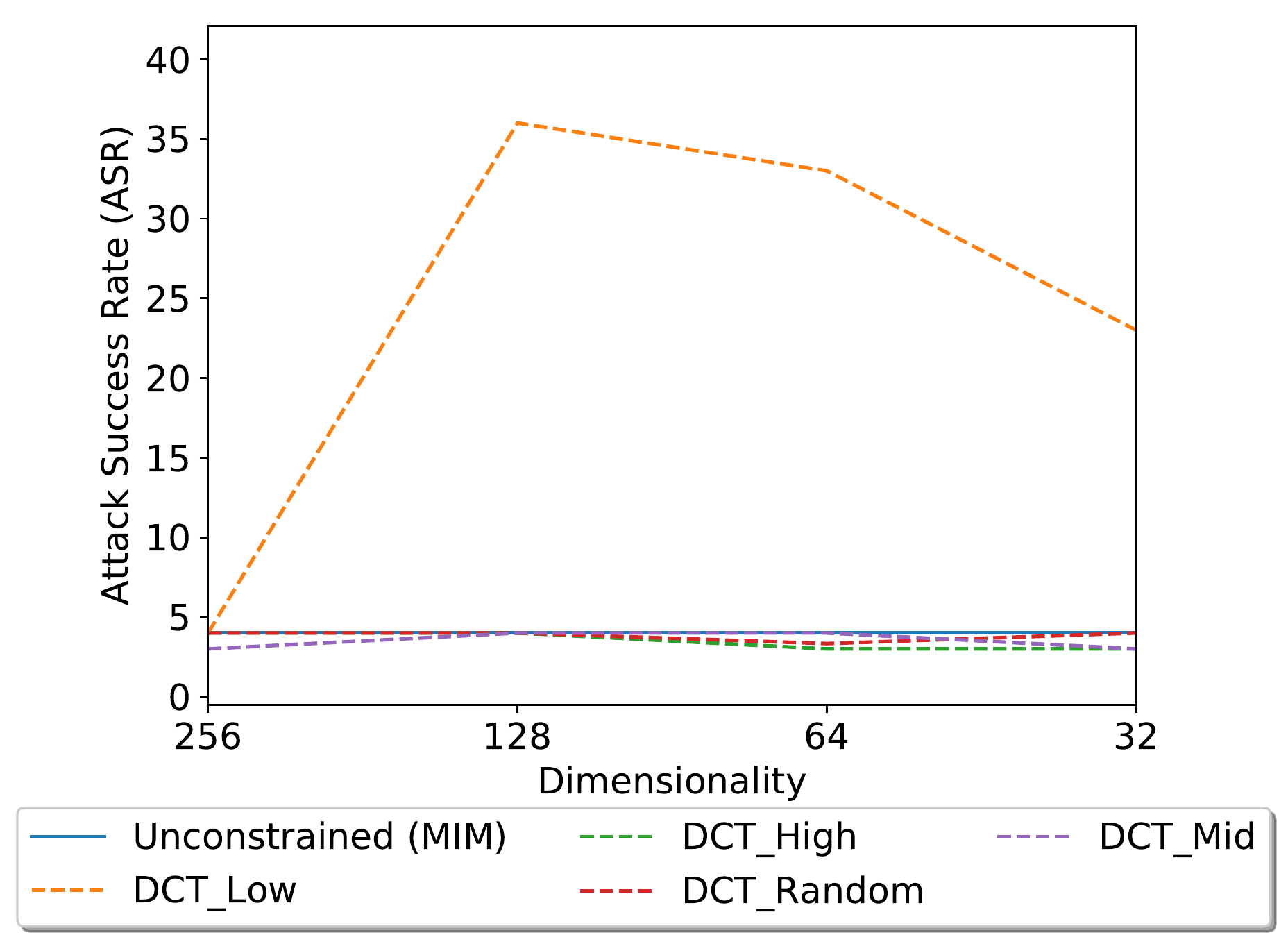}
    \caption{Non-targeted with $\epsilon=16$ and $\text{iterations}=10$.}
    \label{greybox24}
\end{subfigure}
~~~
\begin{subfigure}{0.3\linewidth}
    \centering
    \includegraphics[width=\textwidth]{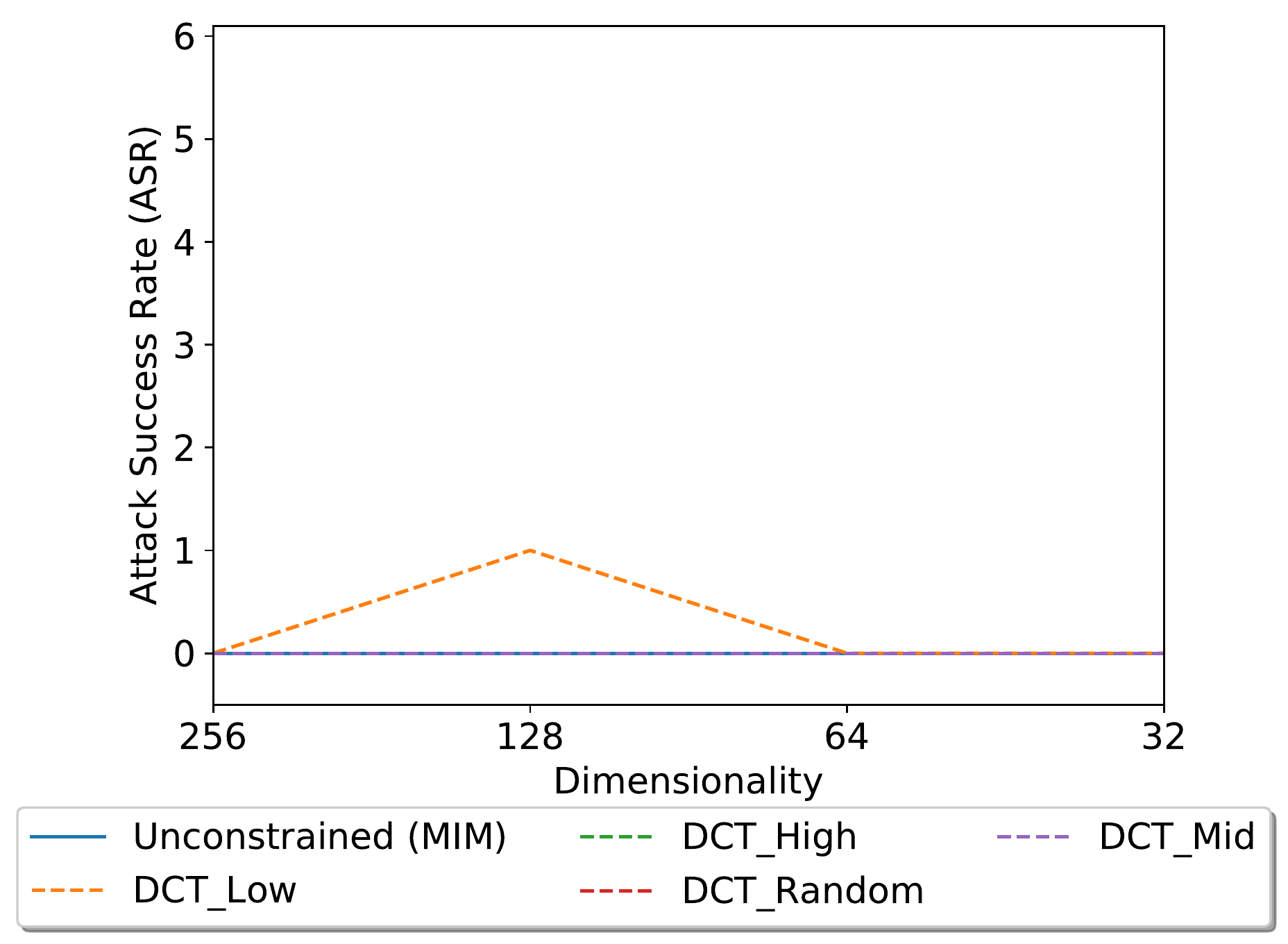}
    \caption{Targeted with $\epsilon=32$ and $\text{iterations}=10$.}
    \label{greybox34}
\end{subfigure}
\caption{\textbf{Black-box} attack from Adv\_1 to D1.}
\label{blackbox32}
\end{figure*}

\begin{figure*}
\begin{subfigure}{0.3\linewidth}
    \centering
    \includegraphics[width=\textwidth]{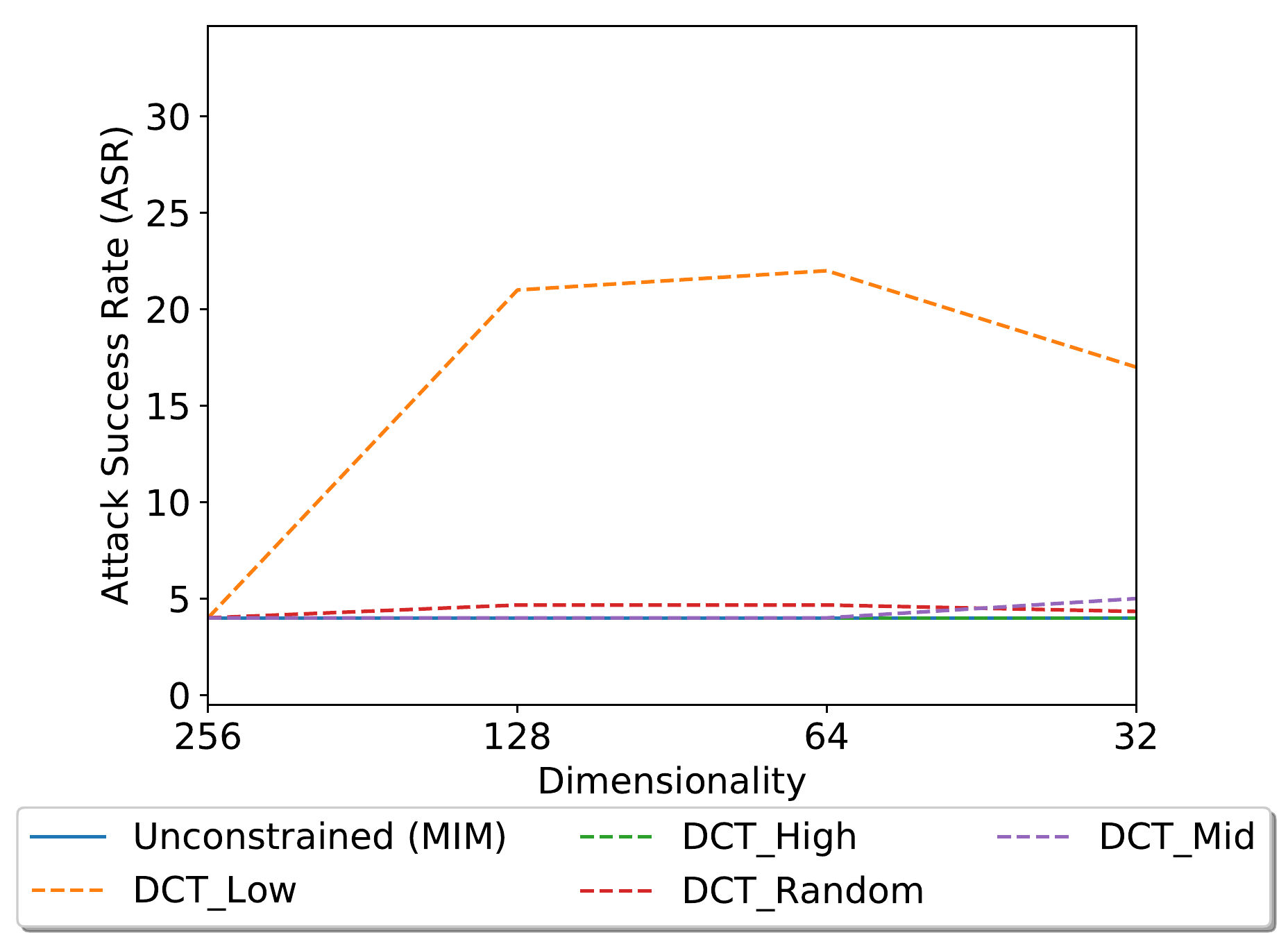}
    \caption{Non-targeted with $\epsilon=16$ and $\text{iterations}=1$.}
    \label{greybox14}
\end{subfigure}
~~~
\begin{subfigure}{0.3\linewidth}
    \centering
    \includegraphics[width=\textwidth]{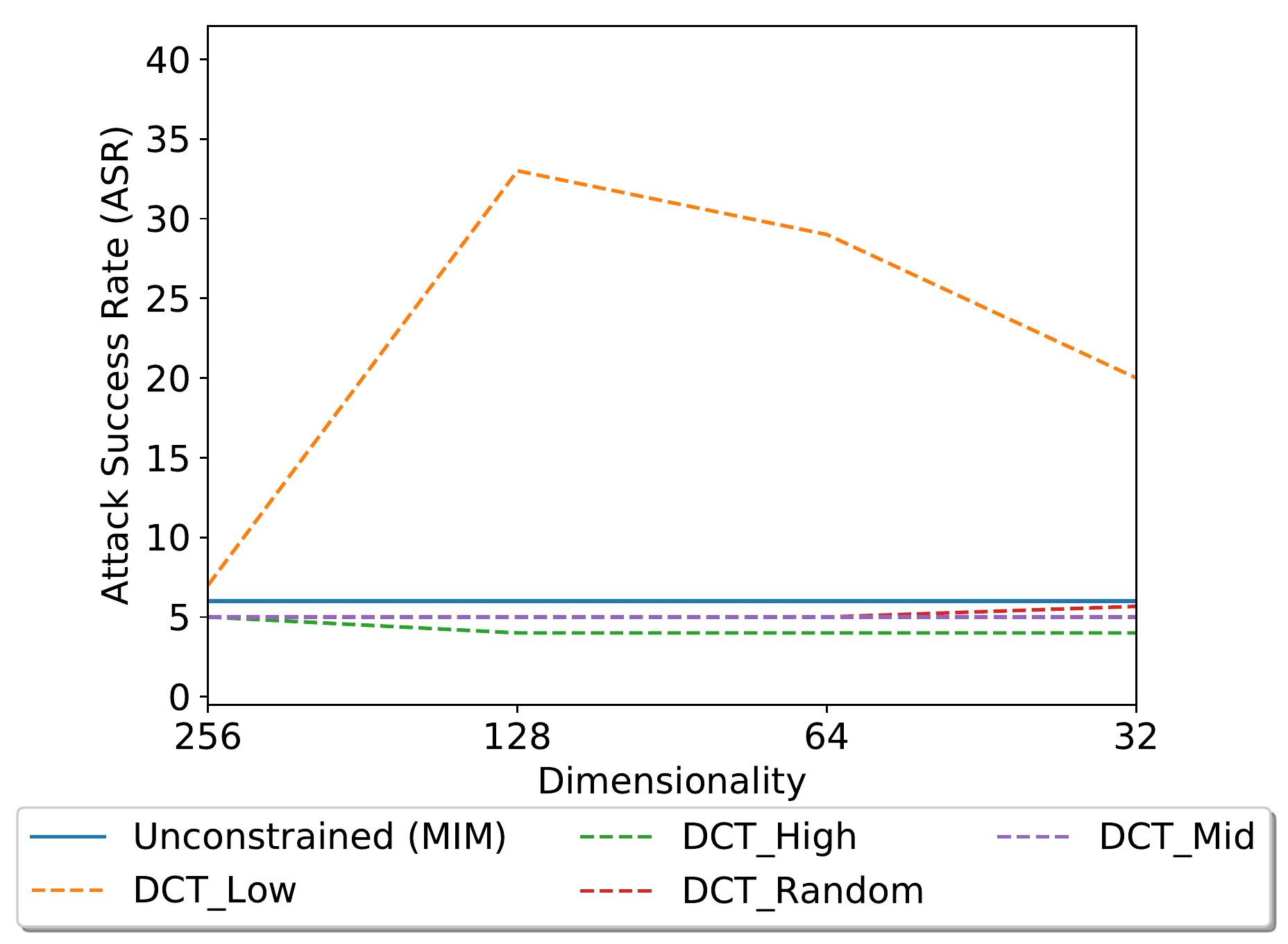}
    \caption{Non-targeted with $\epsilon=16$ and $\text{iterations}=10$.}
    \label{greybox24}
\end{subfigure}
~~~
\begin{subfigure}{0.3\linewidth}
    \centering
    \includegraphics[width=\textwidth]{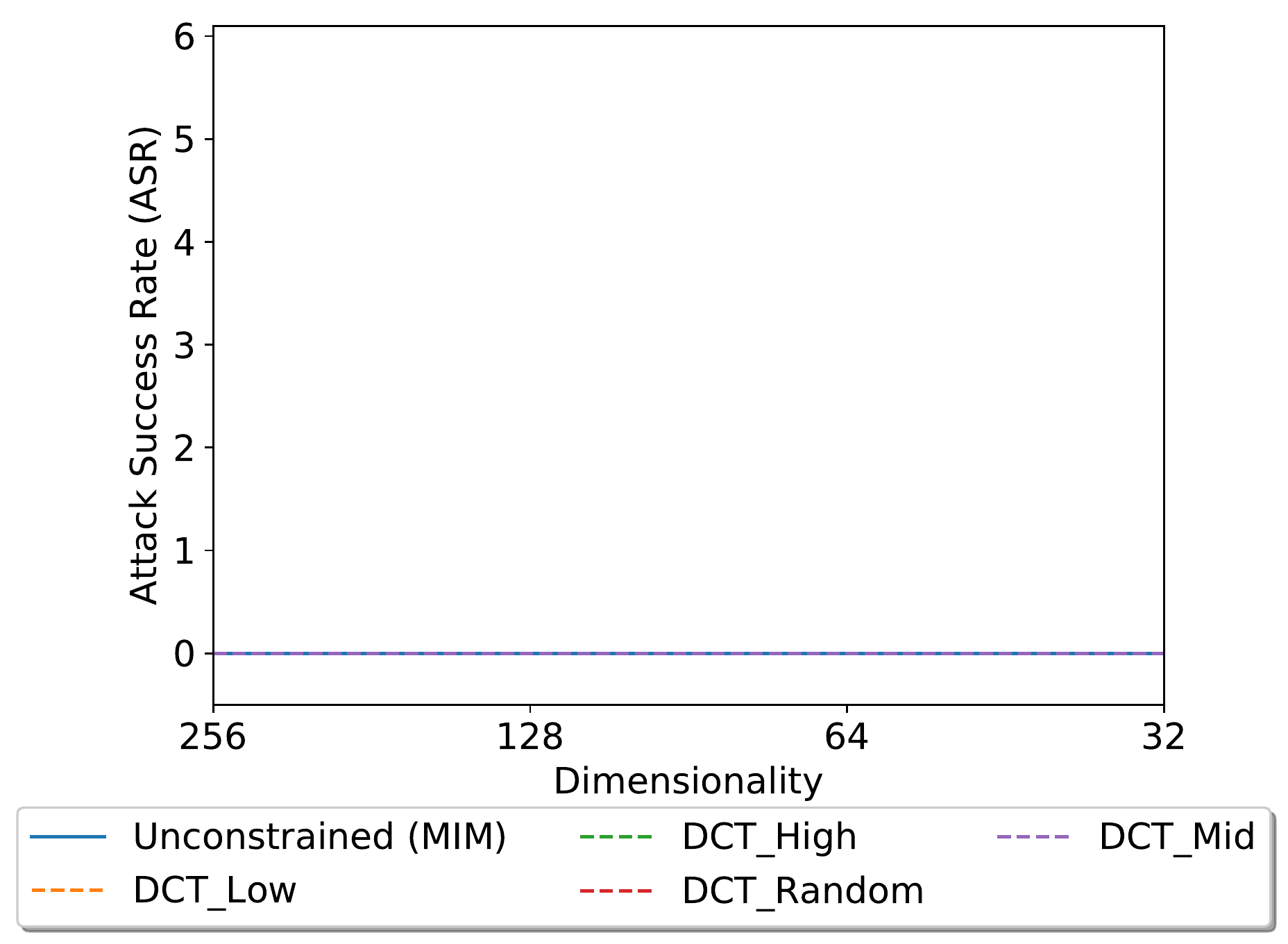}
    \caption{Targeted with $\epsilon=32$ and $\text{iterations}=10$.}
    \label{greybox34}
\end{subfigure}
\caption{\textbf{Black-box} attack from Adv\_1 to D2.}
\label{blackbox33}
\end{figure*}

\begin{figure*}
\begin{subfigure}{0.3\linewidth}
    \centering
    \includegraphics[width=\textwidth]{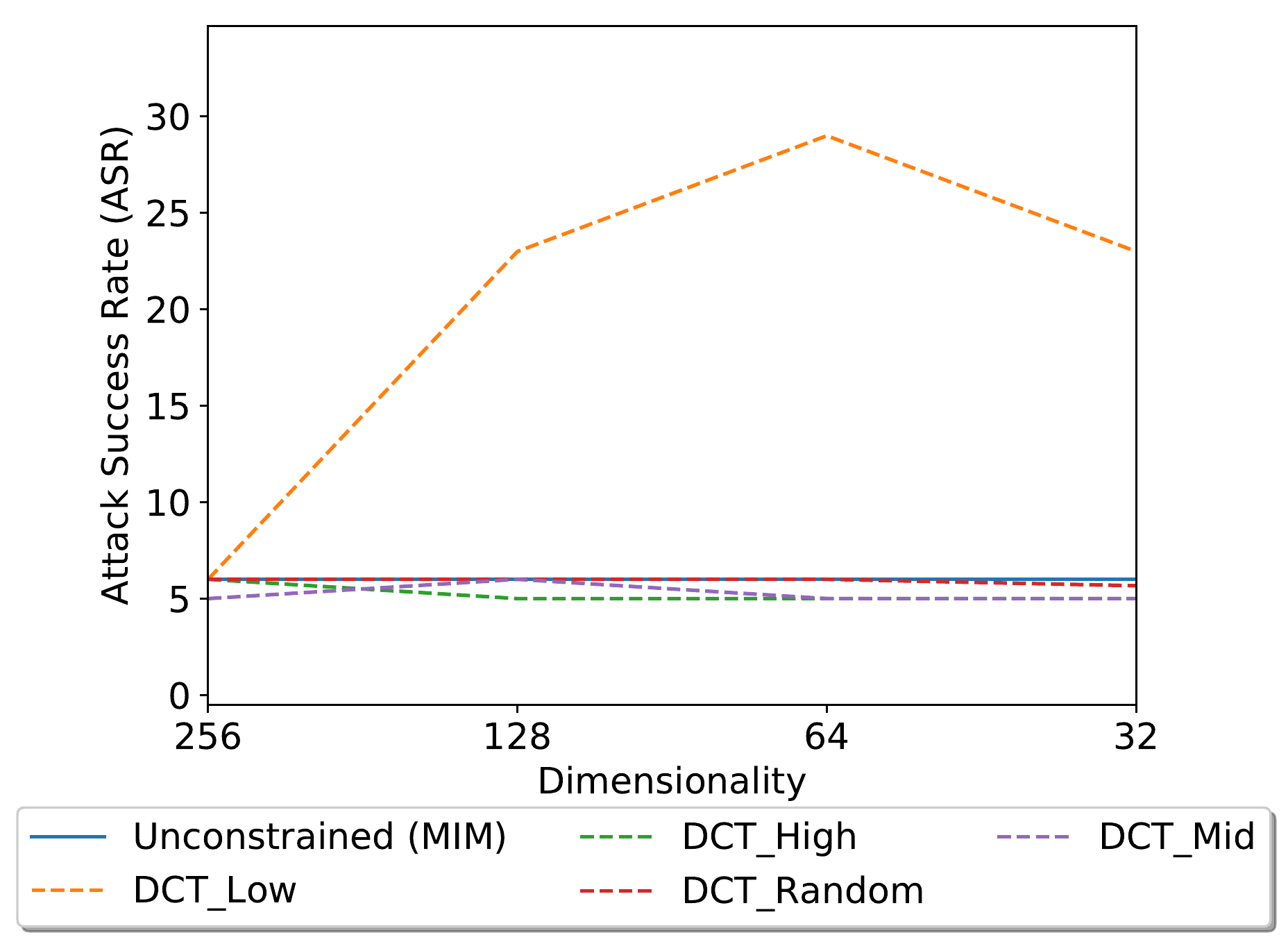}
    \caption{Non-targeted with $\epsilon=16$ and $\text{iterations}=1$.}
    \label{greybox14}
\end{subfigure}
~~~
\begin{subfigure}{0.3\linewidth}
    \centering
    \includegraphics[width=\textwidth]{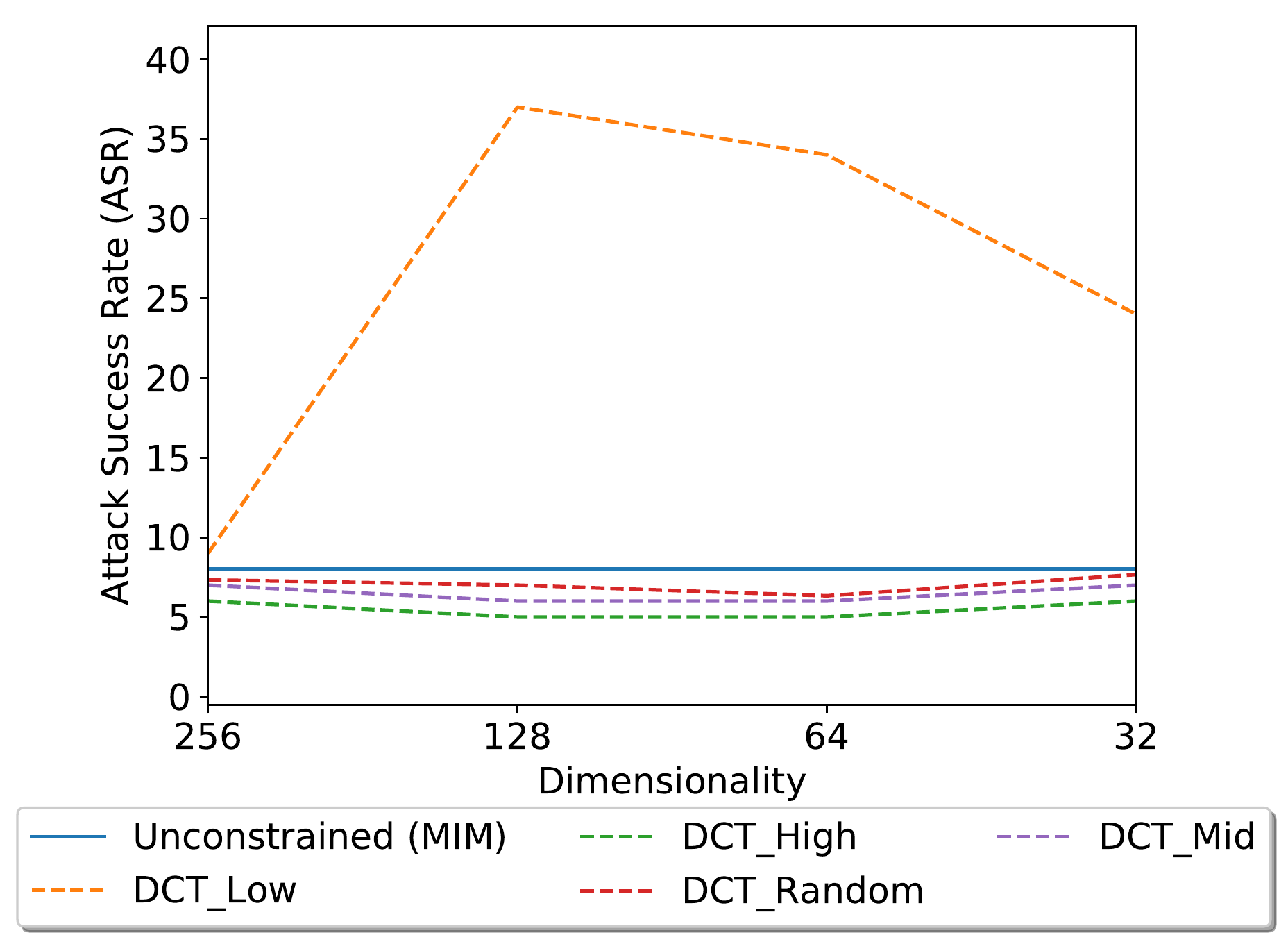}
    \caption{Non-targeted with $\epsilon=16$ and $\text{iterations}=10$.}
    \label{greybox24}
\end{subfigure}
~~~
\begin{subfigure}{0.3\linewidth}
    \centering
    \includegraphics[width=\textwidth]{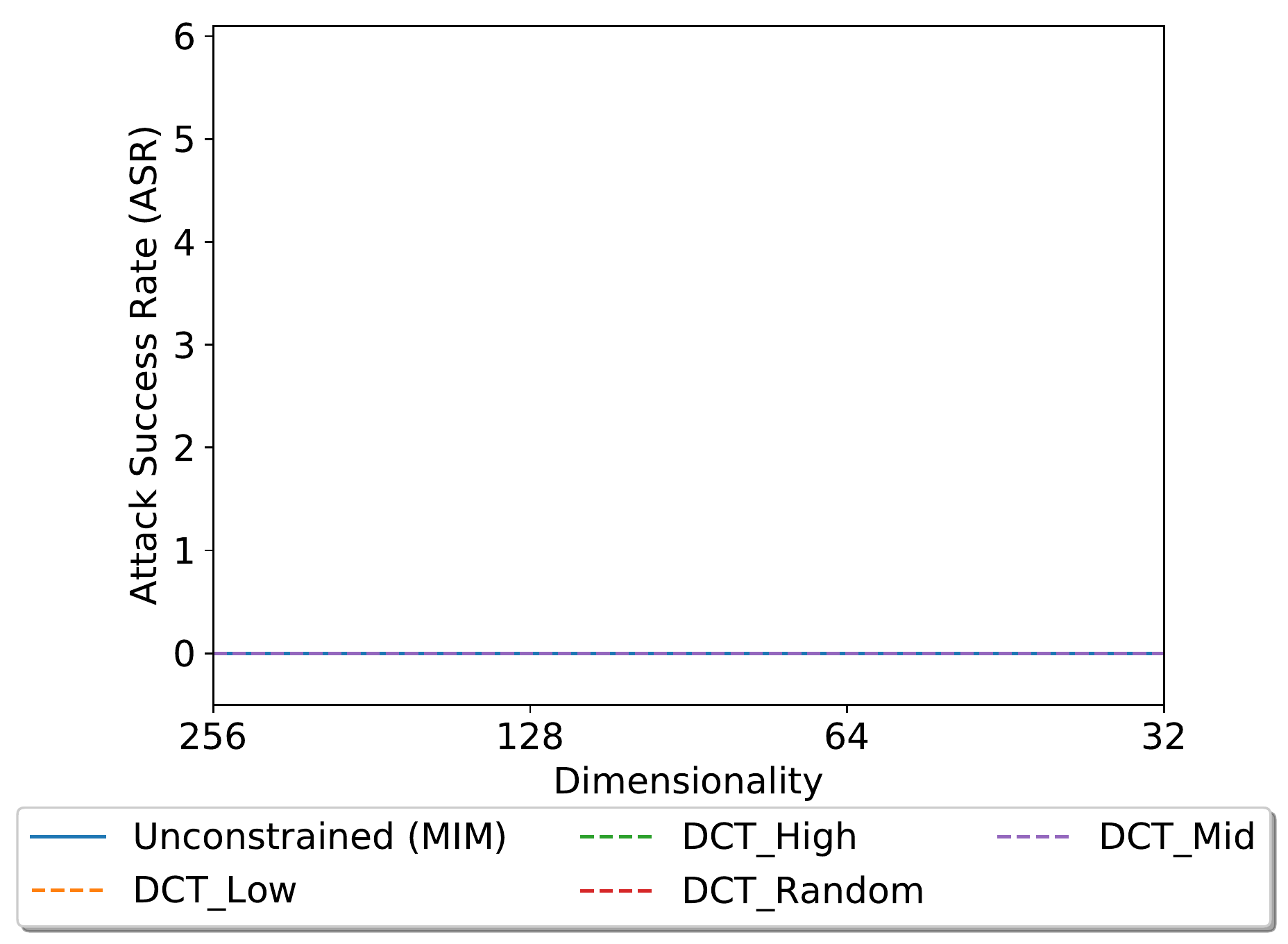}
    \caption{Targeted with $\epsilon=32$ and $\text{iterations}=10$.}
    \label{greybox34}
\end{subfigure}
\caption{\textbf{Black-box} attack from Adv\_1 to D3.}
\label{blackbox34}
\end{figure*}

\begin{figure*}
\begin{subfigure}{0.3\linewidth}
    \centering
    \includegraphics[width=\textwidth]{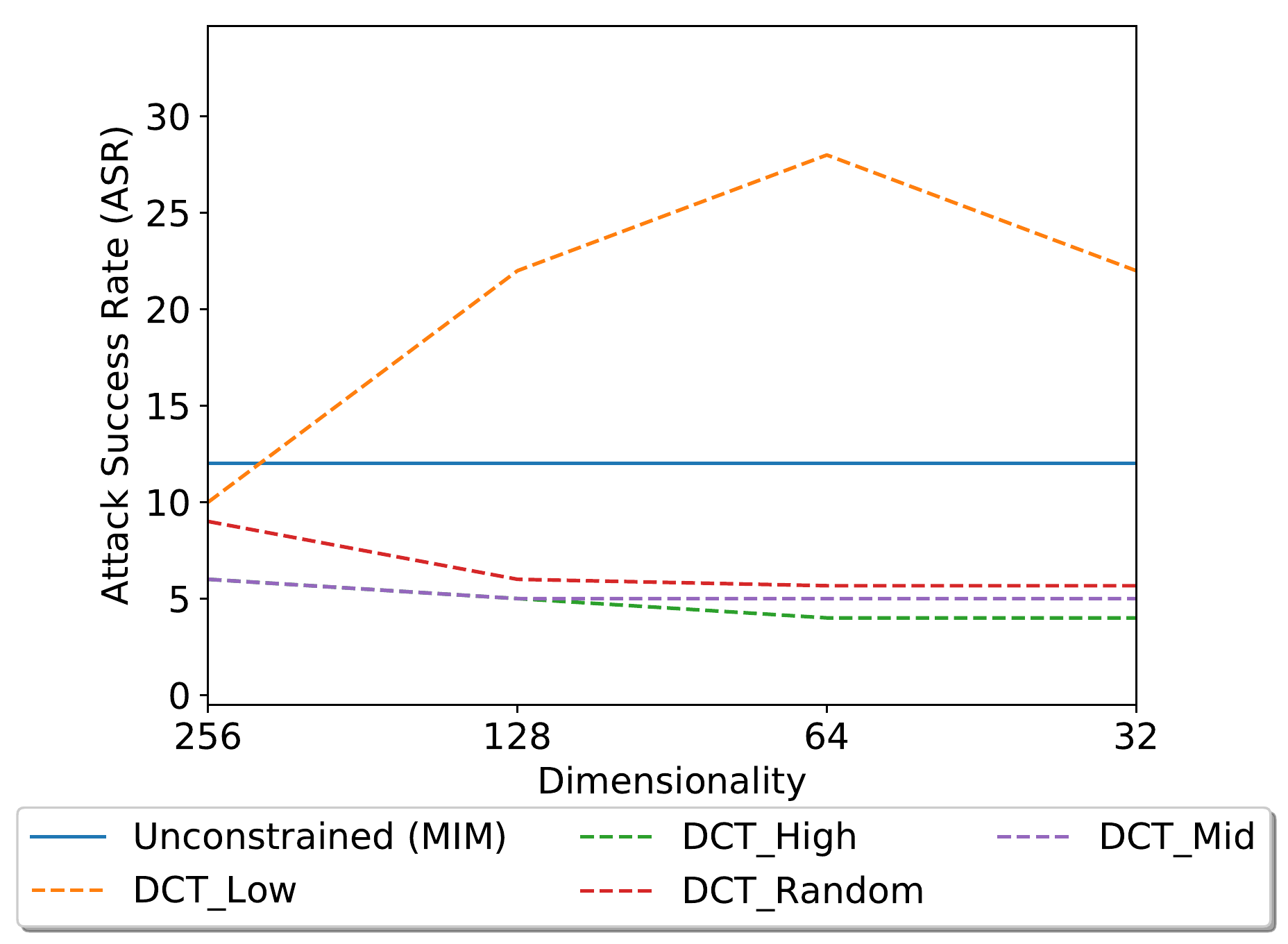}
    \caption{Non-targeted with $\epsilon=16$ and $\text{iterations}=1$.}
    \label{greybox14}
\end{subfigure}
~~~
\begin{subfigure}{0.3\linewidth}
    \centering
    \includegraphics[width=\textwidth]{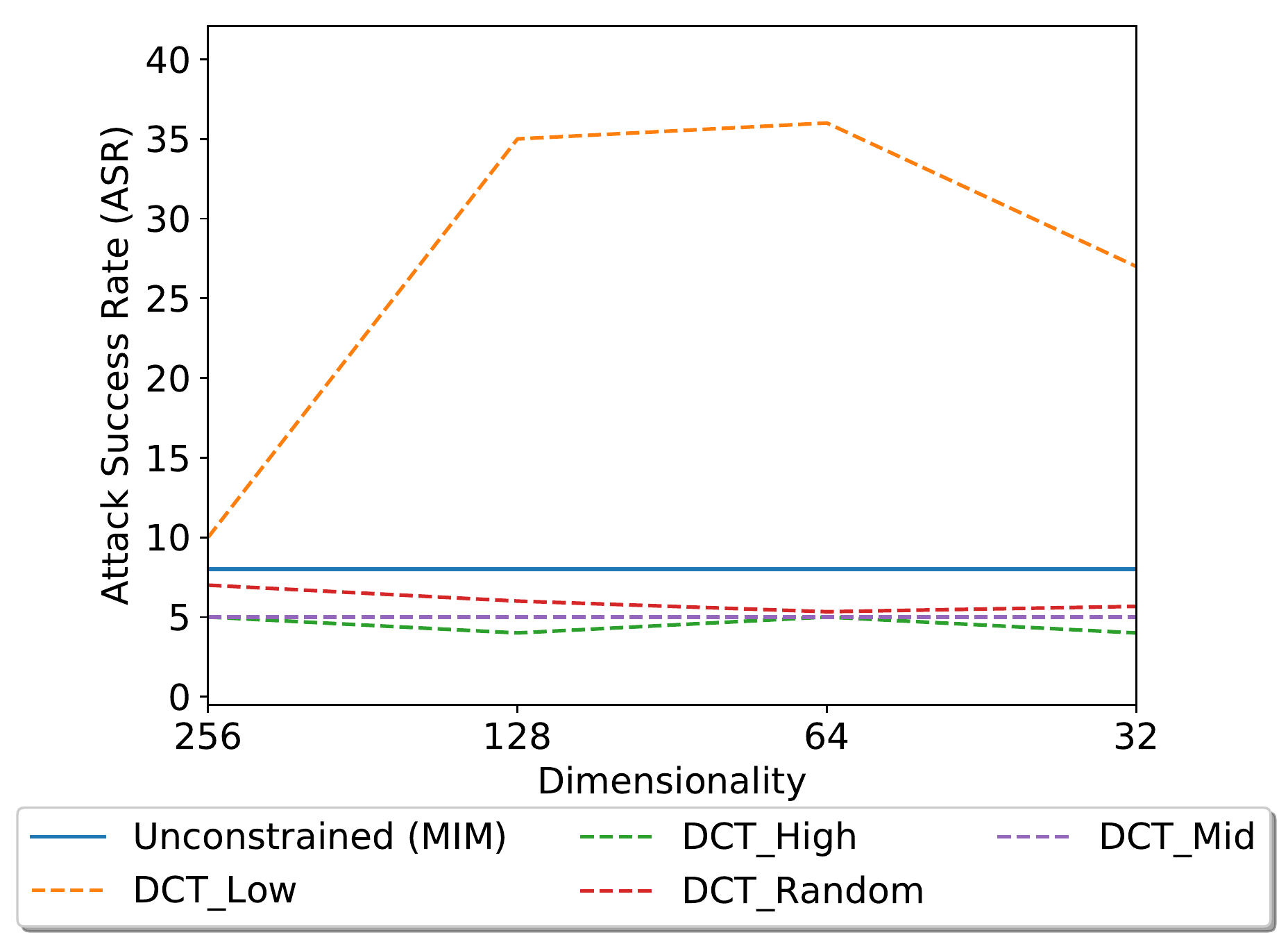}
    \caption{Non-targeted with $\epsilon=16$ and $\text{iterations}=10$.}
    \label{greybox24}
\end{subfigure}
~~~
\begin{subfigure}{0.3\linewidth}
    \centering
    \includegraphics[width=\textwidth]{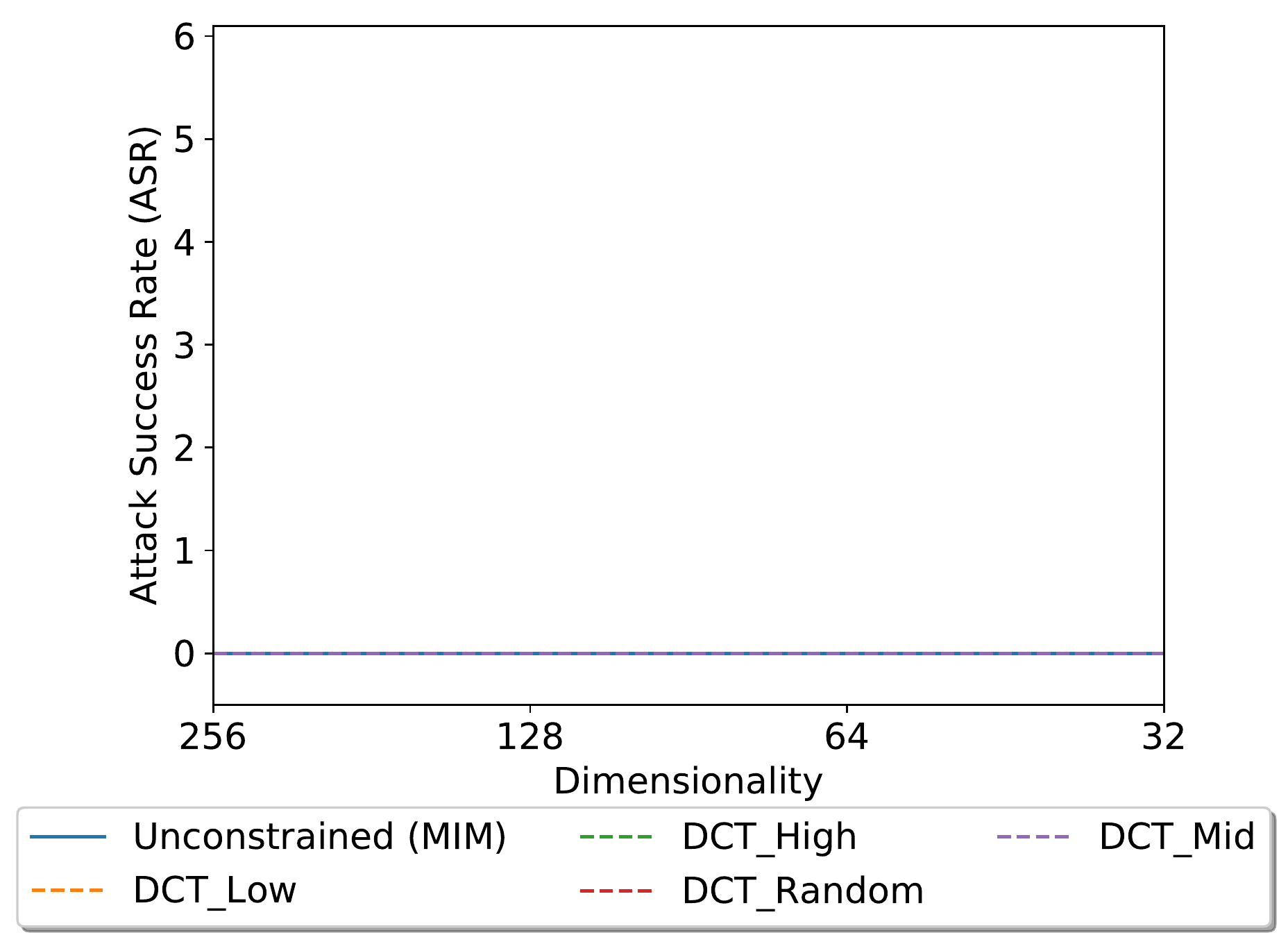}
    \caption{Targeted with $\epsilon=32$ and $\text{iterations}=10$.}
    \label{greybox34}
\end{subfigure}
\caption{\textbf{Black-box} attack from Adv\_1 to D4.}
\label{blackbox35}
\end{figure*}

\begin{figure*}
\begin{subfigure}{0.3\linewidth}
    \centering
    \includegraphics[width=\textwidth]{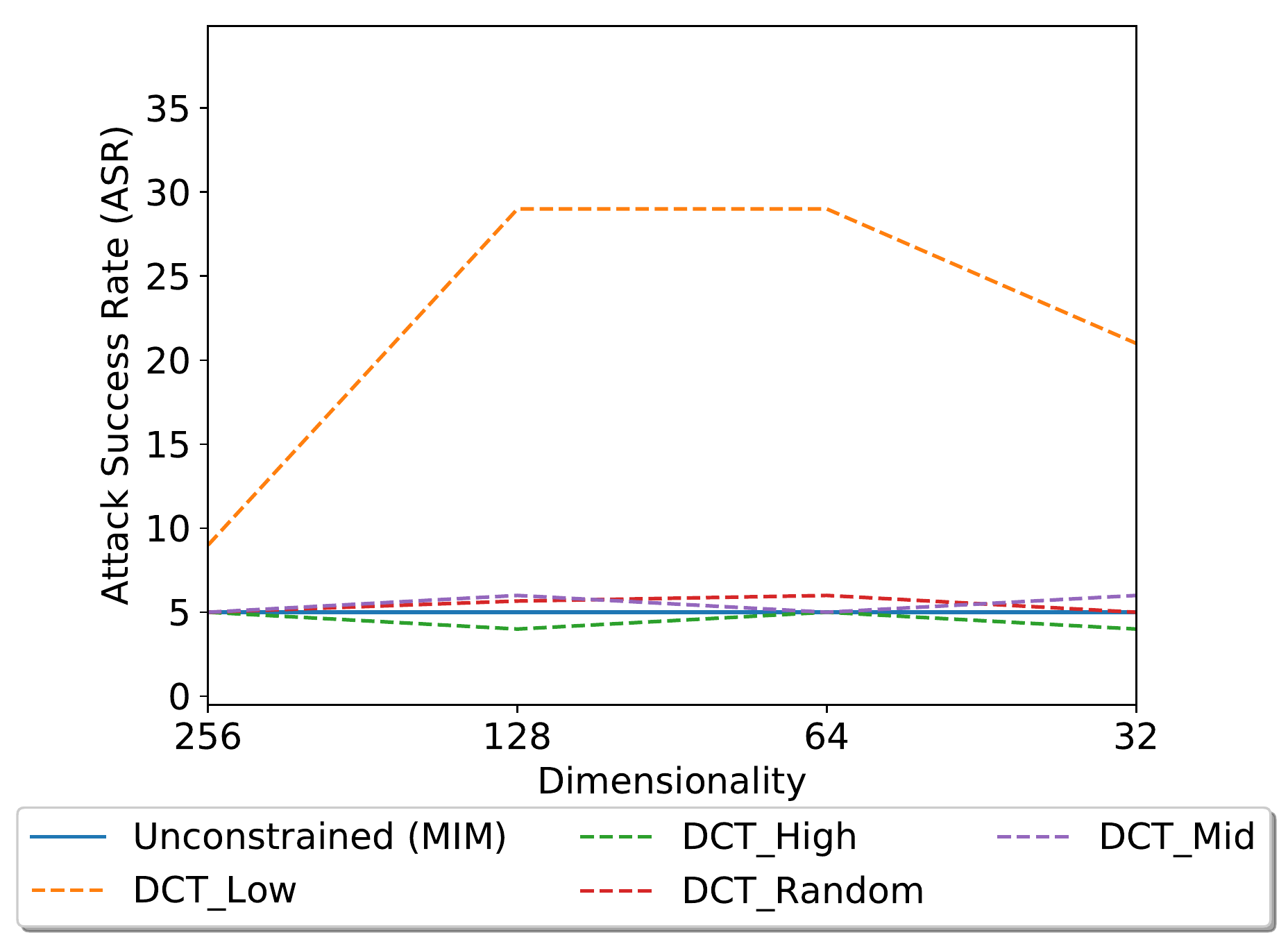}
    \caption{Non-targeted with $\epsilon=16$ and $\text{iterations}=1$.}
    \label{greybox14}
\end{subfigure}
~~~
\begin{subfigure}{0.3\linewidth}
    \centering
    \includegraphics[width=\textwidth]{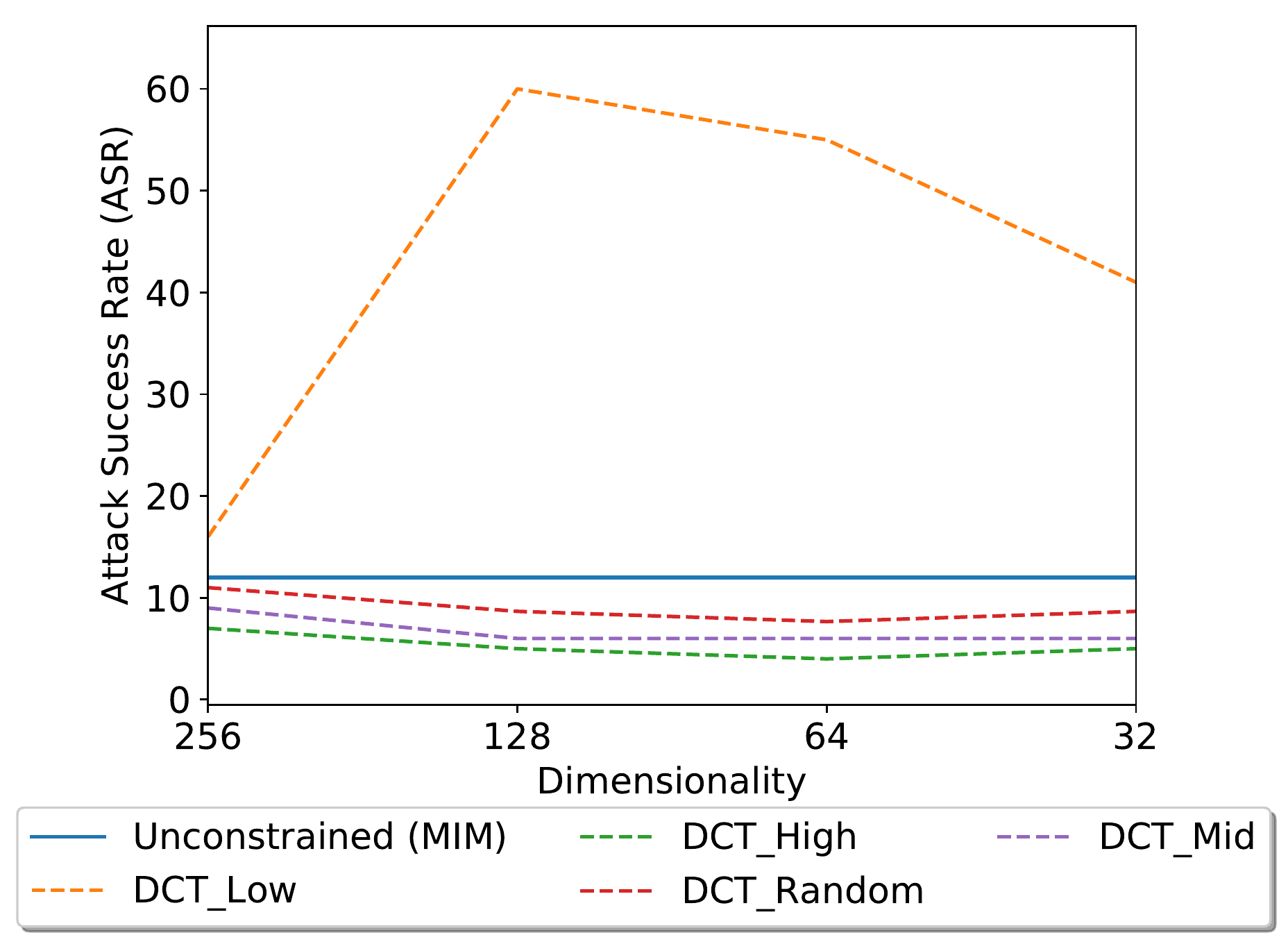}
    \caption{Non-targeted with $\epsilon=16$ and $\text{iterations}=10$.}
    \label{greybox24}
\end{subfigure}
~~~
\begin{subfigure}{0.3\linewidth}
    \centering
    \includegraphics[width=\textwidth]{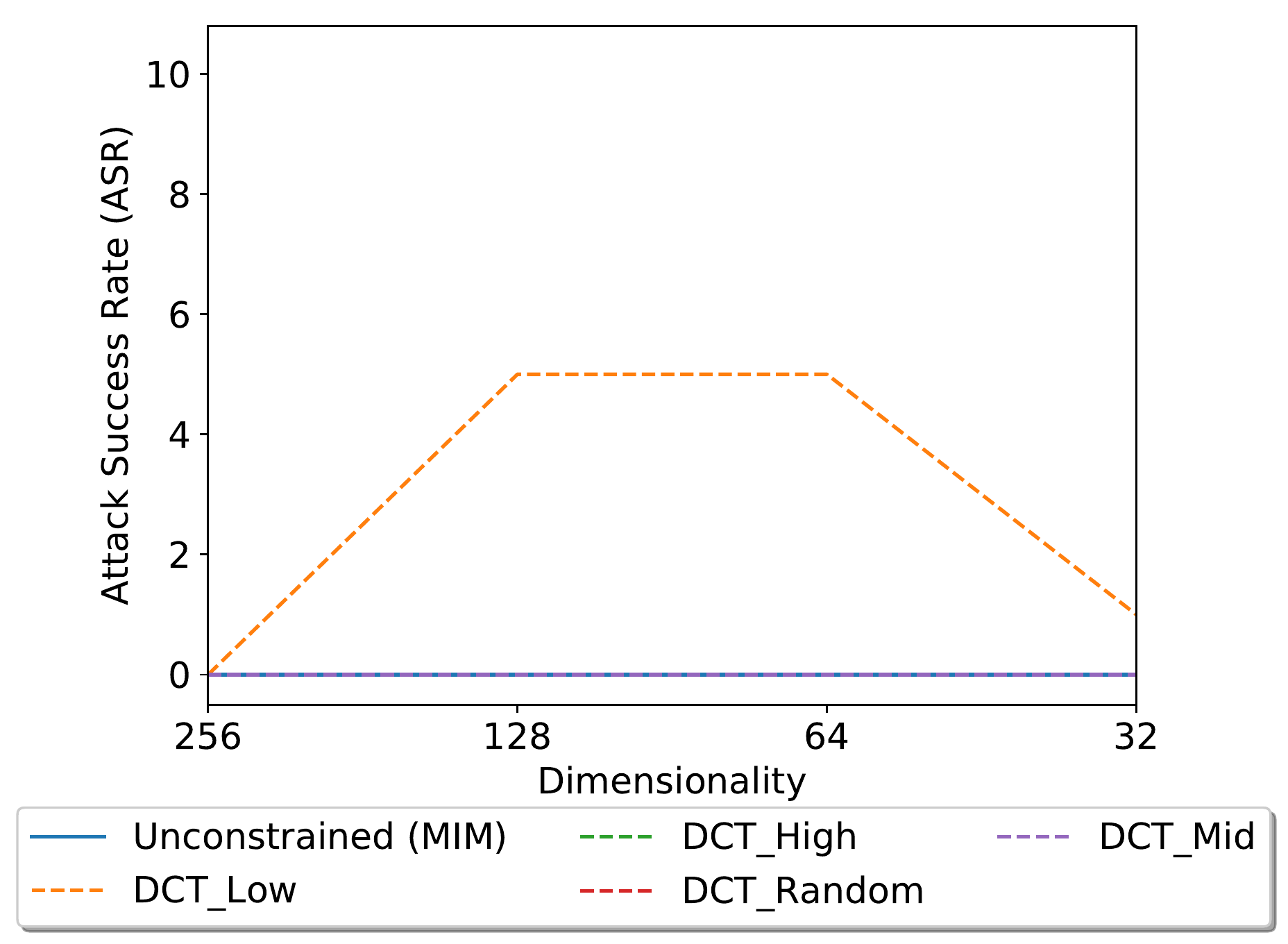}
    \caption{Targeted with $\epsilon=32$ and $\text{iterations}=10$.}
    \label{greybox34}
\end{subfigure}
\caption{\textbf{Black-box} attack from Adv\_3 to EnsAdv.}
\label{blackbox41}
\end{figure*}

\begin{figure*}
\begin{subfigure}{0.3\linewidth}
    \centering
    \includegraphics[width=\textwidth]{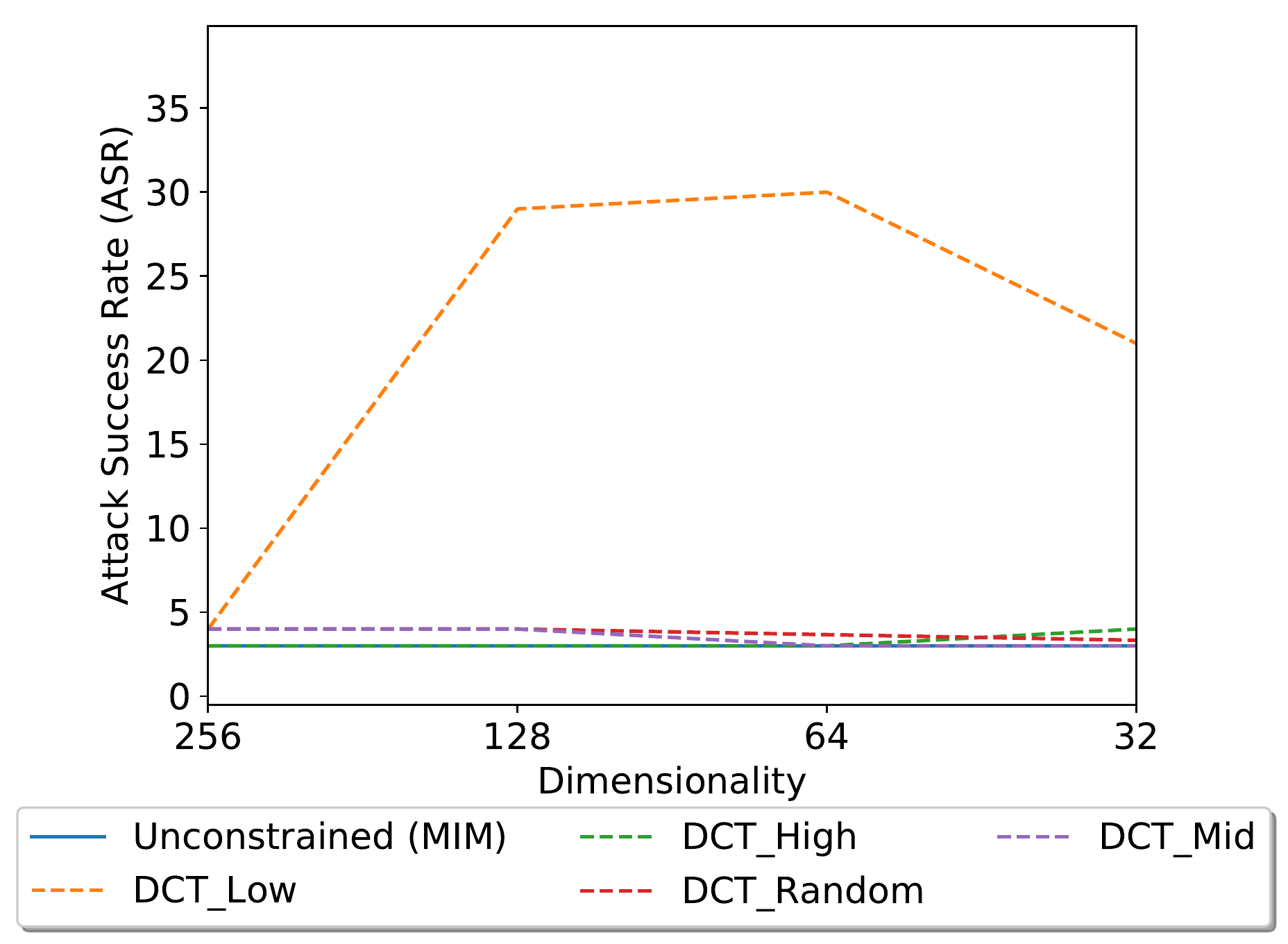}
    \caption{Non-targeted with $\epsilon=16$ and $\text{iterations}=1$.}
    \label{greybox14}
\end{subfigure}
~~~
\begin{subfigure}{0.3\linewidth}
    \centering
    \includegraphics[width=\textwidth]{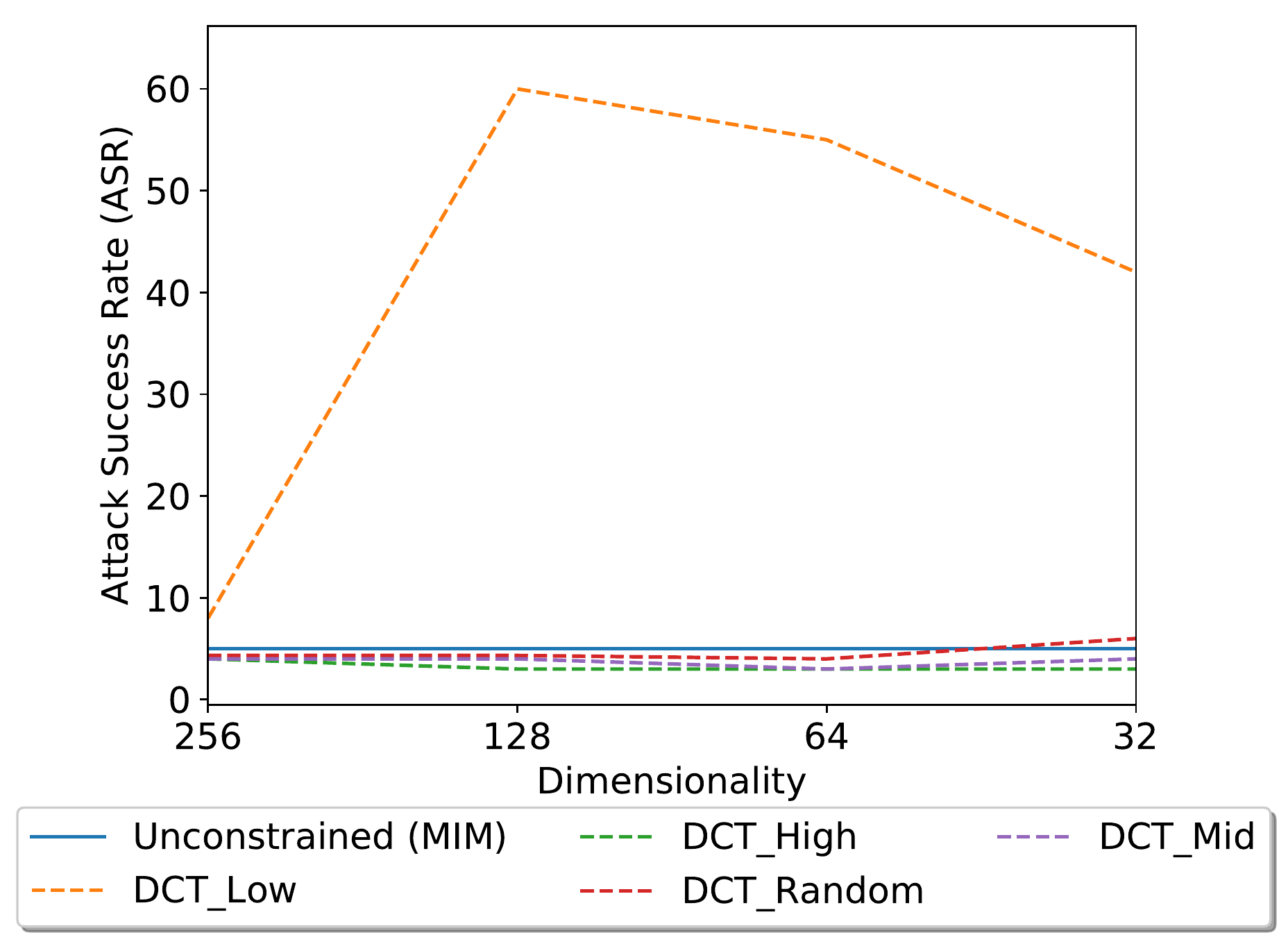}
    \caption{Non-targeted with $\epsilon=16$ and $\text{iterations}=10$.}
    \label{greybox24}
\end{subfigure}
~~~
\begin{subfigure}{0.3\linewidth}
    \centering
    \includegraphics[width=\textwidth]{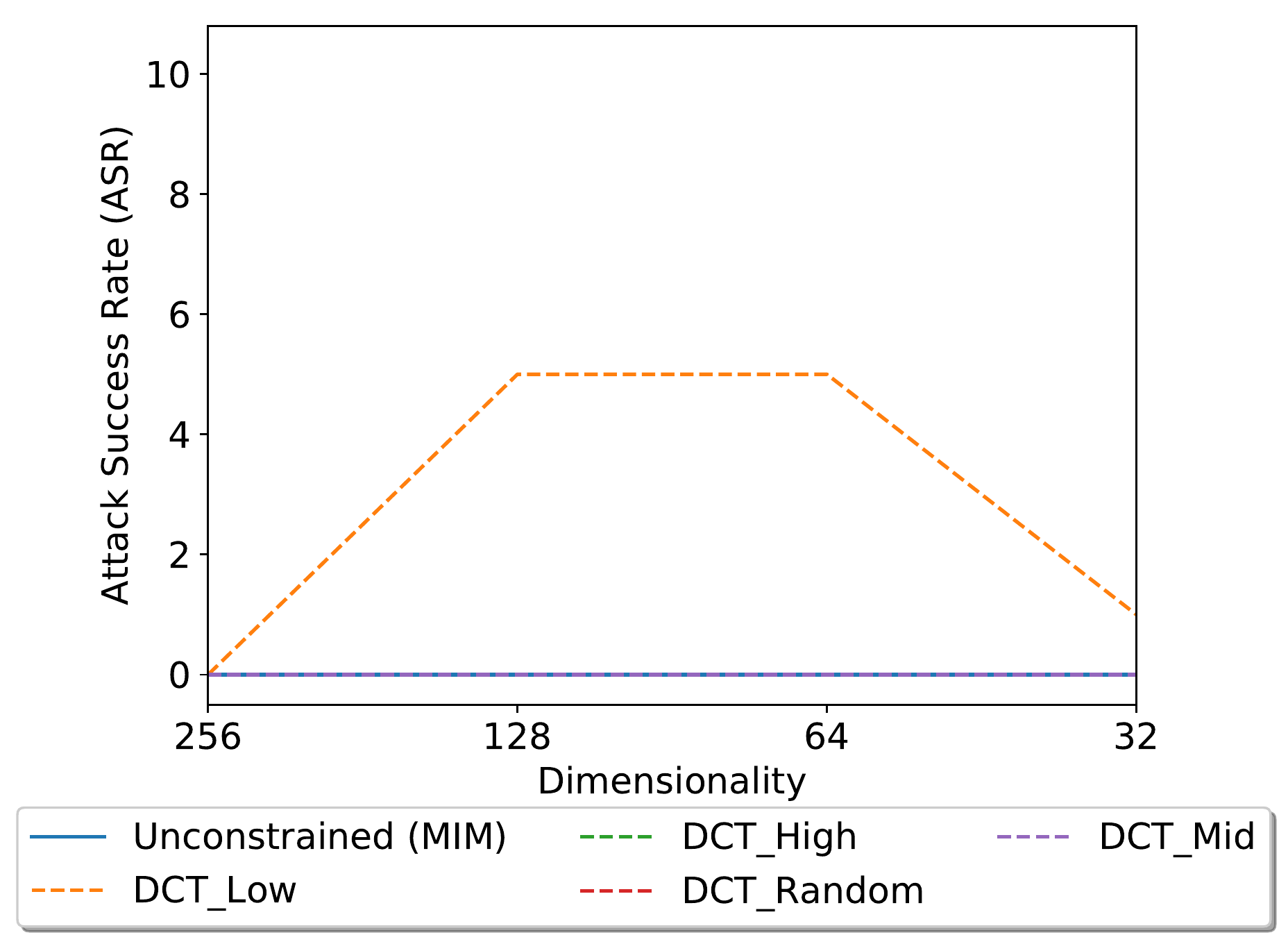}
    \caption{Targeted with $\epsilon=32$ and $\text{iterations}=10$.}
    \label{greybox34}
\end{subfigure}
\caption{\textbf{Black-box} attack from Adv\_3 to D1.}
\label{blackbox42}
\end{figure*}

\begin{figure*}
\begin{subfigure}{0.3\linewidth}
    \centering
    \includegraphics[width=\textwidth]{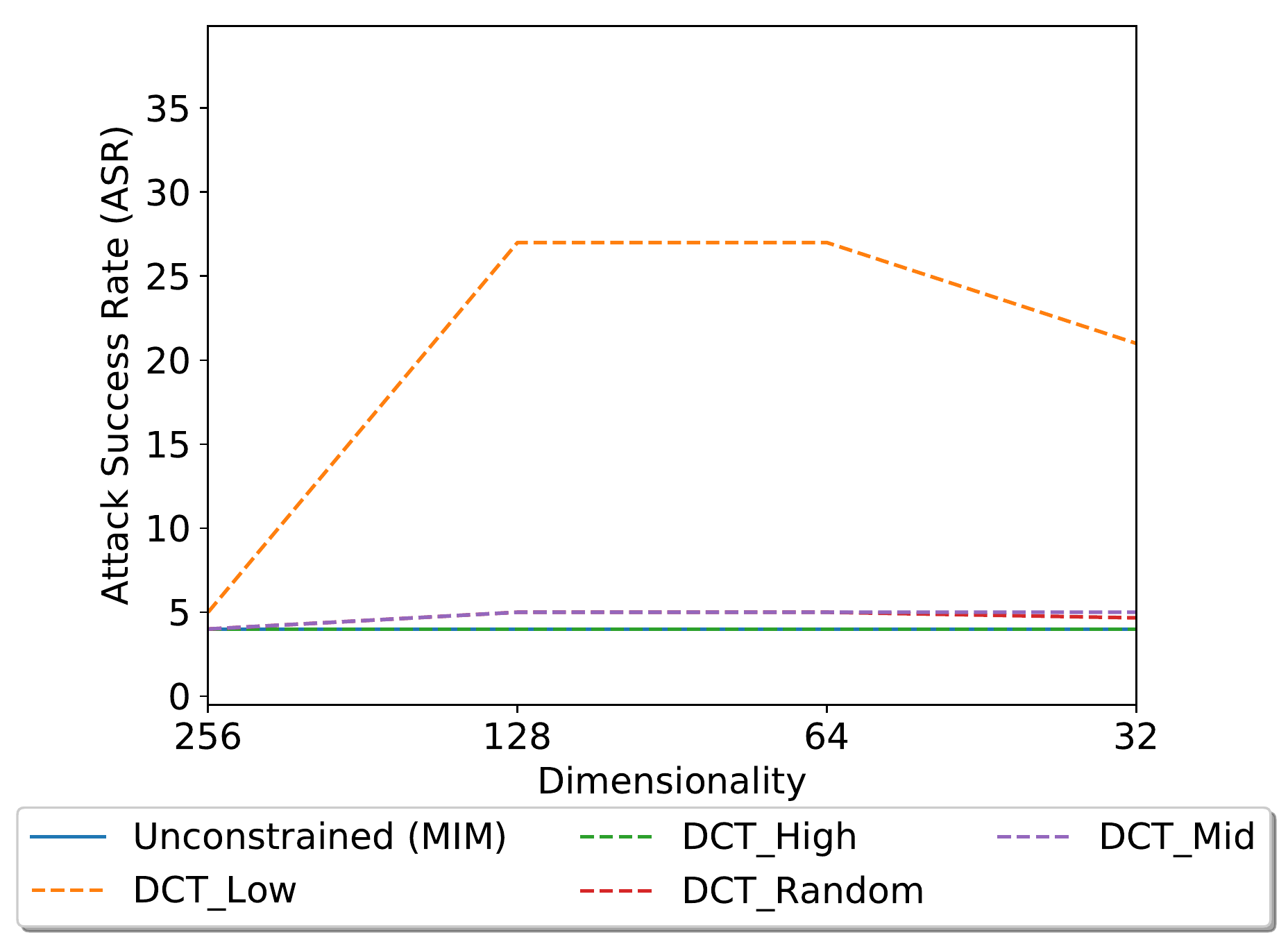}
    \caption{Non-targeted with $\epsilon=16$ and $\text{iterations}=1$.}
    \label{greybox14}
\end{subfigure}
~~~
\begin{subfigure}{0.3\linewidth}
    \centering
    \includegraphics[width=\textwidth]{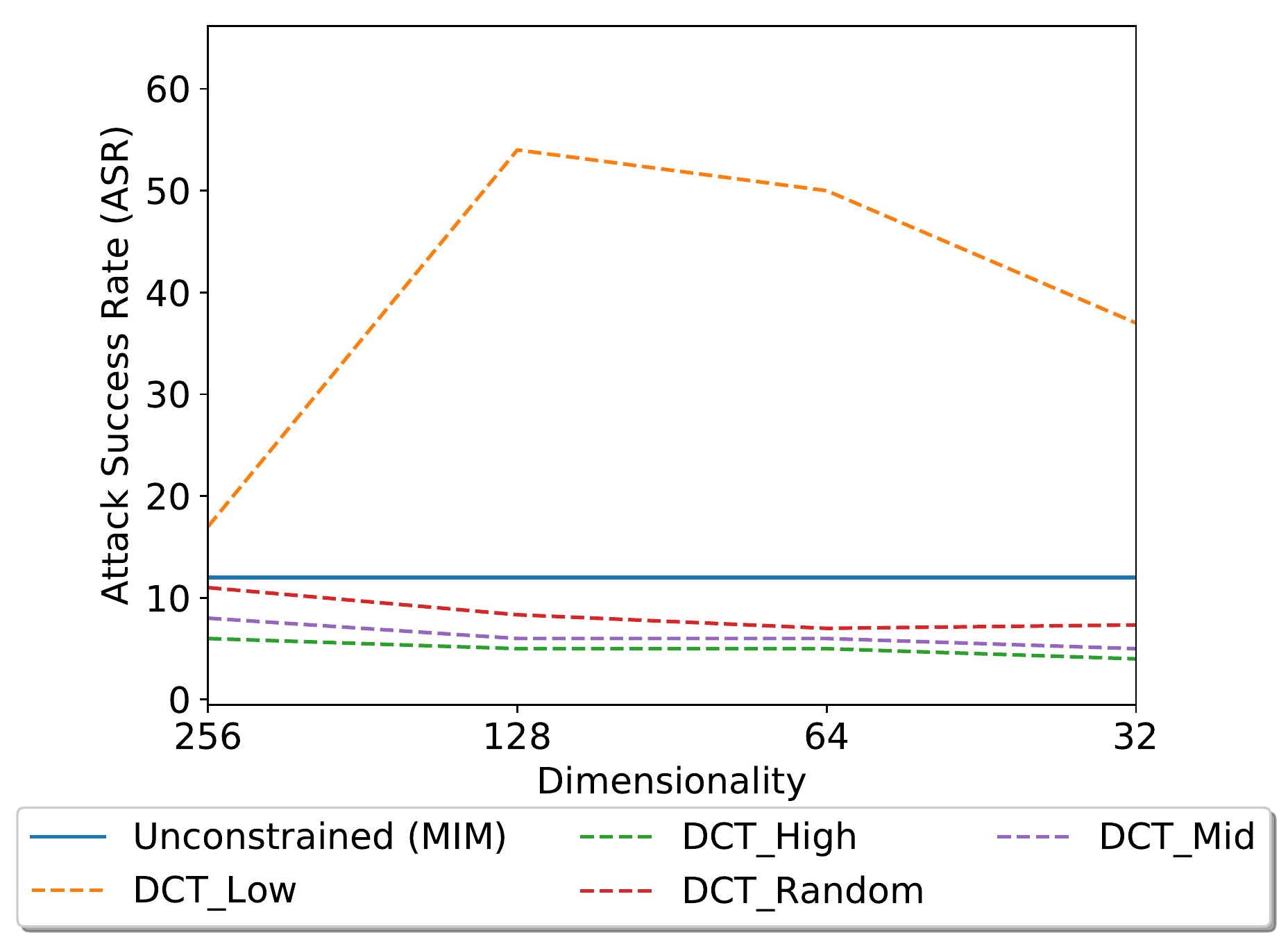}
    \caption{Non-targeted with $\epsilon=16$ and $\text{iterations}=10$.}
    \label{greybox24}
\end{subfigure}
~~~
\begin{subfigure}{0.3\linewidth}
    \centering
    \includegraphics[width=\textwidth]{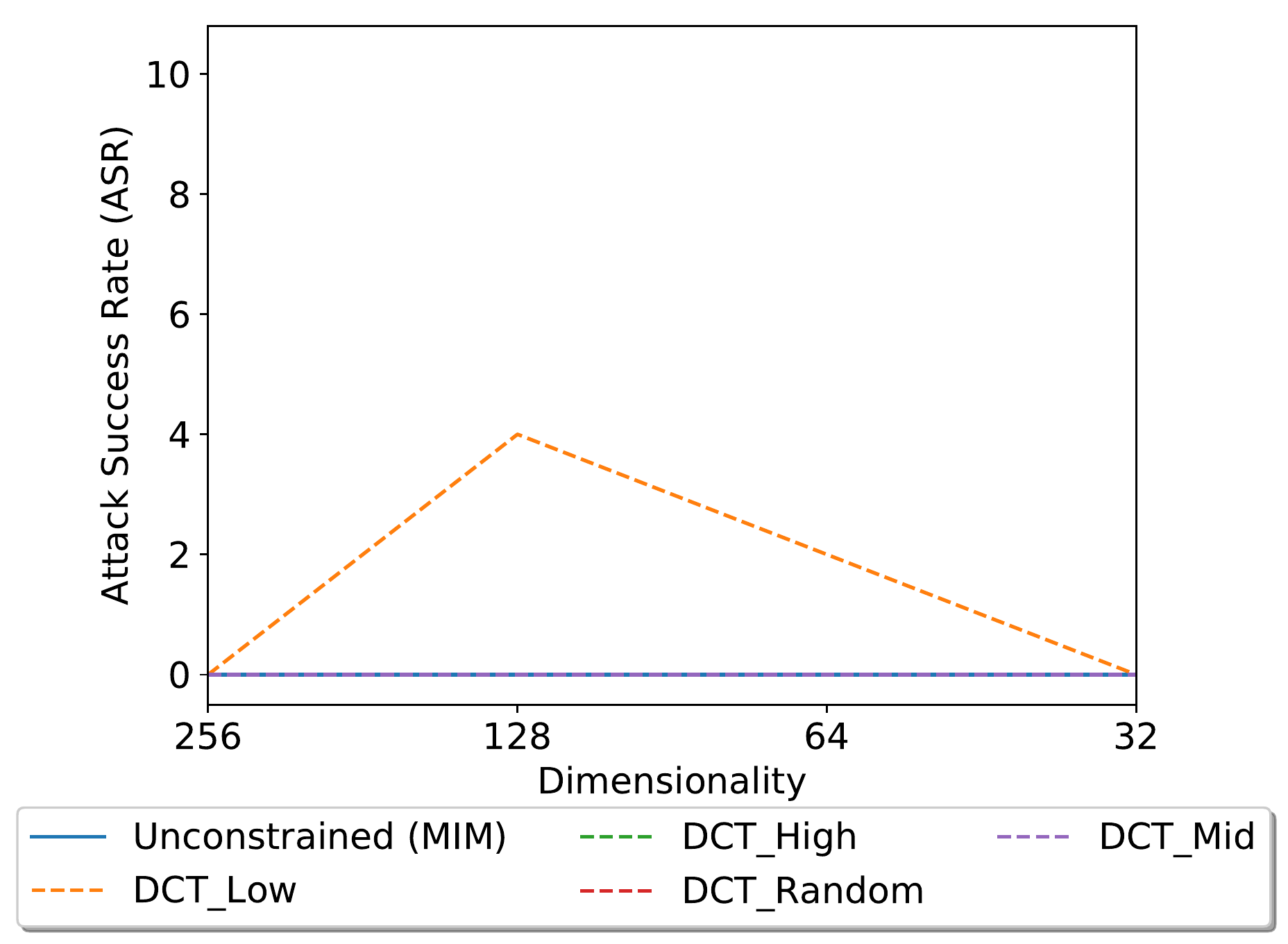}
    \caption{Targeted with $\epsilon=32$ and $\text{iterations}=10$.}
    \label{greybox34}
\end{subfigure}
\caption{\textbf{Black-box} attack from Adv\_3 to D2.}
\label{blackbox43}
\end{figure*}

\begin{figure*}
\begin{subfigure}{0.3\linewidth}
    \centering
    \includegraphics[width=\textwidth]{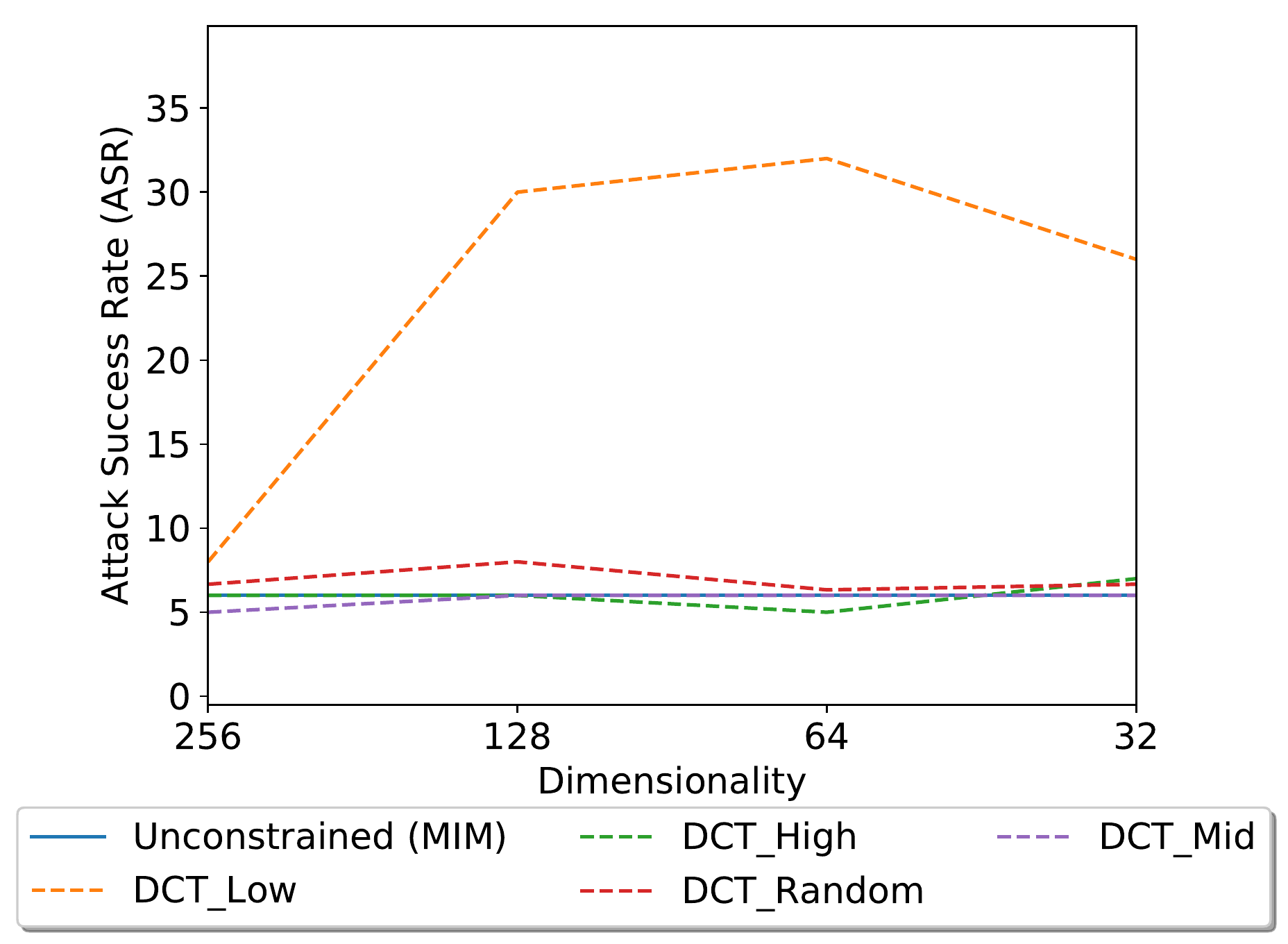}
    \caption{Non-targeted with $\epsilon=16$ and $\text{iterations}=1$.}
    \label{greybox14}
\end{subfigure}
~~~
\begin{subfigure}{0.3\linewidth}
    \centering
    \includegraphics[width=\textwidth]{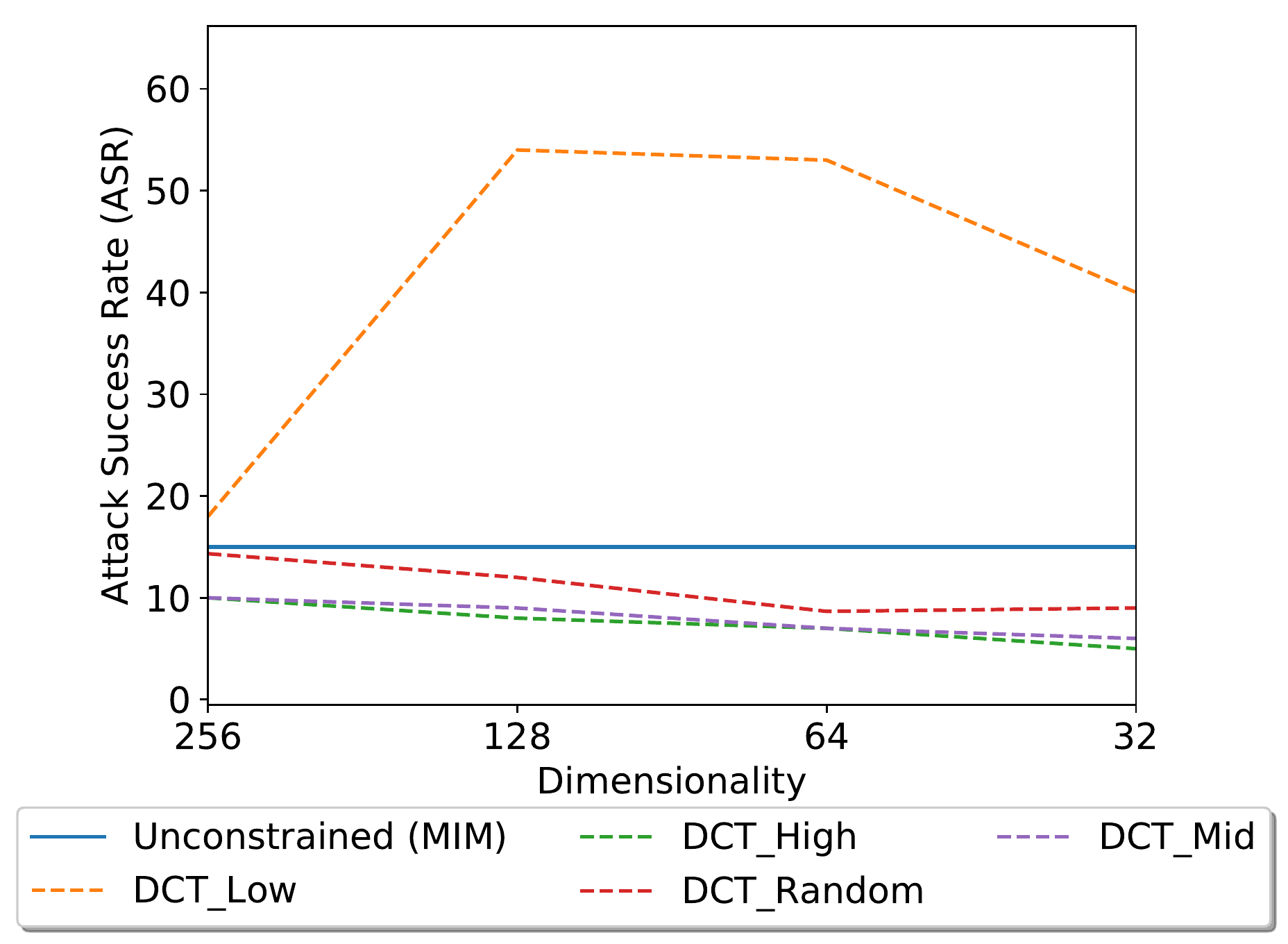}
    \caption{Non-targeted with $\epsilon=16$ and $\text{iterations}=10$.}
    \label{greybox24}
\end{subfigure}
~~~
\begin{subfigure}{0.3\linewidth}
    \centering
    \includegraphics[width=\textwidth]{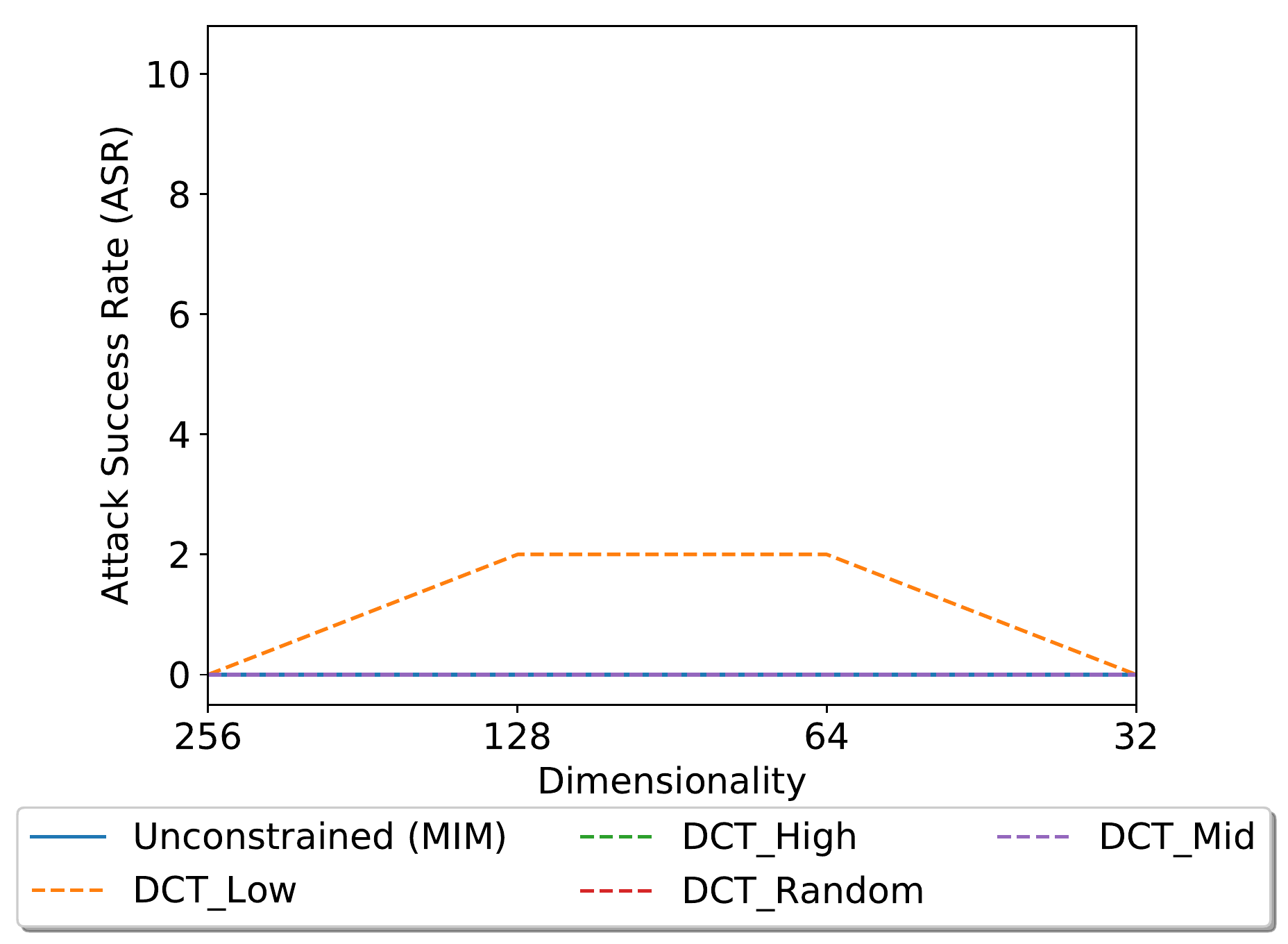}
    \caption{Targeted with $\epsilon=32$ and $\text{iterations}=10$.}
    \label{greybox34}
\end{subfigure}
\caption{\textbf{Black-box} attack from Adv\_3 to D3.}
\label{blackbox44}
\end{figure*}

\begin{figure*}
\begin{subfigure}{0.3\linewidth}
    \centering
    \includegraphics[width=\textwidth]{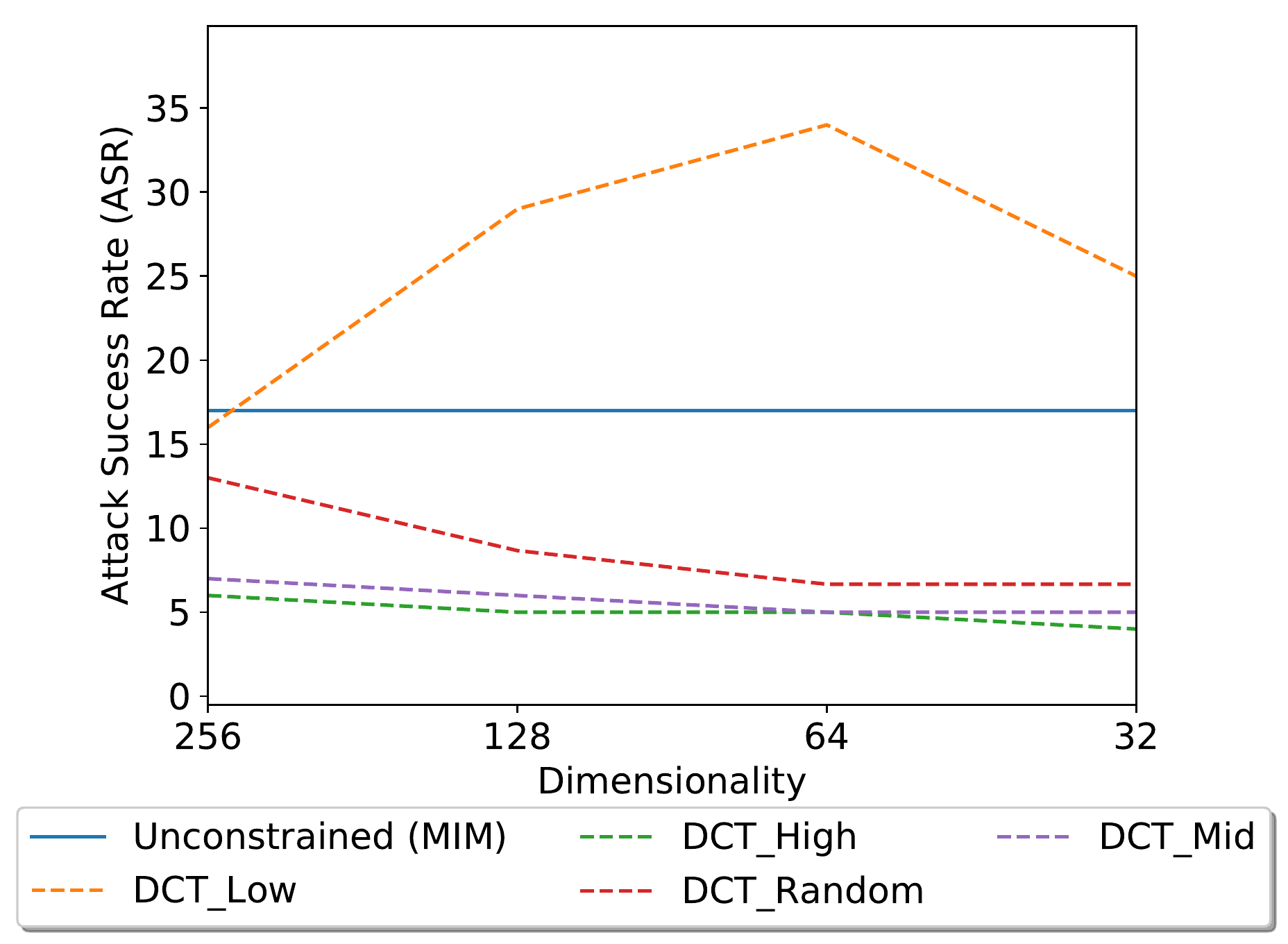}
    \caption{Non-targeted with $\epsilon=16$ and $\text{iterations}=1$.}
    \label{greybox14}
\end{subfigure}
~~~
\begin{subfigure}{0.3\linewidth}
    \centering
    \includegraphics[width=\textwidth]{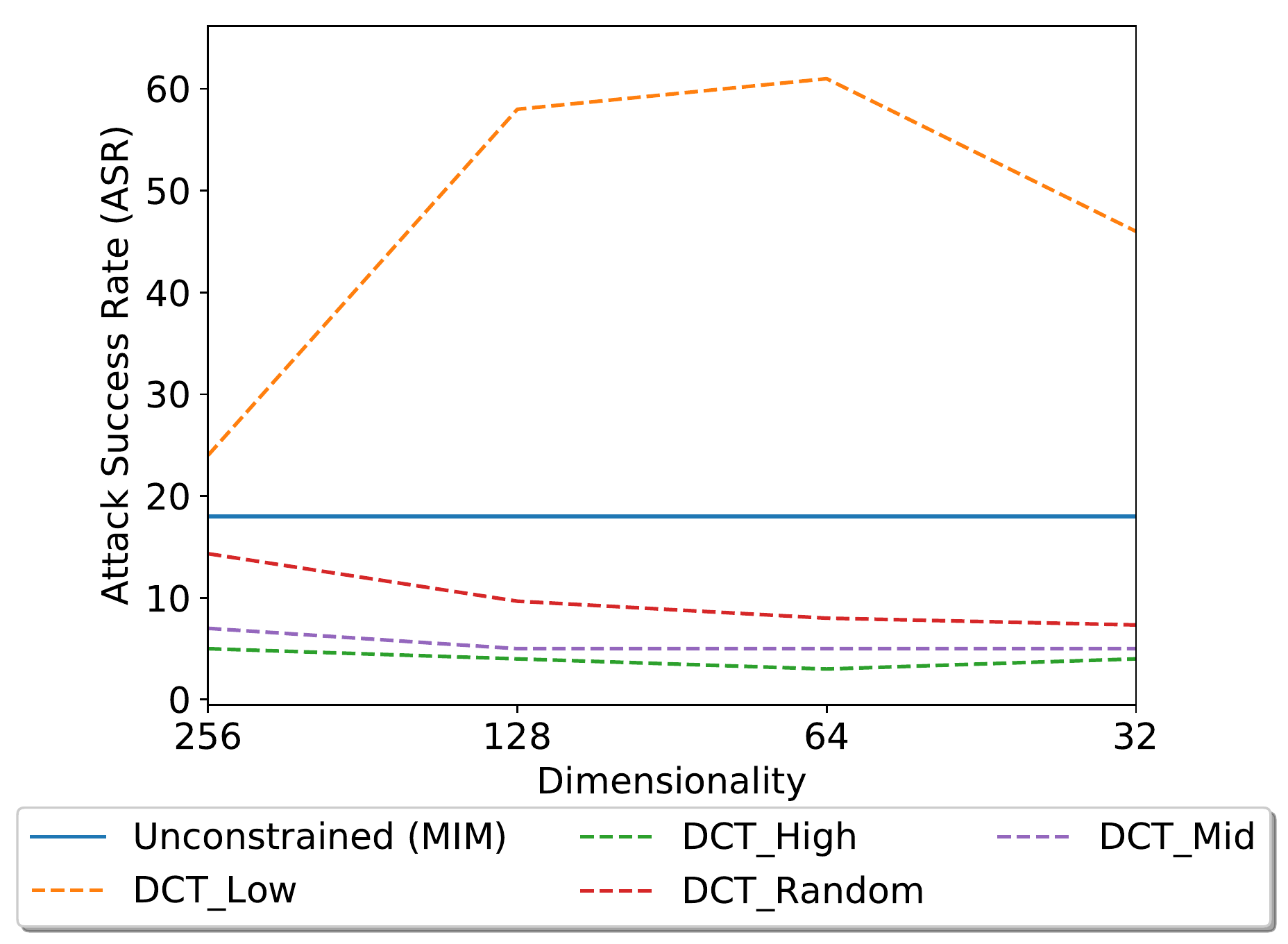}
    \caption{Non-targeted with $\epsilon=16$ and $\text{iterations}=10$.}
    \label{greybox24}
\end{subfigure}
~~~
\begin{subfigure}{0.3\linewidth}
    \centering
    \includegraphics[width=\textwidth]{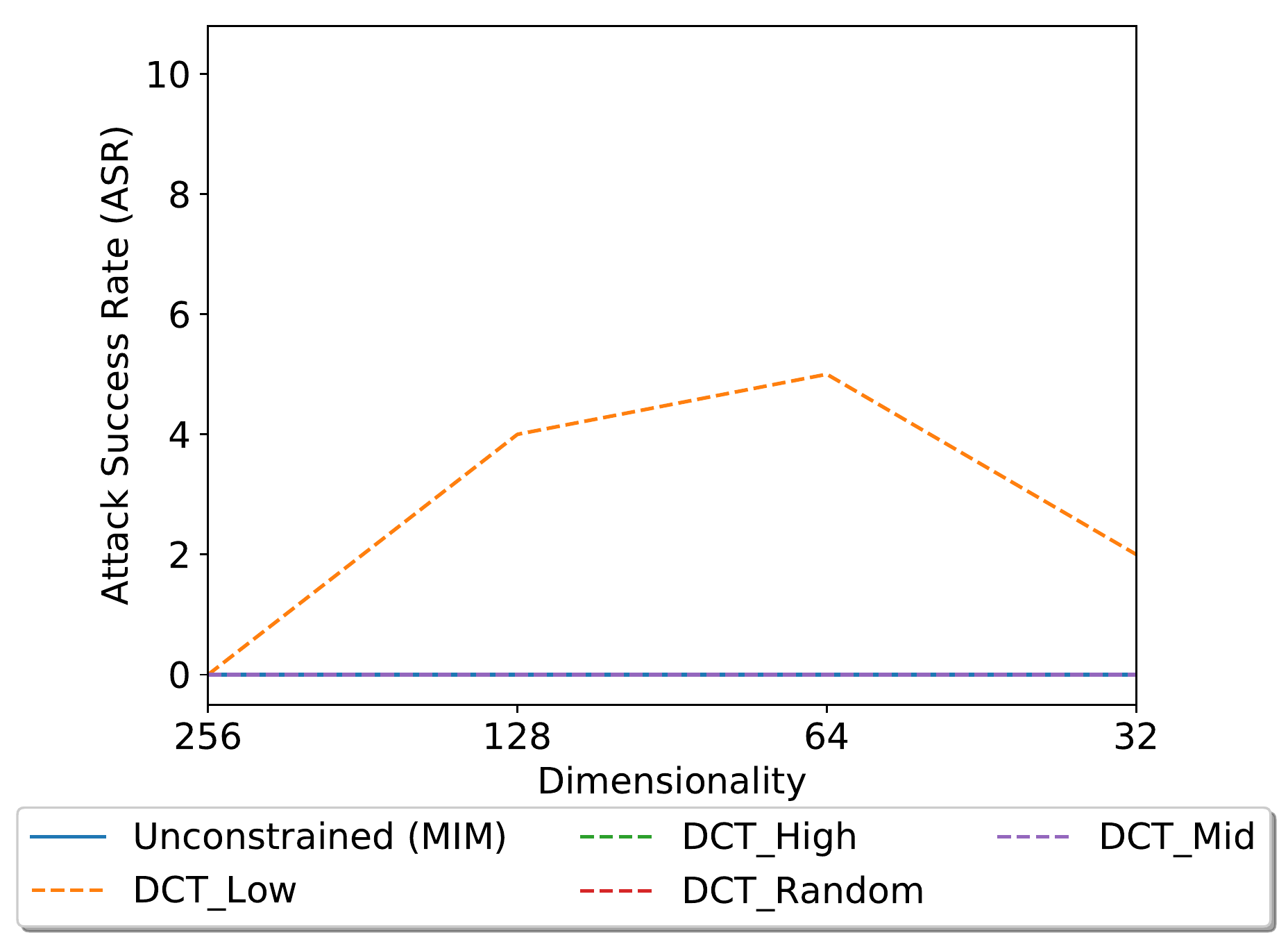}
    \caption{Targeted with $\epsilon=32$ and $\text{iterations}=10$.}
    \label{greybox34}
\end{subfigure}
\caption{\textbf{Black-box} attack from Adv\_3 to D4.}
\label{blackbox45}
\end{figure*}

Figure~\ref{whiteboxs} shows white-box results using \texttt{DCT\_Low} attacking the adversarially trained model. Figures~\ref{greybox1}-\ref{greybox4} shows results against D1, D2, D3, and D4, respectively. Figures [\ref{blackbox11}-\ref{blackbox15}, \ref{blackbox21}-\ref{blackbox25}, \ref{blackbox31}-\ref{blackbox35}, \ref{blackbox41}-\ref{blackbox45}] shows results transferring from each of the source models [Cln\_1,Cln\_3,Adv\_1,Adv\_3] to each of the target defenses [EnsAdv,D1,D2,D3,D4]. 

\begin{figure*}
\begin{subfigure}{0.3\linewidth}
    \centering
    \includegraphics[width=\textwidth]{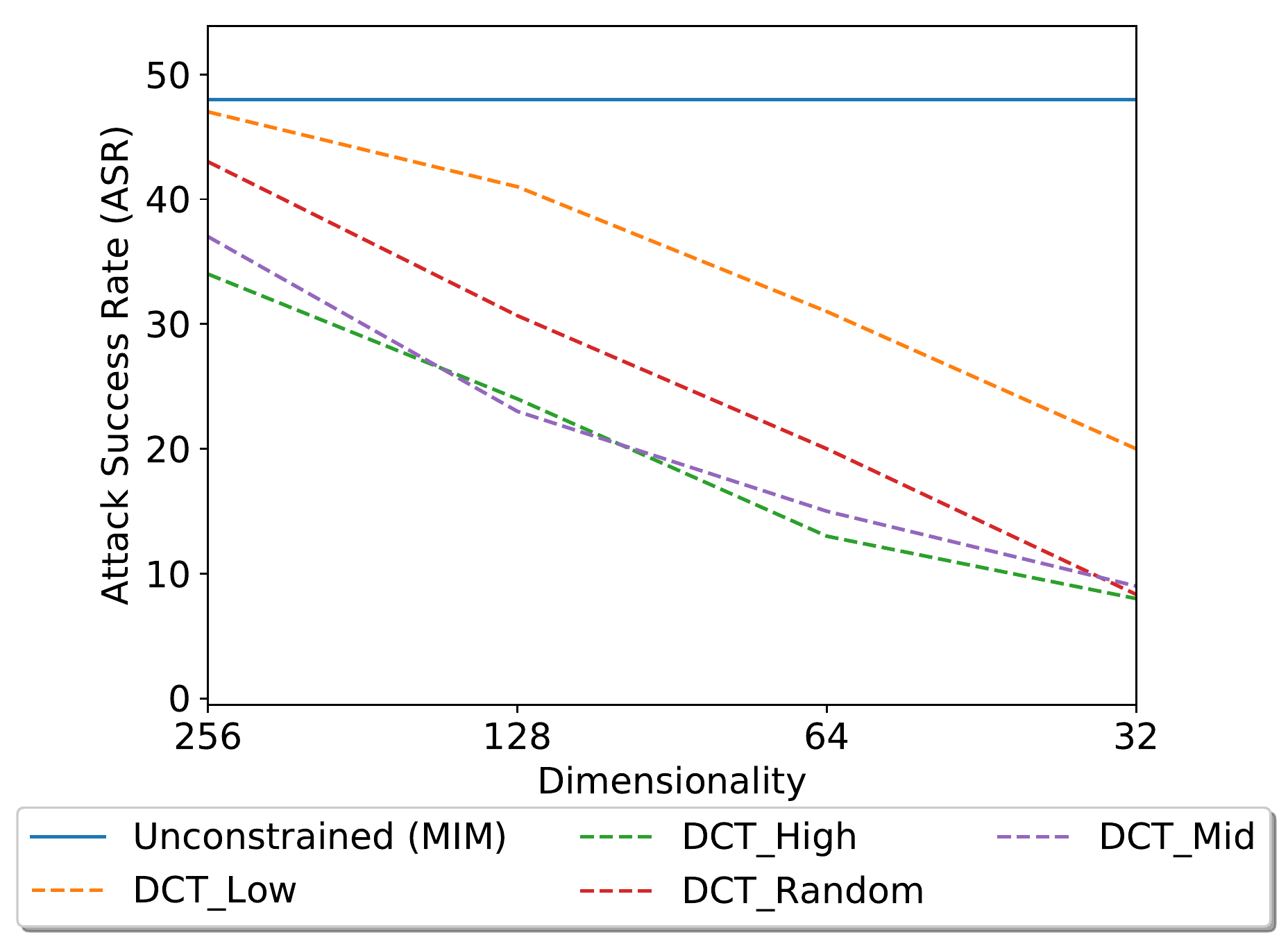}
    \caption{Non-targeted with $\epsilon=16$ and $\text{iterations}=1$.}
    \label{greybox14}
\end{subfigure}
~~~
\begin{subfigure}{0.3\linewidth}
    \centering
    \includegraphics[width=\textwidth]{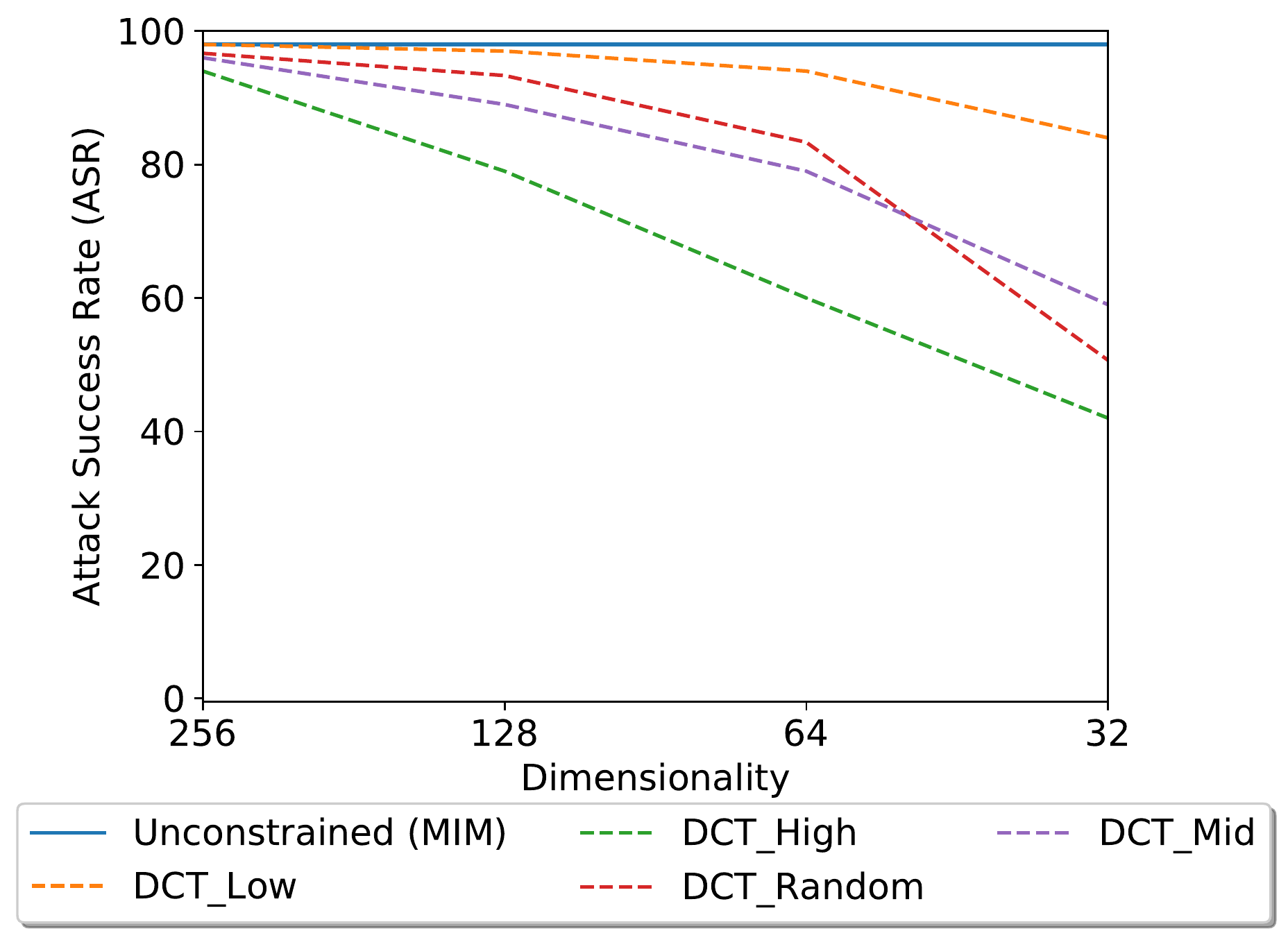}
    \caption{Non-targeted with $\epsilon=16$ and $\text{iterations}=10$.}
    \label{greybox24}
\end{subfigure}
~~~
\begin{subfigure}{0.3\linewidth}
    \centering
    \includegraphics[width=\textwidth]{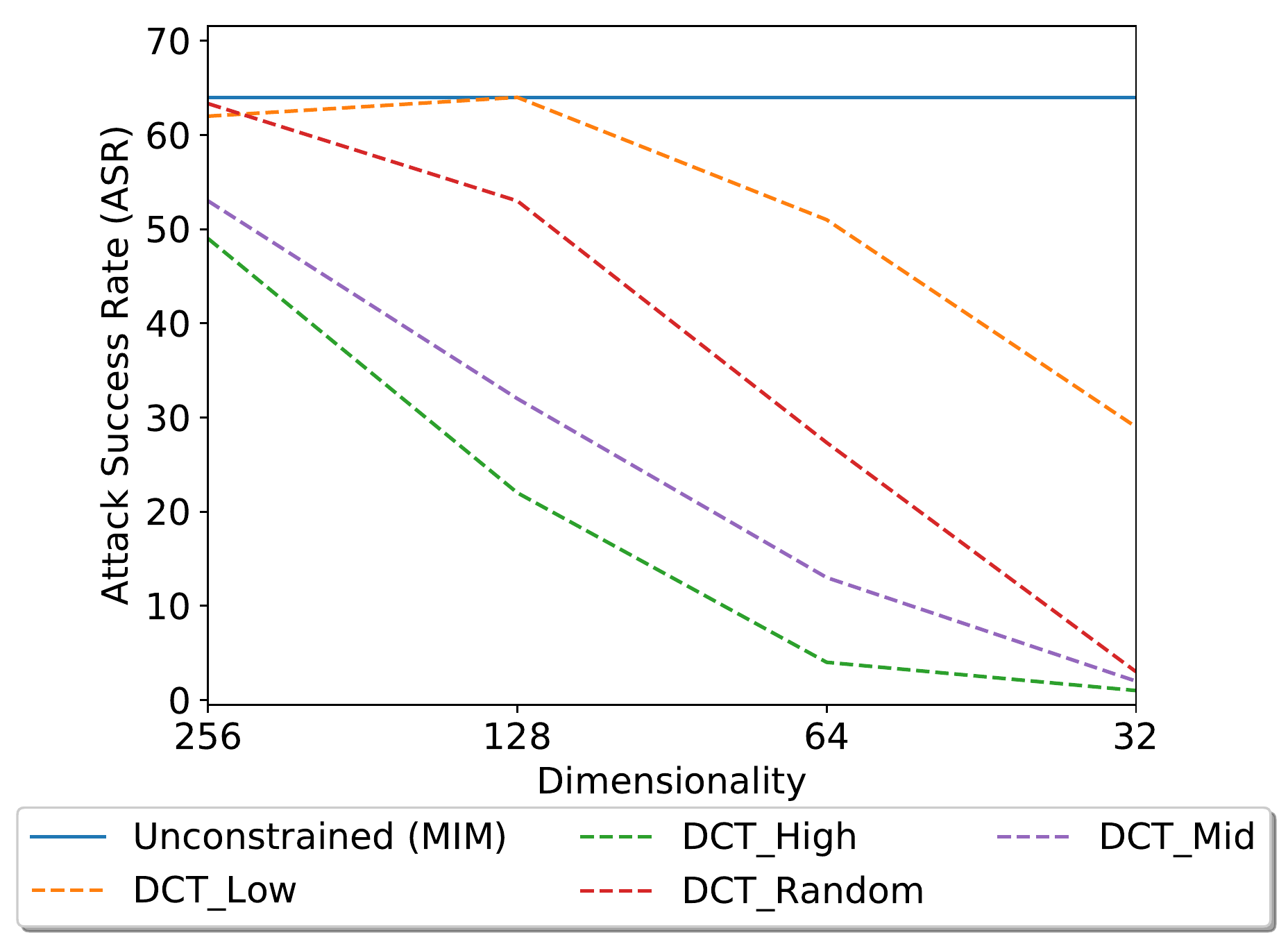}
    \caption{Targeted with $\epsilon=32$ and $\text{iterations}=10$.}
    \label{greybox34}
\end{subfigure}
\caption{\textbf{White-box} attack on cleanly trained model.}
\label{whiteboxc}
\end{figure*}

\begin{figure*}
\begin{subfigure}{0.3\linewidth}
    \centering
    \includegraphics[width=\textwidth]{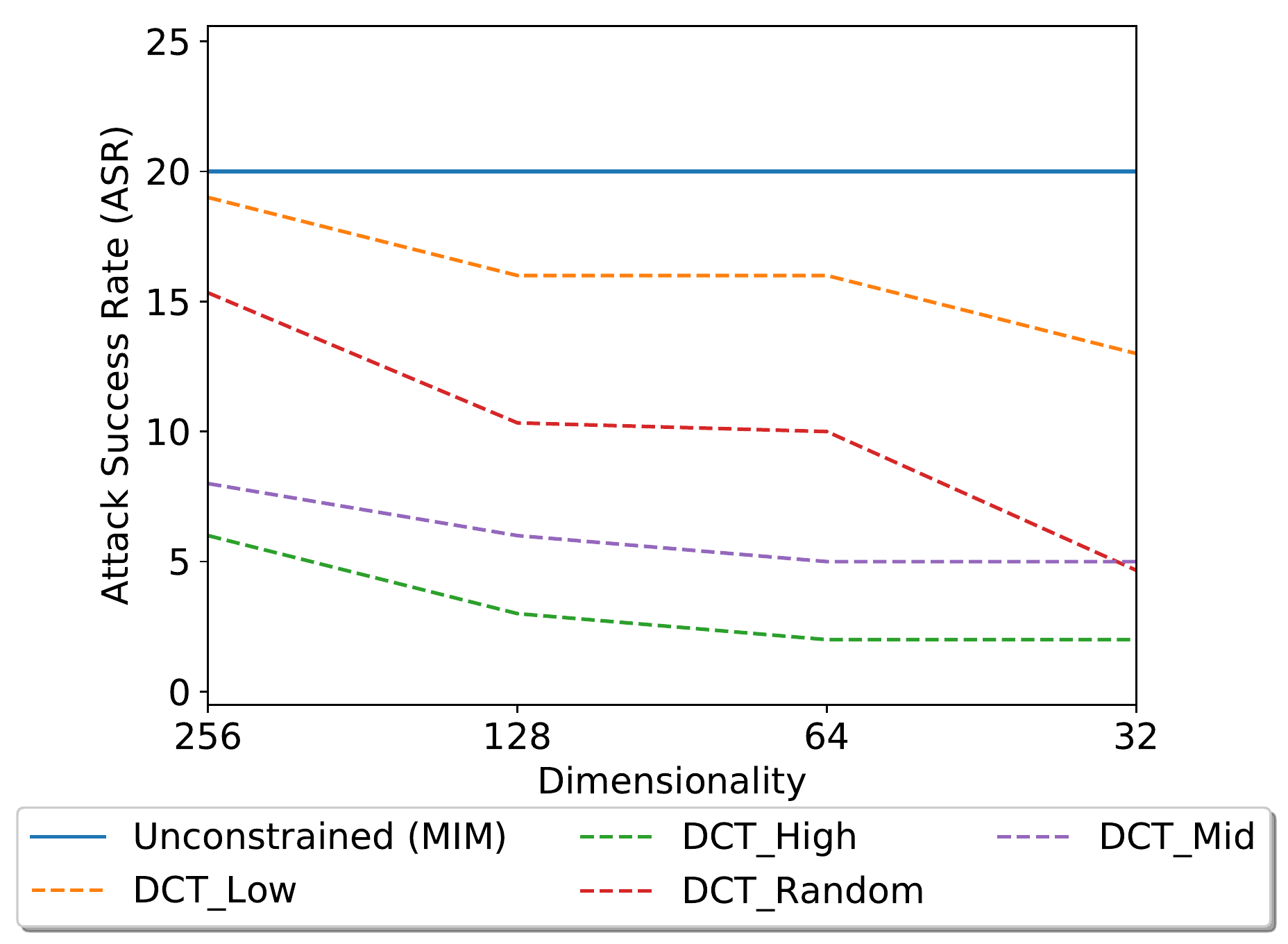}
    \caption{Non-targeted with $\epsilon=16$ and $\text{iterations}=1$.}
    \label{greybox14}
\end{subfigure}
~~~
\begin{subfigure}{0.3\linewidth}
    \centering
    \includegraphics[width=\textwidth]{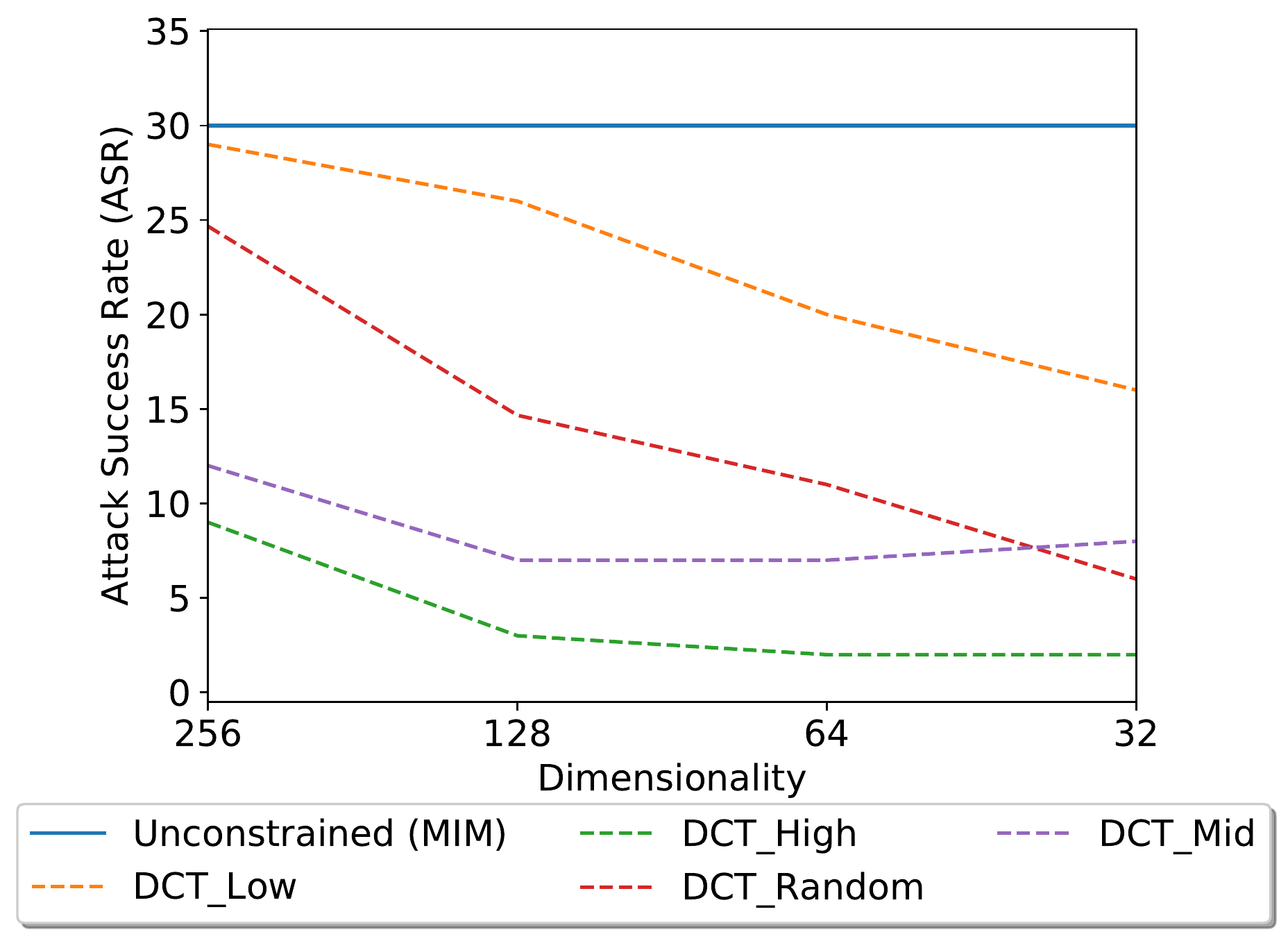}
    \caption{Non-targeted with $\epsilon=16$ and $\text{iterations}=10$.}
    \label{greybox24}
\end{subfigure}
~~~
\begin{subfigure}{0.3\linewidth}
    \centering
    \includegraphics[width=\textwidth]{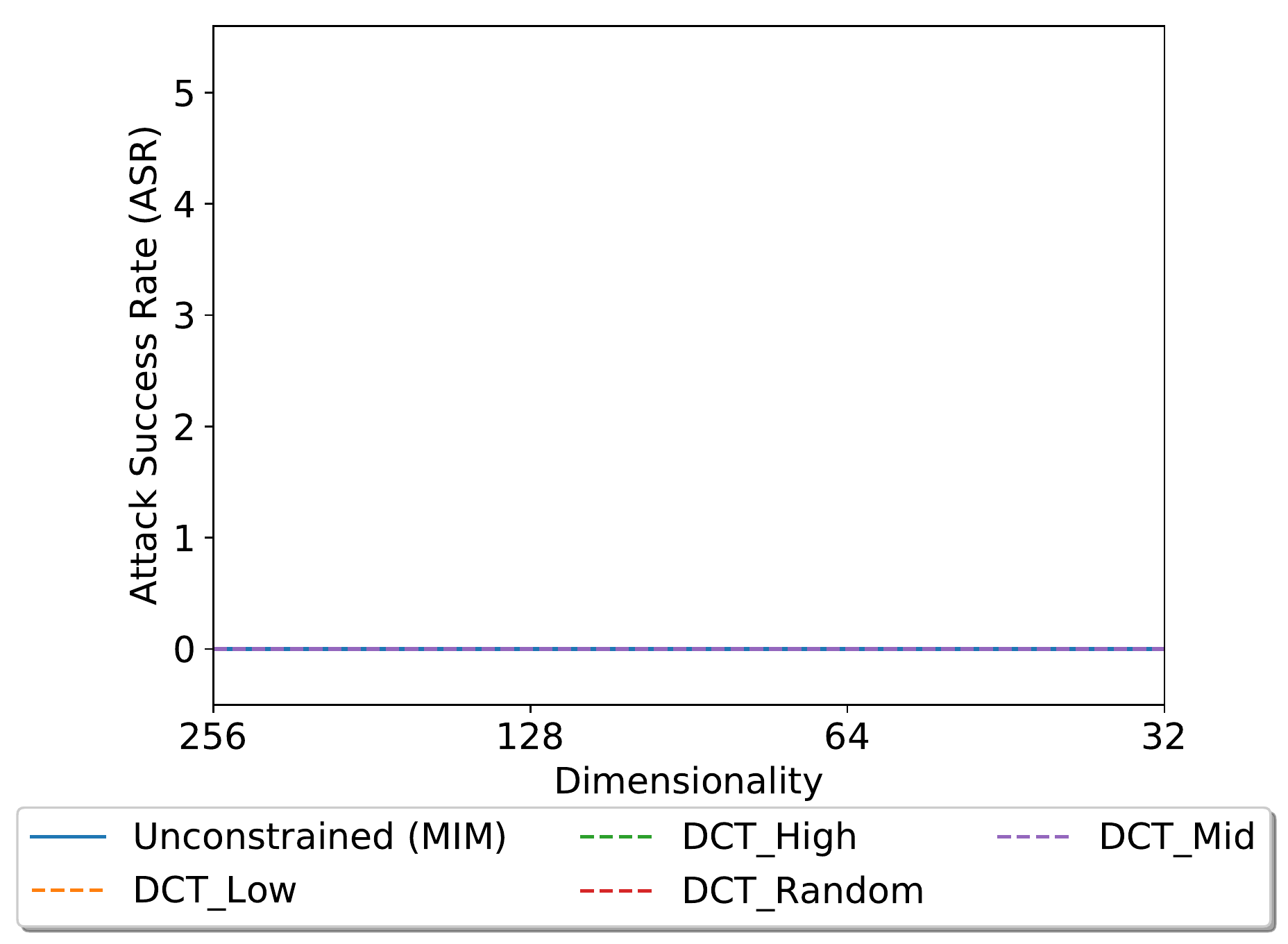}
    \caption{Targeted with $\epsilon=32$ and $\text{iterations}=10$.}
    \label{greybox34}
\end{subfigure}
\caption{\textbf{Black-box} attack from Cln\_1 to Cln.}
\label{blackbox1c}
\end{figure*}

\begin{figure*}
\begin{subfigure}{0.3\linewidth}
    \centering
    \includegraphics[width=\textwidth]{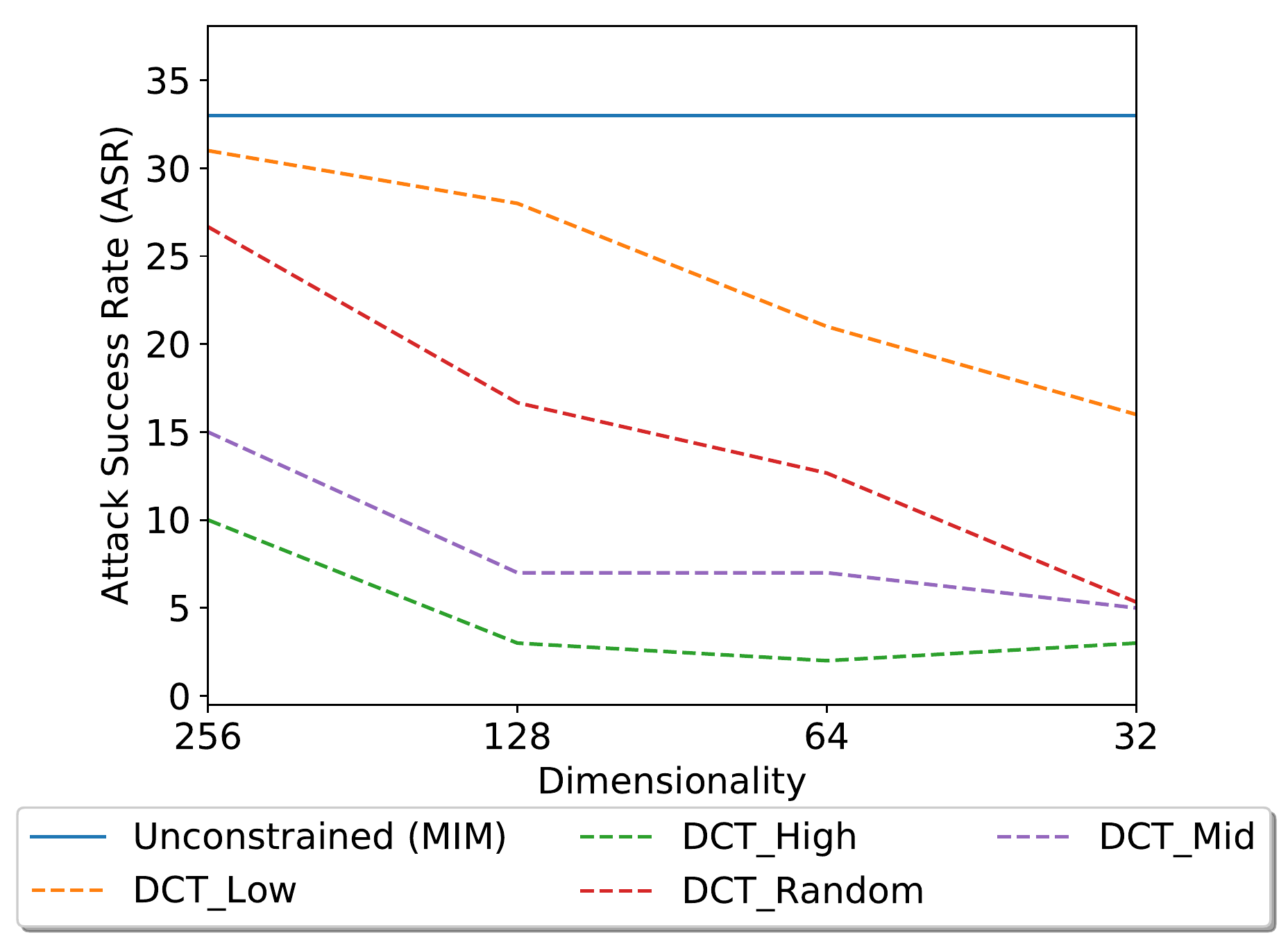}
    \caption{Non-targeted with $\epsilon=16$ and $\text{iterations}=1$.}
    \label{greybox14}
\end{subfigure}
~~~
\begin{subfigure}{0.3\linewidth}
    \centering
    \includegraphics[width=\textwidth]{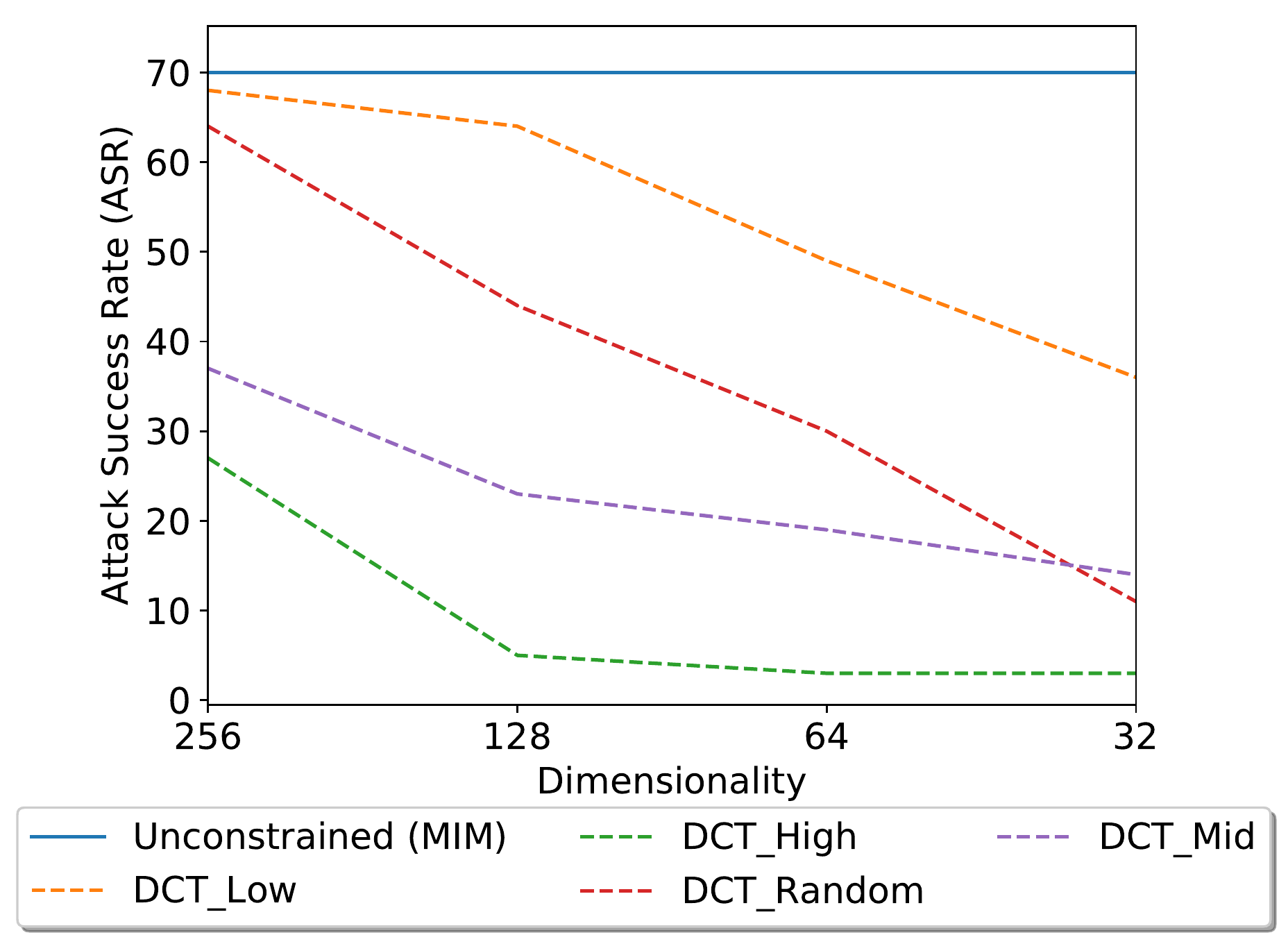}
    \caption{Non-targeted with $\epsilon=16$ and $\text{iterations}=10$.}
    \label{greybox24}
\end{subfigure}
~~~
\begin{subfigure}{0.3\linewidth}
    \centering
    \includegraphics[width=\textwidth]{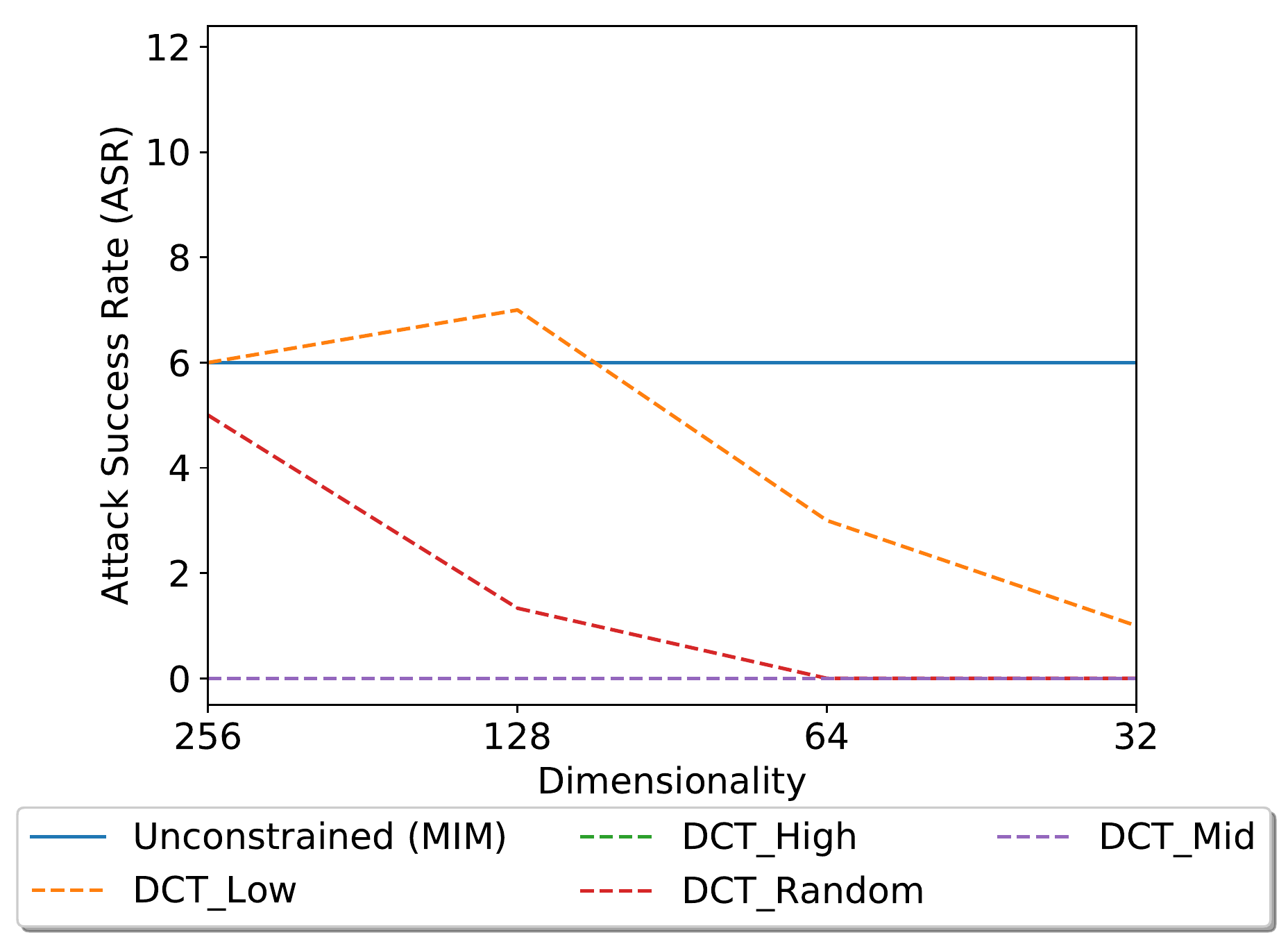}
    \caption{Targeted with $\epsilon=32$ and $\text{iterations}=10$.}
    \label{greybox34}
\end{subfigure}
\caption{\textbf{Black-box} attack from Cln\_3 to Cln.}
\label{blackbox2c}
\end{figure*}

\begin{figure*}
\begin{subfigure}{0.3\linewidth}
    \centering
    \includegraphics[width=\textwidth]{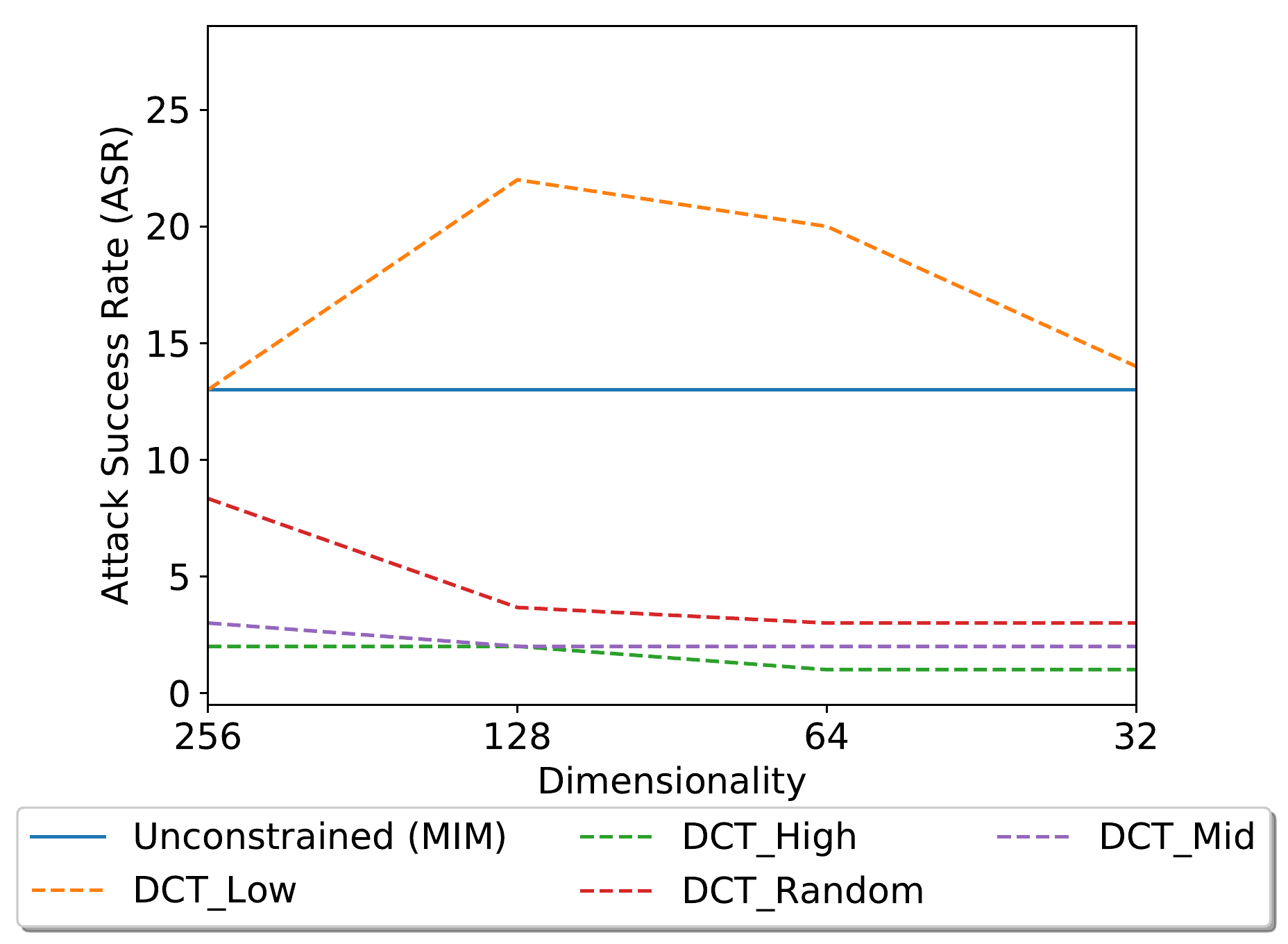}
    \caption{Non-targeted with $\epsilon=16$ and $\text{iterations}=1$.}
    \label{greybox14}
\end{subfigure}
~~~
\begin{subfigure}{0.3\linewidth}
    \centering
    \includegraphics[width=\textwidth]{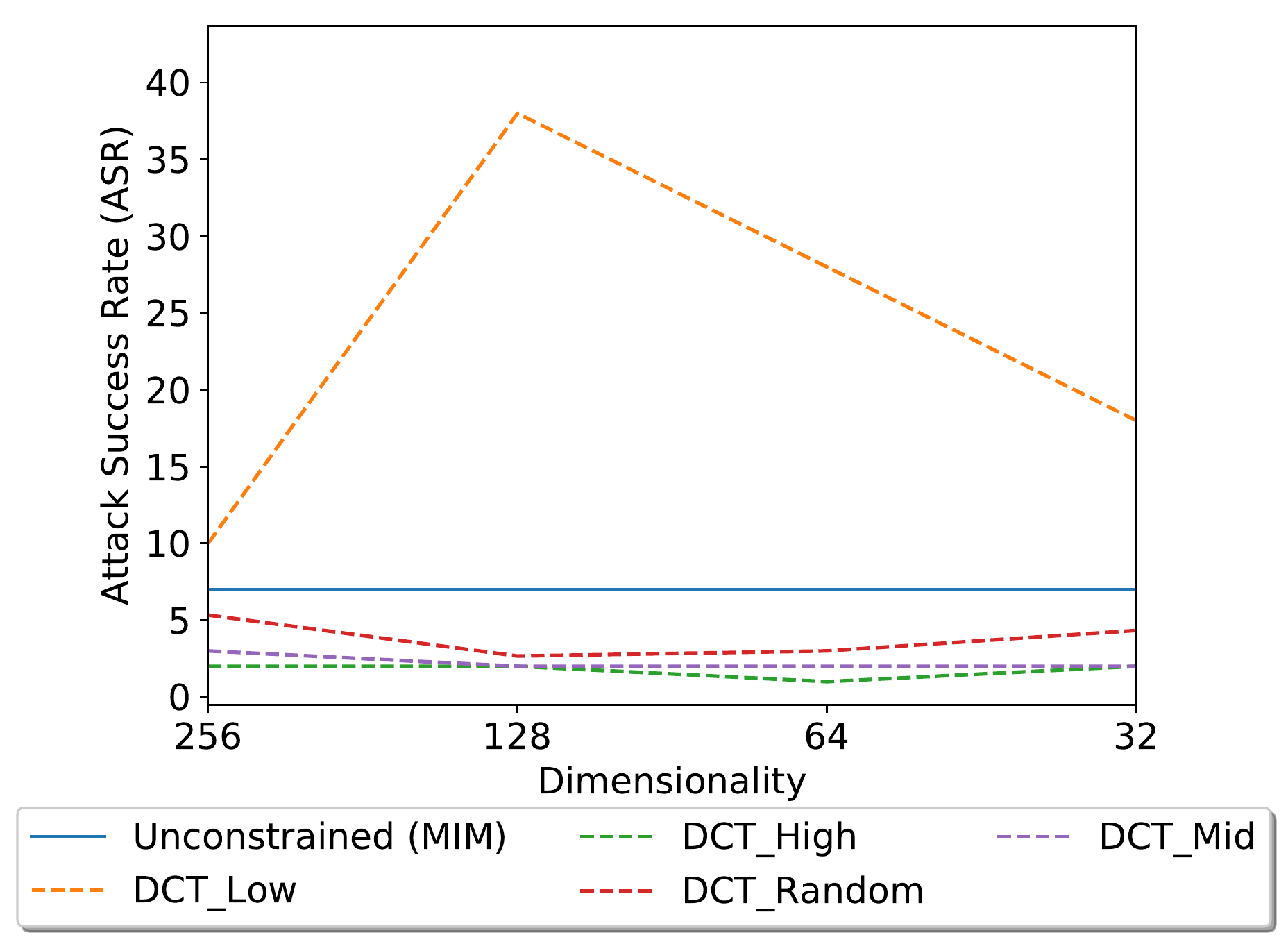}
    \caption{Non-targeted with $\epsilon=16$ and $\text{iterations}=10$.}
    \label{greybox24}
\end{subfigure}
~~~
\begin{subfigure}{0.3\linewidth}
    \centering
    \includegraphics[width=\textwidth]{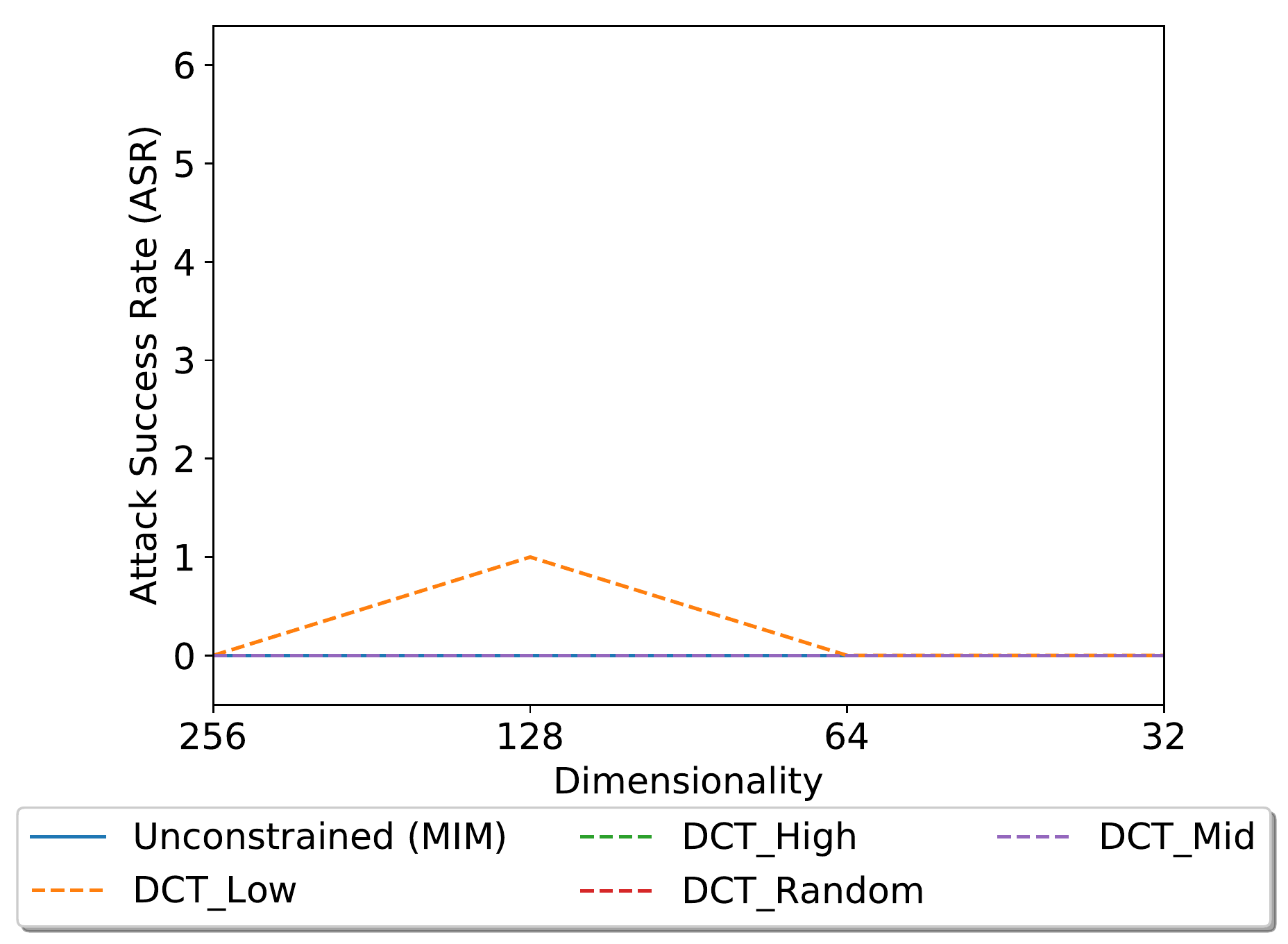}
    \caption{Targeted with $\epsilon=32$ and $\text{iterations}=10$.}
    \label{greybox34}
\end{subfigure}
\caption{\textbf{Black-box} attack from Adv\_1 to Cln.}
\label{blackbox3c}
\end{figure*}

\begin{figure*}
\begin{subfigure}{0.3\linewidth}
    \centering
    \includegraphics[width=\textwidth]{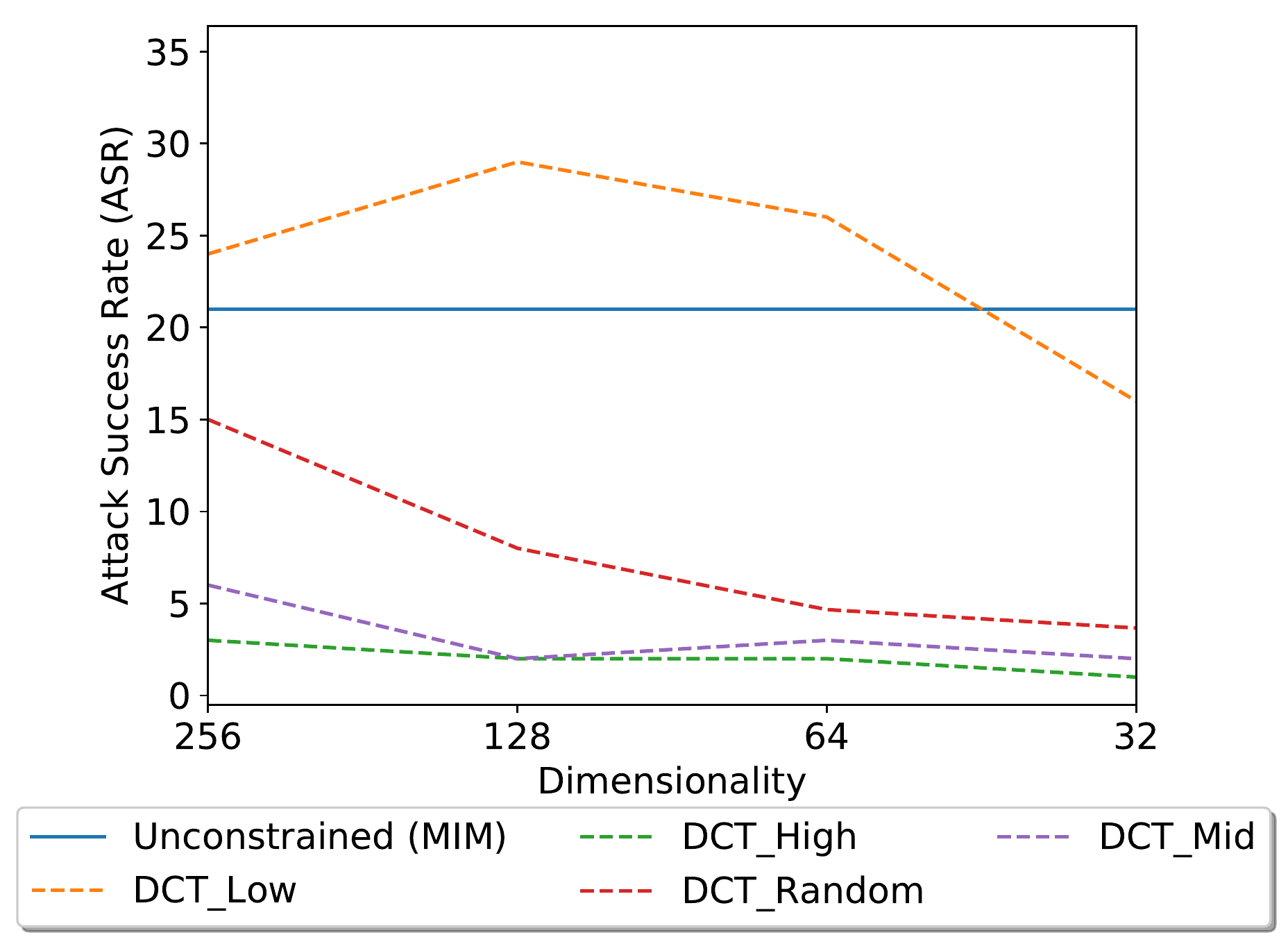}
    \caption{Non-targeted with $\epsilon=16$ and $\text{iterations}=1$.}
    \label{greybox14}
\end{subfigure}
~~~
\begin{subfigure}{0.3\linewidth}
    \centering
    \includegraphics[width=\textwidth]{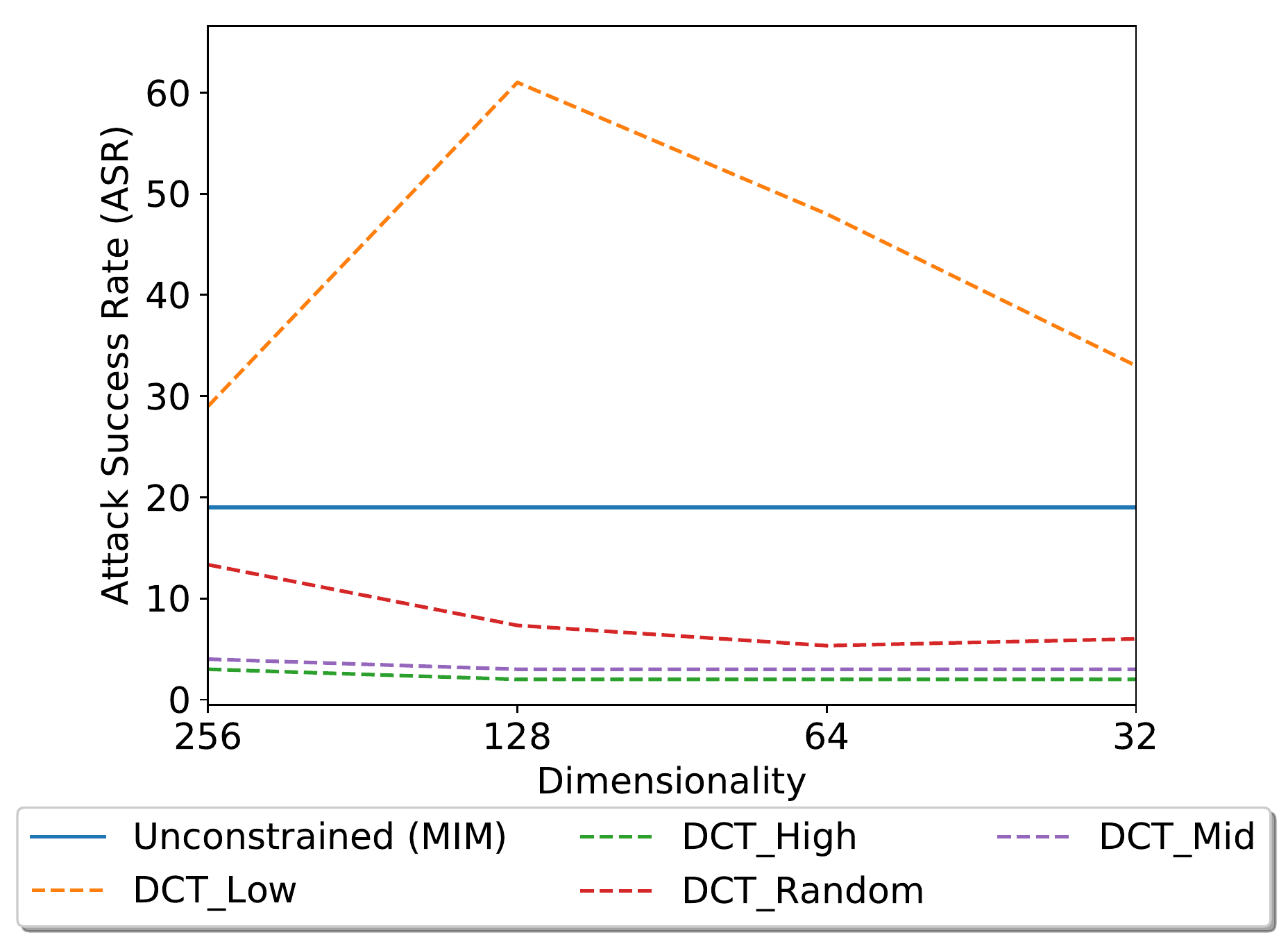}
    \caption{Non-targeted with $\epsilon=16$ and $\text{iterations}=10$.}
    \label{greybox24}
\end{subfigure}
~~~
\begin{subfigure}{0.3\linewidth}
    \centering
    \includegraphics[width=\textwidth]{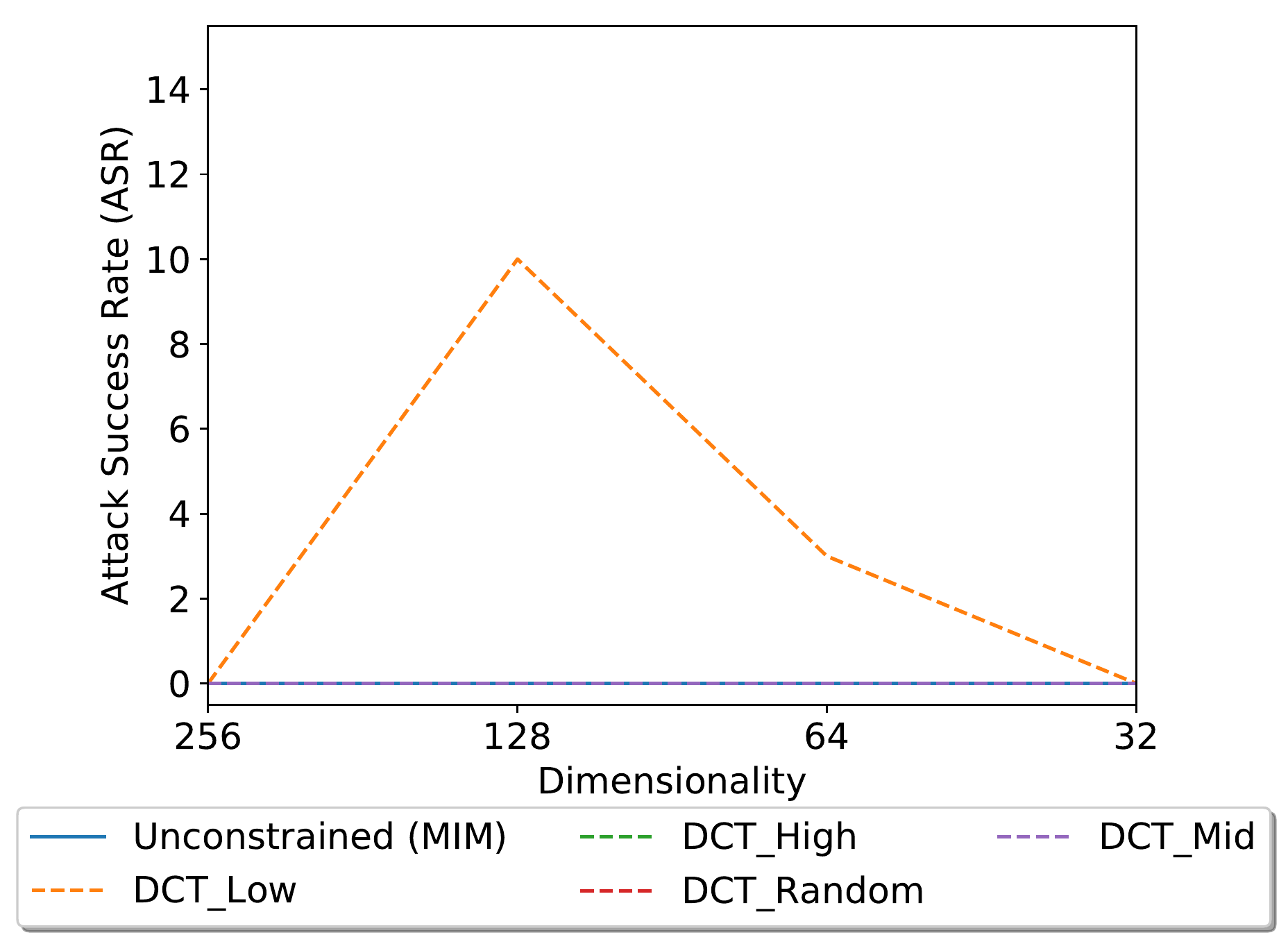}
    \caption{Targeted with $\epsilon=32$ and $\text{iterations}=10$.}
    \label{greybox34}
\end{subfigure}
\caption{\textbf{Black-box} attack from Adv\_3 to Cln.}
\label{blackbox4c}
\end{figure*}

Figure~\ref{whiteboxc} shows white-box results attacking the cleanly trained model. Figures~\ref{blackbox1c}-\ref{blackbox4c} show black-box results transferring from the source models to the cleanly trained model [Cln].
%\clearpage

\end{document}